\documentclass{article}

\usepackage{frExamplee}
\usepackage{apalike}

\usepackage{amsmath}  
\usepackage{amssymb}  
\usepackage{amsthm}
\usepackage{amsfonts}
\usepackage{mathtools}

\usepackage{paralist}

\usepackage{verbatim}

\usepackage{multirow}
\usepackage{makecell}

\usepackage{color}
\usepackage{graphicx}

\usepackage{times}
\usepackage{interval}
\usepackage[super]{nth}

\usepackage{algorithm}
\usepackage[noend]{algpseudocode}
\newcommand\algfunc[1]{\text{#1}}

\usepackage[us]{datetime}
\usepackage{caption}
\usepackage[usenames,dvipsnames,table]{xcolor}

\usepackage{soul}

\usepackage{todonotes}

\usepackage{subcaption}

\usepackage{latexsym}

\usepackage[noadjust]{cite}

\usepackage{float}

\usepackage{tikz}
\usepackage{adjustbox}
\usepackage{pgfplots}
\pgfplotsset{compat=newest}
\usetikzlibrary{shapes,fit,arrows,calc,intersections,positioning,backgrounds,decorations.markings,patterns,external}

\tikzset{external/system call={latex \tikzexternalcheckshellescape -halt-on-error
-interaction=batchmode -jobname "\image" "\texsource";
dvips -o "\image".eps "\image".dvi;
ps2eps "\image.eps"}}

\usepackage[inline]{enumitem} 
\setenumerate[0]{label=\roman*.}

\usepackage{subfiles}
\usepackage{bm}

\tikzset{
  connect/.style args={(#1) to (#2) over (#3) by #4}{
    insert path={
      let \p1=($(#1)-(#3)$), \n1={veclen(\x1,\y1)},
      \n2={atan2(\x1,\y1)}, \n3={abs(#4)}, \n4={#4>0 ?180:-180}  in
      (#1) -- ($(#1)!\n1-\n3!(#3)$)
      arc (\n2:\n2+\n4:\n3) -- (#2)
    }
  },
}

\usepackage{scalerel}
\usetikzlibrary{svg.path}
\definecolor{orcidlogocol}{HTML}{A6CE39}
\tikzset{
  orcidlogo/.pic={
    \fill[orcidlogocol] svg{M256,128c0,70.7-57.3,128-128,128C57.3,256,0,198.7,0,128C0,57.3,57.3,0,128,0C198.7,0,256,57.3,256,128z};
    \fill[white] svg{M86.3,186.2H70.9V79.1h15.4v48.4V186.2z}
    svg{M108.9,79.1h41.6c39.6,0,57,28.3,57,53.6c0,27.5-21.5,53.6-56.8,53.6h-41.8V79.1z M124.3,172.4h24.5c34.9,0,42.9-26.5,42.9-39.7c0-21.5-13.7-39.7-43.7-39.7h-23.7V172.4z}
    svg{M88.7,56.8c0,5.5-4.5,10.1-10.1,10.1c-5.6,0-10.1-4.6-10.1-10.1c0-5.6,4.5-10.1,10.1-10.1C84.2,46.7,88.7,51.3,88.7,56.8z};
  }
}
\newcommand\orcidicon[1]{\href{https://orcid.org/#1}{\mbox{\scalerel*{
  \begin{tikzpicture}[yscale=-1,transform shape]
    \pic{orcidlogo};
  \end{tikzpicture}
}{|}}}}

\makeatletter
\let\NAT@parse\undefined
\makeatother

\usepackage{hyperref} 
\hypersetup{
  colorlinks,
  citecolor=black,
  filecolor=black,
  linkcolor=black,
  urlcolor=black,
  pdfauthor={},
  pdfsubject={},
  pdftitle={}
}

\def\equationautorefname~#1\null{(#1)\null}

\setlist[itemize]{noitemsep, nosep}

\usepackage[binary-units]{siunitx}
\DeclareSIUnit\inch{inch}
\usepackage{ifthen}

\usepackage{booktabs}

\usepackage{tablefootnote}
\makeatletter
\def  \input@path{{./../fig/},{./fig/}}
\makeatother

\definecolor{dark_red}{rgb}{0.4, 0.0, 0.0}
\definecolor{light_red}{rgb}{0.8, 0.1, 0.1}
\definecolor{dark_green}{rgb}{0.0, 0.4, 0.0}
\definecolor{light_green}{rgb}{0.1, 0.8, 0.1}
\definecolor{dark_blue}{rgb}{0.0, 0.0, 0.4}
\definecolor{light_blue}{rgb}{0.1, 0.1, 0.8}
\definecolor{dark_violet}{rgb}{0.4, 0.1, 0.4}
\definecolor{light_violet}{rgb}{0.8, 0.1, 0.8}
\definecolor{dark_orange}{rgb}{0.4, 0.4, 0.1}
\definecolor{light_orange}{rgb}{0.8, 0.6, 0.1}

\usepackage[printonlyused]{acronym}
\acrodef{uav}[UAV]{Unmanned Aerial Vehicle}
\acrodef{ugv}[UGV]{Unmanned Ground Vehicle}
\acrodef{gps}[GPS]{Global Positioning System}
\acrodef{darpa}[DARPA]{Defense Advanced Research Projects Agency}
\acrodef{subt}[SubT]{Subterranean Challenge}
\acrodef{co2}[$\mathrm{CO_2}$]{carbon dioxide}
\acrodef{sar}[S\&R]{Search and Rescue}
\acrodef{ctu}[CTU]{Czech Technical University in Prague}
\acrodef{laval}[Laval]{Laval University}
\acrodef{pc}[PC]{personal computer}
\acrodef{mrna}[mRNA]{messenger ribonucleic acid}
\acrodef{lidar}[LiDAR]{Light Detection and Ranging}
\acrodef{rad}[R\&D]{Research and Development}
\acrodef{imu}[IMU]{Inertial Measurement Unit}
\acrodef{rgb}[RGB]{color}
\acrodef{rgbd}[RGBD]{color-depth}
\acrodef{agl}[AGL]{above ground level}
\acrodef{amsl}[AMSL]{above mean sea level}
\acrodef{loam}[LOAM]{LiDAR Odometry and Mapping}
\acrodef{aloam}[A-LOAM]{Advanced implementation of LOAM}
\acrodef{liosam}[LIO-SAM]{LiDAR Inertial Odometry via Smoothing and Mapping}
\acrodef{dof}[DOF]{Degrees of Freedom}
\acrodef{lkf}[LKF]{Linear Kalman Filter}
\acrodef{mems}[MEMS]{Micro-Electromechanical Systems}
\acrodef{cpu}[CPU]{Central Processing Unit}
\acrodef{gpu}[GPU]{Graphical Processing Unit}
\acrodef{cnn}[CNN]{Convolutional Neural Network}
\acrodef{fov}[FOV]{Field Of View}
\acrodef{nir}[NIR]{Near Infrared}
\acrodef{gan}[GAN]{Generative Adversarial Network}
\acrodef{slam}[SLAM]{Simultaneous Localization and Mapping}
\acrodef{vio}[VIO]{Visual-Inertial Odometry}
\acrodef{tio}[TIO]{Thermal-Inertial Odometry}
\acrodef{lio}[LIO]{LiDAR-Inertial Odometry}
\acrodef{stix}[STIX]{Subterranean Integration Exercise}
\acrodef{niosh}[NIOSH]{National Institute for Occupational Safety \& Health}
\acrodef{icp}[ICP]{Iterative Closest Point}
\acrodef{gicp}[GICP]{Generalized Iterative Closest Point}
\acrodef{ape}[APE]{Absolute Position Error}
\acrodef{ate}[ATE]{Absolute Trajectory Error}
\acrodef{gmm}[GMM]{Gaussian Mixture Models}
\acrodef{csiro}[CSIRO]{CSIRO Data61}
\acrodef{rssi}[RSSI]{Received Signal Strength Indicator}
\acrodef{fcu}[FCU]{Flight Control Unit}
\acrodef{se3}[SE(3)]{Special Euclidean group of dimension 3}
\acrodef{pcb}[PCB]{Printed Circuit Board}
\acrodef{led}[LED]{Light-Emitting Diode}
\acrodef{ltvmap}[LTVMap]{lightweight topological-volumetric map}

\providecommand{\keywords}[1]
{
  \textbf{\textit{Keywords ---}} #1
}

\newcommand{\eg}{{e.g.,}}
\newcommand{\ie}{{i.e.,}}

\newcommand{\heading}{$\eta$}
\renewcommand{\vec}[1]{\mathbf{#1}}
\newcommand{\set}[1]{\mathcal{#1}}
\newcommand{\pnt}[1]{\mathbf{#1}}
\newcommand\mat{\mathbf}

\newcommand{\bemat}[1]
{
  \begin{bmatrix}
    #1
  \end{bmatrix}
}

\newcommand{\norm}[1]{\left\lVert#1\right\rVert}
\newcommand{\abs}[1]{\left\lvert#1\right\rvert}

\newcommand{\tstep}[2][]{_{#1[#2]}}
\newcommand*{\tran}{^{\intercal}}

\DeclareMathOperator*{\argmax}{argmax}
\newcommand{\hyp}{\mathcal{H}}
\newcommand{\dete}{\mathcal{D}}

\newcommand{\tablehdg}[1]{\textbf{#1}}
\newcommand{\tablesize}{\small}

\newcommand{\uavred}{\textit{red}}
\newcommand{\uavgreen}{\textit{green}}
\newcommand{\uavblue}{\textit{blue}}

\title{%
    UAVs Beneath the Surface: Cooperative Autonomy for Subterranean Search and Rescue in DARPA SubT
  }

  \author{%
    Mat\v{e}j Petrl\'{i}k $^{*}$
    \And
    Pavel Petr\'{a}\v{c}ek $^{*}$
    \And
    V\'{i}t Kr\'{a}tk\'{y} $^{*}$
    \And
    Tom\'{a}\v{s} Musil $^{*}$
    \And
    Yurii Stasinchuk $^{*}$
    \And
    Matou\v{s} Vrba $^{*}$
    \And
    Tom\'{a}\v{s} B\'{a}\v{c}a $^{*}$
    \And
    Daniel He\v{r}t $^{*}$
    \And
    Martin Pecka $^{*}$
    \And
    Tom\'{a}\v{s} Svoboda $^{*}$
    \And
    Martin Saska
    \thanks{%
      Authors are with the Faculty of Electrical Engineering, Czech Technical University in Prague, Czech Republic. 
      The corresponding author's e-mail: {\tt matej.petrlik@fel.cvut.cz}.}
    }

\begin{document}
    \maketitle

    \begin{abstract}

This paper presents a novel approach for autonomous cooperating UAVs in search and rescue operations in subterranean domains with complex topology. 
The proposed system was ranked second in the Virtual Track of the DARPA SubT Finals as part of the team CTU-CRAS-NORLAB.
In contrast to the winning solution that was developed specifically for the Virtual Track, the proposed solution also proved to be a robust system for deployment onboard physical UAVs flying in the extremely harsh and confined environment of the real-world competition. 
The proposed approach enables fully autonomous and decentralized deployment of a UAV team with seamless simulation-to-world transfer, and proves its advantage over less mobile UGV teams in the flyable space of diverse environments.
The main contributions of the paper are present in the mapping and navigation pipelines.
The mapping approach employs novel map representations --- SphereMap for efficient risk-aware long-distance planning, FacetMap for surface coverage, and the compressed topological-volumetric LTVMap for allowing multi-robot cooperation under low-bandwidth communication.
These representations are used in navigation together with novel methods for visibility-constrained informed search in a general 3D environment with no assumptions about the environment structure, while balancing deep exploration with sensor-coverage exploitation.
      The proposed solution also includes a visual-perception pipeline for on-board detection and localization of objects of interest in four RGB stream at \SI{5}{Hz} each without a dedicated GPU.
Apart from participation in the DARPA SubT, the performance of the UAV system is supported by extensive experimental verification in diverse environments with both qualitative and quantitative evaluation.
    \end{abstract}




    \subsection*{Support materials}
    \vspace{-1em}
    \label{sec:multimedia_materials}
    The paper is supported by the multimedia materials available at \href{http://mrs.felk.cvut.cz/fr2022darpa}{\texttt{mrs.felk.cvut.cz/fr2022darpa}}.
    Open-source implementation of the core of the \acs{uav} system is available at \href{https://github.com/ctu-mrs/mrs_uav_system}{\texttt{github.com/ctu-mrs/mrs\_uav\_system}}.
    The \acs{slam} datasets are available at \href{https://github.com/ctu-mrs/slam_datasets}{\texttt{github.com/ctu-mrs/slam\_datasets}}.
    The visual detection datasets are available at \href{https://github.com/ctu-mrs/vision_datasets}{\texttt{github.com/ctu-mrs/vision\_datasets}}.

    \pagebreak


    \section{Introduction}

    The research of new robotic technologies and solutions is accelerating at an unprecedented rate mainly in case of aerial robotics.
    Technological development is improving many areas of our lives and, hopefully, even the future of humanity.
    The authors of~\cite{shakhatreh2019unmanned} reviewed current research trends and future insights on potential \ac{uav} use for reducing risks and costs in civil infrastructure.
    The survey of \ac{uav} applications is accompanied by a discussion of arising research challenges and possible ways to approach them.

    Identifying research paths leading to disruptive technologies that can achieve success in the near future is the goal of the US organization \ac{darpa}.
    The most famous technologies that began as \ac{darpa} research are the Internet and \ac{gps}.
    The recent advent of self-driving cars also began as a competition organized by \ac{darpa} --- the \ac{darpa} Grand Challenge and the \ac{darpa} Urban Challenge.
    The latest competition, the \ac{darpa} \ac{subt} focuses on the development of robotic systems to autonomously search subterranean environments.

    The motivation behind searching subterranean environments is to gain situational awareness and assist specialized personnel in specific missions.
    Such missions may include: assessing the structural integrity of collapsed buildings, tunnels, or mines; exploration of a newly discovered branch in a cave network; or searching for lost persons.
    These tasks can often be life-threatening to human workers as many hazards are present in subterranean environments.
    In order to reach survivors quickly in unstable caves or partially collapsed burning buildings, first responders, such as emergency rescuers and firefighters, may potentially put their lives at risk.
    In firefighting tasks, fires can be either localized and reported to personnel by robots or the robots can even directly begin extinguishing flames if the presence of human firefighters is too risky~\cite{spurny2021autonomous,pritzl2021autonomous,martinez2022skyeye}.
    In such scenarios, ceilings can suddenly collapse, toxic gas can appear in a mine, flames can extend to an escape corridor, or a cave cavity can flood with water.
    In distress situations, it is essential to swiftly coordinate the rescue operation as the survivors of a catastrophe might need acute medical assistance or have a limited amount of resources available, namely oxygen and water.
    However, without conducting a proper reconnaissance of the environment and assessing the potential risks prior to the rescue mission, the involved rescuers are exposed to a much higher probability of injury.

    To reduce the possibility of bodily harm or to avoid risks altogether, a robotic system can be sent on-site before the rescuers in order to either quickly scout the environment and report any hazards detected by the onboard sensors, or directly search for the survivors.
    The rescue mission can be further sped up by deploying a team of robots capable of covering larger areas and offer redundancy in case of losses of some robot units in harsh environments.
    Multi-robot teams can also consist of heterogeneous agents with unique locomotion modalities to ensure traversability of various terrains, including muddy ground, stairs, and windows, which is discussed in the overview of collaborative \ac{sar} systems~\cite{queralta2020collaborative}.
    Similarly, sensing modalities can be distributed among individual robots to detect various signs of hazards, such as increased methane levels or the potential presence of survivors deduced from visual or audio cues.
    Mounting all sensors on a single platform would negatively affect its dimensions and, consequently, its terrain traversability as it may not be able to fit into narrow passages, such as crawlspace-sized tunnels or doorways.
    It would also mean a single point of failure for the rescue operation.
    On the other hand, the operation of a single robot can be managed by just one person, while commanding a robot team may be unfeasible for a single operator.
    Assigning each robot to an individual operator would also be an ineffective allocation of resources.
    Moreover, the range of the robot would be limited by the communication link to the operator.
    To provide a valuable tool for the rescue team, the robots must be able to move through the environment on their own and infer about the environment using their sensor data.
    The rescuer can then also act as an operator, providing only high-level inputs to the robotic system to bias their behavior based on a-priori information (\eg{} someone was last seen on the east side of the third floor).
    The research and development of such autonomous systems for assisting first-responders is the primary focus of the \ac{sar} robotics, and also the motivation for the \ac{sar} \ac{uav} system presented in this paper.

    The robotic platforms typically considered for \ac{sar} tasks are categorized into wheeled, tracked, legged, marine, and aerial platforms~\cite{delmerico2019current}.
    Among these locomotive modalities, aerial robots are considered to have the highest traversal capabilities since they can fly over most obstacles which are untraversable by other platforms.
    One example of an autonomous aerial research platform for \ac{sar} is found in ~\cite{tomic2012toward}.
    The mobility of \acp{uav} also surpasses other robot types thanks to its dynamic flight which can achieve large velocities and accelerations.
    These qualities make \acp{uav} ideal for swift environmental scouting for gaining initial knowledge about a situation.
    As such, the aerial platform is predetermined to be deployed as the first robot during the first minutes of the rescue operation. 
    A team deployed in an outdoor multi-\ac{uav} disaster response task~\cite{alotaibi2019lsar} can effectively cover a large search area and minimize the time to find and reach survivors.
    On the other hand, \acp{uav} cannot operate for extended periods of time due to their limited flight time, and the sensory equipment is limited by the maximum payload of the \ac{uav}.
    Some sensing modalities might even be unsuitable for the use on aerial robots due to their propulsion system, \eg{} detecting gas due to the aerodynamic effects of the propellers, or sound detection due to noisy operation.
    Due to the aforementioned pros and cons of \ac{uav} platforms, it is convenient to combine the capabilities of other robot types to form a heterogeneous robotic team.

    This manuscript proposes an autonomous cooperative \ac{uav} system for \ac{sar} as part of the CTU-CRAS-NORLAB team, which participated in the \ac{darpa} \ac{subt}.
    The team consisted of \ac{ctu} and \acl{laval}.

    \subsection{DARPA SubT challenge}
    \label{sec:darpa}

    After major success in accelerating the development of self-driving cars in the Grand Challenges of 2004 and 2005 and the Urban Challenge in 2007, \ac{darpa} announced the \acf{subt}~\cite{orekhov2022darpa} for the years 2017-2021 to advance the state of the art of \ac{sar} robotics.
    Participants had to develop robotic solutions for searching subterranean environments for specific objects that would yield points if reported with sufficient accuracy.
    To achieve the task at hand, the competitors had to develop complex multi-robot systems spanning nearly all research areas of mobile robotics, from design of the robotic platforms to high-level mission planning and decision making.

\begin{figure}
      \newcommand\scale{0.09}
      \centering
     \includegraphics[width=\scale\textwidth,trim={0 0 0 0},clip]{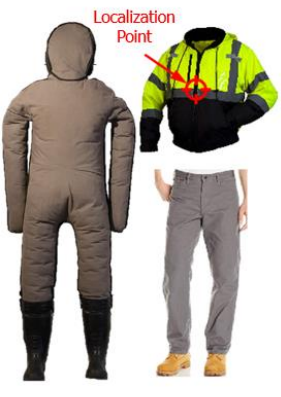}
     \includegraphics[width=\scale\textwidth,trim={0 0 0 0},clip]{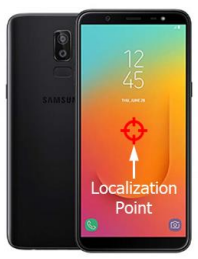}
     \includegraphics[width=\scale\textwidth,trim={0 0 0 0},clip]{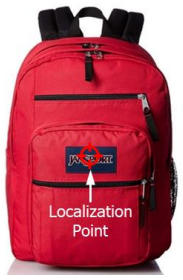}
     \includegraphics[width=\scale\textwidth,trim={0 0 0 0},clip]{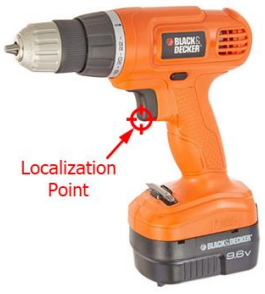}
     \includegraphics[width=\scale\textwidth,trim={0 0 0 0},clip]{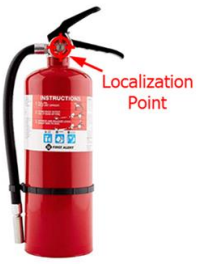}
     \includegraphics[width=\scale\textwidth,trim={0 0 0 0},clip]{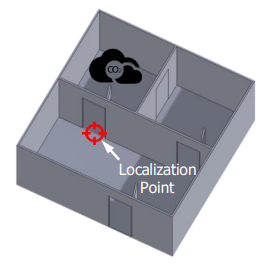}
     \includegraphics[width=\scale\textwidth,trim={0 0 0 0},clip]{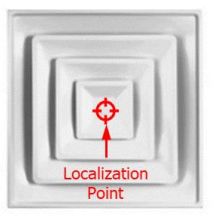}
     \includegraphics[width=\scale\textwidth,trim={0 0 0 0},clip]{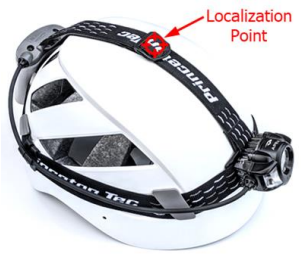}
     \includegraphics[width=\scale\textwidth,trim={0 0 0 0},clip]{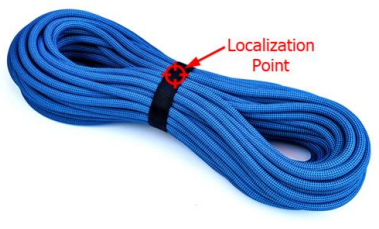}
     \includegraphics[width=\scale\textwidth,trim={0 0 0 0},clip]{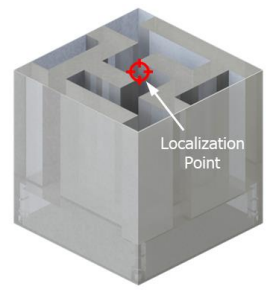}
     \caption{\label{fig:artifacts} 
      All 10 artifacts searched for in the final event of \ac{darpa} \ac{subt} (image courtesy of \acs{darpa}).
      The operator had to submit the position of the identified artifact with accuracy better than \SI{5}{\meter}.
      While the first three artifacts (survivor, cellphone, and backpack) were present in all circuits, the drill and the fire extinguisher were tunnel-specific.
      Similarly, the gas and vent were located in the urban environment, and the helmet with rope could be found in the caves.
      The last artifact (the cube) was introduced only for the final event.
      }
    \end{figure}

\begin{figure}
      \centering
      \includegraphics[width=0.495\textwidth]{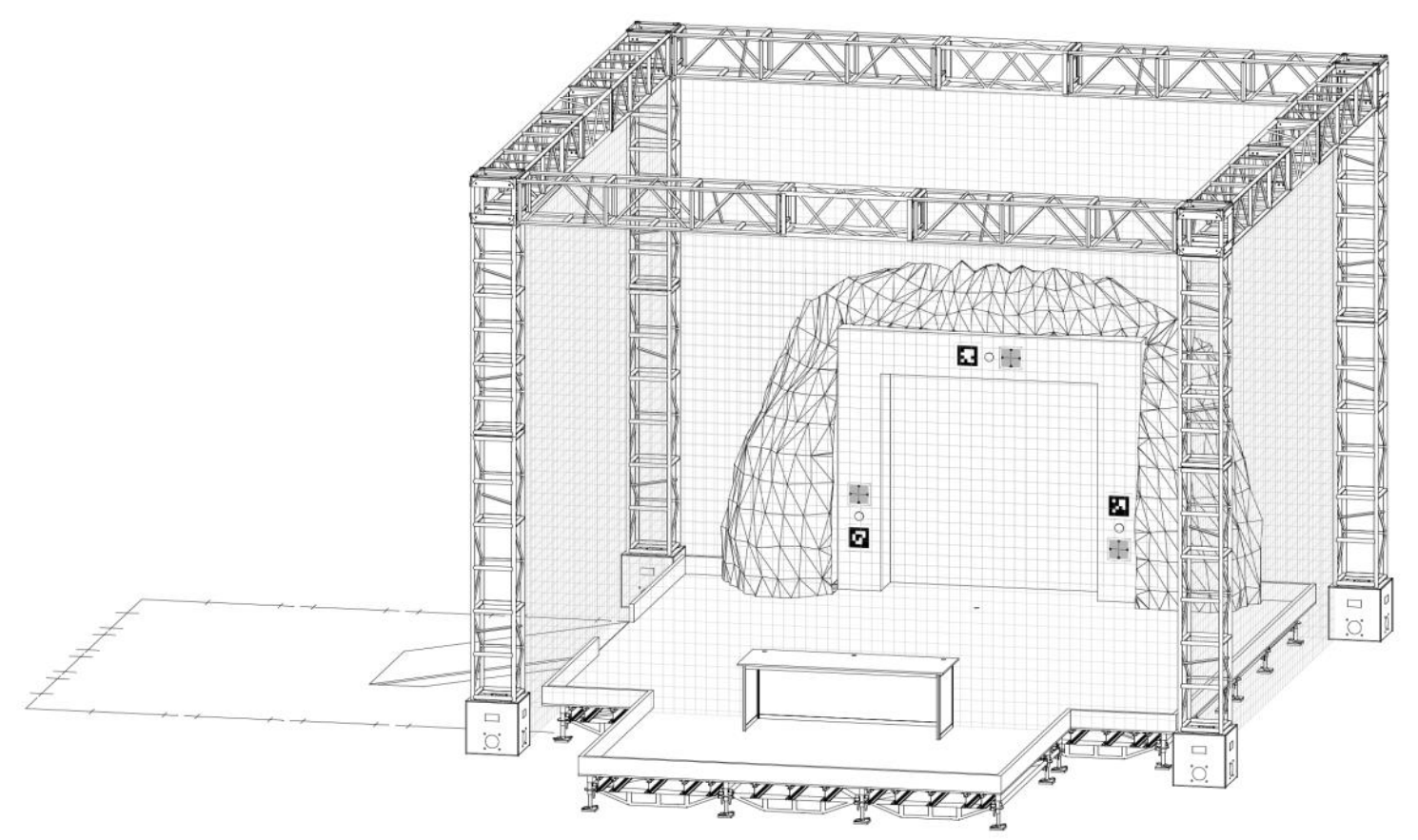}
      \includegraphics[width=0.495\textwidth]{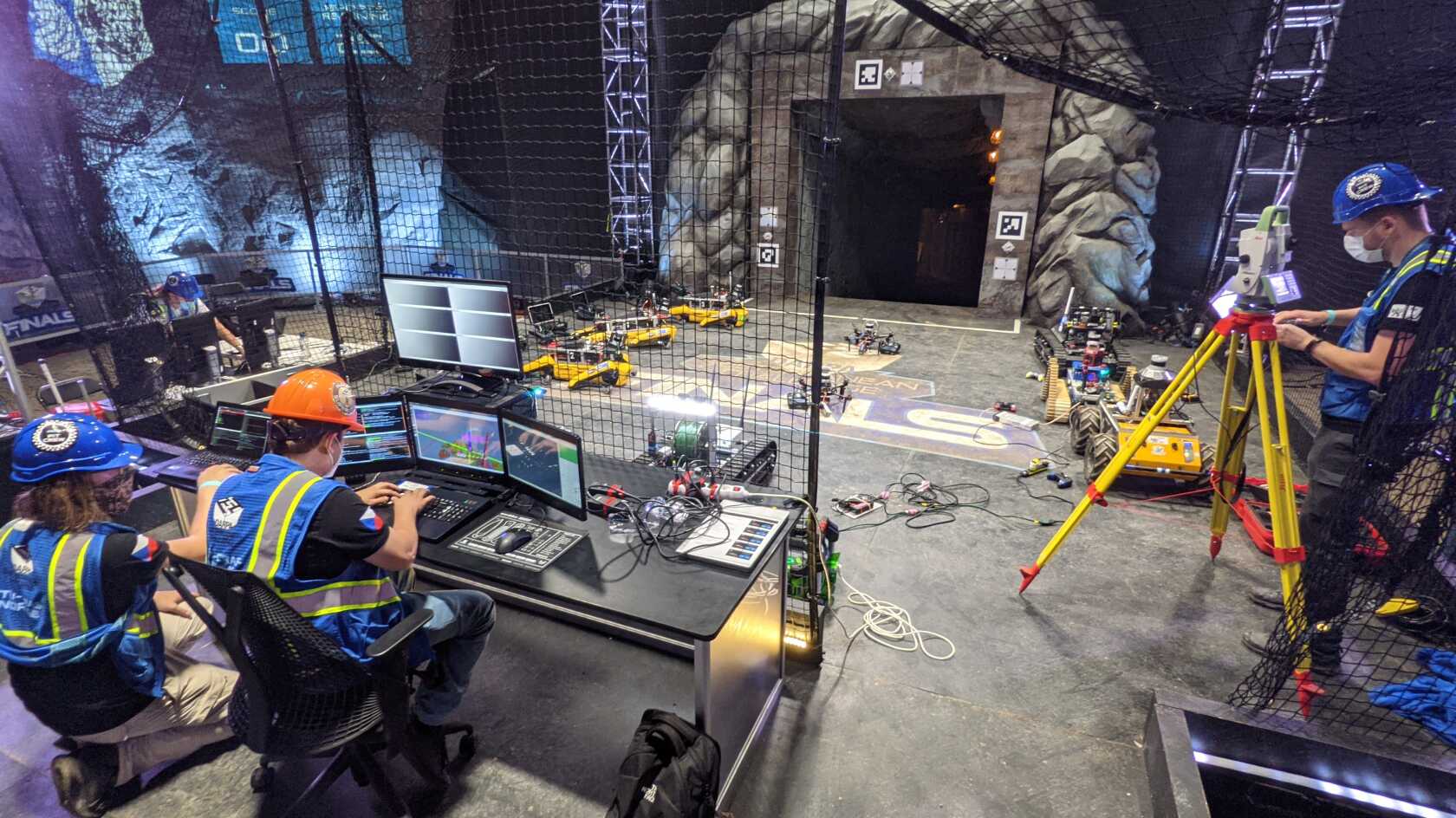}
      \caption{\label{fig:staging_area} 
      The bounded staging area (image courtesy of \acs{darpa}) is the only place where the human crew members can handle the robots. 
      The person sitting behind the displays is the operator who is the only one allowed to issue commands to the team of robots, and also to view and interpret mission data.
      }
    \end{figure}

    The rules of the competition can be summarized in a few points.
    Each team has a dedicated time slot, or \emph{run}, to send their robots into a previously unvisited course and search for specific objects, referred to as artifacts~(\autoref{fig:artifacts}). 
    Each run starts at a predefined time and ends exactly one hour later.
    A single team is present on the course at a time during which they can deploy an unconstrained number of robots of arbitrary size.
    The movement of team personnel and their handling of robots is allowed only in the area in front of the entrance to the course, as shown in~\autoref{fig:staging_area}.
    Only robots can enter the course and only one human operator/supervisor can command the robots and access the data they acquire during the run. 
    These conditions should mimic the conditions of a real \ac{sar} robotic mission.
    The operator can report the type and position of an artifact. If the type was correct and the reported position was not further than \SI{5}{\meter} from the true position, the team was awarded one point.
    The team with the highest score wins the prize according to~\autoref{tab:prize_table}.
    For a more detailed description of the challenge, see~\cite{orekhov2022darpa}.

    \begin{table}
      \parbox{.45\linewidth}{
        \centering
        \caption{\label{tab:prize_table}
        The prize money awarded for achieving the first three places in the final event.
        }
        \centering
        \tablesize
        \begin{tabular}{ccc}
          \toprule
          \tablehdg{Place} & \tablehdg{Systems Track} & \tablehdg{Virtual Track} \\
          \midrule
          1. & \$2M & \$750K \\
          2. & \$1M & \$500K \\
          3. & \$500K & \$250K \\
          \bottomrule
        \end{tabular}
      }
      \hfill
      \parbox{.45\linewidth}{
        \centering
        \caption{\label{tab:env_size}
        Approximate distribution of the environment cross-section as announced by the organizers before the final event.
        }
        \centering
        \tablesize
        \begin{tabular}{cc}
          \toprule
          \tablehdg{Cross-section (\si{\meter\squared})} & \tablehdg{Distribution} \\
          \midrule
          $<$5 & 65\% \\
          5-100 & 20\% \\
          $>$100 & 15\% \\
          \bottomrule
        \end{tabular}
      }
    \end{table}

    To encourage the development of high-level components without worrying about the resilience of the hardware in harsh subterranean conditions and also to enable teams without many resources and/or physical robots to compete, a virtual version (Virtual Track) of the competition was run in parallel to the physical Systems Track.
    The solutions of the Virtual Track were uploaded as Docker images (one image per robot) to the Gazebo-based Cloudsim simulation environment, where the entire run was simulated.
    Every team could use the Cloudsim simulator to test their approaches in practice worlds prior to the actual competition.

\begin{figure}
      \centering
      \includegraphics[width=0.325\textwidth]{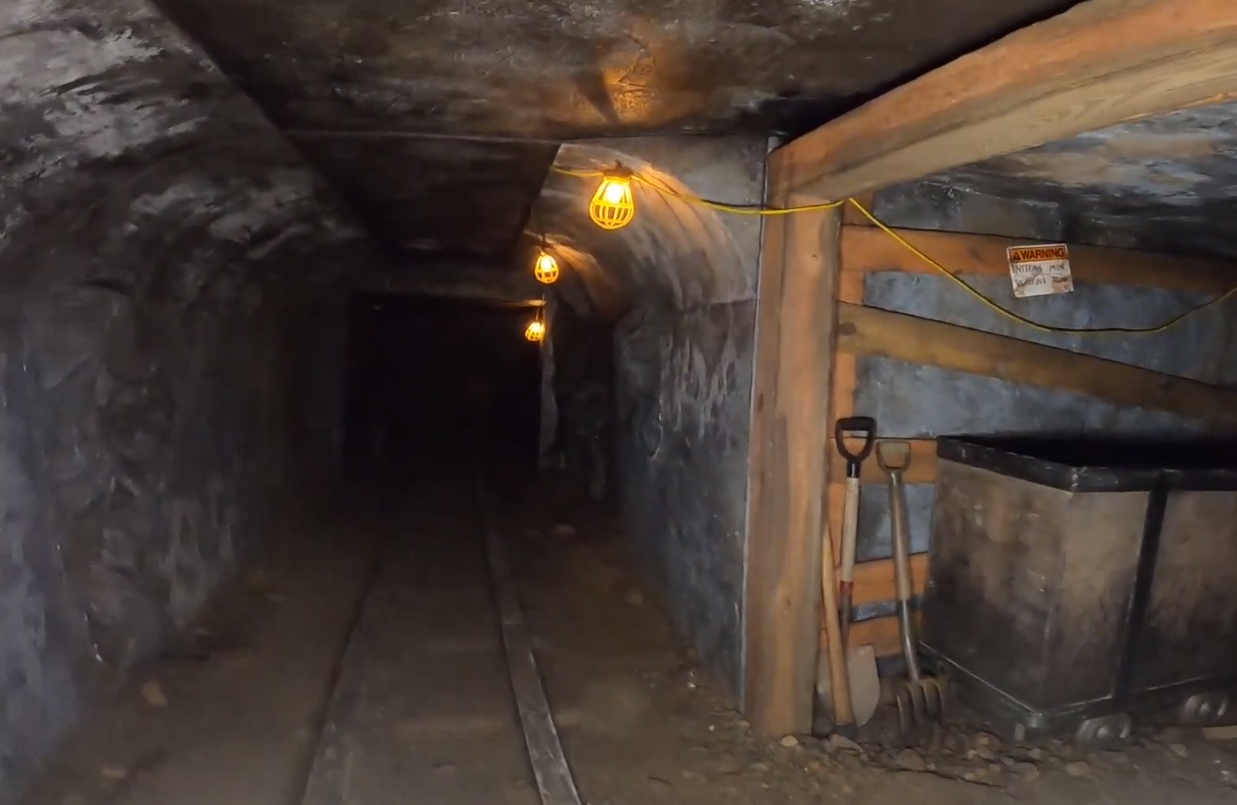}
      \includegraphics[width=0.325\textwidth]{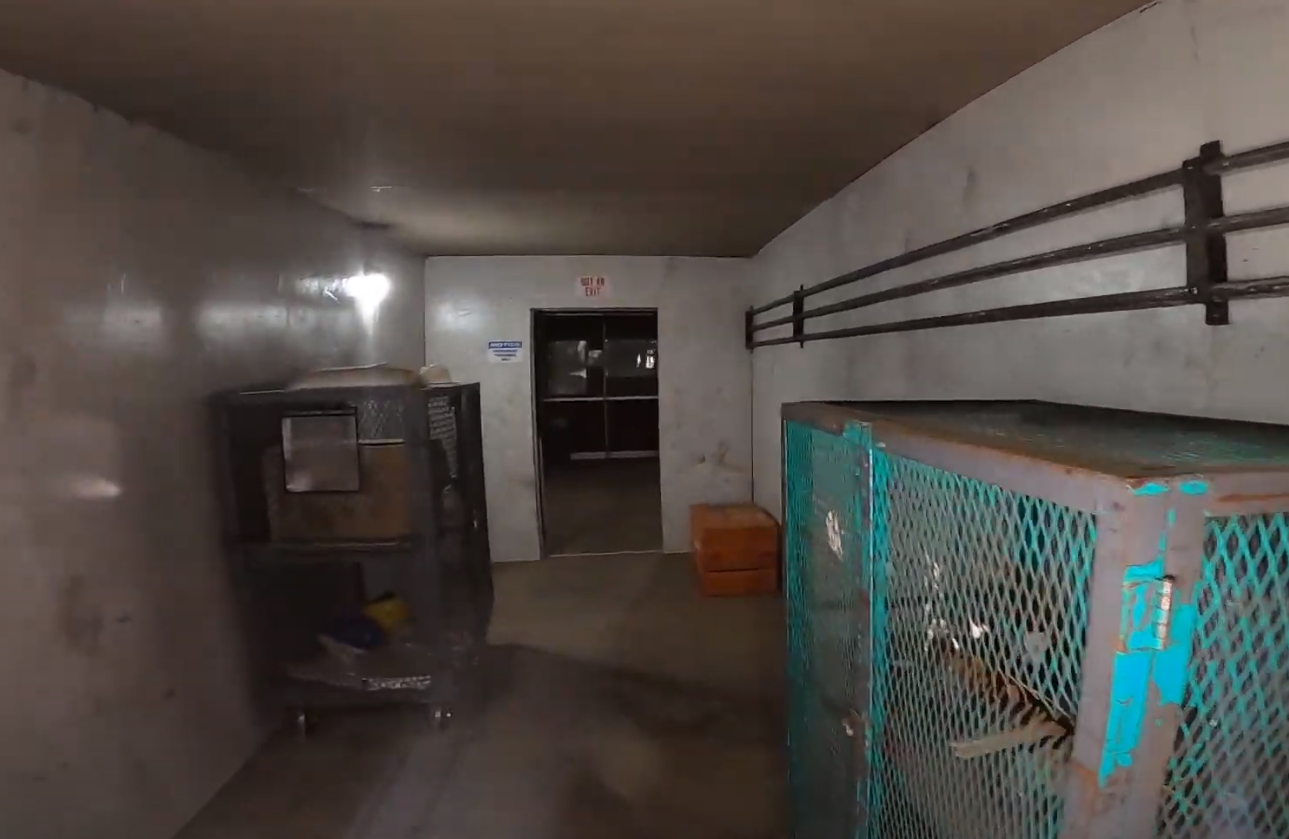}
      \includegraphics[width=0.325\textwidth,trim={0 45pt 0 0},clip]{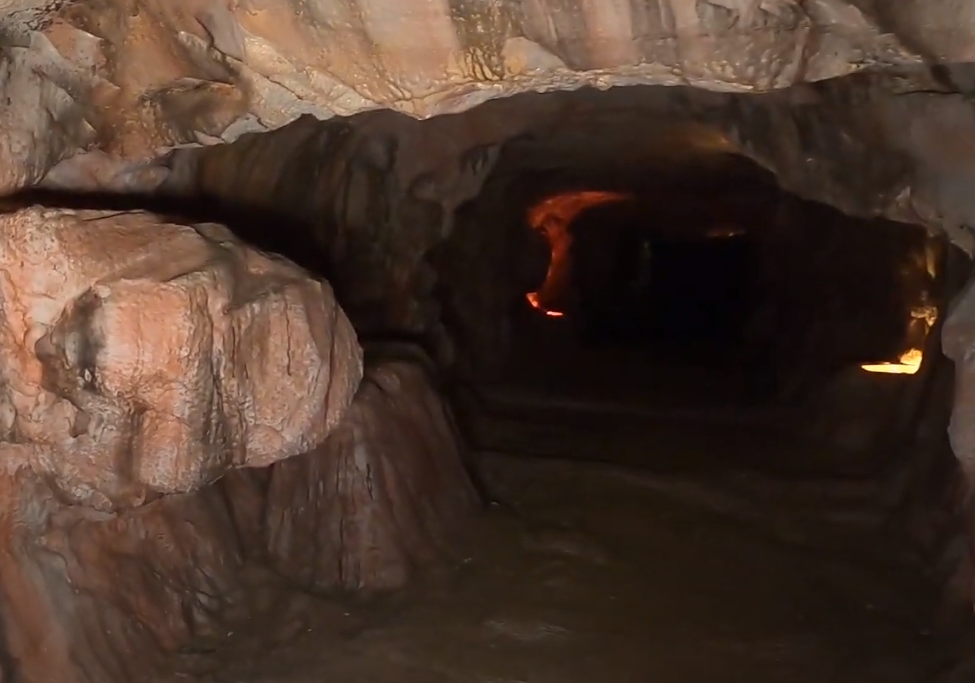}
      \includegraphics[width=0.325\textwidth]{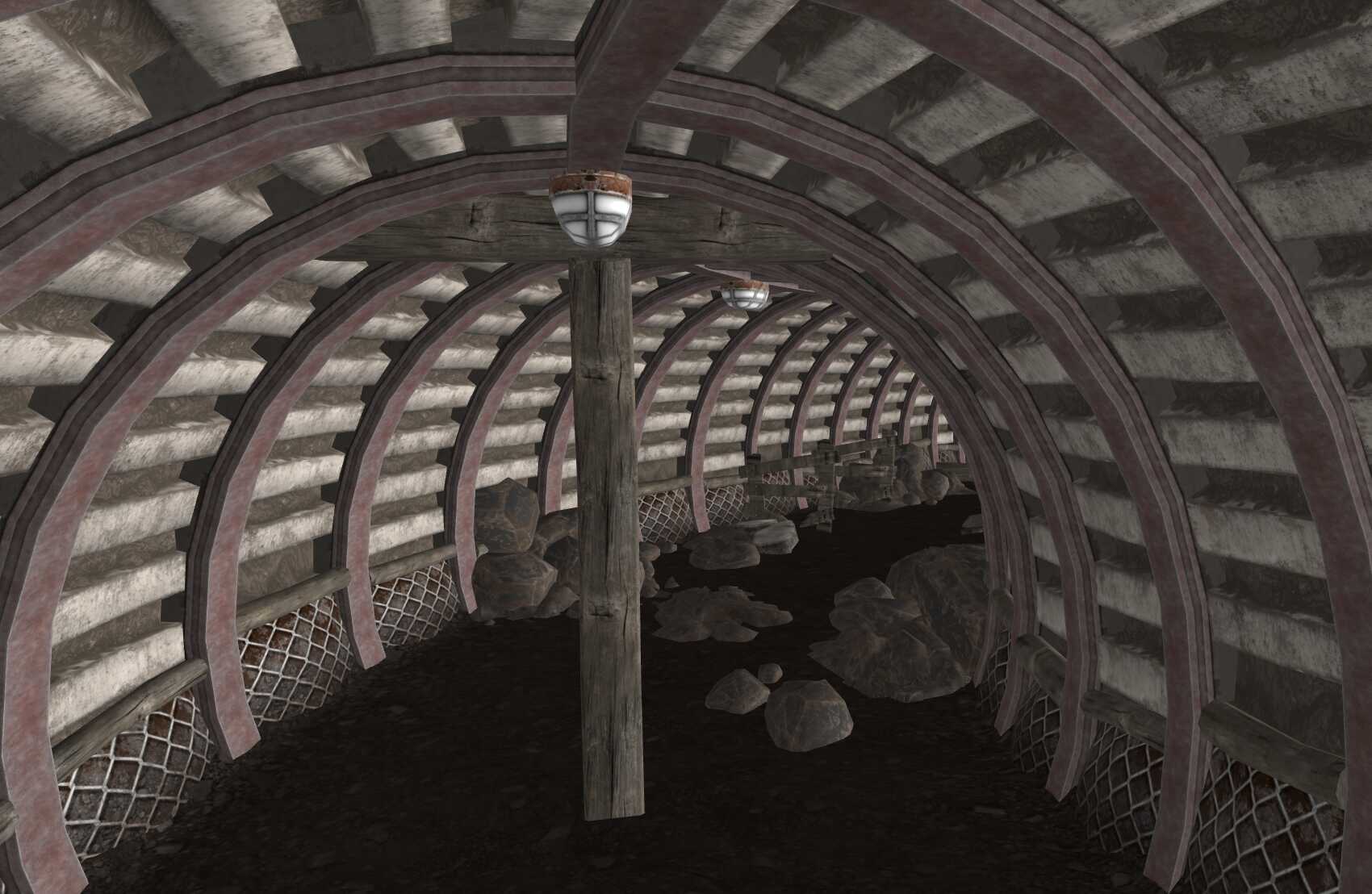}
      \includegraphics[width=0.325\textwidth,trim={0 70pt 0 0},clip]{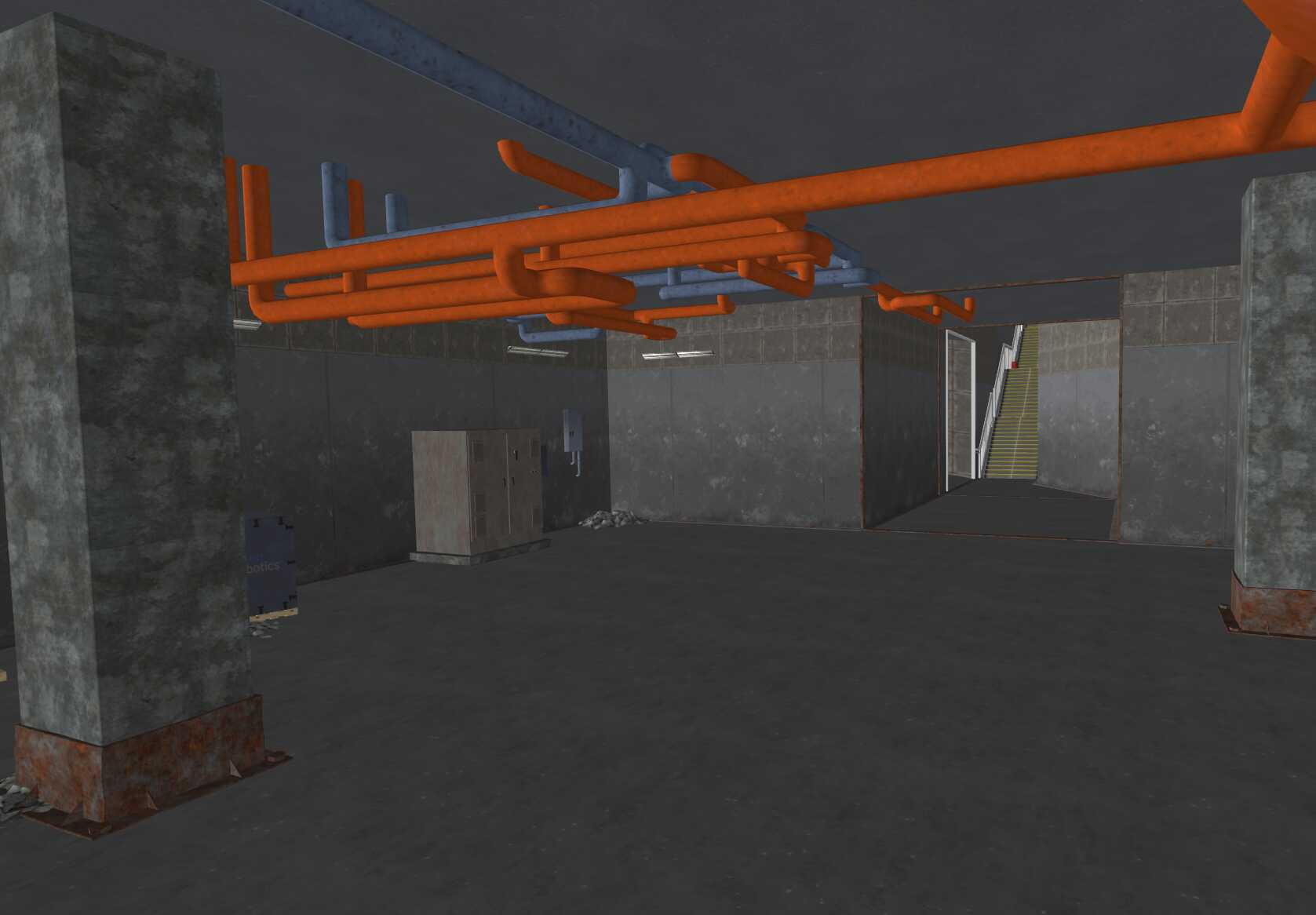}
      \includegraphics[width=0.325\textwidth,trim={0 55pt 0 0},clip]{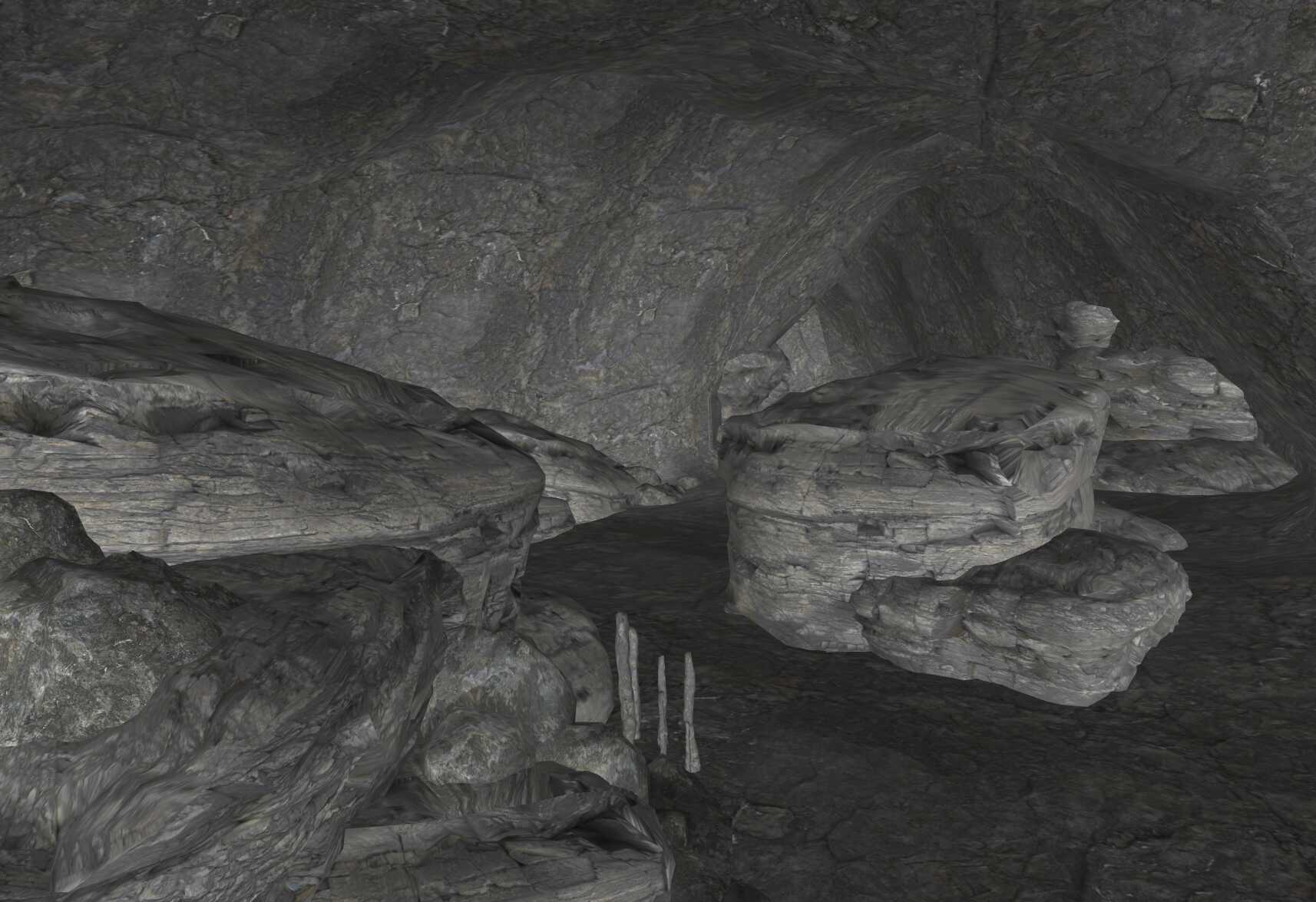}
     \caption{\label{fig:circuit_envs}
      Three types of subterranean environments found in the competition, each challenging for the robot team in a different way.
      From left to right: tunnel, urban, and cave.
      The top row shows examples of environments from the system track of the final event, while the virtual worlds are pictured in the bottom row.
      }
    \end{figure}

    The competition was further subdivided into individual circuits, which were events in the specific subterranean environments of a tunnel, cave, and urban space. Examples of each environment are shown in~\autoref{fig:circuit_envs}. 
    The surroundings were chosen to correlate with typical real \ac{sar} sites to assure the applicability of the systems developed during the competition.
    Every type of environment differs in size, geometric dimensions, traversability conditions, and requirements on perception modalities.
    The specifics of tunnel-like environments are summarized in~\cite{tardioli2019ground} with 10 years of experience in \ac{sar} ground robots research.
    The role of mobile robots in rescue missions after mine disasters is discussed in~\cite{murphy2009mobile}.
    The final event combined all of the previous environments for the ultimate challenge.

    We participated in the competition first as a non-sponsored team.
    In the Tunnel Circuit, we won \nth{1} place among the non-sponsored teams and \nth{3} place in total, which earned us \$200,000.
    This success was repeated in the Urban Circuit with the same place achieved but this with time larger prize money \$500,000.
    Thanks to consistent performance in both circuits, \ac{darpa} awarded our team the funding for the Final Event, which allowed us to acquire more capable hardware.
    The approach presented in this paper is the result of \ac{uav} research, development, and testing over the whole 3-year-long period.



    \section{Related work}
    \label{sec:related_work}

    The state of the art in rescue robotics is coherently summarized in the survey~\cite{delmerico2019current}, which concerns both hardware and software.
    On the hardware side, different robot morphologies, locomotion types, and platform designs are categorized. Regarding software, the survey concerns perception and control algorithms.
    The authors interviewed experts on disaster response and humanitarian aid to understand the situation and needs of rescuers.
    
    Here, we provide an overview of the solutions for perception in adverse conditions of the underground environments, methods of localization and mapping for precise and reliable navigation, and techniques for safe traversal of narrow corridors.
    A summary of systems deployed in previous circuits of \ac{darpa} \ac{subt} follows.
    Finally, relevant datasets are referenced in order to prompt further research effort in the \ac{sar} area.

\subsection{Degraded sensing}
\label{sec:sota_perception}
    Perception in subterranean environments faces constant degradation of the sensor outputs due to the harsh conditions of such places.
    The underground climate is often filled with impervious dust (particularly in mines), where any movement agitates the settled layer of fine dirt and mineral particles.
    On the other hand, caves are typically humid ecosystems, where dense mud replaces the dust layer found in mines.
    However, the elevated humidity forms droplets of fog, which corrupt the measurements of most visible or \ac{nir} light-based sensor modalities, and also causes frequent reflections on wet surfaces.
Radars can reliably penetrate smoke, dust, and fog, and after post-processing using, e.g., \acp{gan}~\cite{goodfellow2014generative}, a 2D occupancy grid for navigation~\cite{lu2020see} can be constructed. 
Another reliable sensing modality for when images from \ac{rgb} cameras are polluted by dust or fog is thermal imaging, which, in~\cite{khattak2019robust}, is used for the localization of robots in areas with airborne obscurants.
Our approach goes beyond these works by employing intensity-based filtering of the \ac{lidar} data, and thus no additional sensors are necessary even in dense clouds of dust.

\subsection{Localization and mapping}
\label{sec:sota_slam}

Recent developments in \ac{sar} robotics sparked the research of more precise local pose estimation algorithms (also referred to as odometry), as well as long-term globally-consistent trajectory and multi-robot map fusion of all agents of the robotic team.
The state-of-the-art methods were published in~\cite{cadena2016past}, where the challenges and future direction of the \ac{slam} development are also identified.
The demands on low control error and robustness to degraded sensor data in the narrow subterranean environments present in the \ac{darpa} \ac{subt} pushed all contesting teams to either adapt and improve an existing method to be usable in the extreme conditions, or to develop a new \ac{slam} tailored to this specific domain.

Team CoSTAR developed a \ac{lidar} odometry solution~\cite{palieri2020locus} based on \ac{gicp} matching of \ac{lidar} scans with initialization from \ac{imu} and wheel odometry, including the possibility of extension to other odometry sources, such as \ac{vio}.
The method is shown to outperform state-of-the-art localization methods on the datasets from Tunnel and Urban circuits. An ablation study presents the influence of individual components on the total \ac{ape}.
All presented experiments are conducted with ground robots.
The localization of aerial vehicles is handled by a resilient HeRo state estimation system~\cite{santamaria2019towards}.
The state estimation stack considers heterogeneity and redundancy in both sensing and state estimation algorithms in order to ensure safe operation, even under the failure of some modules.
Failures are detected by performing confidence tests on both data and algorithm health.
If a check does not pass successfully, the resiliency logic switches to the algorithm with the best confidence, similar to our previous solution published in~\cite{baca2021mrs}.
The local odometry of~\cite{palieri2020locus,santamaria2019towards} is accompanied by loop closure detection and pose graph optimization locally on each robot, as well as globally on the base station. This optimizes the trajectories of all robots for a multi-robot centralized \ac{slam} solution~\cite{ebadi2020lamp}.
A decentralized \ac{slam} solution for \acp{uav}~\cite{lajoie2020door} performs distributed outlier-resilient pose graph optimization when another agent is within communication range.
This method can be used with either a stereo camera or a \ac{lidar}, and is evaluated on a dataset from the Tunnel Circuit.

The long, featureless corridors that are often present in man-made tunnels lead to unobservability of the motion along the degenerate direction, which leads to significant drift.
Promising approaches, such as~\cite{liosam2020shan,xu2022fast}, constrain the solution of the optimization problem using the preintegrated \ac{imu} measurements. This helps to reduce the localization drift under unfavorable environmental geometry.
Nevertheless, the vibrations induced by spinning propellers degrade the inertial measurements, and can thus negatively affect the localization precision.
Approaches, such as those seen in~\cite{ebadi2021dare}, detect the geometrical degeneracy using the ratio of the most observable and the least observable directions. 
This ratio is then used to determine loop closure candidates to reduce the drift along the degenerate direction.
Similarly,~\cite{zhang2016degeneracy} handles environment degeneracy in state estimation by not updating the solution in detected degenerate directions.
Another possibility is to combine the 3D \ac{lidar} method with a direct visual odometry method (e.g.,~\cite{alismail2016direct}), which tracks image patches by minimizing the photometric error.
This approach, which is shown in~\cite{shin2020dvl}, has the advantage over feature-based methods like that of~\cite{zhang2015visual} in that it provides low drift, even when salient image and geometric features are lacking.
The disadvantage is that localization performance is worsened when whirling dust is present in the camera image, as reported in~\cite{petrlik2020robust}.

Team CERBERUS developed a complementary multi-modal sensor fusion~\cite{khattak2020complementary}.
The odometry estimated by visual/thermal inertial odometry is used as a prior for \ac{lidar} scan-to-scan and scan-to-map matching.
The \ac{vio}/\acs{tio} priors constrain the scan matching optimization problem, thus reducing drift in a degenerate environment significantly, which is demonstrated in an experiment conducted in a self-similar environment.

Another multi-modal approach is the Super Odometry~\cite{zhao2021super} of team Explorer, which was deployed on aerial robots in the tunnel and urban circuits of \ac{darpa} \ac{subt}.
The core of the method is the \ac{imu} odometry with biases constrained by \ac{vio} and \ac{lio}, which are initialized with preintegrated inertial measurements of the constrained \ac{imu}.
The relative pose factors of \ac{vio} and \ac{lio} are weighted based on the visual and geometrical degradation, respectively.

Team MARBLE first relied on visual \ac{slam}~\cite{kramer2021vi}, but after \ac{stix}, they transitioned to the \ac{lidar}-based Cartographer~\cite{hess2016real} due to unstable tracking of motion under poor illumination, reflections, dust, and other visual degradation.

Wildcat SLAM~\cite{hudson2021heterogeneous} of the \acl{csiro} team is a multi-agent decentralized solution, where each agent computes a global map using the currently available data shared among the robots.
The odometry of each agent is based on the work of~\cite{bosse2012zebedee}.

Our approach is similar to the other teams' as we also use primarily \ac{lidar} for localization and mapping.
An improvement over the state of the art is the compensation of the delay~\cite{pritzl2022repredictor} caused by the \ac{lidar} scan processing and the delay of the localization itself.

\subsection{Mobility}
\label{sec:sota_mobility}
Deploying aerial robots has one great advantage over ground robots due to their full terrain traversability. 
A \ac{uav} can fly over terrain that would compromise the safety of an \ac{ugv}, e.g., steep decline, mud, water, etc.
The only movement constraint of aerial platforms flying through an enclosed environment is the minimum size of a passage that the robot can safely pass through. 
The dimensions of such passages depend largely on the size of the \ac{uav}, but also on the precision of the pose estimation, the control error of onboard regulators, the map quality, and the reactive behavior in close vicinity of obstacles.
Some platforms also tolerate contact with obstacles in the sense that the contact does not endanger the continuation of the mission~\cite{huang2019duckiefloat}.
Other types of platforms adapt their morphology and/or locomotion modality to their current surroundings and obstacles~\cite{fabris2021soft}.
In voxel-based map representations, the size of a narrow passage is represented too conservatively, i.e., the size of the narrow passage in the voxel map is the lower bound of the true size.
However, in practice, the narrow passage can be up to twice the map resolution larger than its voxel representation, which prevents traversing passages that are well within the physical limits of the \ac{uav}.
To better approximate the true shape of the narrow passage, ~\cite{o2018variable} propose continuous representation based on \ac{gmm}, which is converted to a voxel map of arbitrary resolution when queried.
We took another approach of locally increasing the resolution of the occupancy voxel map when the size of the environment requires it.

\subsection{DARPA SubT approaches}
\label{sec:sota_darpa}
This paper primarily focuses on the approach developed for and experimentally verified in the final event of \ac{darpa} \ac{subt}.
As mentioned, these results are built upon the experience in using the approaches developed for the tunnel and urban circuits.
The practical verification of the developed solutions in challenging environments justifies the robustness of these algorithms. Valuable insights on the future of \ac{sar} robotics can be drawn from lessons learned by the teams.

Team CoSTAR relied on their uncertainty-aware framework, NeBula, in the tunnel and urban circuits~\cite{agha2021nebula}.
The framework supports multi-modal perception and localization including radar, sonar, and thermal cameras.
Aerial robots were part of their heterogeneous team in \ac{stix} and the tunnel circuit, mainly for exploring areas inaccessible to ground robots and data muling with distributed data sharing~\cite{ginting2021chord}.
A reactive autonomy approach COMPRA~\cite{lindqvist2021compra} was also proposed for \ac{uav} underground \ac{sar} missions.
Their solution gained \nth{2} and \nth{1} place in the tunnel and urban circuits respectively.

Team Explorer developed a system~\cite{scherer2022resilient} that achieved \nth{1} place in the tunnel circuit and \nth{2} place in the urban circuit.
Their collision-tolerant platform ``DS" with flight time of~\SI{13}{\minute} was carried on top of a \ac{ugv} and could be launched by the operator when needed.
The authors identified the challenge of combined exploration and coverage problem when their \acp{uav} with limited camera \ac{fov} missed some artifacts along their flight path.
The frontier-based exploration pipeline used a custom OpenVDB mapping structure~\cite{museth2013vdb} for sampling frontier-clearing viewpoints. 
Paths to found viewpoints were planned using bidirectional RRT-Connect.

Team CERBERUS deployed legged ANYMAL robots and aerial DJI Matrice M100 robots in the tunnel circuit.
Their graph-based system for the autonomous exploration of subterranean environments called GBPlanner was deployed in multiple locations.
The exploration of Edgar mine during \ac{stix} and the \ac{niosh} mine during the tunnel circuit are documented in~\cite{dang2020graph}.
Specifically, the exploration method for aerial robots~\cite{dang2019explore} consists of a local fast-response layer for planning short collision-free paths and a global layer that steers the exploration towards unvisited parts of the map.
This method is part of the solution for underground search by aerial robots found in~\cite{dang2020autonomous}.
A mapping and navigation approach~\cite{papachristos2019autonomous} for autonomous aerial robots based on the next-best-view planner~\cite{papachristos2017uncertainty, bircher2016receding} was also proposed, but was later outperformed by the GBPlanner~\cite{dang2020graph}.
The uncertainty in localization and mapping is taken into account during the planning in~\cite{papachristos2019localization} in such a way that among all trajectories arriving to the reference waypoint, the one that minimizes the expected localization and mapping uncertainty is selected.
To unify the exploration framework across both legged and aerial platforms,~\cite{kulkarni2021autonomous} have revised~\cite{dang2020graph} and added a cooperation framework that identifies global frontiers in a global graph built from the sub-maps of individual robots.
The unified strategy for subterranean exploration using legged and aerial robots in tunnel and urban circuits is presented in~\cite{tranzatto2022cerberus}

Team MARBLE presents their system deployed to \ac{stix}, the tunnel circuit, and the urban circuit in~\cite{ohradzansky2021multi}.
The aerial robots relied on direct vision-based local reactive control and map-based global path planning.
Global path planning is common with ground and aerial robots.
Viewpoints are selected based on the frontier voxels covered by the camera \ac{fov} and the approximate travel time.
In the tunnel circuit, the local reactive control generates velocity commands by steering the \ac{uav} towards a look-ahead point from the global path, while being repulsed by nearby obstacles.
With this planner, traversing narrow passages was problematic due to noise in the depth image. Thus, a new planner was developed for the urban circuit based on voxel-based probabilistic tracking of obstacles~\cite{ahmad2021end}.

A heterogeneous team of robots including \acp{uav} was also deployed by team \acl{csiro}~\cite{hudson2021heterogeneous}, both in the tunnel and urban circuits.
The aerial part of the team consisted of a DJI M210 equipped with the commercially available payload of Emesent Hovermap, and a custom gimballed camera.
To explore the environment of the urban circuit, the autonomy utilized an approach based on the direct point cloud visibility~\cite{williams2020online}. 

Although team NCTU did not participate in the final event, their solution~\cite{chenlung2022heterogeneous} to the tunnel and urban circuit showcased originality in the form of autonomous visually-localized blimps~\cite{huang2019duckiefloat}.
Their navigation was based on policies learned by deep reinforcement learning with simulation-to-world transfer.

Our CTU-CRAS-NORLAB team first participated in the \ac{stix} event with a hexarotor platform localized by optic flow~\cite{walter2017mesas} of the downward-facing camera.
The reactive navigation used \ac{lidar} scans to stay in the middle of the tunnel and move forward in a preferred direction at an intersection. 
The predictive controller~\cite{baca2016embedded} was forgiving to imprecise localization caused by strenuous optic flow estimation in the whirling dust of the tunnels.
The heterogeneous team that secured \nth{3} place in the tunnel circuit~\cite{roucek2019darpa} consisted of wheeled, tracked, and aerial robots with different sensor payloads.
Instead of unreliable optic flow, the localization of the \ac{uav} system~\cite{petrlik2020robust} was revamped to rely on 2D \ac{lidar}, HectorSLAM~\cite{kohlbrecher2011flexible}, and state estimation~\cite{petrlik2021lidar}.
The hardware platform was also downscaled to a \SI{450}{\milli\meter} diameter quadrotor.
The vertical element of the urban circuit called for upgrading the \ac{lidar} to a 3D one, which consequently required a redesign of the whole navigation pipeline~\cite{kratky2021exploration} to allow for six \ac{dof} mobility through the 3D environment.
Physically, the platform was based on the same frame as what was used in the tunnel circuit, however prop guards were added to reduce the chance of destructive collision while flying through doors.
The CTU-CRAS-NORLAB approach to the urban circuit, which we completed in \nth{3} place, is described in~\cite{roucek2020urban}.
Although the cave circuit was canceled, extensive preparations were still performed in the sizable Bull Rock cave in South Moravia~\cite{petracek2021caves}. 
The exploration depth of the \ac{uav} team was greatly extended by a multi-robot coordinated homing strategy that focused on extending the communication range of the base station by landing the returning \acp{uav} on the edge of the signal.
Based on the lessons learned during these competition and testing deployments (during the 3 years of development \acp{uav} of the CTU-CRAS-NORLAB team achieved $>400$ flights and traveled $>$\SI{50}{\kilo\meter} in demanding real world environments) the new approaches presented in this paper were designed.

\subsection{Datasets}
\label{sec:sota_datasets}
Due to the challenging nature of the subterranean environments, such as narrow passages, degenerate geometry, and perception degradation, datasets that were collected by the competing teams are valuable to the community as the algorithms can be evaluated on demanding data degraded by the previously mentioned issues.
In contrast to the verification often conducted under artificially ideal lab conditions, these datasets present a fair way to compare algorithms in realistic conditions.
A \ac{slam} dataset~\cite{rogers2020test} collected during the tunnel circuit and \ac{stix} consists of \ac{lidar} scans, images from a stereo camera and thermal camera, \ac{imu} measurements, and \ac{rssi}, together with a professionally surveyed ground truth map and measured artifact positions.
The dataset from the urban circuit~\cite{rogers2020darpa} was recorded using the same sensors with the exception of an added \ac{co2} sensor and the lack of a thermal camera.
Another dataset~\cite{kasper2019benchmark} for comparison of \ac{vio} methods contains outdoor, indoor, tunnel, and mine sequences, with ground truth poses obtained by laser tracking the sensors rig.
Aerial datasets consisting of unsynchronized \ac{lidar} scans and \ac{imu} measurements from \acp{uav} flying in the cave, tunnel, and mine environments are included in this paper\footnote{\href{https://github.com/ctu-mrs/slam_datasets}{\texttt{github.com/ctu-mrs/slam\_datasets}}}, with ground truth poses estimated using a professionally surveyed ground truth map.
We also publish the labeled visual detection datasets\footnote{\href{https://github.com/ctu-mrs/vision_datasets}{\texttt{github.com/ctu-mrs/vision\_datasets}}} consisting of images from both \ac{uav} and \ac{ugv} cameras that were used for training of the artifact detection \ac{cnn}.
Images from the Tunnel and Urban circuits, Bull Rock Cave, and industrial buildings are included.


    \section{Contributions}
    \label{sec:contributions}

    An approach for cooperative exploration of demanding subterranean environments by a team of fully autonomous \acp{uav} in \ac{sar} tasks is presented in this paper.
    Deployment of this approach in the \ac{darpa} \ac{subt} virtual competition was awarded by \nth{2} place. The simulation model of the \ac{uav} platform designed by our team was used by seven out of nine teams.
    The crucial contributions of the developed system can be summarized in the following list:
    \begin{itemize}
      \item 
        A complex approach that can serve as a guide for building a system for \ac{gps}-denied operations.
        The proposed approach was extensively verified in numerous simulated worlds and real physical environments ranging from vast caves, industrial buildings, tunnels, and mines to large outdoor openings.
            Most importantly, the \acp{uav} were deployed into the intentionally harsh conditions of the \ac{darpa} \ac{subt} to push them to their limits.
            The experience gained from hundreds of flights in such conditions are condensed into the lessons learned presented in this paper, which we deem valuable for the field robotics community.
      \item Novel mapping structures are proposed for safety-aware reactive planning over large distances, for compact volumetric inter-robot information sharing, for storing coverage of surfaces by onboard sensors, and for finding a suitable landing spot.
      \item Maximization of the probability of detecting a nearby artifact by searching not only the unexplored space, but also visually covering known surfaces while respecting the limited field of view of the onboard sensors.
      The detection is coupled with probabilistic estimation of artifact positions based on multi-target tracking and detection-to-hypothesis association, which improves the precision of artifact localization while the robot is moving around the artifact.  
      \item A novel safety-aware approach to planning that considers the risk of planned trajectories in addition to the path length in the optimized cost function. 
        In contrast to the state-of-the-art methods, longer paths are selected if the estimated risk of collision is lower than the risk of a shorter path.
      \item Full autonomy of the \ac{uav} allows for scalability of the size of the deployed fleet without placing additional workload on the operator. 
        Nevertheless, the operator can override the autonomy with one of the available special commands to change the default search behavior when the \ac{uav} is in communication range.
      \item The multi-robot autonomous search benefits from a higher number of deployed \acp{uav} that share their topological representations of the environment to cooperatively cover a larger area by biasing the search towards parts unvisited by other agents.
    \end{itemize}



    \section{System architecture overview}
    \label{sec:architecture}

\begin{figure}
      \includegraphics[width=1.0\textwidth]{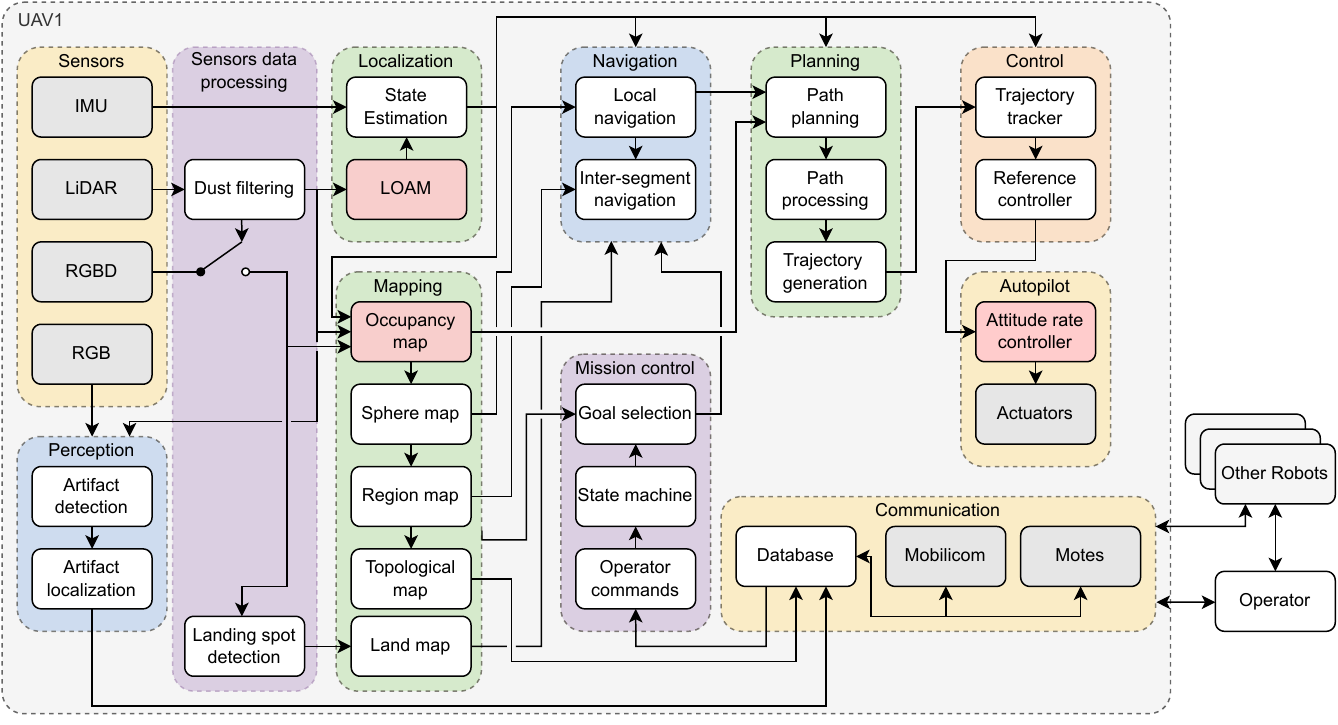}
      \caption{\label{fig:system} 
      The diagram shows individual modules of the \ac{uav} system architecture grouped into logical categories.
      Hardware modules are filled with gray, and red distinguishes open source modules not developed by us.}
    \end{figure}

    The whole autonomous system of a single \ac{uav} consists of software modules, each with different inputs, outputs, and purpose. 
    These modules and their interconnections are depicted in~\autoref{fig:system} with the individual modules grouped into more general logical categories.
    The first category includes the physical \textit{Sensors}~(\autoref{sec:sensory_perception}) of the \ac{uav} --- the \ac{imu}, \ac{lidar}, \acs{rgb}, and \acs{rgbd} cameras.
    The description of the important parameters of the used sensors is available in~\autoref{sec:hardware}.
    Measurements from \ac{imu} and \ac{lidar} enter the \textit{Localization} group~(\autoref{sec:localization}), where a full-state estimate of the \ac{uav} is obtained.
    \ac{lidar} is also used in combination with the \acs{rgbd} camera for building maps in the \textit{Mapping} module group~(\autoref{sec:mapping}).
    The \textit{Perception}~(\autoref{sec:object_detection}) category focuses on detection and localization of artifacts using all the available sensor data.

    Autonomous search through the environment is governed by the \textit{Mission control} category~(\autoref{sec:mission_control}), which selects goals~(\autoref{sec:exploration}) based on the current state of the state machine, models of the environment from the \textit{Mapping} group, and possibly also commands from the operator.
    A coarse path consisting of waypoints to the selected goals is found by the \textit{Navigation}~(\autoref{sec:navigation}) and further refined and time-parametrized in the \textit{Planning} modules~(\autoref{sec:planning}) in order to produce a safe and dynamically feasible trajectory.
    The \textit{Control} blocks~\cite{baca2021mrs} track the trajectory and generate attitude rate references for the low-level \textit{Autopilot} that controls the actuators~(\autoref{sec:hardware}).

    The operator receives crucial mission status data, topological maps, and, most importantly, detected artifacts through the \textit{Communication} layer \cite{roucek2020urban}. This also allows the operator to influence or override the autonomous behavior of the \ac{uav}.
    All transmitted data is received by other \acp{uav} (or other robots, in the case of a heterogeneous team) in the communication range, which serves two purposes:
    one, the receiving agent can propagate the message further down the network, and, two, the topological maps allow penalizing goals already visited by other robots to better allocate resources over a large area.

    \section{Spatial perception}
\label{sec:sensory_perception}

The equipment on board \ac{uav} platforms within our research group is modular and replaceable to support a wide spectrum of research areas~\cite{hert2022hardware}.
In the proposed system for agile subterranean navigation, however, the aerial platform is fixed to ease fine-tuning of the on-board-running algorithms.
From the point of perception, it relies heavily on 3D \ac{lidar} from Ouster (SLAM, dense mapping, and artifact localization), and utilizes vertically-oriented \ac{rgbd} cameras for filling space out of \ac{fov} of the primary \ac{lidar} sensor, and uses two \ac{rgb} Basler cameras for artifact detection, supported by powerful LEDs illuminating the scene.
The flow of sensory data within the entire system is shown directly in~\autoref{fig:system}.

  \subsection{Sensors calibration}
  \label{sec:sensors_calibration}

  The intrinsics of \ac{lidar} sensor and \ac{rgbd} cameras are factory-calibrated whilst monocular \ac{rgb} cameras are calibrated with standard OpenCV calibration tools, assuming the pinhole camera model.
  To find the extrinsics of the sensors, all the cameras are one-by-one calibrated with respect to the \ac{lidar} sensor whereas the extrinsics of the \ac{lidar} with respect to the flight control unit are given by the CAD model of the robot.
  The camera-to-lidar extrinsics are calibrated using a checkerboard camera calibration pattern with known dimensions.
  The calibration pipeline detects the pattern in both modalities (\ac{lidar} data and \ac{rgb} image), finds mutual correspondences, and estimates the extrinsics directly by defining the problem as perspective-n-point optimization which minimizes the reprojection error of the mutual correspondences.
  The calibration process for a single camera is described in~\autoref{alg:calibrate_camera_to_lidar}.

  \begin{algorithm}
    \algdef{SE}[SUBALG]{Indent}{EndIndent}{}{\algorithmicend\ }%
    \algtext*{Indent}
    \algtext*{EndIndent}

    \algnewcommand\AND{\textbf{and}~}
    \algnewcommand\Not{\textbf{not}~}
    \algnewcommand\Or{\textbf{or}~}
    \algnewcommand\Continue{\textbf{continue}~}
    \algnewcommand\Input{\State{\textbf{Input:~}}}%
    \algnewcommand\Output{\State{\textbf{Output:~}}}%
    \algnewcommand\Parameters{\State{\textbf{Parameters:~}}}%
    \algnewcommand\Begin{\State\textbf{Begin:~}}%
      \algnewcommand{\LineComment}[1]{\State \(\triangleright\) #1}

      \caption{Calibrating extrinsic parameters of \ac{rgb} cameras with respect to \ac{lidar} sensor.
      }\label{alg:calibrate_camera_to_lidar}

      \begin{algorithmic}[1]
        \footnotesize

        \Input
        \Indent

        \State $\set{P} = \left\{\mathcal{L}_i,\mathcal{I}_i\right\}$
        \Comment{set of synchronized pairs of \ac{lidar} and camera frames}

        \State $\mathbf{K} \in \mathbb{R}^{3\times3}$
        \Comment{camera matrix of intrinsic parameters}

        \State $\mathbf{D} \in \mathbb{R}^{5}$
        \Comment{camera distortion coefficients}

        \EndIndent

        \Output
        \Indent

        \State $\mathbf{R},~\mathbf{t}$
        \Comment{extrinsics (rotation and translation) of the camera in the \ac{lidar} frame}

        \EndIndent

        \Parameters
        \Indent

        \State $\mathcal{O}$
        \Comment{checkerboard camera calibration pattern (dimensions, square count, square size)}

        \EndIndent

        \Begin
        \Indent

        \State $\set{S} \coloneqq \emptyset$
        \Comment{initialize set of camera-\ac{lidar} correspondences}

        \For{$ \left\{\mathcal{L}_i,\mathcal{I}_i\right\} \in \set{P} $}

        \State $ \set{C}_{L} \coloneqq \algfunc{findCalibPatternCornersInLidar}\left(\mathcal{L}_i, \mathcal{O}\right) $
        \Comment{determine four square-corners of calibration pattern in \ac{lidar} data}
        \State $ \set{C}_{I} \coloneqq \algfunc{findCalibPatternCornersInImage}\left(\mathcal{I}_i, \mathcal{O}, \mathbf{K}, \mathbf{D}\right) $
        \Comment{determine four square-corners of calibration pattern in \ac{rgb} image}

        \If{$\Not~|\set{C}_{L}| = 4~\Or~\Not~|\set{C}_{I}| = 4$}
        \State \Continue
        \Comment{skip pair}
        \EndIf

        \State $\set{S} \coloneqq \set{S} \cup \left\{ \set{C}_{L},~\set{C}_{I} \right\}$

        \EndFor

        \State $\hat{\mathbf{R}},~\hat{\mathbf{t}} \coloneqq \algfunc{solvePnP}\left(\set{S}\right)$
        \Comment{compute rough estimate of extrinsics~\cite{lepetit2008EPnPAA}}

        \State $\mathbf{R},~\mathbf{t} \coloneqq \algfunc{refinePnPWithLM}\left(\set{S}, \hat{\mathbf{R}}, \hat{\mathbf{t}}\right)$
        \Comment{numerically optimize estimate of extrinsics with Levenberg-Marquardt method}

        \EndIndent

      \end{algorithmic}

    \end{algorithm}

\subsection{Filtering observation noise}
\label{sec:filtering_observation_noise}

The aerodynamic influence of a multi-rotor \ac{uav} on the environment is not negligible, particularly in confined settings.
The fast-rotating propellers generate airflow lifting up light particles of dust and whirling them up in clouds.
In environments where the clouds are not blown away but are rather rebounded back to the \ac{uav}, the effect on sensory performance might be crippling.
To minimize deterioration in perception and its dependent systems (e.g., mapping, localization), the incident noise is filtered out from local \ac{lidar} data.

\begin{figure}
  \centering

  \begin{subfigure}[t]{0.275\textwidth}
    \centering
    \includegraphics[width=\textwidth]{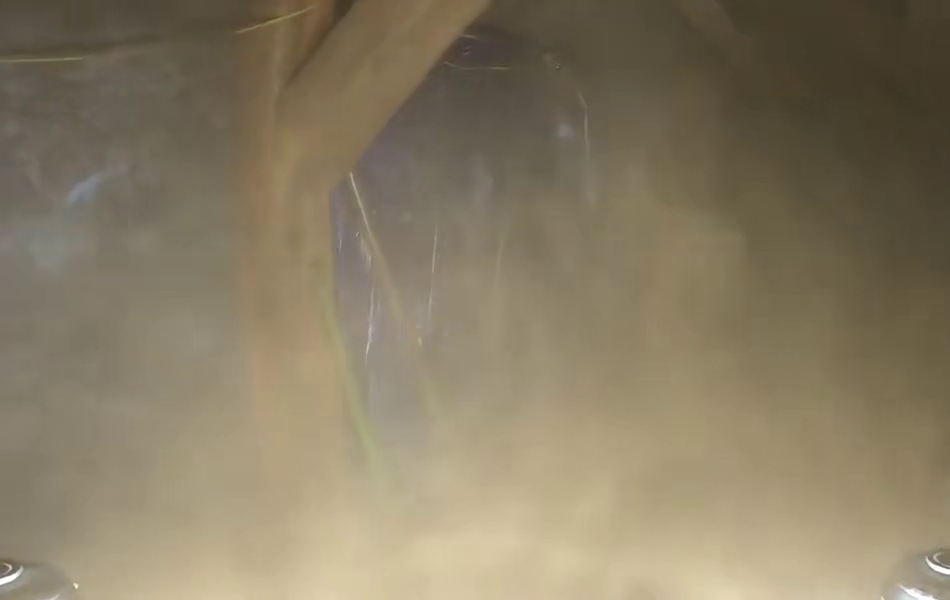}
    \caption{Dense cloud dust around the \ac{uav} as viewed in onboard \ac{rgb} camera at time \SI{330}{\second}.}
      \label{fig:dust_filtering_rgb}
  \end{subfigure}%
  \begin{subfigure}[t]{0.325\textwidth}
    \centering
    \includegraphics[trim=0 50 0 85,clip,width=\textwidth]{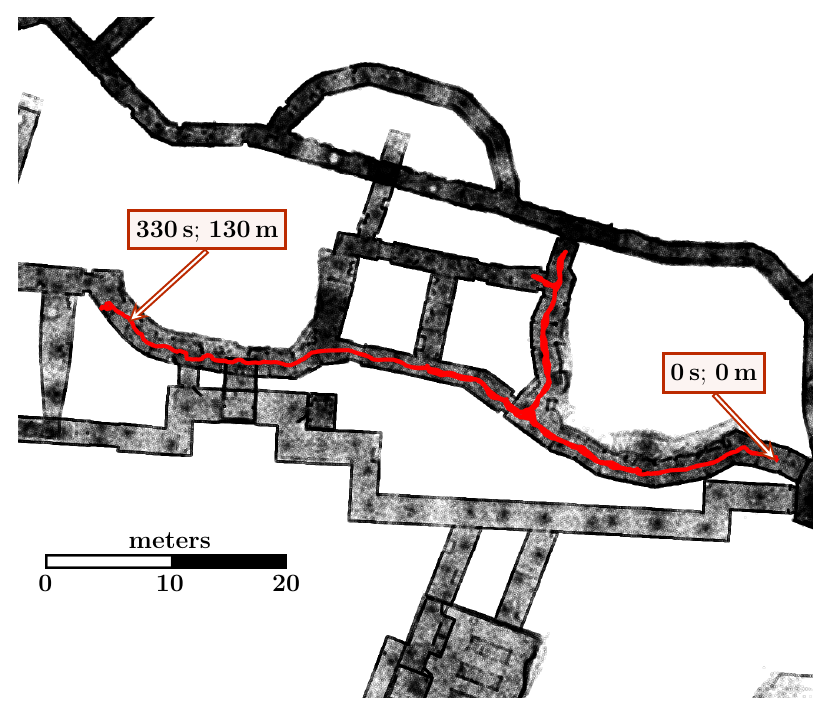}
    \caption{%
      Top view on the \ac{uav} trajectory.}
      \label{fig:dust_filtering_traj}
  \end{subfigure}%
  \begin{subfigure}[t]{0.4\textwidth}
    \centering
    \includegraphics[width=\textwidth]{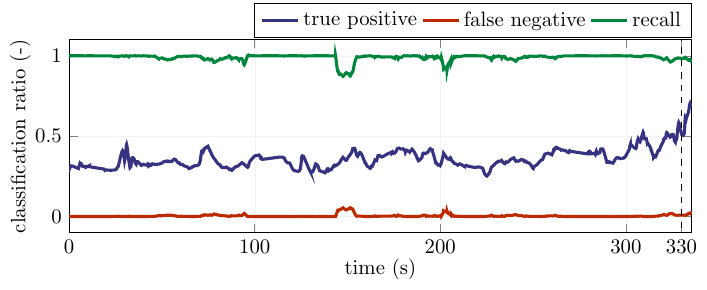}
    \caption{%
      Performance of noise classification in \ac{lidar} data in \SI{3}{\metre} local radius from the sensor.
      Average recall reached \SI{99}{\percent}.}
      \label{fig:dust_filtering_classification}
  \end{subfigure}%

  \caption{%
    \ac{lidar}-data noise filtration running onboard a \ac{uav} during a \SI{154}{\metre} flight in the mine part (the dustiest part) of the \ac{darpa} \ac{subt} finals environment.
    The true positive classification in (c) denotes the ratio of correctly classified noise whereas the false negative represents the ratio of noise preserved after the filtration process (i.e., the unfiltered noise) to the size of the point cloud.
    The data for the classification analysis (c) were obtained by spatially comparing the sensor measurements with the map of the environment provided by the organizers.
    }
    \label{fig:dust_filtering}
\end{figure}

The idea of robust filtering of dust is based on the method presented in~\cite{kratky2021exploration} in which \ac{lidar} data are sorted by the intensity field (measured intensity of the reflected light for a given point) and \SI{10}{\percent} of the lowest-intensity data in a local radius from the sensor are removed.
In contrast to the baseline method, simpler thresholding is adopted such that a subset $\mathcal{P}_F \subseteq \mathcal{P}$ of \ac{lidar} data $\mathcal{P}$ is preserved.
The absence of data sorting lowers the computational load and reduces delay in data processing.
The set is given as $\mathcal{P}_F = \mathcal{P}_D \cup \mathcal{P}_I$, where
\begin{align}
  \mathcal{P}_{D} &= \left\{\pnt{p}\,|\, \norm{\pnt{p}} \geq \kappa,\,\pnt{p} \in \mathcal{P}\right\},\\
  \mathcal{P}_{I} &= \left\{\pnt{p}\,|\,\mathcal{I}(\pnt{p}) > \Upsilon,\,\pnt{p} \in \mathcal{P} \setminus \mathcal{P}_{D}\right\}.
\end{align}
$\mathcal{I}(\pnt{p})$ $\left(\si{\watt\per\meter\squared}\right)$ is the intensity of the reflected light from a point $\pnt{p}$, $\kappa$ $\left(\si{\meter}\right)$ is a local radius of a filtering sphere with \ac{lidar} data origin at its center, and $\Upsilon$ $\left(\si{\watt\per\meter\squared}\right)$ is the minimal intensity of preserved data points.
With $n$ data points within a radius $\kappa$, the computational complexity is reduced to $\mathcal{O}\left(n\right)$ from baseline $\mathcal{O}\left(n\,\log(n)\right)$.
Although to achieve optimal performance the method requires calibration to given environmental conditions, a set of reasonable parameters ($\kappa = \SI{5}{\metre}$ and $\Upsilon = \SI{30}{\watt\per\meter\squared}$ throughout many of our real-world deployments in the harshest dust conditions) suffices in the majority of applications.
The performance of the dust filtering is analyzed in \autoref{fig:dust_filtering} on an example \ac{uav} flight in the mine part (the dustiest zone) of the \ac{darpa} \ac{subt} finals environment.

\subsubsection{Detecting artificial fog in the virtual environment}
\label{sec:fog_detection}

\begin{figure}
  \centering

  \begin{subfigure}[t]{0.33\textwidth}
    \centering
    \captionsetup{width=1.0\textwidth}
    \adjincludegraphics[width=\textwidth,trim={0.1\width} {0\height} {0.1\width} {0\height},clip,]{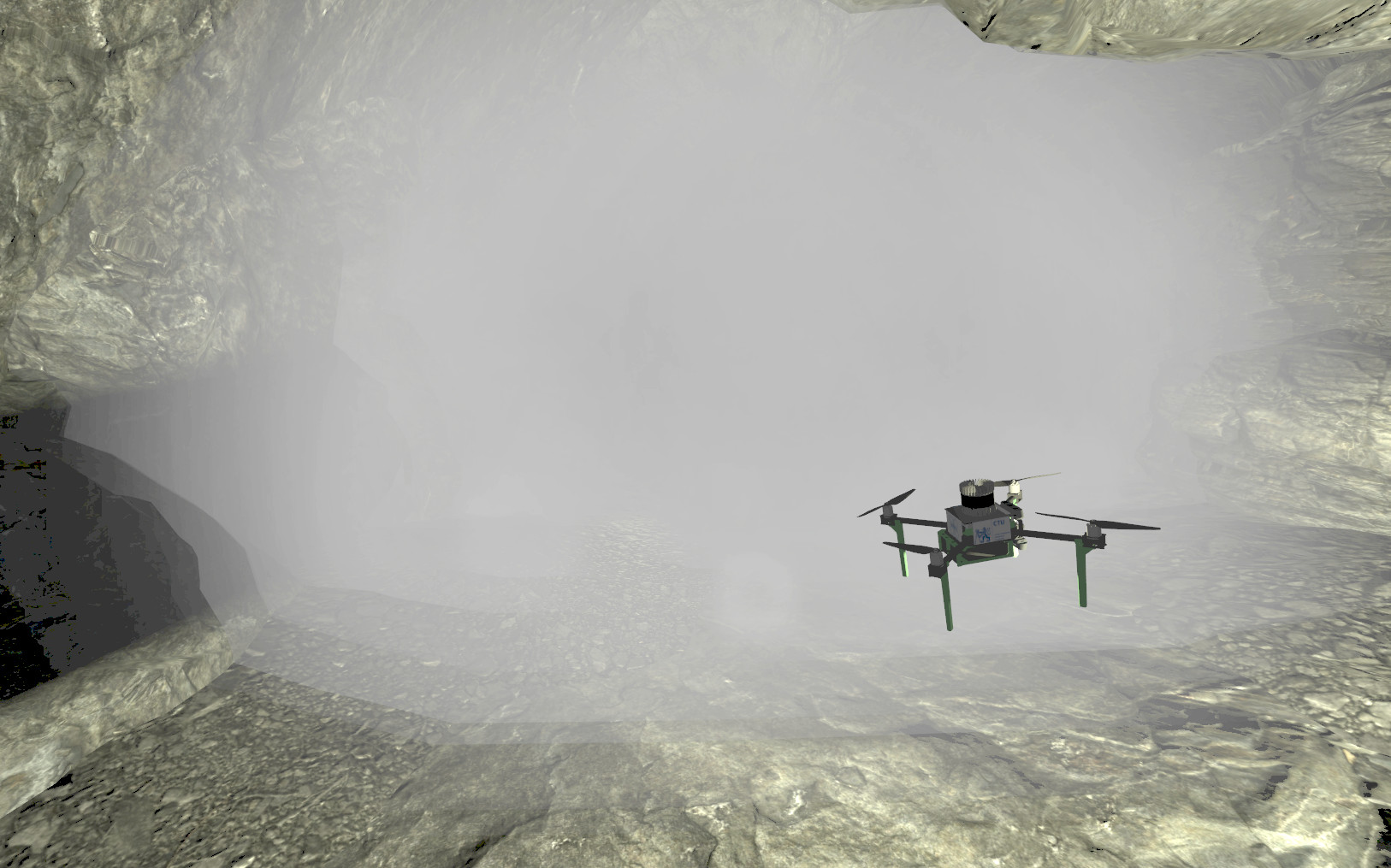}
    \caption{Visualization of virtual fog in Ignition Gazebo.}
  \end{subfigure}%
  \begin{subfigure}[t]{0.33\textwidth}
    \centering
    \captionsetup{width=1.0\textwidth}
    \includegraphics[width=\textwidth]{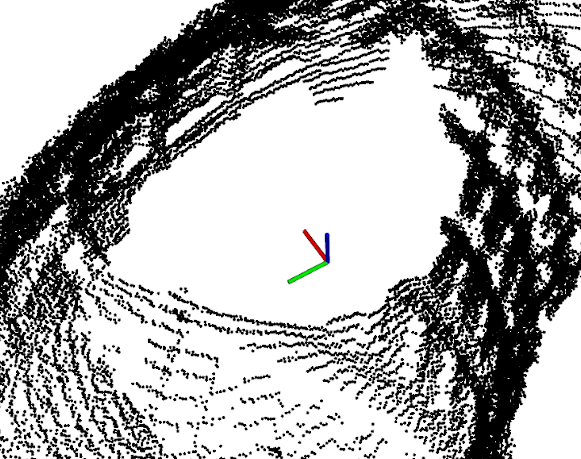}
    \caption{Example 3D \ac{lidar} data outside fog.}
  \end{subfigure}%
  \begin{subfigure}[t]{0.33\textwidth}
    \centering
    \captionsetup{width=1.0\textwidth}
    \includegraphics[width=\textwidth]{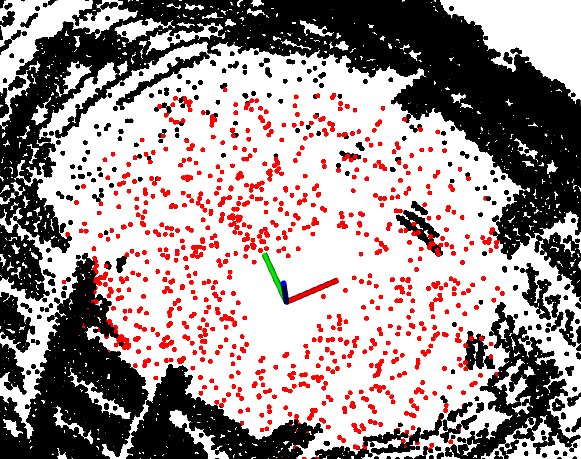}
    \caption{Example 3D \ac{lidar} data inside fog (fog colored locally in red).}
    \label{fig:fog_detection_infog}
  \end{subfigure}%
  \caption{%
    Simulated fog and its effect on sensory perception in the virtual environment.
    A fog cloud (a) spawns when a robot reaches its proximity.
    The cloud then affects the sensory inputs such that a uniform-distribution noise emerges in \ac{lidar} data corresponding to the fog (c).}
    \label{fig:fog_detection}
\end{figure}

The virtual competition contained a fog emitter plugin (see \autoref{fig:fog_detection}) to mimic environmental perception degradation arising from observing smoke, dust, and fog.
The plugin spawned a fog cloud when a robot reached the proximity of the emitter.
Although our localization pipeline was able to cope with local noise, the inability to filter out the fog particles in a robust way led to a degradation of the local DenseMap, and consequently to blocking local planning which respects strict requirements on collision-free path planning.
Thus in our setup for the virtual challenge, the navigation stack did not try to enter through the fog areas but detected them, maneuvered out of them, and blocked the areas for global planning.

To detect the presence of the \ac{uav} within such a fog cloud, a discretized occupancy voxel grid is built from a set of data within a local radius (example data within a radius are shown in \autoref{fig:fog_detection_infog}).
Within this radius is compared the occupancy ratio $r$ (number of occupied voxels to all voxels in the local grid) with maximum occupancy $R$ given by the field of view of the sensor producing the data.
For each \ac{lidar} or depth sensor, the sensor is classified as being in fog if
\begin{equation}
  r > \lambda R, 
\end{equation}
where $\lambda \in \langle0, 1\rangle$ is a unitless multiplier converting $\lambda R$ to a maximal occupancy ratio threshold.
The multiplier was set empirically to $\lambda = 0.7$ in our final setup.

For depth cameras that are not used for self-localization of the \ac{uav}, the in-fog classification solely controls whether the depth data are integrated within the mapping pipeline.
However, if a localization-crucial 3D \ac{lidar} is classified to be in fog, a backtracking behavior is triggered within the mission supervisor (see~\autoref{sec:mission_control}).
The primary purpose of the backtracking is to prevent being stuck in fog and thus the \ac{uav} is blindly navigated out of the fog through the recent history of collision-free poses, ignoring occupied cells in the DenseMap (including possible noise from fog measurements).
Lastly, detection of fog in a 3D \ac{lidar} blocks the area in global planning.

\subsection{Detecting spots safe for landing}
\label{sec:landing_spot_detection}

\begin{figure}
  \centering

  \begin{subfigure}[t]{0.31\textwidth}
    \centering
    \captionsetup{width=0.99\textwidth}
    \includegraphics[width=\textwidth]{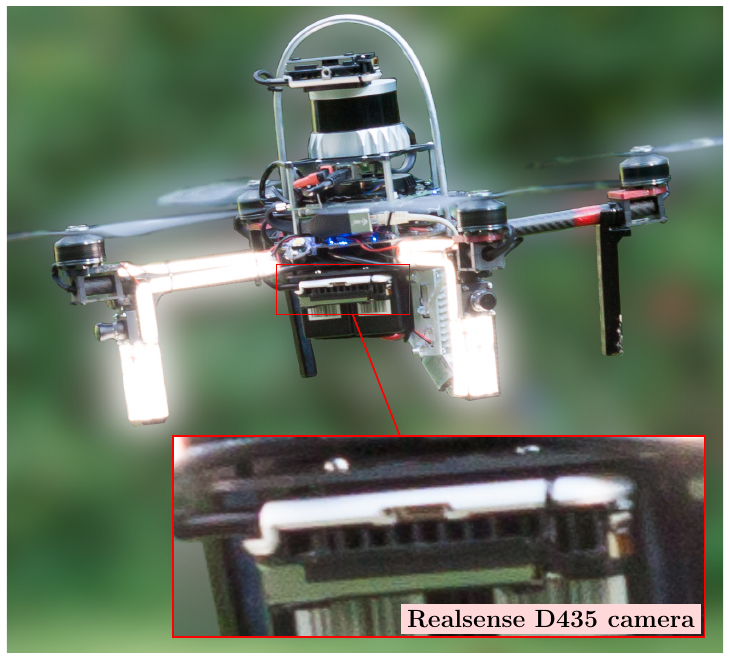}
    \caption{Downward-facing \ac{rgbd} camera used for landability detection mounted on our \ac{uav} platform.}
  \end{subfigure}%
  ~
  \begin{subfigure}[t]{0.22\textwidth}
    \centering
    \captionsetup{width=0.95\textwidth}
    \includegraphics[width=\textwidth]{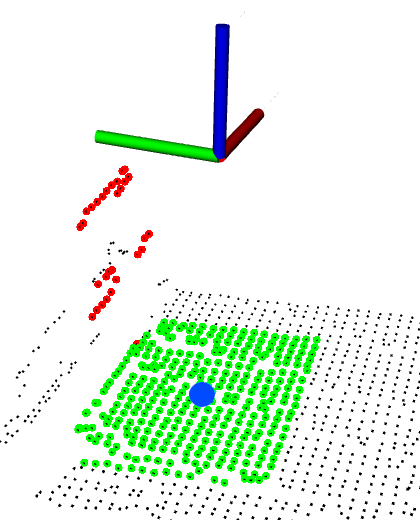}
    \caption{Even planar surface: safe for landing.}
  \end{subfigure}%
  ~
  \begin{subfigure}[t]{0.22\textwidth}
    \centering
    \captionsetup{width=0.95\textwidth}
    \includegraphics[width=\textwidth]{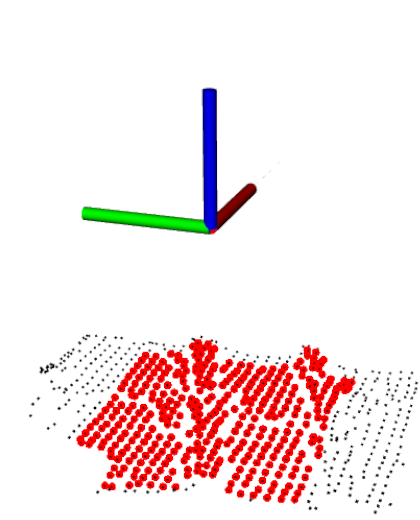}
    \caption{Non-planar surface (rails): unsafe for landing.}
  \end{subfigure}%
  ~
  \begin{subfigure}[t]{0.22\textwidth}
    \centering
    \captionsetup{width=0.95\textwidth}
    \includegraphics[width=\textwidth]{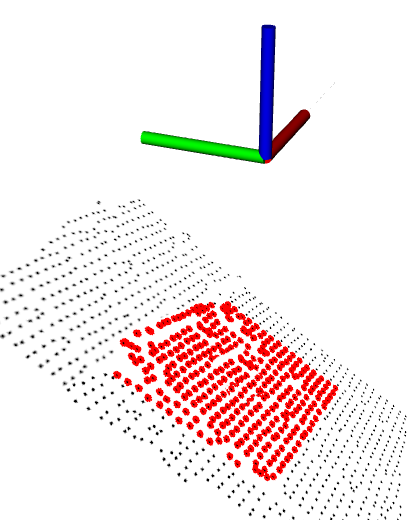}
    \caption{Uneven surface: unsafe for landing.}
  \end{subfigure}%

  \caption{%
    Deciding on landability of a \ac{uav} from downward-facing depth data --- binary classification to safe (b) and unsafe (c-d) landing areas.
    In (b-d), the \ac{uav} is represented by Cartesian axes whereas the depth data are colored in black.
    The blue sphere in the safe classification (b) denotes the centroid of the plane-inliers (colored in green) passed as a feasible landing position to LandMap (see~\autoref{sec:landmap}).}
    \label{fig:landing_spot_detection}
\end{figure}

Depth data of downward-facing \ac{rgbd} camera are used in locating areas safe for landing throughout the \ac{uav} flight.
The depth data are fitted with a plane model whose coefficients are used in the binary classification of safe or unsafe landability respecting the plane-fit quality and deviation of its normal vector from the gravitational vector.
The process of deciding on safe landability given a single depth-data frame is visualized in~\autoref{fig:landing_spot_detection} and described in~\autoref{alg:landing_spot_detection}.
The classification assumes that the data frame can be transformed into a gravity-aligned world coordinate frame.

\begin{algorithm}
  \algdef{SE}[SUBALG]{Indent}{EndIndent}{}{\algorithmicend\ }%
  \algtext*{Indent}
  \algtext*{EndIndent}

  \algnewcommand\AND{\textbf{and}~}
  \algnewcommand\Not{\textbf{not}~}
  \algnewcommand\Or{\textbf{or}~}
  \algnewcommand\Input{\State{\textbf{Input:~}}}%
  \algnewcommand\Output{\State{\textbf{Output:~}}}%
  \algnewcommand\Parameters{\State{\textbf{Parameters:~}}}%
  \algnewcommand\Begin{\State\textbf{Begin:~}}%
    \algnewcommand\RETURN{\State\textbf{return:~}}%
    \algnewcommand{\LineComment}[1]{\State \(\triangleright\) #1}

    \caption{Detecting spots safe for \ac{uav} landing  in downward-facing \ac{rgbd} camera}
    \label{alg:landing_spot_detection}

    \begin{algorithmic}[1]
      \footnotesize

      \Input
      \Indent

      \State $\set{D}$
      \Comment{Depth-data frame in sensor coordinate frame}

      \EndIndent

      \Output
      \Indent

      \State $\set{L}$
      \Comment{Binary classification for landing: $\{\text{SAFE}, \text{UNSAFE}\}$}

      \State $\mathbf{p}_W$
      \Comment{Position of landing area in the world coordinate frame}

      \EndIndent

      \Parameters
      \Indent

      \State $s$
      \Comment{Square-size of safe landing spot in meters}

      \State $I_{min}$
      \Comment{Minimal ratio of inliers in plane fitting}

      \State $N_{min}^z$
      \Comment{Minimal z-axis component of the normalized plane-normal vector}

      \EndIndent

      \Begin
      \Indent

      \State{$\set{S} \coloneqq \algfunc{cropFrameAtCenter}\left(\set{D}, s\right)$}
      \Comment{Crop frame-centered square with size $s$}

      \If{$\algfunc{height}\left(\set{S}\right) < s~\Or~\algfunc{width}\left(\set{S}\right) < s$}

      \RETURN{$\{\set{L} = \text{UNSAFE},~\mathbf{p}_W = \text{N/A}\}$}
      \Comment{Not safe to land: too close to the ground to decide}

      \EndIf

      \State{$\set{P} \coloneqq \algfunc{fitPlaneWithRANSAC}\left(\set{S}\right)$}
      \Comment{Fit data with plane using RANSAC}

      \If{$\algfunc{inliers}\left(\set{P}\right)~/~\algfunc{count}\left(\set{S}\right) < I_{min}$}

      \RETURN{$\{\set{L} = \text{UNSAFE},~\mathbf{p}_W = \text{N/A}\}$}
      \Comment{Not safe to land: data are not planar}

      \EndIf

      \State $\set{P}_W \coloneqq \algfunc{transformToWorldFrame}\left(\set{P}\right)$
      \Comment{Transform plane to gravity-aligned frame}

      \If{$|\algfunc{normal}\left(\set{P}_W\right).z| < N_{min}^z$}

      \RETURN{$\{\set{L} = \text{UNSAFE},~\mathbf{p}_W = \text{N/A}\}$}
      \Comment{Not safe to land: ground is too steep for landing}

      \EndIf

      \State{$\set{S}_W \coloneqq \algfunc{transformToWorldFrame}\left(\set{S}\right)$}
      \State $\mathbf{p}_W \coloneqq \algfunc{centroid}\left(\set{S}_W\right)$
      \Comment{Express landing spot as the centroid of the depth data in the world}

      \RETURN{$\{\set{L} = \text{SAFE},~\mathbf{p}_W\}$}

      \EndIndent

    \end{algorithmic}

  \end{algorithm}


    \section{Localization}
    \label{sec:localization}


    Accurate and reliable localization is critical for most other parts of the system.
    The ability of the reference controller to track the desired state depends largely on the quality of the available state estimate. 
    In the narrow environments which are often present in subterranean environments (see~\autoref{tab:env_size} for cross-section distribution in the final event), minimizing the control error is crucial to avoid collisions.
    Multi-robot cooperation assumes the consistency of maps created by individual robots.
    If the maps of two robots are not consistent due to errors in localization, the multi-robot search might be suboptimal. 
    For example, an unvisited goal can be rejected as already reached by a robot with an inconsistent map.
    Moreover, the localization accuracy influences the position error of a reported artifact.
    A \ac{uav} with localization drift over \SI{5}{\meter} can detect and perfectly estimate the position of an artifact. Nevertheless, the report may never score a point since the position of the \ac{uav} itself is incorrect.

    Our approach relies on a \ac{lidar} sensor for localization as the laser technology proved to be more robust to the harsh conditions of the subterranean environment than the vision-based methods.
    We have been using \ac{lidar} since the Tunnel circuit~\cite{petrlik2020robust} where a lightweight 2D \ac{lidar} aided by a rangefinder for measuring \ac{agl} height was sufficient for navigation in tunnels with a rectangular cross-section.
    The more vertical environment of the urban circuit required redesigning the localization system to use 3D \ac{lidar} for navigating in 3D space~\cite{kratky2021exploration}.

\begin{figure}
      \centering
      \includegraphics[width=1.0\textwidth]{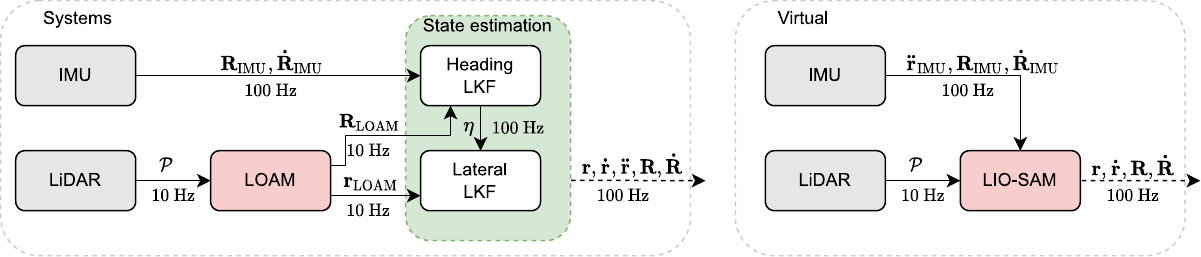}
      \caption{\label{fig:diagram_localization} 
      The diagram shows the flow of data among individual localization modules for the Systems Track (left) and Virtual Track (right). 
      The 3D \ac{lidar} supplies \acs{aloam} or \acs{liosam} with the laser scans in the point cloud form $\mathcal{P}$.
      Assisted by the orientation $\mathbf{R}$ from the \ac{imu}, \acs{aloam} produces a position estimate $\mathbf{r}=\left[x, y, z\right]^T$ that is fed into the \textit{State estimation} block, which outputs the full state estimate.
      In the case of the virtual pipeline, the \ac{imu} data fusion is executed in \ac{liosam}, and thus the state estimation module is not needed thanks to the sufficient accuracy of both lateral and heading components.
      }
    \end{figure}

    The localization system deployed in the final event and presented in this manuscript builds upon the solution proposed in~\cite{kratky2021exploration} and is divided into two modules: the localization algorithm and the state estimation method.
    \autoref{fig:diagram_localization} shows the data flow in the localization pipeline.
    We have based the localization on the \acs{aloam} implementation of the \ac{loam} algorithm~\cite{zhang2014loam} for the Systems Track and the \ac{liosam}~\cite{liosam2020shan} for the Virtual Track.
    Our implementation\footnote{\href{https://github.com/ctu-mrs/aloam}{\texttt{github.com/ctu-mrs/aloam}}} has been tested in a real-time \ac{uav} control pipeline throughout multiple experimental deployments as part of our preliminary works~\cite{kratky2021exploration,petracek2021caves} and in the \ac{darpa} \ac{subt} competition.

    \subsection{A-LOAM}
    \label{sec:aloam}

    The \acs{aloam} implementation of the \ac{loam}~\cite{zhang2014loam} algorithm utilizes the laser scans from a multi-line \ac{lidar} to obtain its 6-\ac{dof} pose.
    To achieve real-time performance and accurate pose estimation at the same time, the method is divided into two parts.

    The first part of the algorithm processes the incoming data at the rate of their arrival and estimates the rigid motion between the consecutive point clouds $\mathcal{P}_k$ and $\mathcal{P}_{k+1}$ obtained at the timestamps $t_k$ and $t_{k+1}$, respectively.
    The process starts with finding geometric features in the input point cloud $\mathcal{P}_{k+1}$.
    The points are first sorted by the smoothness of their local neighborhood, and then those which are the least and most smooth are selected as edge and planar features, respectively.
    To achieve a more uniform distribution of features, the point cloud is divided into regions of the same size, and each region can contain only a limited number of edge and planar feature points.
    A point cannot be chosen as a feature point if there is already a feature point in its local neighborhood.
    A correspondence is found in $\mathcal{P}_k$ for each edge/planar point from $\mathcal{P}_{k+1}$.
    These correspondences are then weighted by their inverse distance, and correspondences with the distance larger than a threshold are discarded as outliers.
    Finally, the pose transform $\mathbf{T}^L_{k+1}$ between $\mathcal{P}_{k+1}$ and $\mathcal{P}_k$ is found by applying the Levenberg-Marquardt method to align the correspondences.

    The second part estimates the pose of the sensor in the map $\mathcal{M}_k$, which is continuously built from the feature points found by the first part of the algorithm. 
    First, $\mathcal{P}_{k+1}$ is projected into the map coordinate system to obtain $\mathcal{P}_{k+1}^W$.
    Then, feature points are searched similarly to as is done in the first part, with the difference being that 10 times more features are found.
    Their correspondences are found in $\mathcal{M}_k$, which is divided into cubes with \SI{10}{\meter} edges.
    The correspondences are searched for only in the cubes intersected by the $\mathcal{P}_{k+1}^W$ to keep the run-time bounded.
    The transform $\mathbf{T}^W_{k+1}$ between $\mathcal{P}_{k+1}^W$ and $\mathcal{M}_k$ is obtained with the same steps as in the first part.
    Due to the 10-times greater amount of correspondences and search through a potentially larger map, this is a much slower process than the first part.

    Thanks to the combination of both parts, the algorithm outputs the pose estimate of the rate of the \ac{lidar}, with drift bounded by slower corrections that snap the pose to the map.
     
    \begin{figure}
      \includegraphics[width=1.0\textwidth]{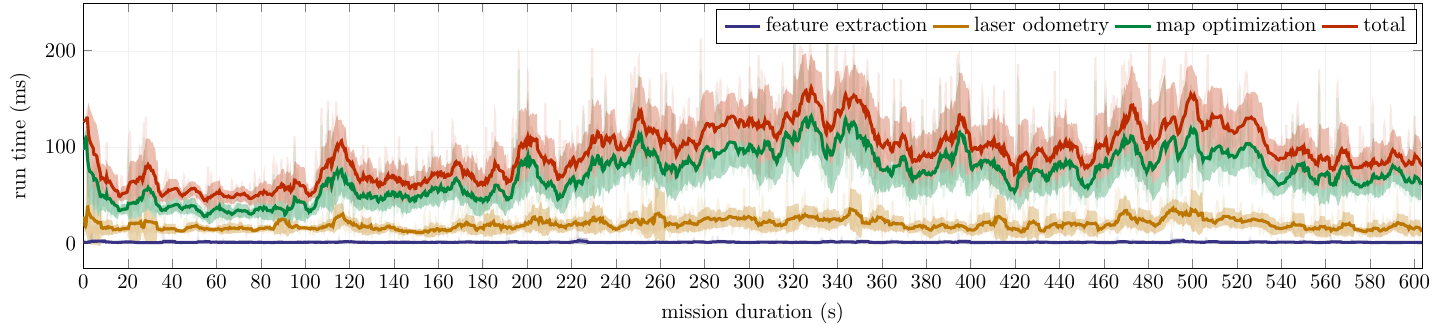}
      \caption{\label{fig:aloam_runtime} 
      The computation time of the most demanding parts of the \ac{aloam} algorithm is plotted with respect to the time in the mission.
      The total time is the sum of all three parts. 
      The darkest colors depict moving mean, the medium dark bands represent the moving standard deviation, and raw data is shown by the lightest colors.
      The moving statistics are calculated over \SI{1}{\second} long time window.
      On average, the feature extraction takes \SI{1}{\milli\second}, the laser odometry \SI{19}{\milli\second}, the map optimization \SI{91}{\milli\second}, and, in total, the pose estimate is obtained in \SI{111}{\milli\second}.
      }
    \end{figure}


    \subsection{LIO-SAM}
    \label{sec:liosam}
 
    \ac{liosam}~\cite{liosam2020shan} utilizes \ac{imu} integration on top of dual factor-graph optimization.
    The first factor-graph optimization is similar to the \ac{aloam} mapping pipeline as it first extracts geometrical features out of raw \ac{lidar} data and registers them to a feature map, with the motion prior given by the second optimization pipeline.
    The second factor-graph optimization fuses the mapping output with \ac{imu} measurements and outputs fast odometry used in the state estimation pipeline.
    The first graph is maintained consistently throughout the run, whereas the second graph optimization is reset periodically to maintain real-time properties.

\begin{figure}[htb]
      \centering
      \includegraphics[width=0.95\textwidth]{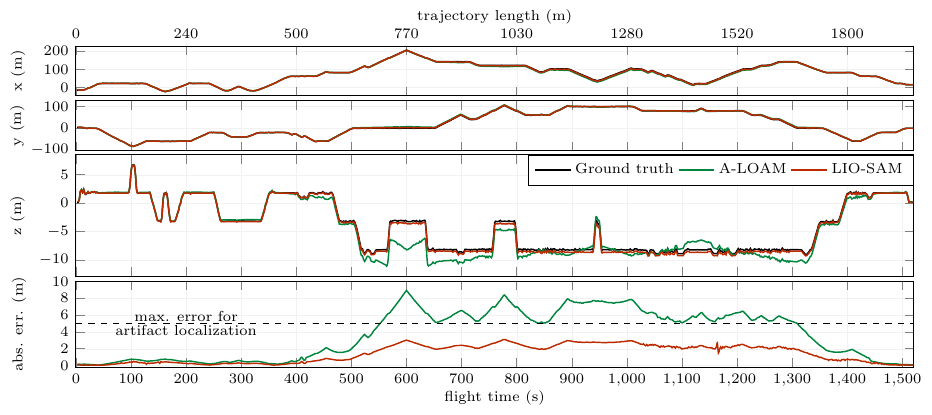}
      \caption{\label{fig:slam_accuracy} 
        The performance of \acs{aloam} and \acs{liosam} during a single flight within \textit{Finals Prize Round World 01} (see~\autoref{fig:virtual_worlds}) of the \ac{darpa} \ac{subt} virtual environment.
        \acs{aloam} does not fuse the inertial measurements which assist \acs{liosam} during \ac{lidar}-scan matching in areas of the environment where such matching suffers from geometric degeneration, in the context of solving optimization problems.
        The selected environment contains a variety of narrow vertical passages where the performance of narrow-\acs{fov} \ac{lidar} perception is limited, leading to drift in the ego-motion estimation that is clearly visible in the \acs{aloam} method.
        The \ac{liosam} method was shown to achieve sufficient accuracy and low drift during long-term and arbitrary 3D navigation within a simulated environment.%
      }
    \end{figure}

    In a simulated environment, \acs{liosam} yields greater accuracy than \acs{aloam} for its fusion of inertial measurements with precisely modeled and known characteristics.
    A comparison of both the methods within the simulated environment is summarized in \autoref{fig:slam_accuracy}.
    In the real world, the measurements of an \ac{imu} rigidly mounted on board a \ac{uav} contain a wide spectrum of large stochastic noise.
    During empirical testing, the integration method in \ac{liosam} was shown to not be robust towards the unfiltered noise while frequency-band and pass filters induced significant time delays, destabilizing the pipeline completely.
    For the inability to accurately model the noise, real-world laser-inertial fusion is done manually by smoothing over a short history of past measurements (see~\autoref{sec:estimation}).

    \subsection{State estimation}
    \label{sec:estimation}

    For precise and collision-free navigation through a cluttered narrow environment, which typically appears in subterranean \ac{sar} scenarios, the control stack requires a smooth and accurate state estimate at a high rate (\SI{100}{\hertz}).
    The \textit{State estimation} module provides such an estimate through the fusion of data from \ac{aloam} and \ac{imu}. It also does this by applying filtering, rejection, and prediction techniques.
    We provide only a brief description of the estimation process as it is not viewed as the primary contribution and has already been presented in~\cite{baca2021mrs}.

    The state vector of the \ac{uav} is defined as $\mathbf{x}=[\mathbf{r}, \mathbf{\dot{r}}, \mathbf{\ddot{r}}, \mathbf{R}, \mathbf{\dot{R}}]^T$.
    The position $\mathbf{r}=[x, y, z]^T$, its first two derivatives of $\mathbf{\dot{r}}$ and $\mathbf{\ddot{r}}$, the orientation in the world frame $\mathbf{R}$, and the angular velocities $\mathbf{\dot{R}}$ include all the dynamics required by other onboard algorithms. 
    Even though the position $\mathbf{r}$ is provided by the \ac{aloam} algorithm, the rate of the position updates is too low for the control loop.
    Furthermore, the velocity and acceleration vector is not known, and must thus be estimated.
    A \ac{lkf} of a point mass model with position, velocity, and acceleration states is employed to estimate the unknown variables at the desired rate.

   While the \ac{imu} of the onboard autopilot provides the orientation $\mathbf{R}$, the heading\footnote{Heading is the angle between the heading vector and the first world axis. 
   The heading vector is the direction of the forward-facing body-fixed axis projected onto the plane formed by the horizontal axes of the world frame, as formally defined in~\cite{baca2021mrs}.} \heading{} is prone to drift due to the bias of the gyroscopes in \ac{mems} \acp{imu}.
   We correct this drift in a standalone heading filter, which fuses $\mathbf{\dot{R}}$ gyro measurements with \ac{aloam} \heading{} corrections.
   Corrections from the magnetometer are not considered, due to the often-occurring ferromagnetic materials and compounds in subterranean environments.

\begin{figure}
    \centering
    \includegraphics[width=0.6\textwidth]{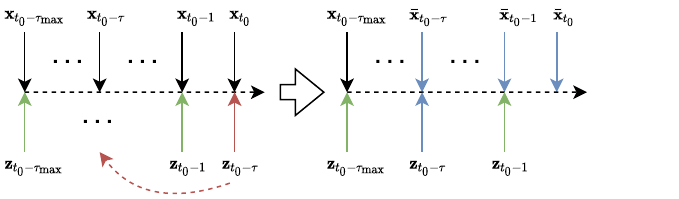}
    \caption{\label{fig:repredictor} 
    The left time sequence shows the situation in the filter after the arrival of delayed correction $\mathbf{z}_{t_0-\tau}$ at time $t_0$.
    The green arrows represent corrections applied at the correct time.
    The delayed $\mathbf{z}_{t_0-\tau}$ would be fused at $t_0$ in a traditional filter, resulting in a suboptimal state estimate.
    However, thanks to the buffering of state and correction history, it is fused into the correct state at time $t_0-\tau$.
    The states after $t_0-\tau$ had to be recalculated to reflect the correction $\mathbf{z}_{t_0-\tau}$, which is shown by the blue color in the right time sequence.
    }
  \end{figure}

\begin{figure}
    \centering
    \includegraphics[width=1.0\textwidth]{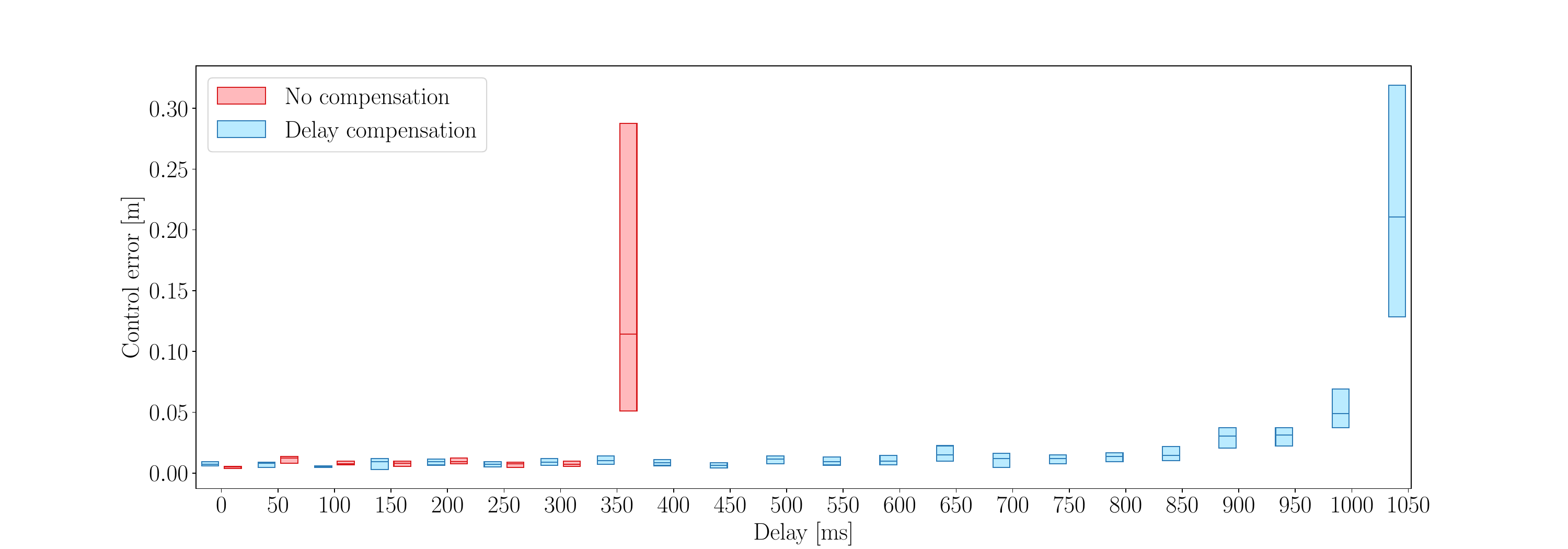}
    \caption{\label{fig:delay_analysis} 
    The box plot shows the median with lower and upper quartiles of the control error with respect to the delay of the position estimate used in the feedback loop. 
    The data was obtained in simulation by artificially increasing the delay of ground truth position in \SI{50}{\milli\second} increments.
    Without compensation, the system becomes unstable after exceeding \SI{300}{\milli\second} delay, which results in oscillation-induced control error at \SI{350}{\milli\second}.
    The control error for the longer delay is not shown, because the high amplitude of oscillations led to a collision of the \ac{uav}.
    The highest delay with compensation is \SI{1000}{\milli\second} when the system has over a \SI{5}{\centi\meter} control error, but is still stable.
    The \ac{uav} stability is lost at \SI{1050}{\milli\second} delay.
    }
  \end{figure}

   The processing of a large quantity of points from each scan and matching them into the map takes \SI{111}{\milli\second} on average (see~\autoref{fig:aloam_runtime} for run time analysis) for the onboard \ac{cpu}.
   The empirical evaluation shows that the controller of the \ac{uav} becomes increasingly less stable when the state estimate is delayed for more than \SI{300}{\milli\second}. 
   To reduce the negative effect of the delay on the control performance, we employ the time-varying delay compensation technique~\cite{pritzl2022repredictor}.
   We define the delay as $\tau=t_{\mathbf{T}_{k+1}} - t_{\mathcal{P}_{k+1}}$, \ie{} the time it took \ac{loam} to compute the pose transform after receiving the point cloud from \ac{lidar}.
   The core of the method is a buffer $\mathbf{Q_x}$ containing the past states $\mathbf{x}_{\langle t_0-\tau_{\text{max}}, t_0\rangle}$, and buffer $\mathbf{Q_z}$ having the past corrections $\mathbf{z}_{\langle t_0-\tau_{\text{max}}, t_0\rangle}$ of the filter.
   The length of the buffer is not fixed, but data older than the expected maximum delay $\tau_{\text{max}}$ are discarded to keep the buffer size bounded.
   When a new delayed measurement $\mathbf{z}_{t_0-\tau}$ arrives at time $t_0$, it is applied as a correction to the state $\mathbf{x}_{t_0-\tau}$ in $\mathbf{Q_x}$. 
   The corrected state $\mathbf{\bar{x}}_{t_0-\tau}$ replaces $\mathbf{x}_{t_0-\tau}$.
   All subsequent states $\mathbf{x}_{(t_0-\tau,t_0\rangle}$ are discarded from $\mathbf{Q_x}$, and replaced by the states $\mathbf{\bar{x}}_{(t_0-\tau,t_0\rangle}$ propagated from $\mathbf{\bar{x}}_{t_0-\tau}$, using regular prediction steps of the filter with all corrections from $\mathbf{Q_z}$. 
   \autoref{fig:repredictor} visualizes the sequence of performed actions.
   Thus, we acquire a time-delay compensated state estimate which, when used in the feedback loop of the \ac{uav} controller, allows for stable flight with a delay of up to \SI{1}{\second}.
   The effect that increasing the delay has on the control error is plotted in~\autoref{fig:delay_analysis}.

    \section{Mapping}
    \label{sec:mapping}
\begin{figure} [ht]
    \newcommand{\imheight}{10.8em}
    \newcommand{\xcap}{1.0em}
    \newcommand{\ycap}{-1.0em}
    \newcommand{\fillopa}{0.3}
    \centering

    \begin{tikzpicture}
      \node[anchor=north west,inner sep=0] (b) at (0,0) {\adjincludegraphics[height=\imheight,trim={{0.0\width} {0.0\height} {0.35\width} {0.0\height}},clip]{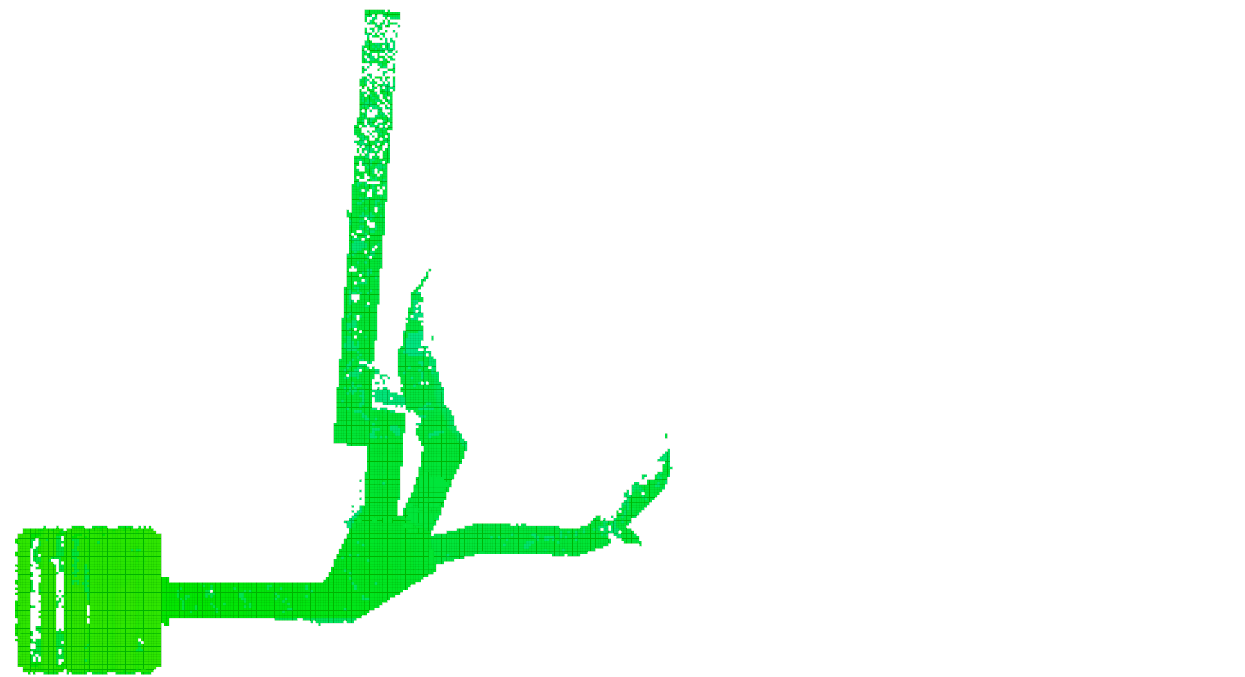}};%
    \begin{scope}[x={(b.south east)},y={(b.north west)}]
      \node[fill=black, fill opacity=\fillopa, text=white, text opacity=1.0] at (\xcap, \ycap) {\textbf{(a)}};
      \end{scope}
    \end{tikzpicture}%
    \begin{tikzpicture}
      \node[anchor=north west,inner sep=0] (b) at (0,0) {\adjincludegraphics[height=\imheight,trim={{0.1\width} {0.1\height} {0.4\width} {0.05\height}},clip]{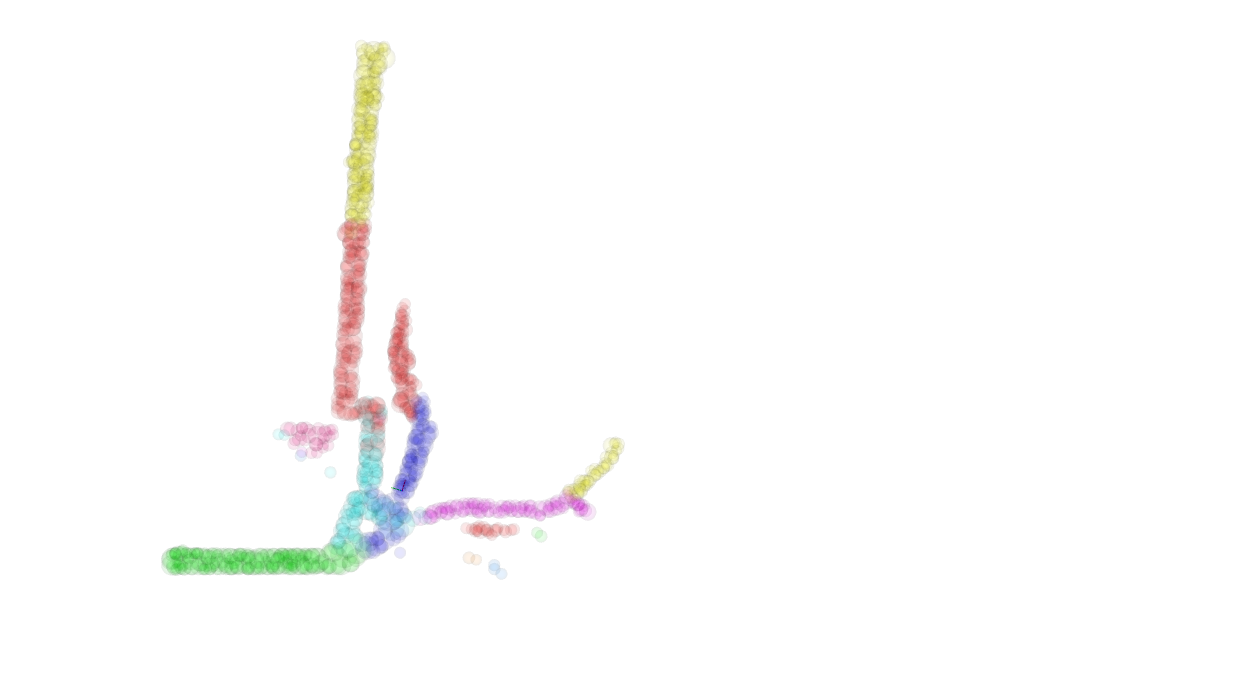}};%
    \begin{scope}[x={(b.south east)},y={(b.north west)}]
      \node[fill=black, fill opacity=\fillopa, text=white, text opacity=1.0] at (\xcap, \ycap) {\textbf{(b)}};
      \end{scope}
      \label{fig:spheremap_spheres}
    \end{tikzpicture}%
    \begin{tikzpicture}
      \node[anchor=north west,inner sep=0] (b) at (0,0) {\adjincludegraphics[height=\imheight,trim={{0.1\width} {0.05\height} {0.40\width} {0.15\height}},clip]{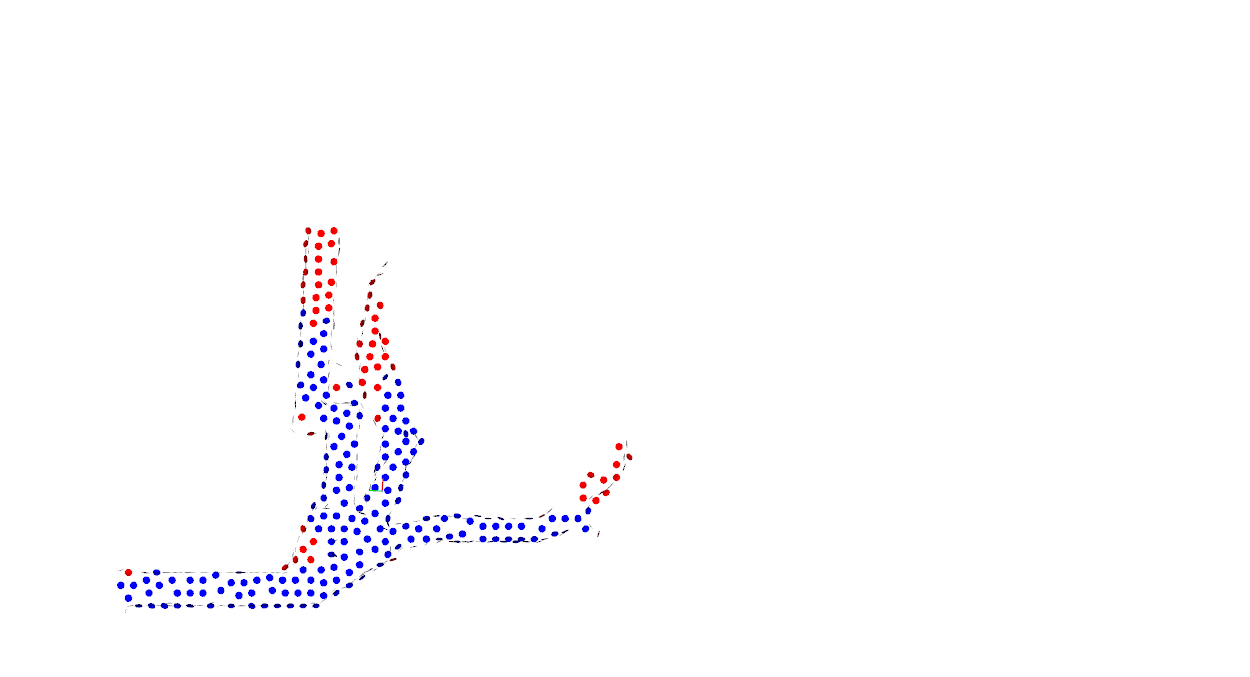}};%
    \begin{scope}[x={(b.south east)},y={(b.north west)}]
      \node[fill=black, fill opacity=\fillopa, text=white, text opacity=1.0] at (\xcap, \ycap) {\textbf{(c)}};
      \end{scope}
    \end{tikzpicture}%
    \begin{tikzpicture}
      \node[anchor=north west,inner sep=0] (b) at (0,0) {\adjincludegraphics[height=\imheight,trim={{0.1\width} {0.10\height} {0.45\width} {0.00\height}},clip]{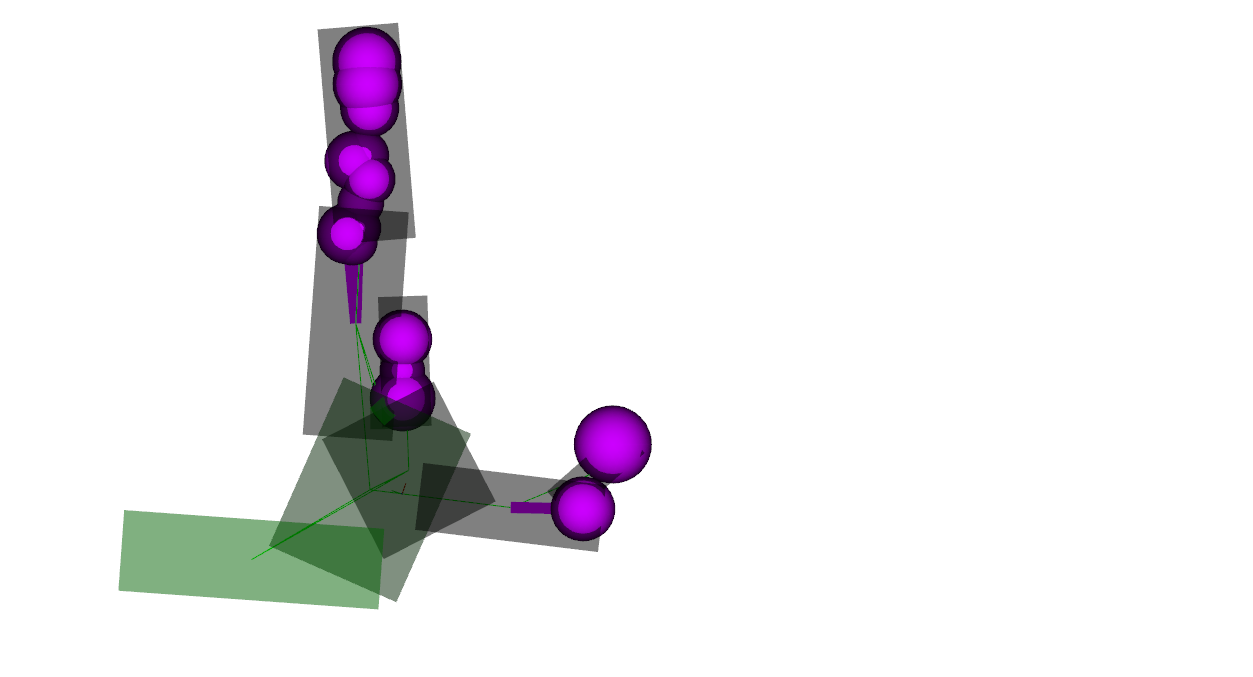}};%
    \begin{scope}[x={(b.south east)},y={(b.north west)}]
      \node[fill=black, fill opacity=\fillopa, text=white, text opacity=1.0] at (\xcap, \ycap) {\textbf{(d)}};
      \end{scope}
    \end{tikzpicture}%
    \caption{
      Top view of the used mapping structures from the intersection of the final event map.
      DenseMap (a) is used  for short-distance planning, SphereMap (b) for safety-aware long-distance planning, FacetMap (c) for storing surface coverage, and LTVMap (d) for compact topological information sharing among robots.
  }
    \label{fig:map_types}
  \end{figure}
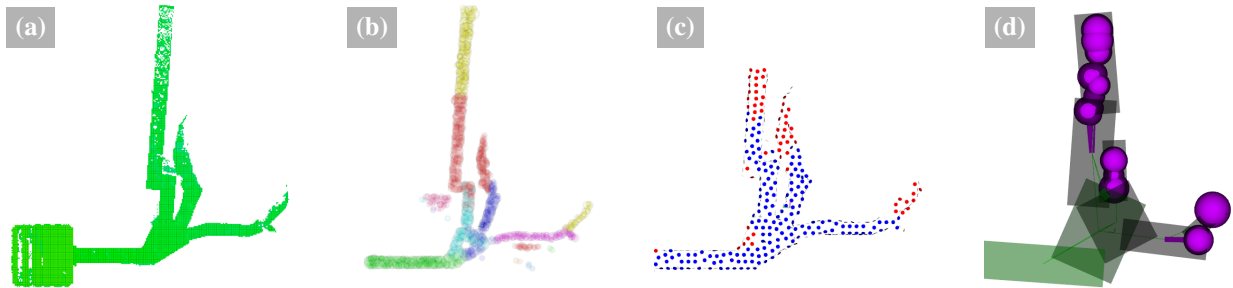

    In this section, we present our approach to mapping the explored environments. 
    As each task has specific requirements on the map properties, we designed multiple spatial representations, each of which is structured for a particular task.
    In particular, DenseMap~(\autoref{fig:map_types}a) is utilized for short-distance path planning; FacetMap~(\autoref{fig:map_types}b) for surface coverage tracking; SphereMap~(\autoref{fig:map_types}c) for fast and safe long-distance path planning; \ac{ltvmap}~(\autoref{fig:map_types}d) for compressed, topological, and mission-specific information sharing between robots in low bandwidth areas; and LandMap~(\autoref{fig:landmap}) for representing feasible spots for safe \ac{uav} landing.
    These maps and the methods for building them are presented in this section.

    \subsection{DenseMap}
    \label{sec:densemap}
    
    Local information of the \ac{uav} is combined within a dense map to serve as the basis for the entire navigation stack, as described in~\cite{kratky2021exploration}.
    The map integrates information in a dense, probabilistic manner using an efficient octree structure implemented within the OctoMap~\cite{hornung2013octomap} library.
    During the map update, the data of each input modality producing spatial measurements are used to update the map with respect to the pose estimate correlating to the timestamp of the respective measurement.
    The data to be integrated are first cleared of any observation noise (see~\autoref{sec:sensory_perception}). The ray of each remaining spatial measurement is integrated within a discretized representation of the environment using the Bayes rule and ternary classification to the unknown, free, and occupied voxels.  
    The output of dense mapping is convertible to other navigation representations and serves as the fundamental structure for local planning and dynamic obstacle detection.

    To retain maximum information under constraints on real-time performance, the voxelization resolution is selected such that a scan insertion is processed at \SI{5}{\hertz}, at worst.
    The resolution can be locally increased if path planning demands a decrease in discretization errors. This is a useful feature for improving safety and repeatability in navigating highly narrow passages.
    To maintain the map structure, the local resolution is controlled by a factor $n$ such that the local resolution equals ${r}/{2^n}$ with $r$ being the default resolution of the dense map.
    In our sensory and computation setup, the default resolution is empirically set to \SI{20}{\centi\meter}, reduced by a factor of $n = 2$ to \SI{5}{\centi\meter} for navigating narrow passages, if required.
    The integrated data consist of \ac{lidar} measurements and depth estimates of two \ac{rgbd} cameras.
    These sensors are mounted on-board \acp{uav} so that the spatial observations cover roughly all directions around the robot, enabling almost arbitrary \ac{uav}-motion planning in collision-free 3D space.
    

    \subsection{SphereMap}
    \label{sec:spheremap}

    To enable the UAV to quickly evaluate the travel time and risk caused by flying near obstacles while also pursuing any given goal, we developed a multi-layer graph structure that uses volumetric segmentation and path caching, called SphereMap~\cite{musil2022spheremap}.
    All three layers of the SphereMap are updated near the UAV in every update iteration, which runs at approximately \SI{2}{\hertz}.

    Path planning in the SphereMap depends on only one parameter $c_R$, which we call \textit{risk avoidance}. It is used to trade path safety for path length.
    For long-distance planning, we disregard UAV dynamics and only take into account the path length and obstacle clearance along the path. 
    We define the path cost between points $\mathbf{p}_1$ and $\mathbf{p}_2$ as
    \begin{equation}
      \label{eq:spheremap_path_cost}
      D(\mathbf{p}_1,\mathbf{p}_2) = L + c_R R,
    \end{equation}
    where $L$ is the path Euclidean length summed over all edges of the path in the sphere graph, and $R\in[0,L]$ is a risk value computed by examining the radii of the spheres along the path.
    For example, a path with all spheres with radii at the minimal allowed distance from obstacles would have $R=L$, and a path through open space with large sphere radii would have $R=0$.

    The lowest layer of the SphereMap is a graph of intersecting spheres, shown in~\autoref{fig:map_types}b. 
    It is constructed by filling the free space of an obstacle $k$-d tree built from the DenseMap with spheres at randomly sampled points. 
    The graph is continuously built out of intersecting spheres, and then by pruning the spheres that become unsafe or redundant.
    The radii of the spheres carry obstacle clearance information, which is used for path risk evaluation. 

    The second layer of the SphereMap is a graph of roughly convex segments of the sphere-graph. It is updated after every update of the sphere graph by creating and merging segments until every sphere in the graph belongs to a segment.

    The third and last layer of the SphereMap is a navigation graph.
    For every two adjacent segments, we store one sphere-sphere connection, which we call a \textit{portal} between the segments, as in~\cite{blochtlinger2018topomap}.
    These portals form the vertices of the navigation graph.
    At the end of every SphereMap update iteration, we compute which paths are optimal according to the path cost from~\autoref{eq:spheremap_path_cost} between all pairs of portals of a given segment.
    The paths are computed only inside that given segment. If the segments are kept small (tens of meters in length), the recomputation is reasonably fast.
    The optimal portal-portal paths form the edges of the navigation graph.
    The UAV uses the navigation graph to quickly find long-distance paths between any two points in the known space by planning over the edges of the navigation graph, and then by only planning over the sphere graph in the first and last segments of the path.


    \subsection{FacetMap}
    \label{sec:facetmap}
    \begin{figure}
      \centering
      \includegraphics[width=0.495\textwidth]{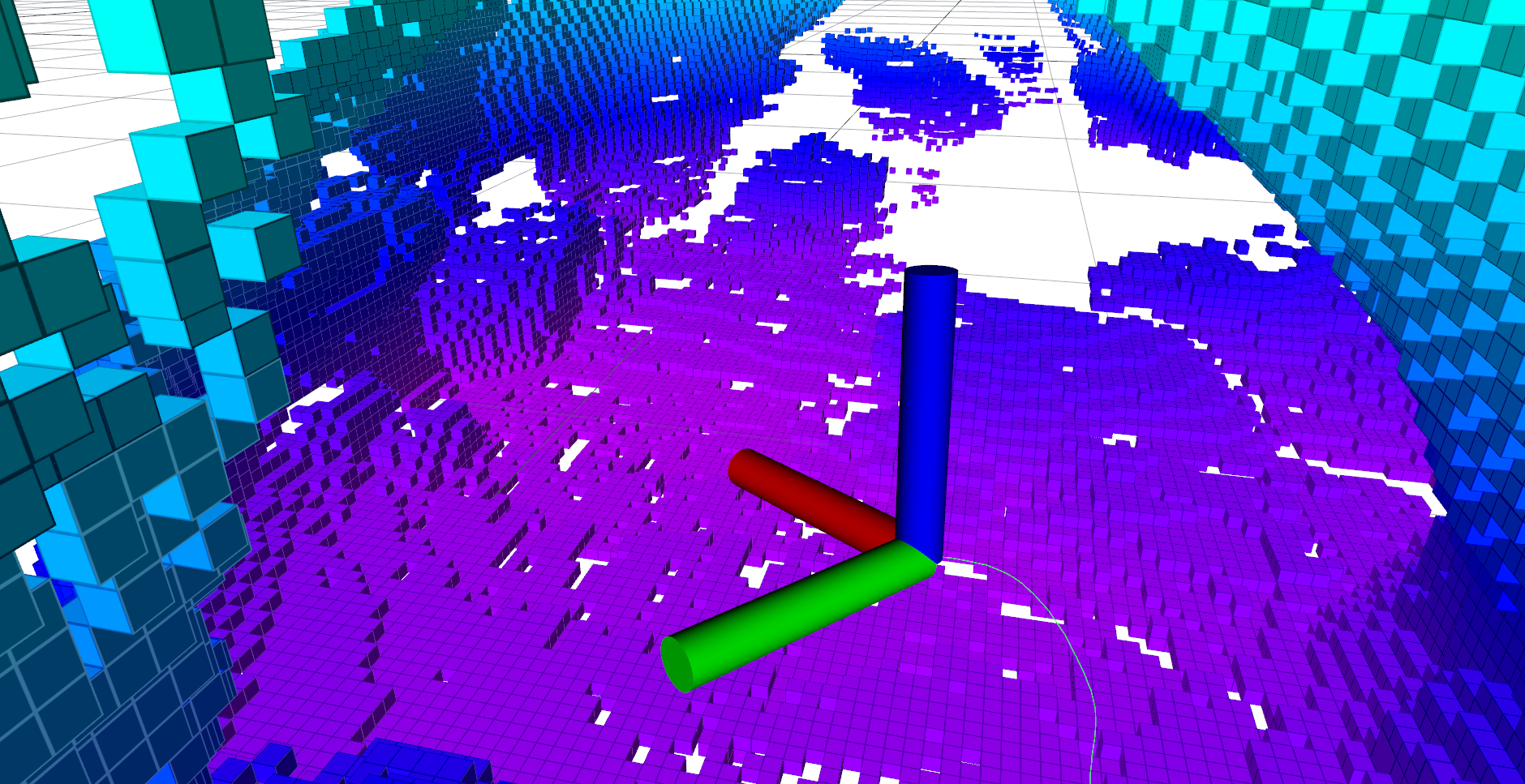}
      \includegraphics[width=0.495\textwidth]{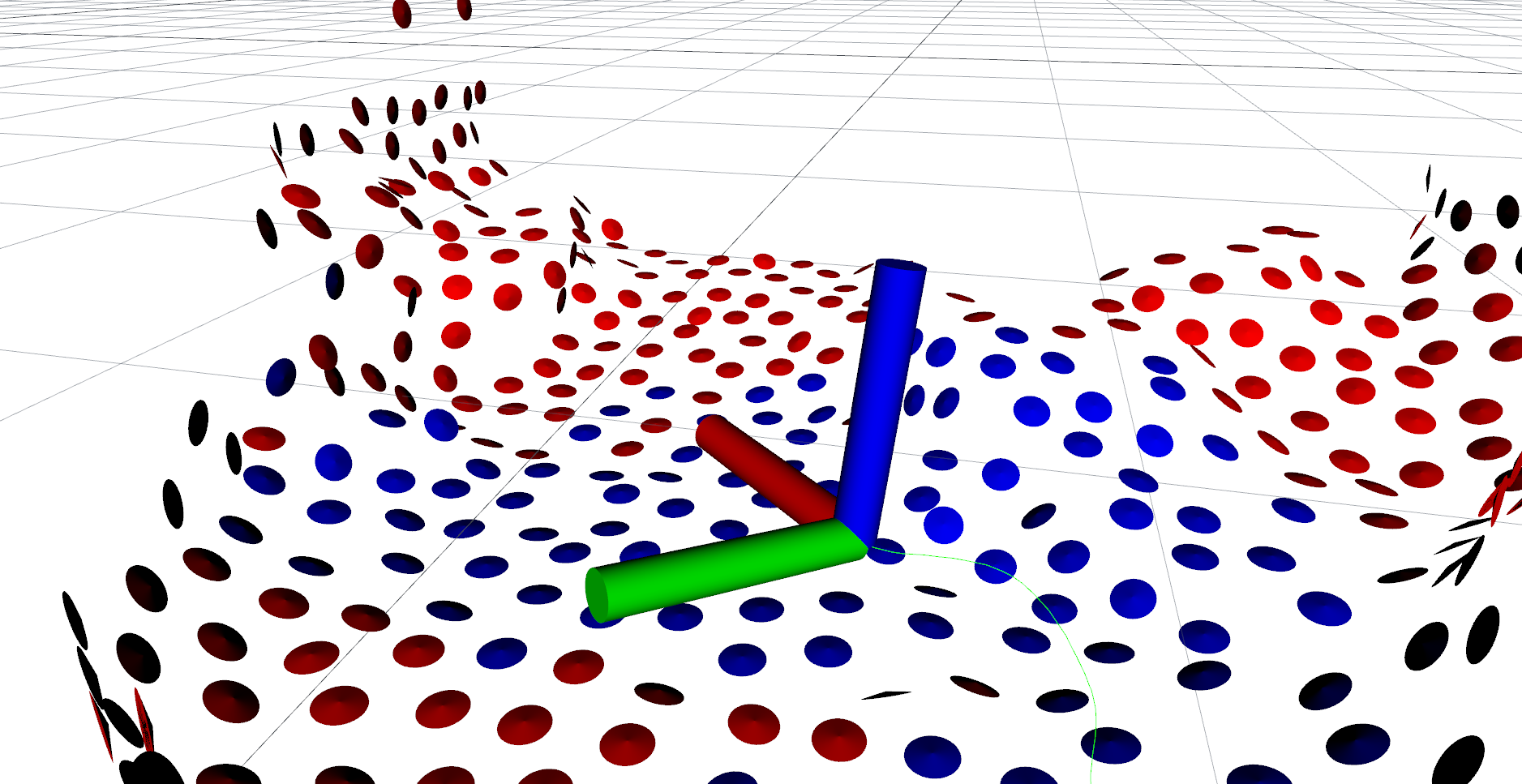}
     \caption{\label{fig:facetmap_demo} 
      Illustration of the FacetMap as described in~\autoref{sec:facetmap}.
      The map is built from the DenseMap (left) by finding normals of sampled points.
      The orientation of the visualization discs (right) is determined by the facet's normal, and the color by whether the facet was covered by the UAV's front-facing cameras or not.
      }
    \end{figure}
    The occupancy octree and SphereMap maps are sufficient for volumetric exploration.
    However, the goal of the DARPA SubT challenge was to locate artifacts, most of which could be detected only from cameras. Because the \ac{fov} of our UAVs' cameras did not cover the entire \ac{fov} of the \ac{lidar} and depth cameras, not all occupied voxels in the occupancy map could be considered as ``covered by cameras". 
    For this reason, we developed another map, called FacetMap, illustrated in~\autoref{fig:facetmap_demo}.
    This map is a simple surfel map, with the facets stored in an octree structure, each having an orientation, a coverage value, and a fixed size.
    The FacetMap is built by computing the normals of the occupancy map at sampled occupied points, and creating facets with a set resolution if there are no existing facets with a similar normal nearby. 
    The facets are updated (i.e. added or deleted) periodically at approximately \SI{2}{\hertz} in a cube of pre-defined size around the \ac{uav}.

    Each facet holds a coverage value that is, for simplicity, defined as binary.
    A facet is marked as covered if the facet center falls into the FOV of any camera, and the ray from the camera to the facet center is at a certain angle from the facet's normal, so as to not mark surfaces as covered if they are viewed at a very skewed angle.
    The covered facets stay in the map even if the underlying occupancy map shifts (e.g. when an obstacle moves).
    As described in~\autoref{sec:viewpoint_path_enhancement}, one strategy used in our system uses this map to cover as much of the surface as possible while flying between volumetric exploration viewpoints. The strategy in~\autoref{sec:dead_end_inspection} uses this map to completely cover surfaces of certain areas.
    Coverage of entire regions of the SphereMap can also be easily computed, and then stored in the \ac{ltvmap}, as described in~\autoref{sec:lsegmap}


    \subsection{\acs{ltvmap}}
    \label{sec:lsegmap}
    Distributing all of the maps described in this chapter among the \acp{uav} would be highly demanding for the communication network. 
    As such, we have developed the \acf{ltvmap}, which combines the necessary mission-related information from the other maps and can be quickly extracted from the SphereMap and sent at any time.

    This map consists of an undirected graph, where each vertex is created from a free-space segment in the original SphereMap and the edges are added for all of its adjacent segments.
    Each vertex holds an approximation of the segment's shape.
    In our implementation, we use four \ac{dof} bounding boxes (with variable size and rotation along the vertical axis) for shape approximation, though any other shape could be used.

    For cooperative exploration purposes, the frontier viewpoints (described in~\autoref{sec:viewpoint_computation}) found by a given UAV are also sent in the \ac{ltvmap}, with each viewpoint being assigned an information value and segment from which the viewpoint is reachable.
    For surface coverage purposes, every segment in the \ac{ltvmap} also holds a single numerical value representing the percentage of relevant surfaces covered in that segment.
    This value is computed by projecting points from the facets of the FacetMap and counting the points that fall into every segment.
    Further description and analysis of \acp{ltvmap} can be found in~\cite{musil2022spheremap}.
    These \acp{ltvmap} are shared among robots, and are used for cooperative search planning onboard \acp{uav}, as described in~\autoref{sec:coop_ss}.
    
    \subsection{LandMap}
    \label{sec:landmap}

\begin{figure}[htb]
      \centering
      \includegraphics[width=0.9\textwidth]{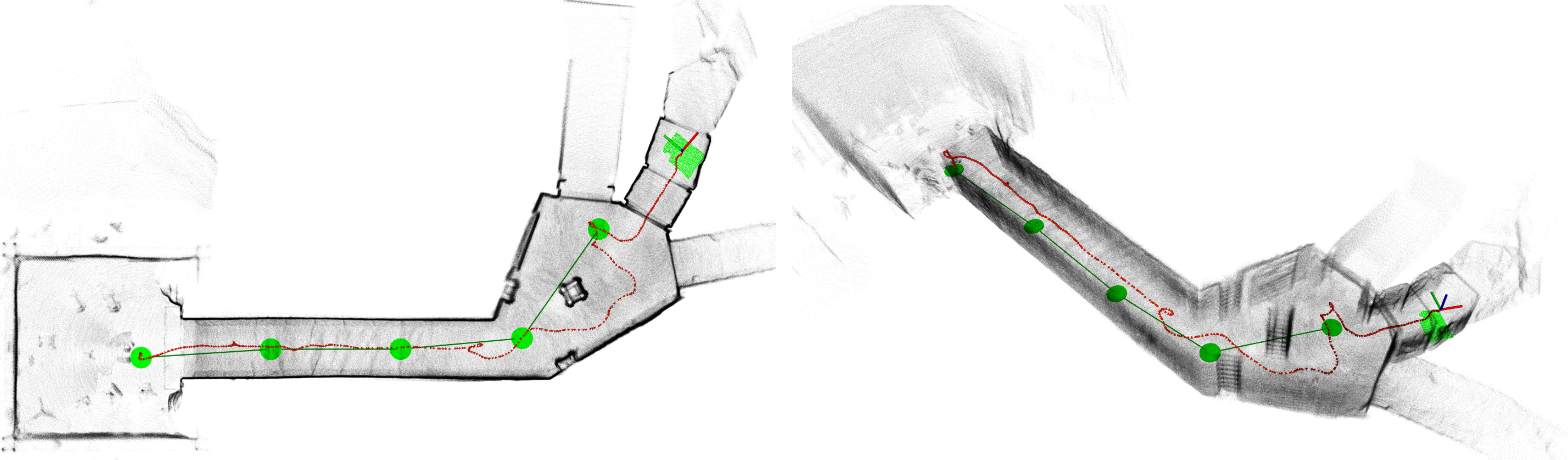}
      \caption{\label{fig:landmap} 
        Example of the LandMap with resolution of \SI{5}{\meter} built in the beginning of the \ac{darpa} \ac{subt} final event after \SI{70}{\second} of a \ac{uav} flight. The \ac{uav} is represented by the Cartesian axes with its trajectory colored in red.
        The LandMap incorporates the spots classified as safe for \ac{uav} landing (green circles) which are used during the \ac{uav} homing phase of the mission to ensure safety during the landing procedure.%
      }
    \end{figure}

\begin{figure}
      \centering
      \adjincludegraphics[width=1.0\textwidth,trim={{0.00\width} {0.0\height} {0.00\width} {0.00\height}},clip]{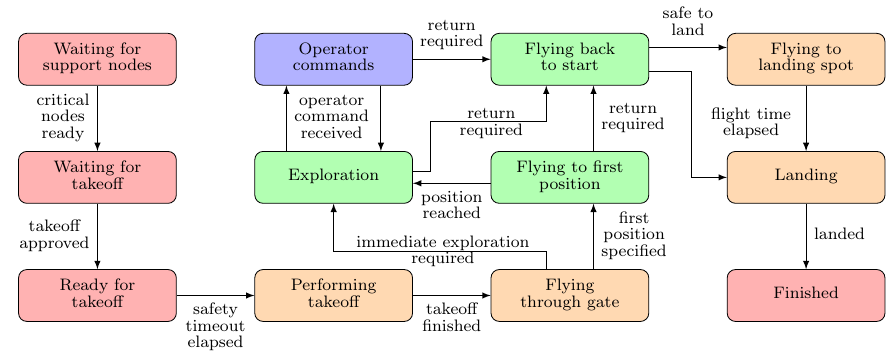}
      \caption{\label{fig:mission_sm} 
        Simplified version of the state machine governing the autonomous mission in SubT System Finals.
      }
    \end{figure}

    As described in \autoref{sec:landing_spot_detection}, a downward-facing \ac{rgbd} camera detects areas safe for landing.
    These areas are continuously collected within an unconnected set and stored in a sparse point-cloud manner under a certain resolution, usually low enough to keep the LandMap memory-light.
    An example of the LandMap is shown in~\autoref{fig:landmap}.
    During the homing phase of the mission, the \ac{uav} navigates to an area connected to the ground station via the communication network (see~\autoref{sec:communication}). 
    After reaching this area, the \ac{uav} navigates towards a safe landing spot as indicated by the LandMap, which is closest to its current pose (see mission state machine in~\autoref{fig:mission_sm}).
    While flying towards the LandMap-selected spot, the \ac{uav} lands sooner if the ground below the \ac{uav} is classified as safe-for-landing in the current \ac{rgbd} data.
    The landing spots previously identified as safe are, once more, verified before landing in order to ensure safety in dynamic environments.
    If the spot is no longer safe for landing, it is invalidated and the \ac{uav} is navigated to the next closest landing spot.

    \section{Autonomous search}
    \label{sec:exploration}

    Since communication between robots in subterranean environments can never be ensured, the UAVs in our system operate completely autonomously and only use information from other robots to update their goal decision (e.g. blocking frontiers leading to areas explored by other robots).
    The system can also be controlled at a very high level by the human operator, which is described in~\autoref{sec:operator_commands}. 
    This section describes the high-level search autonomy of our system.

    \subsection{Informative viewpoint computation and caching}
    \label{sec:viewpoint_computation}
    For exploration purposes, the UAVs in our system do not consider the information gain along trajectories, but rather sequences of discrete viewpoints, so that we can have a unified goal representation for both local and global search planning.
    These viewpoints are divided into places at which a UAV could obtain some volumetric information, called \textit{frontier viewpoints}, and the points at which a UAV could cover some not-yet-covered surfaces with its cameras, called \textit{surface coverage viewpoints}.
    Each viewpoint $\xi$, comprising of position $\vec{p}_\xi$ and heading $\varphi_\xi$, is therefore assigned some information value $I(\xi)$.
  In our approach, the information gain of frontier viewpoints $\xi_F$ and surface viewpoints $\xi_S$ is computed as
  \begin{equation}
    {I}( \xi_F) = c_{F} \frac{n_{\textrm{unk}}}{n_{\textrm{rays}}}, \quad \quad I(\xi_S) =c_{S}n_{\textrm{unc}} + c_{SF},
  \end{equation}
  where ${n_{\textrm{unk}}}/{n_{\textrm{rays}}}$ is the ratio of rays cast in the UAV's depth cameras' and LIDAR's FOVs that hit an unknown cell of the occupancy map before hitting an occupied one or going out of range.
  Similarly, $n_{\textrm{unc}}$ is equal to the number of uncovered facets of the FacetMap, hit by rays that are cast in the UAV's RGB cameras' FOVs.
  The constants $c_F$, $c_S$, $c_{SF} $ are empirically tuned to alter the UAV's next viewpoint selection and hence, its behavior.

  The UAV does not sample and evaluate viewpoints on-demand after reaching some viewpoint, rather it continually samples viewpoints in its vicinity at a given rate and stores them into a map of cached viewpoints.
    Only viewpoints that have $I(\xi)$ above some threshold, are safe, not too close to another informative viewpoint, and not blocked by mission control are stored.
    The viewpoints are also pruned from the map if they become uninformative or if a better viewpoint is nearby.
    Lastly, viewpoints that were found in a previous update and are now outside the local update box, are kept as global goals and are pruned more aggressively than the local goals.
    This approach continually produces a map of informative viewpoints that is denser near the UAV and sparse in the rest of the environment.

    \subsection{Single-UAV autonomous search planning}
    \label{sec:single_uav_goal_eval}
    In our approach, the UAV can be in three states of autonomous search --- \textit{locally searching}, \textit{traveling to goal} or \textit{returning}, and the goal planning and evaluation is divided into local and global planning, as in~\cite{dang2019gbplanner}.
    The transitions between these states are fairly simple --- if there are informative and reachable viewpoints near the UAV, the UAV is in the locally searching state and tries to always keep a sequence of two viewpoints. 
    These are given to the trajectory planning pipeline so that the UAV doesn't stop at each viewpoint and compute the next best one.
    This is done by performing a local replanning of the sequence whenever the UAV is getting close to a viewpoint.

    When there are no reachable viewpoints near the UAV or when new information is received from the operator or other robots, a global replanning is triggered.

    The global replanning computes paths to all stored informative viewpoints (not only in the local search box) and evaluates them. 
    The best viewpoint is then set as a goal to the long-distance navigation pipeline described in~\autoref{sec:long_distance_navigation}.
    Finally, the \textit{returning} state is triggered when the global planning does not find any reachable goals, or if the operator demands it, or if $t_{\textrm{home}} < c_R t_{\textrm{battery}}$, where $t_{\textrm{home}}$ is the estimated time of flight needed to return to the base station, $t_{\textrm{battery}}$ is the estimated remaining flight time, and $c_R$ is an empirically tuned constant.
    The value of $t_{\textrm{home}}$ is computed from the UAV's average flight speed, and a path found through the SphereMap to the base station.
    If there is no path to the base station, the UAV will instead try to return along a tree of visited positions, which is built specifically for this purpose, so that for example if a path is only temporarily blocked, the UAV will fly to the roadblock, and if it is removed, will continue flying to the base station.
    The UAV can also recover from this state, if it is returning due to having found no reachable goals, and suddenly some goals become reachable again.
    When the UAV gets close to the goal, it switches back to the \textit{locally searching} state.

    The reward functions used to evaluate goals govern the behavior of the UAV while searching the environment, and as such, they define the search strategy of the UAV.
    For simplicity, we made the local planner and global planner use the same reward function in a given strategy, with only one difference, that the local planner can add a penalty to local goals, based on the UAV's current momentum and heading, to allow for smoother local search.
    These strategies and their corresponding reward functions were utilized in the challenge:

    \subsubsection{Greedy search strategy (GS)}
      The simplest reward function for selecting the next best viewpoint $\xi$ from the current UAV viewpoint $\xi_{\textrm{UAV}}$ (the UAV's current position and heading) can be written as
    \begin{equation}
      R_{\textrm{GS}}(\xi_{\textrm{UAV}}, \xi) = {I(\xi)} - { D(\xi_{\textrm{UAV}}, \xi)},
    \end{equation}
        where $I(\xi)$ is the information value of the viewpoint (described in~\autoref{sec:viewpoint_computation}) and $D$ is the best path cost computed in the SphereMap (described in~\autoref{sec:spheremap}).
        This reward function for controlling the next best goal selection thus depends on the constants $c_F$, $c_S$, $c_{FS}$ described in~\autoref{sec:viewpoint_computation} and the risk-awareness constant $c_R$ used in path planning, which can be used to tune the search based on the user's needs.

    This reward function is very simple and can take the UAV in various directions, leaving behind uncovered surfaces in faraway places.
        The next strategy aims to solve this.

    \subsubsection{Dead end inspection strategy (DEI)}
    \label{sec:dead_end_inspection}
      A more thorough reward function can be written as
    \begin{equation}
      R_{\textrm{DEI}}(\xi_{\textrm{UAV}}, \xi) = I(\xi) - D(\xi_{\textrm{UAV}}, \xi) + (D(\mathbf{p}_{\textrm{HOME}}, \xi) - D(\mathbf{p}_{\textrm{HOME}}, \xi_{\textrm{UAV}})).
    \end{equation}
    This strategy adds the difference in path costs to the base station position $\mathbf{p}_{\textrm{HOME}}$ from the evaluated viewpoint and from the UAV. 
    This greatly increases the value of viewpoints that are deeper in the environment, relative to the UAV.
    Using this reward function, the UAV will most likely first explore frontiers until reaching a dead-end, and then thoroughly cover surfaces from the dead end back to the base, analogous to a depth-first search.

    \subsubsection{Viewpoint path enhancement strategy (VPE)}
    \label{sec:viewpoint_path_enhancement}
      The third strategy used on the UAVs is not a change of the reward function, but rather a simple way to increase surface coverage when the UAV is flying through long stretches of explored but not perfectly covered space, either in the DEI or GS strategy.
      If VPE is enabled and the UAV is flying to a distant goal, then we periodically take the short-distance trajectory from the local path planner (described in~\autoref{sec:planning}), sample it into multiple viewpoints, and try to perturb these viewpoints to increase surface coverage, while not increasing the flight time too much.
      Thus, we fully utilize the agility of quadcopter UAVs, as they can easily turn from side to side while flying in a given direction.


    \subsection{Probabilistic cooperative search planning}
    \label{sec:coop_ss}
    Our approach to multi-UAV search planning was to make the UAVs completely autonomous and decentralized by default, while also being able to share important information and use it for their own planning.
    Each UAV always keeps the latest version of the LTVMap (described in~\autoref{sec:lsegmap}) received from a given UAV.
    When a new LTVMap is received, every newest received map currently being stored onboard the UAV is updated by every other newest received map, as well as by the LTVMap constructed from the UAV's own SphereMap.
    \begin{figure}
      \centering
      \includegraphics[clip, trim=4cm 4.0cm 0cm 0cm, width=0.80\textwidth]{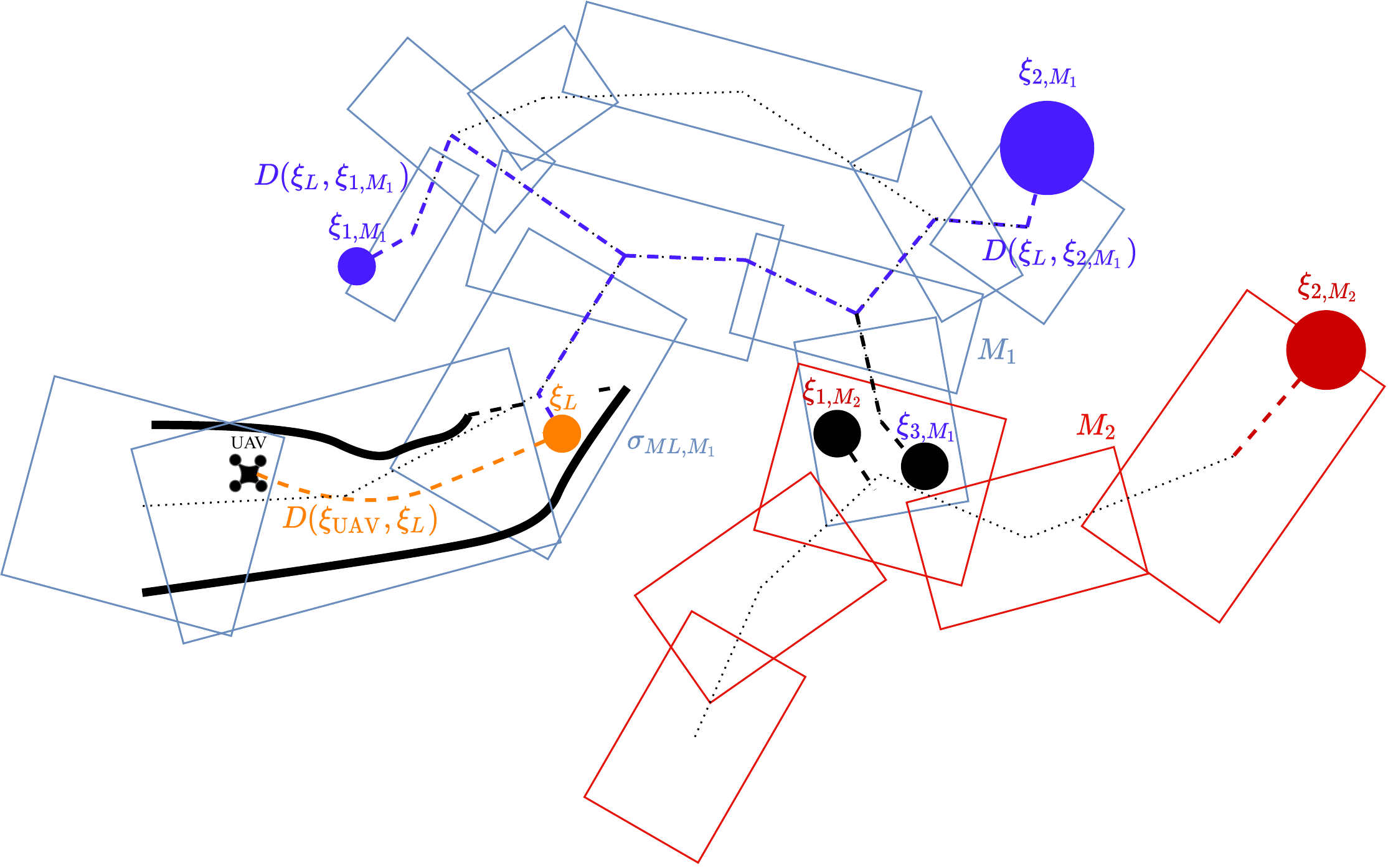}
      \caption{%
        Diagram illustrating the computation of the cooperative exploration reward function, as described in~\autoref{eq:coop_planning}.
        The image shows a UAV evaluating a frontier viewpoint $\xi_L$ (orange) in its local occupancy map (black lines).
        The UAV has received two LTVMaps $M_1, M_2$ from two other UAVs.
        As the local map frontier $\xi_L$ falls into one of the free space segments $\sigma_{ML,M_1}$ of $M_1$, it is assigned as belonging to that segment and acts as an edge in planning paths between the local map and the received map $M_1$.
        Therefore, the frontier viewpoints $\xi_{1, M1}, \xi_{2,M2}$ should be reachable through $\xi_L$. A path to them is estimated across the centroids of the segments of $M_1$.
        The viewpoints $\xi_{3,M1}, \xi_{1,M2}$ (black) are marked as having $l(\xi \in V_{\mathrm{exp}}) = 1$, since they fall deep into the explored space of the other received map, and are therefore not considered.
      }
      \label{fig:coop_schematic}
    \end{figure}

\begin{figure}
    \newcommand{\imheight}{12.50em}
    \newcommand{\xcap}{0.95em}
    \newcommand{\ycap}{0.8em}
    \newcommand{\fillopa}{0.3}
    \newcommand{\clipv}{0.11}
    \centering
    \begin{tikzpicture}
      \node[anchor=south west,inner sep=0] (b) at (0,0) {\adjincludegraphics[width=0.245\textwidth,trim={{0.00\width} {0.0\height} {0.00\width} {\clipv\height}},clip]{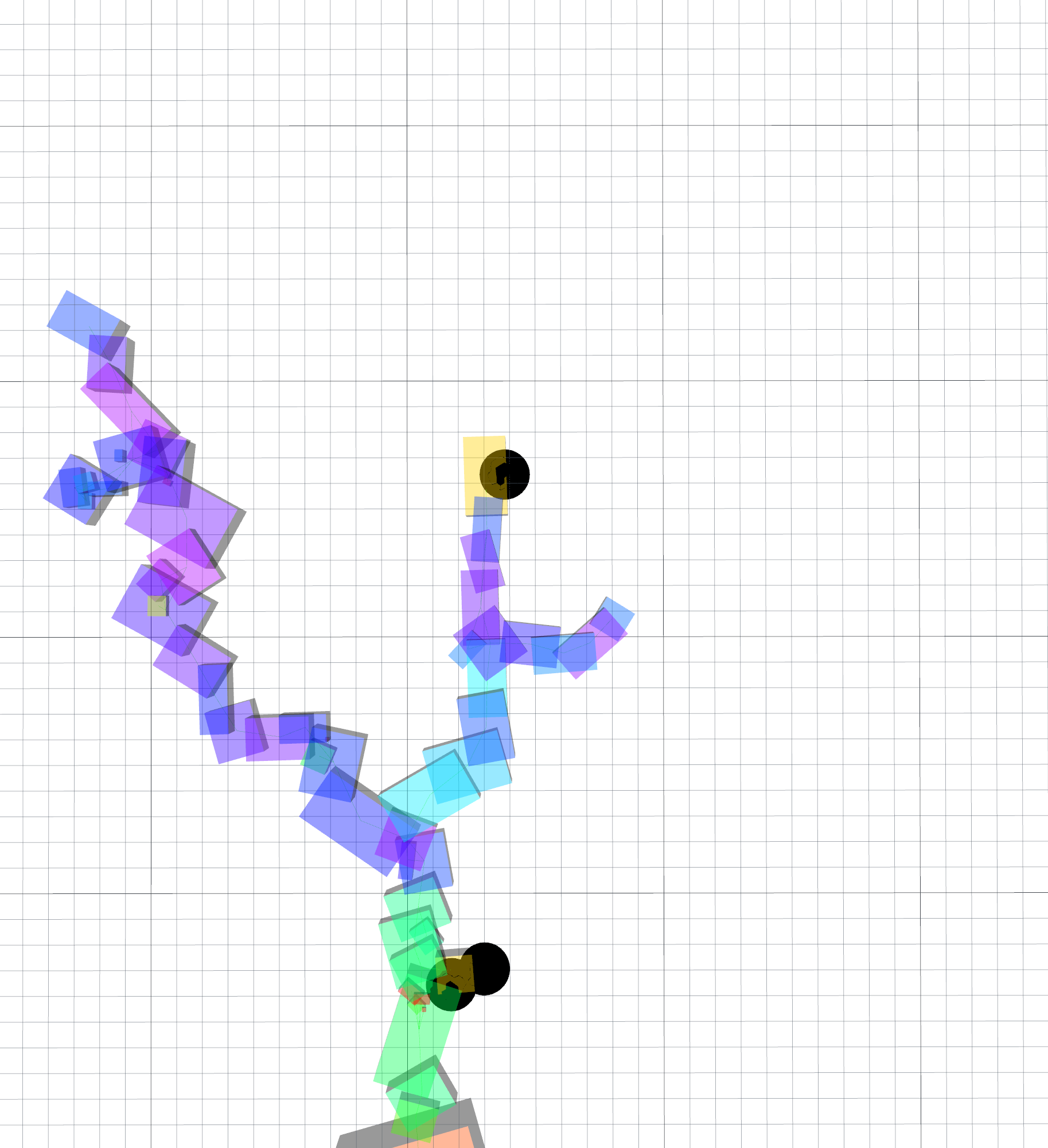}};%
    \begin{scope}[x={(b.south east)},y={(b.north west)}]
  \node[fill=black, fill opacity=\fillopa, text=white, text opacity=1.0] at (\xcap, \ycap) {\textbf{(a)}};
      \end{scope}
    \end{tikzpicture}
    \begin{tikzpicture}
      \node[anchor=south west,inner sep=0] (b) at (0,0) {\adjincludegraphics[width=0.245\textwidth,trim={{0.00\width} {0.0\height} {0.00\width} {\clipv\height}},clip]{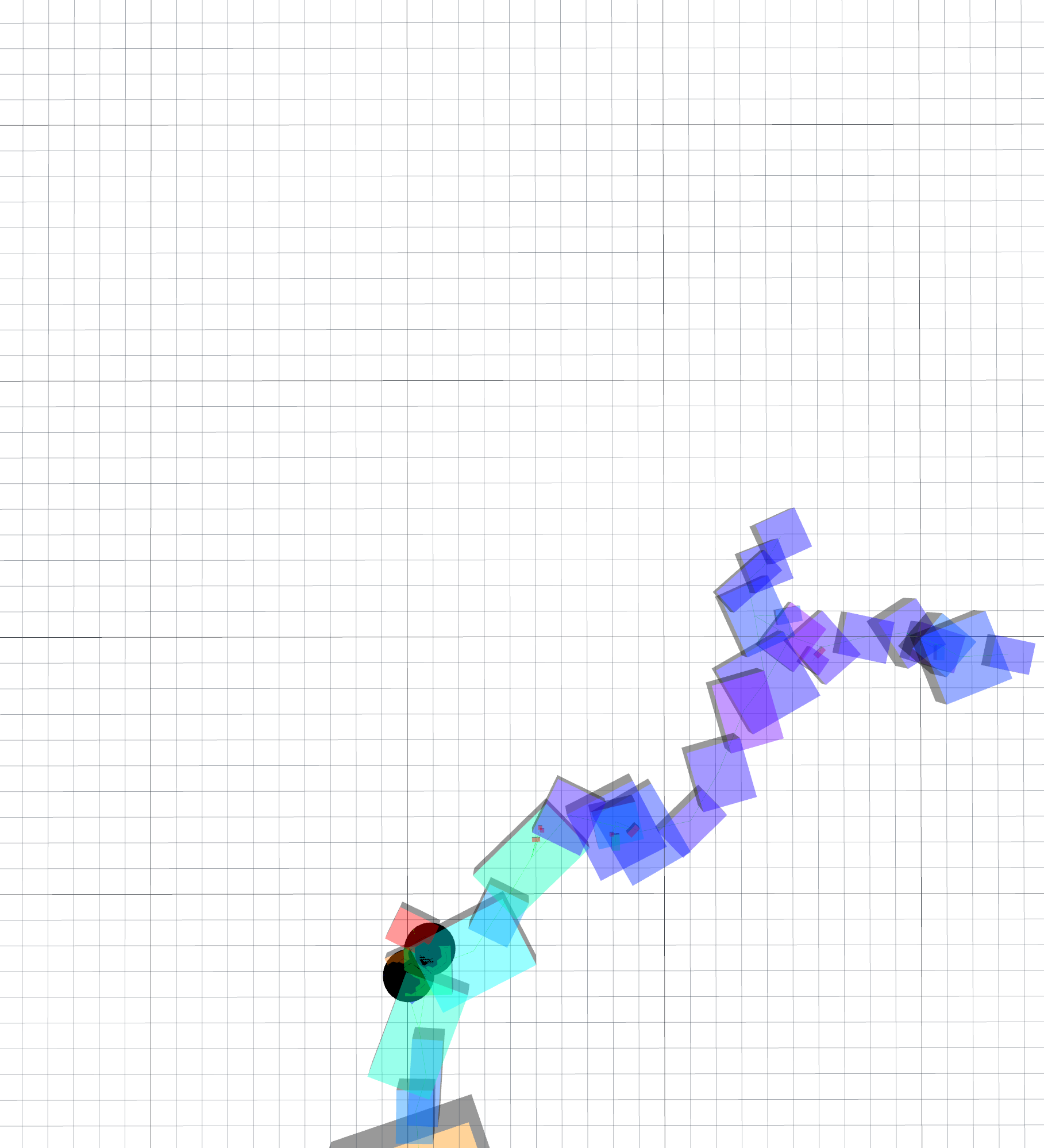}};%
    \begin{scope}[x={(b.south east)},y={(b.north west)}]
  \node[fill=black, fill opacity=\fillopa, text=white, text opacity=1.0] at (\xcap, \ycap) {\textbf{(b)}};
      \end{scope}
    \end{tikzpicture}
    \begin{tikzpicture}
      \node[anchor=south west,inner sep=0] (b) at (0,0) {\adjincludegraphics[width=0.245\textwidth,trim={{0.00\width} {0.0\height} {0.00\width} {\clipv\height}},clip]{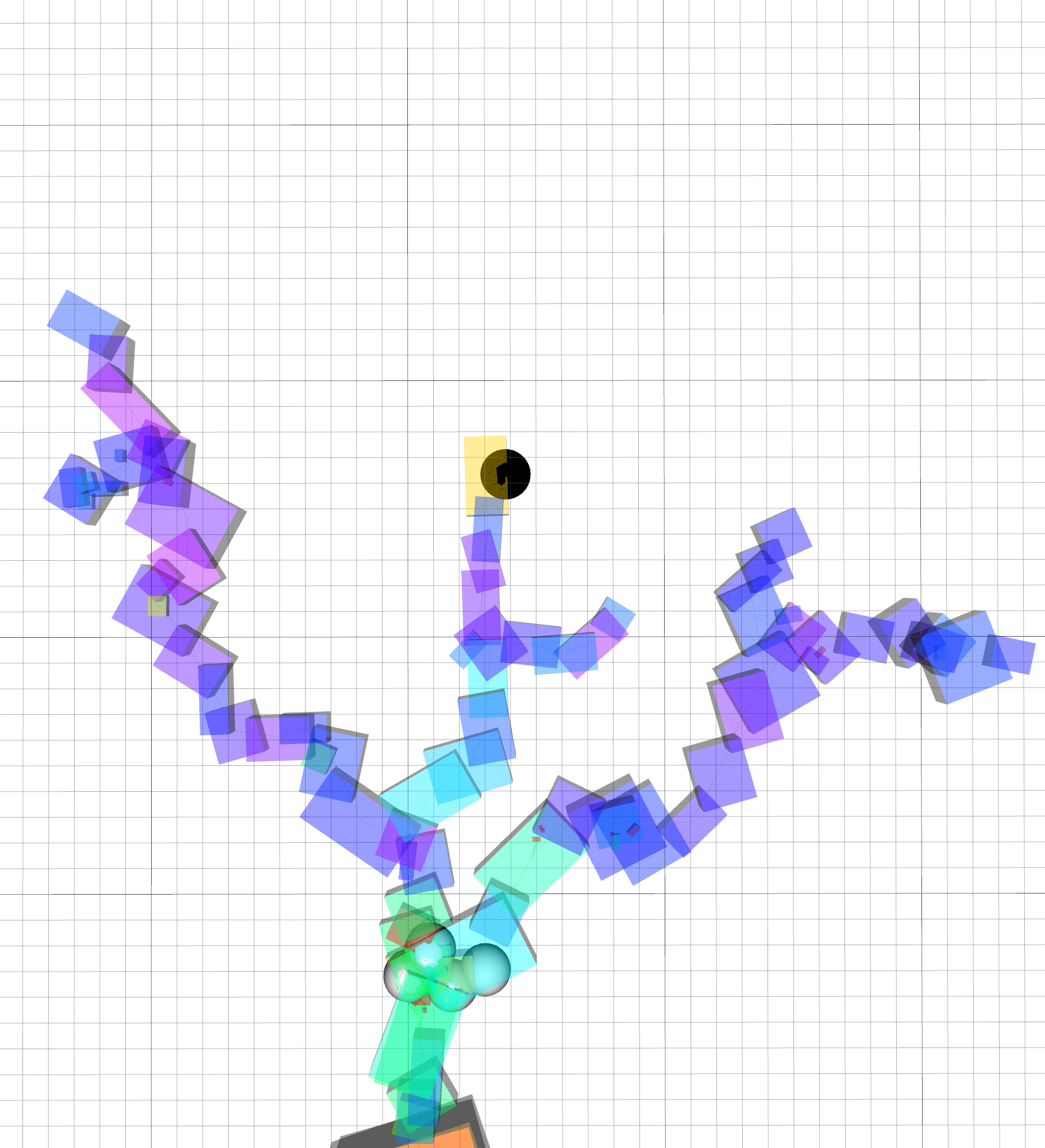}};%
    \begin{scope}[x={(b.south east)},y={(b.north west)}]
  \node[fill=black, fill opacity=\fillopa, text=white, text opacity=1.0] at (\xcap, \ycap) {\textbf{(c)}};
      \end{scope}
    \end{tikzpicture}
    \begin{tikzpicture}
      \node[anchor=south west,inner sep=0] (b) at (0,0) {\adjincludegraphics[width=0.245\textwidth,trim={{0.00\width} {0.0\height} {0.00\width} {\clipv\height}},clip]{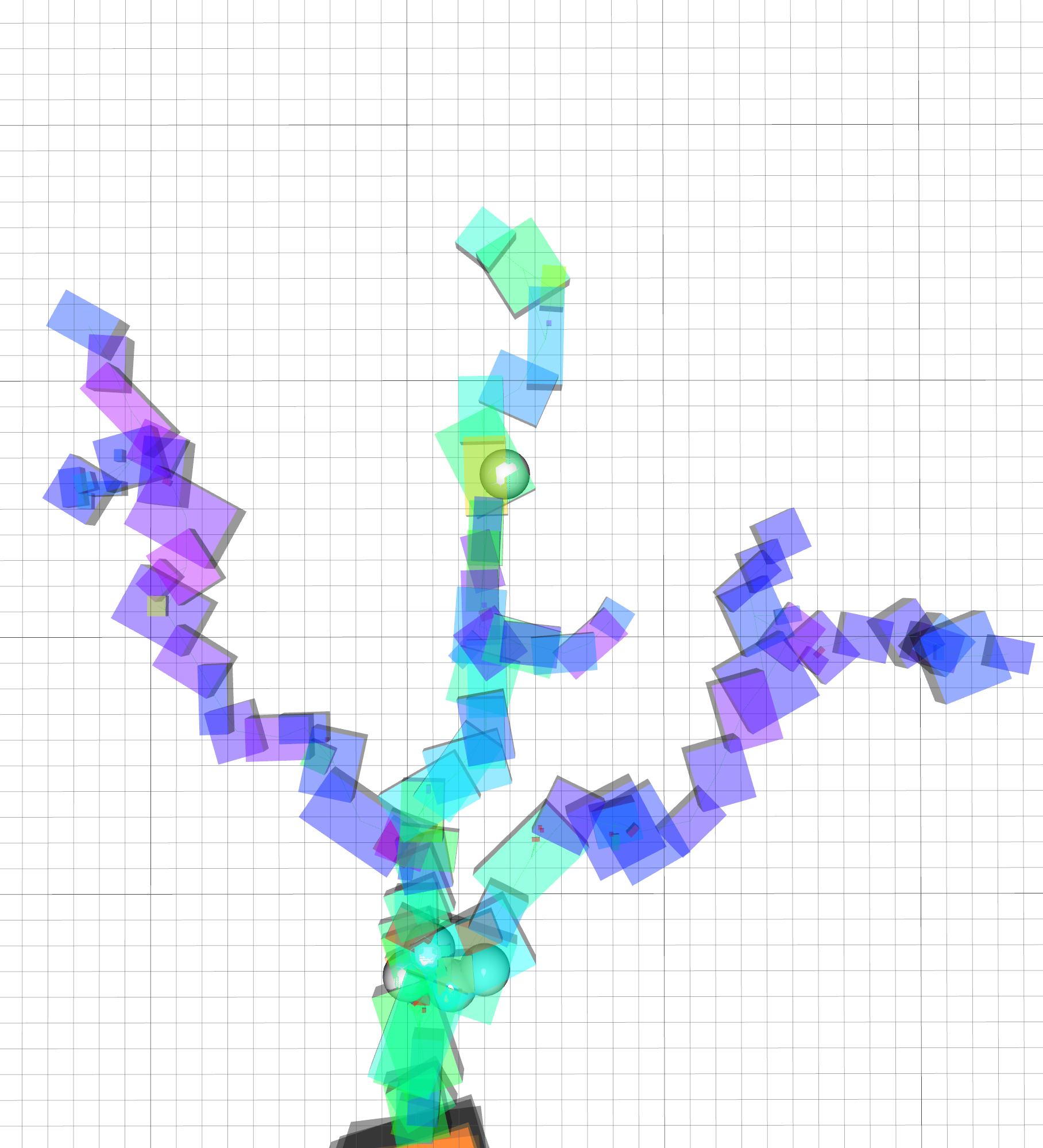}};%
    \begin{scope}[x={(b.south east)},y={(b.north west)}]
  \node[fill=black, fill opacity=\fillopa, text=white, text opacity=1.0] at (\xcap, \ycap) {\textbf{(d)}};
      \end{scope}
    \end{tikzpicture}
     \caption{\label{fig:use_of_received_maps} 
      Illustration of \ac{ltvmap} sharing and utilization during a cave exploration mission in simulation with three UAVs running the DEI strategy (described in~\autoref{sec:dead_end_inspection}).
      The heatmap color of the \ac{ltvmap} segments shows surface coverage of the individual segments, with purple signifying complete coverage. The colors of the exploration viewpoints signify their $l(\xi \in V_{\textrm{exp}})$ value, with white having a value equal to 1 and black being 0.
      Image (a) shows the \ac{ltvmap} sent by UAV1 after returning to communication range with the base station.
     This map is given to UAV2, which then launches and chooses to explore the nearest unexplored frontier in the map of UAV1.
      Image (b) shows the \ac{ltvmap} sent by UAV2 when it is returning.
      Image (c) then shows how the maps are co-updated onboard UAV3, which launches after receiving the \ac{ltvmap} from UAV2.
     The only non-explored viewpoint remaining is in the top part of the image.
      Image (d) shows the maps received by the base station from all three UAVs at the end of the mission with no unexplored viewpoints remaining.
      }
    \end{figure}
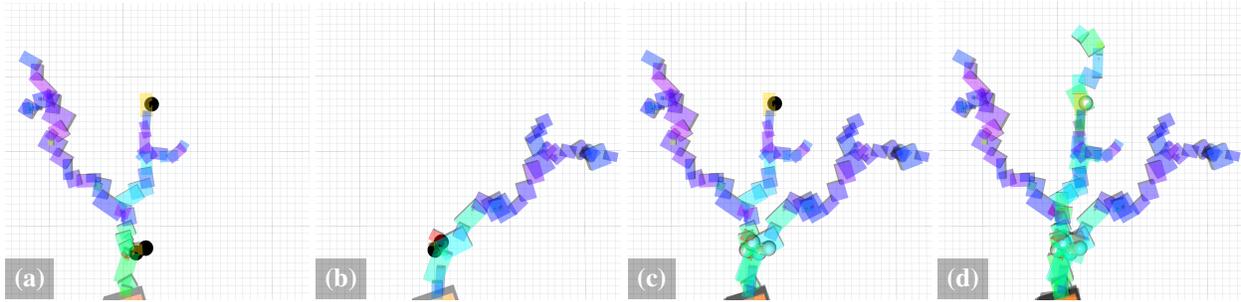

    The updating is done so that the frontier viewpoints, sent along with each LTVMap, which fall into explored space in other LTVMaps, are blocked. 
    This is difficult to do in a deterministic manner due to map drift and other inaccuracies. Therefore, each frontier viewpoint in any LTVMap is assigned a likelihood $l(\xi \in V_{\textrm{exp}})$ to represent how likely it is that the viewpoint has already been visited by any other UAV.
    The $l(\xi \in V_{\textrm{exp}})$ of any viewpoint is the maximum of a function describing the likelihood that the point lies in a given segment's bounding box, computed over all segments of all the other received LTVMaps.
    This likelihood function can be selected arbitrarily; for our approach, we selected a function, which is equal to 0 outside of the segment's bounding box, and grows linearly to 1 the closer it is to the center of the bounding box.
    The updates of these $l(\xi \in V_{\textrm{exp}})$ values for a three \ac{uav} mission can be seen in~\autoref{fig:use_of_received_maps}.

    For a frontier viewpoint $\xi_L$ in the UAV's local map, which has $l(\xi_L \in V_{\textrm{exp}}) > 0$, the reward function changes into
    \begin{equation}
      R(\xi_{\textrm{UAV}}, \xi_L, \mathbb{M}) = l(\xi_L \in V_{\textrm{exp}})R_R (\xi_{\textrm{UAV}}, \xi_L, \mathbb{M}) + (1 - l(\xi_L \in V_{\textrm{exp}})) R_L (\xi_{\textrm{UAV}}, \xi_L),
    \end{equation}
    where $R_L$ is the reward function defined by the employed single-\ac{uav} search strategy described in~\autoref{sec:single_uav_goal_eval}. This does not take into account any information from other UAVs. $R_R$ is a reward function which considers other frontiers in received LTVMaps that could be reachable through $\xi_L$.
    This function is defined as

    \begin{equation}
      \label{eq:coop_planning}
      R_{R}(\xi_{\textrm{UAV}}, \xi_L, \mathbb{M}) = \max_{M \in \mathbb{M}} \max_{\xi_R \in M} {I(\xi_R) - D(\xi_{\textrm{UAV}}, \xi_L)} - \frac{D_R(\xi_L, {\xi_{R}}, \sigma_{ML,M})}{1- l(\xi_R \in V_{\textrm{exp}} )},
    \end{equation}
    where $\mathbb{M}$ is the set of all received LTVMaps, and $\sigma_{ML,M}$ is the most likely segment that $\xi_L$ belongs to in a map $M$.
    The function $D_R$ is a special path cost function computed as a sum of Euclidean distances of segment centers in a given map, spanning from $\xi_L$, through the center of $\sigma_{ML,M}$, and towards a given frontier viewpoint $\xi_R$.
    The value of $D_R$ is also scaled by a user-defined parameter. This is done so as to increase the cost of viewpoints in received maps as there is more uncertainty about the path to these viewpoints.
    The division by $1-l(\xi_R \in V_{\textrm{exp}})$ serves to gradually decrease the reward of exploring the viewpoint up to $-\infty$ when the viewpoint was surely explored by another \ac{uav}.
    Computation of this reward function is illustrated in~\autoref{fig:coop_schematic}.

    The percentage of covered surfaces inside segments received in the LTVMap is used for blocking the surface coverage viewpoints in segments, where the percentage is above a user-defined threshold.
    The segments with low surface coverage could be used as additional goals in a similar manner as shared frontiers in~\autoref{fig:coop_schematic}. However, for simplicity, this was not implemented.

    \subsection{Autonomy robustness enhancements}
    \label{sec:autonomy_robustness}
    One important problem is that in the case of dark and non-reflective surfaces (common in the DARPA SubT Finals course) the LiDAR beam does not return with enough energy.
    Such surfaces will not be marked as occupied and essentially become permanent frontiers, which means that some informative viewpoints, as defined in~\autoref{sec:single_uav_goal_eval}, are non-informative.
    To solve this, the UAV builds a map of visited positions. 
    With time spent near a visited position, we linearly decrease the value of nearby viewpoints.
    After some time, the sampling is blocked near those positions completely. 

    Another problem arising is due to highly dynamic obstacles in the occupancy map, such as other robots, fog, or very narrow corridors where the discretization of occupancy can oscillate. As such, the reachability of a given viewpoint can oscillate. 
    This was solved by putting a timeout on trying to reach a given viewpoint and was triggered if the \ac{uav} did not get closer to the goal within a defined time.
    After this timeout, an area around the viewpoint is blocked until the end of the mission, or until a manual reset by the operator.
    This approach may cause the \ac{uav} to block some goals that are only temporarily blocked by another robot in narrow passages, but it was deemed preferable rather than having the \ac{uav} permanently oscillate in such passages.

    The autonomy system can be easily controlled by operator commands (described in~\autoref{sec:operator_commands}) which can block viewpoints in a set cylinder in space, force the \ac{uav} to explore towards some goal, or simply move to a given position and stay there.
    In this way, problematic situations not covered by our solution, such as organizing multiple robots in a tight corridor, can be resolved by the operator.



    \section{Path planning, trajectory generation and tracking}
    \label{sec:planning}

    Planning collision-free paths and generating dynamically feasible trajectories is another vital component of the presented \ac{uav} system operating in a constrained environment.
    The sequence of waypoints that efficiently guides the \ac{uav} through the environment is produced by the long-distance navigation module, described in~\autoref{sec:navigation}. 
    Given the navigation waypoints, a computationally undemanding multi-stage approach is applied to obtain a trajectory lying at a safe distance from obstacles, while also respecting dynamic constraints and minimizing the time of trajectory following.
    In particular, the solution can be divided into three modules: path planning to obtain the local reference path, path processing to increase the safety margin of the path, and the trajectory generation to obtain a time-parametrized trajectory respecting the dynamic constraints of the \ac{uav}. 

    \subsection{Safety-aware long-distance navigation}
    \label{sec:long_distance_navigation}
    When a goal, or a sequence of goals, is set to the navigation stack, the long-distance navigation module computes a path through the SphereMap, optimal according to~\autoref{eq:spheremap_path_cost}.
    The module then keeps this path and utilizes the trajectory planning and tracking modules to follow it.
    This is done simply by a ``carrot and stick" approach, where the trajectory planning module is given a near point (approx. \SI{20}{\meter} away from the \ac{uav} at maximum, to keep planning time short) on the path. This point is then slid across the path towards the goal.

    If the trajectory planning and tracking modules cannot advance along the SphereMap path for a specified amount of time, which can be caused by a dynamic obstacle such as a rockfall, fog, or another robot, the SphereMap path following is stopped and an unreachability flag is raised. 
    The \ac{uav} then chooses a different goal or tries to find a new path to the same goal based on the current state of mission control.

    When the search planning requires the \ac{uav} to fly through multiple nearby viewpoints, such as when covering the surfaces in a room with cameras or when visiting multiple viewpoints while traveling and using the VPE strategy described in~\autoref{sec:viewpoint_path_enhancement}, the local planning is instead given a sequence of viewpoints. It then plans a trajectory that reaches each up until a maximum distance in order to keep planning time short.
    Thus, the output of this module is always a sequence of one or more waypoints, which may or may not require heading alignment, and through which the local path planning module should find a path in a short time, which we can control by changing the look-ahead distance.

    \label{sec:navigation}

      \subsection{Local path planning}
      
      The grid-based path planner coupled with iterative path processing was adopted from~\cite{kratky2021exploration} to obtain the primary reference path. 
      The proposed approach presents a path planning and processing algorithm, which is based on the traditional A* algorithm applied on a voxel grid with several modifications to decrease the computational demands.
      The first modification lies in avoiding the computationally demanding preprocessing of the map representation (e.g., obstacle dilation by Euclidean distance field), which often requires more time than the actual planning on the grid. 
      This holds true especially for shorter direct paths that leave a significant portion of the previously processed environment unexploited. 
      For this reason, the presented approach builds a $k$-d tree representation of the environment which is then used to conclude the feasibility of particular cells, based on their distance to the nearest obstacle.
      As a result, the computational demands are partially moved from the preprocessing phase to the actual planning phase.
      This approach is particularly efficient in the case of paths that do not require exploiting a significant part of the environment.
      The second important modification is applying node pruning, similar to the jump point search algorithm.  
      This modification helps to decrease the number of unnecessarily expanded nodes. As such, it lowers the computational time required for obtaining the solution. 
      A detailed analysis of the influence of particular modifications on the performance of the planning algorithm is provided in~\cite{kratky2021exploration}. 

      To allow the generated paths to lead through narrow passages, the limits on safety distance are set to the dimension of the narrowest opening that is supposed to be safely traversable by the \ac{uav}. 
      However, setting this distance to a value that ensures safety in the event of the maximum possible deviation from the path caused by any external or internal source would lead to the preclusion of entering narrow passages of the environment. 
      On the contrary, setting this distance to a minimum value without considering safety margins would increase the probability of collision along the whole path. 
      To balance the traversability and safety of the generated path, the minimum required \ac{uav}-obstacle distance applied in the planning process is set to the lowest traversability limit, and iterative path post-processing is applied to increase the \ac{uav}-obstacle distance in wider parts of the environment. 
      The employed post-processing algorithm proposed in~\cite{kratky2021exploration} iteratively shifts the path towards the free part of the environment, while continually maintaining the path's connectivity.
      As such, this anytime algorithm increases the average \ac{uav}-obstacle distance throughout the flight, which significantly improves the reliability of the navigation with respect to imprecisions in the reference trajectory tracking. 

      The generated path is periodically replanned at a rate of \SI{0.5}{\hertz} to exploit the newly explored areas of the environment and handle dynamic obstacles.
      The continuous path following is achieved by using the predicted reference generated by the MPC tracker~\cite{baca2018model} to identify the starting position for the planner at time $T_s$ in the future.  
      Apart from the periodic replanning, the planning is also triggered by the detection of a potential collision on the prediction horizon of the trajectory reference produced by the MPC tracker.
      The potential collisions are checked at a rate of \SI{5}{\hertz} by comparing the distance of particular transition points of the predicted trajectory to the nearest obstacle in the most recent map of the environment.
      Depending on the time left to the time instant of a potential collision, the \ac{uav} is either requested to perform a stopping maneuver or to trigger replanning with the most up-to-date map.

      \subsection{Trajectory generation}

      The path generated by the path planning pipeline is a series of waypoints, each consisting of a 3D position and heading.
      A trajectory (a series of dense time-parameterized waypoints) is generated for each new path, so that the motion of the \ac{uav} satisfies translational dynamics and dynamic constraints up to the 4th derivative of position.
      The trajectory generation system is based on the polynomial trajectory generation approach~\cite{richter2016polynomial, burri2015realtime}, but it was significantly extended to perform in a constrained, real-world environment~\cite{baca2021mrs}.
      This approach was modified to minimize the total flight time while still satisfying the dynamic constraints.
      Furthermore, an iterative sub-sectioning algorithm was added to force the resulting trajectory into a feasible corridor along the original path.
      Moreover, a fallback solver was added to cope with invalid QP solver results caused by numerical instabilities.
      Finally, a dynamic initialization mechanism and a time-outing system were added to cope with the non-zero trajectory generation and path planning computation times.
      Even though the path planning and the trajectory generation can last for several hundreds of milliseconds, the resulting trajectory always smoothly connects to the currently tracked trajectory. Therefore, no undesired motion of the \ac{uav} is produced.
      The updated trajectory generation approach was released and is maintained as part of the MRS UAV System~\cite{baca2021mrs}.

      \subsection{Trajectory tracking and feedback control}

      The low-level guidance of the \ac{uav} is provided by a universal \ac{uav} control system, as developed by the authors of~\cite{baca2021mrs}.
      The onboard control system supports modular execution of \ac{uav} reference generators, feedback controllers, and state estimators.
      During the SubT Finals, the system exclusively utilized the geometric tracking control on \emph{SE(3)}~\cite{lee2010geometric} to follow the desired states generated by the MPC Tracker~\cite{baca2018model}.


    \section{Artifact detection, localization and reporting}
    \label{sec:object_detection}

\begin{figure}[htb]
      \centering
      \includegraphics[width=0.8\textwidth]{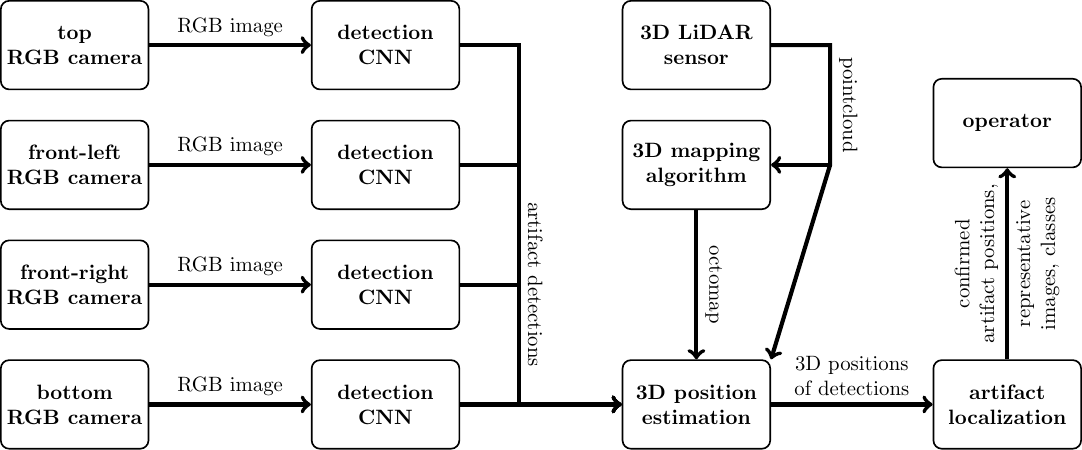}
      \caption{%
        Schematic of the artifact detection and localization pipeline.%
      }
      \label{fig:detection_schematic}
    \end{figure}

    Objects of interest (artifacts) in the explored area are detected visually using a \ac{cnn} that processes images from several onboard RGB cameras covering the frontal, top, and bottom sectors of the \ac{uav}.
    The \ac{cnn} detector is trained on our manually-labeled dataset and outputs predicted bounding boxes and corresponding classes of the artifacts in the input images.
    To estimate the 3D positions of the detections, we have leveraged the onboard 3D \ac{lidar} sensor and the mapping algorithm described in~\autoref{sec:mapping}.
    These positions are processed by an artifact localization filter based on our previous work~\cite{vrba_ral2019}, which fuses the information over time to filter out sporadic false positives and improve the localization precision.
    The artifact detection, localization, and filtering pipeline is illustrated in~\autoref{fig:detection_schematic}.

    
    \subsection{Artifact detection}

    The artifact detection is executed in parallel on image streams from all cameras at the same time, which would require a dedicated \ac{gpu} onboard the \ac{uav}.
    Therefore, we have chosen the lightweight MobileNetV2 \ac{cnn}~\cite{mobilenet}, in order to achieve a high detection rate and keep the load on the onboard computer as low as possible.

    The \ac{cnn} is running on the Intel UHD \ac{gpu} that is integrated within the onboard \ac{cpu} of the \ac{uav}.
    The integrated Intel \ac{gpu} interfaces with our pipeline using the OpenVino\footnote{\href{https://docs.openvino.ai/latest/index.html}{\texttt{docs.openvino.ai/latest/index.html}}} framework.
    The OpenVino framework together with the Intel \ac{gpu} achieves more than \SI{5}{\hertz} detection rate on 4 cameras in parallel but due to fixed resource allocation, we are locking the camera rates to \SI{5}{\hertz}.
    This artificial throttling of the detection rate avoids issues when the integrated GPU locks the memory resources for the \ac{cpu}, which might lead to lag in the control pipeline.

    \begin{table}
    \setlength{\tabcolsep}{3pt}
    \centering
    \footnotesize
    \begin{tabular}{l c c c c c c c c c c c}
    \toprule
      \tablehdg{Input} & $384^2$×3 & $1122^2$×32 & $1122^2$×16 & $562^2$×24 & $282^2$×32 & $142^2$×64 & $142^2$×96 & $72^2$×160 & $72^2$×320 & $72^2$×1280 & 1×1×1280 \\
      \tablehdg{Type} & conv2d & bottleneck & bottleneck & bottleneck & bottleneck & bottleneck & bottleneck & bottleneck & conv2d 1×1  & avgpool 7×7 & conv2d 11 \\
      \tablehdg{$n$} & 1 & 1 & 2 & 3 & 4 & 3 & 3 & 1 & 1 & 1 & - \\
    \bottomrule
    \end{tabular}
    \caption{The architecture of MobileNetV2.
      Each column of the table describes a sequence of 1 or more identical layers, repeated $n$ times.}
    \label{table:mobilenet}
    \end{table}


    The MobileNetV2 base model (see~\autoref{table:mobilenet}) is modified for training using the OpenVino open-source tools.
    The evaluation of the model is based on the mean average precision metric (mAP) and recall.
    The mAP metric is a standard metric for object detection models since it provides information about how accurate the prediction is.
    Recall provides an understanding what is the ratio between true positive predictions and the total number of positive samples in the dataset.

\begin{figure}
      \centering
    \newcommand{\xcap}{0.95em}
    \newcommand{\ycap}{0.8em}
    \newcommand{\fillopa}{0.3}
    \newcommand{\imheight}{10em}
    \centering
    \begin{tikzpicture}
      \node[anchor=south west,inner sep=0] (b) at (0,0) {\adjincludegraphics[height=\imheight,trim={{0.00\width} {0.0\height} {0.00\width} {0.00\height}},clip]{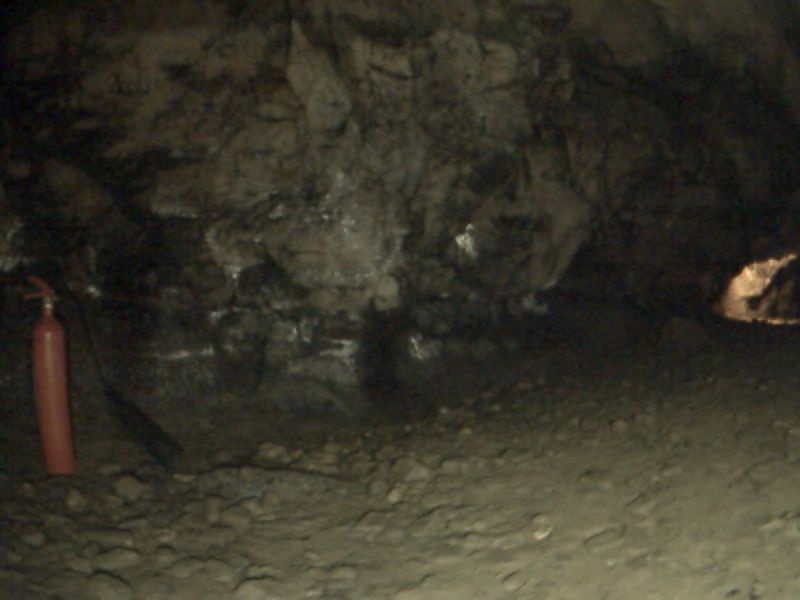}};%
    \begin{scope}[x={(b.south east)},y={(b.north west)}]
  \node[fill=black, fill opacity=\fillopa, text=white, text opacity=1.0] at (\xcap, \ycap) {\textbf{(a)}};
      \end{scope}
    \end{tikzpicture}%
    \begin{tikzpicture}
      \node[anchor=south west,inner sep=0] (b) at (0,0) {\adjincludegraphics[height=\imheight,trim={{0.00\width} {0.0\height} {0.10\width} {0.00\height}},clip]{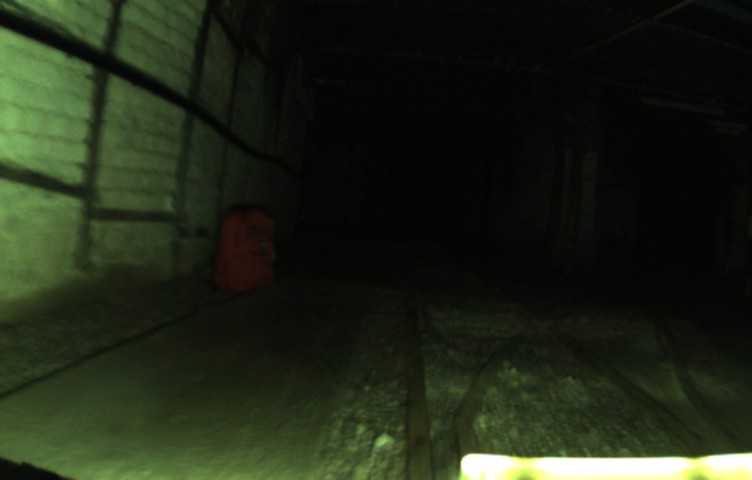}};%
    \begin{scope}[x={(b.south east)},y={(b.north west)}]
  \node[fill=black, fill opacity=\fillopa, text=white, text opacity=1.0] at (\xcap, \ycap) {\textbf{(b)}};
      \end{scope}
    \end{tikzpicture}%
    \begin{tikzpicture}
      \node[anchor=south west,inner sep=0] (b) at (0,0) {\adjincludegraphics[height=\imheight,trim={{0.00\width} {0.0\height} {0.10\width} {0.00\height}},clip]{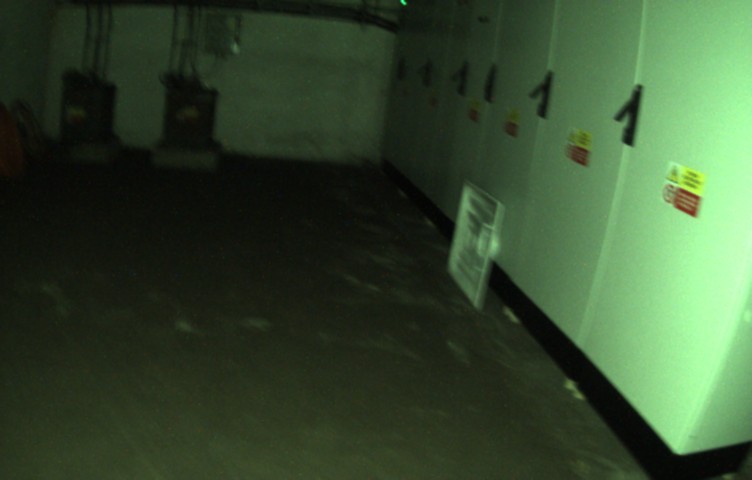}};%
    \begin{scope}[x={(b.south east)},y={(b.north west)}]
  \node[fill=black, fill opacity=\fillopa, text=white, text opacity=1.0] at (\xcap, \ycap) {\textbf{(c)}};
      \end{scope}
    \end{tikzpicture}%
      \caption{Training images containing artifacts captured by the onboard cameras in cave (a), tunnel (b), and urban (c) environments.}
      \label{fig:pics_exa}
    \end{figure}
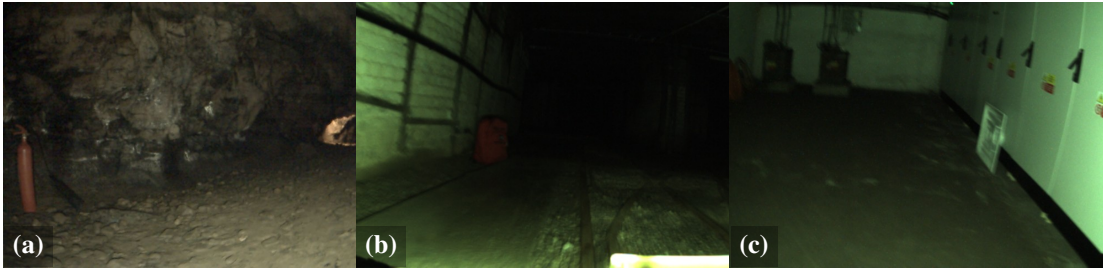

    The main challenge for the model is to adapt to different domains --- mine, urban, and cave environments have different lighting and backgrounds (see~\autoref{fig:pics_exa}), which affect the detection performance.
    Moreover, the angle from which the images were taken is different as part of the images in the dataset were taken by ground vehicles and the rest by UAVs.
    As the model was trained only on part of the dataset, we had to train it incrementally whenever data from a new type of environment or camera angle was gathered to ensure all cases were represented uniformly in the training data.

    For training the model on a growing dataset, we used a variety of learning schedulers from the MMdetection toolbox~\cite{mmdetection}.
    The Cosine scheduler designed by~\cite{cosineLR} is used for warm-restarts of the training pipeline to overcome the loss of learned features.
    The main challenge of transfer learning is to overcome the loss of learned distribution on the previous dataset when training the model on the new dataset (in this case the new dataset is a combination of the previous dataset and newly collected data).

    In our experience, different learning rate schedulers should be used depending on the size of newly added data:
    \begin{itemize}
      \item \textit{Cosine scheduler} is used during clean model training on the initial dataset.
      \item \textit{Cyclic scheduler} \cite{cyclicLR} is used when the size of new data is more than \SI{15}{\percent} of the size of the initial dataset.
      \item \textit{Step decal scheduler} is used when a small portion of data is added.
    \end{itemize}

    This method resulted in a score of \SI{49.1}{\percent} mAP on the whole dataset. 
    Such a value is acceptable for the current solution since it is a trade-off between accuracy and speed of detection.

    The dataset was collected using the off-the-shelf objects that were specified by the organizers, see~\autoref{fig:artifacts}.
    The data has been recorded from the onboard cameras on the UAVs and UGVs, in particular:
    \begin{itemize}
      \item Intel RealSense D435
      \item Basler Dart daA1600 
      \item Bluefox MLC200w
    \end{itemize}
    The Basler cameras do not have an IR filter installed to maximize the amount of captured captured light.
    Altogether the dataset has 37820 labeled images, sometimes with multiple objects in one frame, an example of images from the dataset is shown in~\autoref{fig:pics_exa}.

    We publish the labeled detection datasets that were used for training of the neural network at \href{https://github.com/ctu-mrs/vision_datasets}{\texttt{github.com/ctu-mrs/vision\_datasets}}.
    In addition we also publish the tools to convert it into PASCAL VOC or COCO formats for immediate usage on most of the open-source models.

    \begin{figure}
      \centering
    
      \begin{subfigure}[t]{0.23\textwidth}
        \centering
        \includegraphics[width=\textwidth]{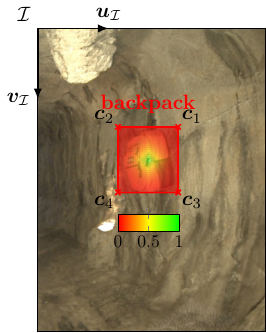}
        \caption{%
          Example of an artifact detection with an overlay visualization of the sample weighting function $f_{\text{w}}$.
        }
        \label{fig:det_pos_weights}
      \end{subfigure}%
      ~
      \begin{subfigure}[t]{0.75\textwidth}
        \centering
        \includegraphics[width=\textwidth]{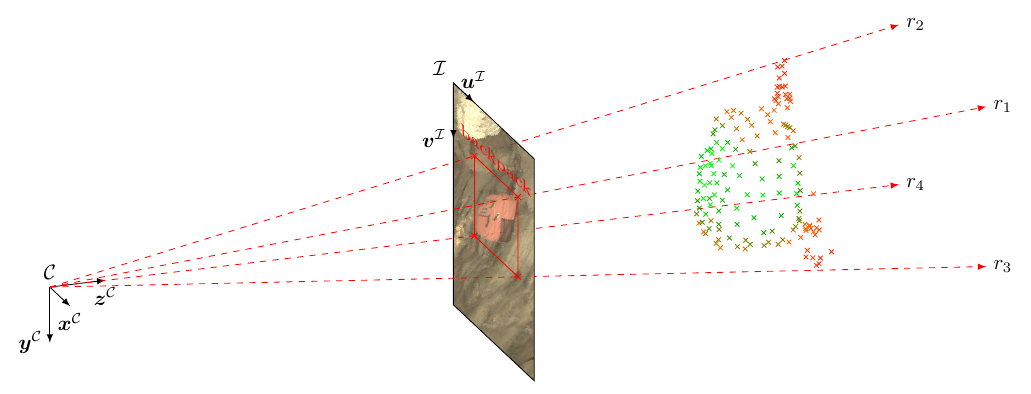}
        \caption{%
          Model of the camera and the point cloud-based sampling method. Rays $r_1, r_2, r_3, r_4$ are projections of $c_1, c_2, c_3, c_4$, respectively. Only the points within the area defined by these rays are selected. The selected points are colored based on their weight. Non-selected points are not drawn for clarity.
        }
        \label{fig:det_pos}
      \end{subfigure}%
    
      \caption{
        Illustration of the point sampling for 3D position estimation of detected artifacts with an example detection of a backpack.
      }
    \end{figure}
    

    
    \subsection{Estimation of 3D position of detections}
    Positions of the detected objects are estimated using data from the onboard \ac{lidar} sensor and the mapping algorithm.
    Each detection is represented by four corner points $\pnt{c}_1, \pnt{c}_2, \pnt{c}_3, \pnt{c}_4$ of its bounding rectangle in the image plane of the corresponding camera, as estimated by the detector (see~\autoref{fig:det_pos_weights}).
    These points are expressed as undistorted pixel coordinates in the image frame $\mathcal{I}$.
  
    The mathematical projection model of the camera $f_{\text{proj}} : \mathbb{R}^3 \to \mathbb{R}^2$ is assumed to be known.
    In our case, we have used the standard pinhole camera model formulated as
    \begin{equation}
      k \bemat{ u \\ v \\ 1 } = \bemat{ f_u & 0 & u_0 \\ 0 & f_v & v_0 \\ 0 & 0 & 1 } \bemat{ x \\ y \\ z },
    \end{equation}
    where $f_u, f_v, u_0, c_0$ are parameters of the model (focal length and image center), $\bemat{ x, y, z }\tran$ is a 3D point in the camera coordinate frame $\mathcal{C}$, and $u, v$ are distortion-free pixel coordinates in the image frame $\mathcal{I}$, corresponding to the 3D point (see~\autoref{fig:det_pos} for illustration).
    To model the distortion of the real-world camera, we have used a standard radial-tangential polynomial distortion model.
    It is worth noting that the output of $f_{\text{proj}}^{-1}$ is a 3D ray and not a single point, which is represented in the model by the free scaling factor $k \in \mathbb{R}$.
  
    The input \ac{lidar} scan is represented as a set of 3D points $\set{S} = \left\{ \pnt{p}_i \right\}$ expressed in the camera coordinate frame~$\mathcal{C}$.
    The occupancy map is represented using the DenseMap data structure that is described in~\autoref{sec:mapping}, and which provides a raycasting function $f_{\text{raycast}} : \set{R} \to \mathbb{R}^3$ where $\set{R}$ is the set of all 3D rays.
    The function $f_{\text{raycast}}$ returns the point, corresponding to the first intersection of the specified ray with an obstacle in the environment (or nothing if there is no such intersection).
  
    The position of each detected object is estimated from a number of points that are sampled using two methods: a primary one that utilizes the latest available point cloud from the \ac{lidar} and a secondary backup method using the latest DenseMap estimated by the mapping algorithm.
    The primary method is more accurate and less computationally intensive, while the secondary method ensures that 3D positions of artifacts lying outside of the \ac{fov} of the \ac{lidar} scan can still be estimated.
    For each sampled point $\pnt{s}_i \in \set{S}$, its weight $w_i$ is calculated.
    The position estimate $\pnt{d}$ and its corresponding uncertainty covariance matrix $\mat{Q}_{\pnt{d}}$ are obtained as a weighted mean of the sampled points:
    \begin{align}
      \pnt{d} = \sum_{i = 1}^{|\set{S}|} \pnt{s}_i w_i, && \mat{Q}_{\pnt{d}} = \frac{1}{1 - \sum_{i = 1}^{|\set{S}|} w_i^2} \sum_{i = 1}^{|\set{S}|} w_i \left( \pnt{s}_i - \pnt{d} \right)\left( \pnt{s}_i - \pnt{d} \right)\tran,
    \end{align}
    where $\set{S}$ is the set of sampled points and the weights $w_i$ are normalized so that $\sum_{i=1}^{\set{S}}w_i = 1$.
  
    The weight of a point $\pnt{s}$ is obtained based on the distance of its reprojection to the image coordinates $\pnt{s}' = \bemat{s_u, s_v}\tran = f_{\text{proj}}\left( \pnt{s} \right)$ from the center of the detection's bounding box $\pnt{c}_0 = \bemat{c_u, c_v}\tran$ using the function
    \begin{equation}
      f_{\text{w}}\left(\pnt{s}', \pnt{c}_0\right) = \left( 1 - \frac{2\abs{s_u - c_u}}{w_{\text{bb}}} \right)^2 \left( 1 - \frac{2\abs{s_v - c_v}}{h_{\text{bb}}} \right)^2,
    \end{equation}
    where $w_{\text{bb}}, h_{\text{bb}}$ are the width and height of the bounding box, respectively.
    The weighting function serves to suppress points further from the center of the bounding box.
    This is based on our empirical observation that the center provides the most reliable estimate of the detected object's position, while the bounding box's corners typically correspond to the background and not the object, as illustrated in~\autoref{fig:det_pos_weights}.
    The whole 3D position estimation algorithm is presented in~\autoref{alg:posest}.
    The \texttt{sampleRectangle} routine used in~\autoref{alg:posest} is described in~\autoref{alg:sampleRectangle}.
  
    The estimated positions and the corresponding covariance matrices serve as an input to the \textit{artifact localization filter} described in the next section (refer to~\autoref{fig:detection_schematic}).
    To avoid bias and numerical singularities in the filter, some special cases of the covariance calculation have to be handled.
    Namely, these are:
    \begin{enumerate}
        \item \textit{All extracted points lie on a plane.}
        This happens, e.g. when all the cast rays of the secondary position estimation method intersect the same voxel of the DenseMap.
        The covariance matrix is then singular, which causes numerical problems with integrating the measurement.
  
        \item \textit{All extracted points are too close to each other.}
        This typically happens when the detected object is too far or too small.
        The covariance matrix's eigenvalues are then too small, biasing the fused position estimate of the artifact.
    \end{enumerate}
    To avoid these problems, the estimated covariance matrix is rescaled, so that all eigenvalues conform to a specified minimal threshold before being processed by the artifact localization filter.
  
  
  \begin{algorithm}
    \algdef{SE}[SUBALG]{Indent}{EndIndent}{}{\algorithmicend\ }%
    \algtext*{Indent}
    \algtext*{EndIndent}
  
    \algnewcommand\AND{\textbf{and}~}
    \algnewcommand\Not{\textbf{not}~}
    \algnewcommand\Or{\textbf{or}~}
    \algnewcommand\Input{\State{\textbf{Input:~}}}%
    \algnewcommand\Output{\State{\textbf{Output:~}}}%
    \algnewcommand\Parameters{\State{\textbf{Parameters:~}}}%
    \algnewcommand\Begin{\State\textbf{Begin:~}}%
    \algnewcommand{\LineComment}[1]{\State \(\triangleright\) #1}
  
    \caption{Algorithm for the estimation of a detection's position and covariance.}\label{alg:posest}
  
    \begin{algorithmic}[1]
      \footnotesize
  
      \Input
      \Indent
  
      \State $\set{D} = \left\{ \pnt{c}_1, \pnt{c}_2, \pnt{c}_3, \pnt{c}_4 \right\},~\pnt{c}_i \in \mathbb{R}^2 $
      \Comment{undistorted coordinates of the detection's bounding box}
  
      \State $f_{\text{proj}} : \mathbb{R}^2 \to \set{R} $
      \Comment{the projection model of the camera}
  
      \State $\set{P} = \left\{ \pnt{p}_1, \pnt{p}_2, \dots, \pnt{p}_{|\set{P}|} \right\},~\pnt{p}_i \in \mathbb{R}^3 $
      \Comment{the latest point cloud from the \ac{lidar}}
  
      \State $f_{\text{raycast}} : \set{R} \to \mathbb{R}^3 $
      \Comment{the raycasting function of the occupancy map}
  
      \State $n_{\text{desired}} \in \mathbb{N}$
      \Comment{the desired number of sampled points}
  
      \EndIndent
  
      \Output
      \Indent
  
      \State $\pnt{d} \in \mathbb{R}^3$
      \Comment{estimated position of the detection}
  
      \State $\mat{Q}_{\pnt{d}} \in \mathbb{R}^{3\times 3}$
      \Comment{covariance matrix of the position estimate}
  
      \EndIndent
  
      \Begin
      \Indent
  
      \LineComment{First, the desired number of points is sampled using the primary and secondary methods.}
  
      \State $r_1 \coloneqq f_{\text{proj}}^{-1}\left(\pnt{c}_1\right),~r_2 \coloneqq f_{\text{proj}}^{-1}\left(\pnt{c}_2\right),~r_3 \coloneqq f_{\text{proj}}^{-1}\left(\pnt{c}_3\right),~r_4 \coloneqq f_{\text{proj}}^{-1}\left(\pnt{c}_4\right)$
      \Comment{project the corners of the bounding box to 3D rays}
  
      \State $\set{S}_1 \coloneqq \left\{ \pnt{p}_i \in \set{P} \mid \pnt{p}_i \text{ within the area defined by edges } r_1, r_2, r_3, r_4 \right\}$
  
      \State $\set{S}_2 \coloneqq \text{sampleRectangle}\left( \left\{ \pnt{c}_1, \pnt{c}_2, \pnt{c}_3, \pnt{c}_4 \right\}, n_{\text{desired}} - |\set{S}_1|, f_{\text{proj}}, f_{\text{raycast}} \right)$
      \Comment{sample any remaining points from the occupancy map}
  
      \State $\set{S} \coloneqq \set{S}_1 \cup \set{S}_2$
  
      \LineComment{Then, the weight of each sampled point is calculated using the weighting function $f_{\text{w}}$.}
  
      \State $\vec{c}_0 \coloneqq \text{mean}\left( \pnt{c}_1, \pnt{c}_2, \pnt{c}_3, \pnt{c}_4 \right)$
      \Comment{calculate the center of the bounding box}
  
      \For{each $ \pnt{s}_i \in \set{S} $}
  
        \State $\vec{s}_i' \coloneqq f_{\text{proj}}\left( \pnt{s}_i \right) $
        \Comment{project the point back to the image frame $\mathcal{I}$}
  
        \State $w_i \coloneqq f_{\text{w}}\left( \vec{s}_i', \vec{c}_0 \right)$
        \Comment{calculate its weight}
  
      \EndFor
  
      \LineComment{Finally, the position and its uncertainty are calculated as a weighted mean and covariance and returned.}
  
      \State $\pnt{d} \coloneqq \sum_{i = 1}^{|\set{S}|} \pnt{s}_i w_i $
  
      \State $\mat{Q}_{\pnt{d}} = \frac{1}{1 - \sum_{i = 1}^{|\set{S}|} w_i^2} \sum_{i = 1}^{|\set{S}|} w_i \left( \pnt{s}_i - \pnt{d} \right)\left( \pnt{s}_i - \pnt{d} \right)\tran$
  
      \State\textbf{return~} $\pnt{d},~\mat{Q}_{\pnt{d}}$
  
      \EndIndent
  
    \end{algorithmic}
  
  \end{algorithm}
  
  \begin{algorithm}
    \algdef{SE}[SUBALG]{Indent}{EndIndent}{}{\algorithmicend\ }%
    \algtext*{Indent}
    \algtext*{EndIndent}
  
    \algnewcommand\AND{\textbf{and}~}
    \algnewcommand\Not{\textbf{not}~}
    \algnewcommand\Or{\textbf{or}~}
    \algnewcommand\Input{\State{\textbf{Input:~}}}%
    \algnewcommand\Output{\State{\textbf{Output:~}}}%
    \algnewcommand\Parameters{\State{\textbf{Parameters:~}}}%
    \algnewcommand\Begin{\State\textbf{Begin:~}}%
    \algnewcommand{\LineComment}[1]{\State \(\triangleright\) #1}
  
    \caption{The \texttt{sampleRectangle} routine for sampling a number of 3D points from the occupancy map.}\label{alg:sampleRectangle}
  
    \begin{algorithmic}[1]
      \footnotesize
  
      \LineComment{This routine samples points within a rectangle in the image plane $\mathcal{I}$ by raycasting pixels on inscribed ellipses with an increasing radius.}
      \State\textbf{Routine } $\text{sampleRectangle}$\textbf{:}
      \Indent
  
      \Input
      \Indent
  
      \State $\left\{ \pnt{c}_1, \pnt{c}_2, \pnt{c}_3, \pnt{c}_4 \right\},~\pnt{c}_i \in \mathbb{R}^2 $
      \Comment{corners of the rectangle to be sampled in the image frame $\mathcal{I}$}
  
      \State $n_{\text{remaining}} \in \mathbb{N}$
      \Comment{the desired number of samples}
  
      \State $f_{\text{proj}} : \mathbb{R}^2 \to \set{R} $
      \Comment{the projection model of the camera}
  
      \State $f_{\text{raycast}} : \set{R} \to \mathbb{R}^3 $
      \Comment{the raycasting function of the occupancy map}
  
      \EndIndent
  
      \Output
      \Indent
  
      \State $\set{S} = \left\{ \pnt{s}_i \right\}$
      \Comment{a set of sampled points in the image frame $\mathcal{I}$ such that $|\set{S}| \le n_{\text{remaining}}$}
  
      \EndIndent
  
      \Parameters
      \Indent
  
      \State $n_r \in \mathbb{N}$, $n_{\alpha} \in \mathbb{N}$
      \Comment{number of radial sampling steps and number of circumferential steps per unit circumference}
  
      \EndIndent
  
      \Begin
      \Indent
  
      \State $ w \coloneqq c_{1,u} - c_{3,u},~h \coloneqq c_{1,v} - c_{3,v} $
      \Comment{calculate the width and height of the rectangle}
  
      \State $ r_{\text{step}} \coloneqq 1/n_{r} $
  
      \For{$ r \in \left\{ 0, r_{\text{step}}, 2r_{\text{step}}, \dots, 1 \right\} $}
  
        \State $ \alpha_{\text{step}} \coloneqq r / n_{\alpha} $
  
        \State $ {}_\Delta\alpha \coloneqq u,~u \sim \mathcal{U}\left(-\pi, \pi\right) $
        \Comment{generate a random angular offset to avoid biasing some directions}
  
        \For{$ \alpha \in \left\{ 0, \alpha_{\text{step}}, 2\alpha_{\text{step}}, \dots, 2\pi \right\} $}
  
          \State $\pnt{s}' \coloneqq \bemat{ w r\cos\left( \alpha + {}_\Delta\alpha \right)/2,~h r\sin\left( \alpha + {}_\Delta\alpha \right)/2 }\tran$
          \Comment{calculate a sample point on an ellipse}
  
          \State $r \coloneqq f_{\text{proj}}\left( \pnt{s}' \right)$
          \Comment{project the point to a 3D ray}
  
          \State $\set{S} \coloneqq \set{S} \cup f_{\text{raycast}}\left( r \right)$
          \Comment{find an intersection of the ray with an obstacle and add it to $\set{S}$}
  
          \If{$|\set{S}| = n_{\text{remaining}}$}
  
            \State\textbf{return~} $\set{S}$
  
          \EndIf
  
        \EndFor
  
      \EndFor
  
      \State\textbf{return~} $\set{S}$
  
      \EndIndent
      \EndIndent
  
    \end{algorithmic}
  
  \end{algorithm}
  
    

    
    \subsection{Artifact localization filter}
    \label{sec:artifact_localization}
    
\begin{figure}
      \centering
    
      \begin{subfigure}[t]{0.44\textwidth}
        \centering
        \includegraphics[width=\textwidth]{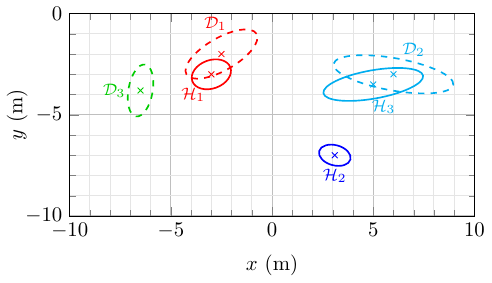}
        \caption{%
          Situation before the update step.
          The detections $\dete_1$ and $\dete_2$ are associated to the hypotheses $\hyp_1$ and $\hyp_3$, respectively.
          The detection $\dete_3$ is not associated to any hypothesis. The hypothesis $\hyp_2$ has no detection associated.
        }
      \end{subfigure}%
      ~~
      \begin{subfigure}[t]{0.44\textwidth}
        \centering
        \includegraphics[width=\textwidth]{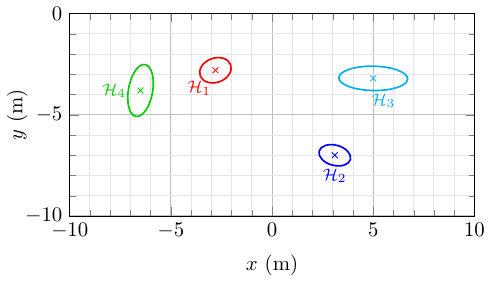}
        \caption{%
          Situation after the update step.
          The detections $\dete_1$ and $\dete_2$ updated the hypotheses $\hyp_1$ and $\hyp_3$, respectively.
          The detection $\dete_3$ initialized a new hypothesis $\hyp_4$ and the hypothesis $\hyp_2$ remained unchanged.
        }
      \end{subfigure}%
    
      \caption{%
        Illustration of one step of the artifact localization filter (a top-down view).
        Hypotheses $\hyp_i$ are shown as covariance ellipsoids with the mean $\hat{\vec{x}}_i$ marked by an `×' symbol.
        Detections $\dete_i$ are represented in the same way using dashed lines.
        Associations between hypotheses and detections are highlighted using color.
      }
      \label{fig:art_filter}
    \end{figure}

    Artifact detections are filtered using an approach based on our previous work, where a multi-target tracking algorithm was employed for detection, localization, and tracking of micro aerial vehicles~\cite{vrba_ral2019}.
    The filtering serves to improve the precision of the artifacts' estimated positions and to reject false positives.
    Only artifacts that are consistently detected multiple times with sufficient confidence are confirmed, and only the confirmed artifacts are then reported to the operator to save the limited communication bandwidth.
    A single step of the algorithm is illustrated in~\autoref{fig:art_filter}.
    
    The filter keeps a set of hypotheses about objects in the environment.
    Each hypothesis $\hyp$ is represented by an estimate of the object's position $\hat{\vec{x}}$, its corresponding covariance matrix $\mat{P}$, and a probability distribution of the object's class $p_{\hyp} : \mathcal{C} \to \interval{0}{1}$, where $\mathcal{C}$ is the set of considered classes.
    For every hypothesis $\hyp$, up to one detection $\dete_\hyp$ is associated according to the rule
    \begin{equation}
      \dete_\hyp = \begin{cases}
        \argmax_{\dete} l\left(\dete \mid \hyp \right), &\text{ if }\max_{\dete} l\left(\dete \mid \hyp \right) > l_{\text{thr}}, \\ 
        \emptyset, &\text{ else,}
      \end{cases} \label{eq:art_assoc}
    \end{equation}
    where $l\left(\dete \mid \hyp\right)$ is the likelihood of observing $\dete$ given that it corresponds to $\hyp$, and $l_{\text{thr}}$ is a likelihood threshold.
    The associated detections are used to update the corresponding hypotheses. The detections that are not associated initialize new hypotheses.
    
    The position estimate $\hat{\vec{x}}$ of a hypothesis $\hyp$ and its covariance $\mat{P}$ are updated using the Kalman filter's update equation and an associated detection $\dete_\hyp$ at time step $t$ as
    \begin{align}
      \mat{K}\tstep{t}   &= \mat{P}\tstep{t} \mat{H}\tran \left( \mat{H} \mat{P}\tstep{t} \mat{H}\tran + \mat{Q}\tstep[\vec{d}]{t} \right)^{-1}, \label{eq:art_kf1}\\
      \hat{\vec{x}}\tstep{t+1} &= \hat{\vec{x}}\tstep{t} + \mat{K}\tstep{t}\left( \vec{d}\tstep{t} - \mat{H}\hat{\vec{x}}\tstep{t} \right), \\
      \mat{P}\tstep{t+1} &= \left( \mat{I} - \mat{K}\tstep{t} \mat{H} \right)\mat{P}\tstep{t},\label{eq:art_kf3}
    \end{align}
    where $\mat{K}\tstep{t}$ is a Kalman gain, $\mat{I}$ is an identity matrix, $\mat{H}$ is an observation matrix (in our case, equal to $\mat{I}$), $\vec{d}\tstep{t}$ and $\mat{Q}\tstep[\vec{d}]{t}$ are the estimated position of $\dete\tstep[\hyp]{t}$ and its corresponding covariance matrix, respectively. 
    The class probability distribution $p_\hyp$ is updated as
    \begin{equation}
      p\tstep[\hyp]{t+1}\left(c\right) = \frac{ n\tstep[\text{dets}]{t} p\tstep[\hyp]{t}\!\left(c\right) + p\tstep[\dete_\hyp]{t}\!\left(c\right) }{ n\tstep[\text{dets}]{t} + 1},
    \end{equation}
    where $c \in \mathcal{C}$ is an object's class and $n\tstep[\text{dets}]{t}$ is the number of detections, associated to $\hyp$ thus far.
    
    Because the artifacts are assumed to be immobile, the Kalman filter's prediction step is not performed, which has the effect that the uncertainty of a hypothesis (represented by $\mat{P}$) can decrease without bounds.
    This can cause the likelihood $l\left(\dete \mid \hyp \right)$ of new measurements corresponding to the same object to be below the association threshold, breaking the association algorithm.
    To avoid this, the covariance matrix $\mat{P}$ is rescaled after each update so that its eigenvalues are larger than a specified minimal value, which enforces a lower bound on the position uncertainty of the hypotheses.

    \subsubsection{Association likelihood}
    To calculate the likelihood $l\left( \dete\tstep{t} \mid \hyp\tstep{t} \right)$ of observing a detection $\dete \equiv \left\{ \vec{d},~ \mat{Q}_{\vec{d}} \right\}$ given that it corresponds to a hypothesis $\hyp = \left\{ \hat{\vec{x}},~ \mat{P} \right\}$ at time step $t$, we use a measurement model
    \begin{align}
      \vec{d}\tstep{t} = \mat{H} \vec{x} + \vec{\xi}\tstep{t}, && \vec{\xi}\tstep{t} \sim \mathcal{N}\left(\vec{0},~\mat{Q}\tstep[\vec{d}]{t}\right), \label{eq:art_meas_model}
    \end{align}
    where $\mat{H}$ is the observation matrix, $\vec{x}$ is a hidden state (the real position of the artifact), $\vec{\xi}\tstep{t}$ is measurement noise, and $\mathcal{N}\left(\vec{0},~\mat{Q}\tstep[\vec{d}]{t}\right)$ denotes the Gaussian probability distribution with zero mean and covariance matrix $\mat{Q}\tstep[\vec{d}]{t}$.
    Using this model, the probability density function of the expected measurement given $\vec{x}$ is
    \begin{align}
      p\left( \vec{d}\tstep{t} \mid \vec{x} \right) = f\left( \vec{d}\tstep{t} \mid \mat{H}\vec{x}, \mat{Q}\tstep[\vec{d}]{t} \right),
    \end{align}
    where $f\left(~\cdot \mid \vec{\mu}, \mat{\Sigma} \right)$ denotes the density function of the Gaussian distribution with mean $\vec{\mu}$ and covariance matrix $\mat{\Sigma}$.

    The Kalman filter described by equations \eqref{eq:art_kf1} to \eqref{eq:art_kf3} can be interpreted as an estimator of the probability density of the hidden state given previous measurements.
    This probability density is represented as a random variable with a Gaussian distribution:
    \begin{equation}
      p\left( \vec{x} \mid \vec{d}\tstep{1}, \dots, \vec{d}\tstep{t} \right) = f\left( \vec{x} \mid \hat{\vec{x}}\tstep{t}, \mat{P}\tstep{t} \right). \label{eq:art_state_model}
    \end{equation}

    The likelihood $l\left(\vec{d}\tstep{t}\right)$ of observing a new measurement $\vec{d}\tstep{t}$ given previous measurements $\vec{d}\tstep{1}, \dots, \vec{d}\tstep{t-1}$ is the value of a probability density function $p\left( \vec{d} \mid \vec{d}\tstep{1}, \dots, \vec{d}\tstep{t-1} \right)$ at $\vec{d}\tstep{t}$.
    By combining equations \eqref{eq:art_meas_model} and \eqref{eq:art_state_model}, the likelihood may be expressed as
    \begin{equation}
      \begin{split}
        l\left(\vec{d}\tstep{t}\right) = p\left( \vec{d}\tstep{t} \mid \vec{d}\tstep{1}, \dots, \vec{d}\tstep{t-1} \right) &= \int p\left( \vec{d}\tstep{t} \mid \vec{x} \right) p\left( \vec{x} \mid \vec{d}\tstep{1}, \dots, \vec{d}\tstep{t-1} \right) d\vec{x}, \\
        &= \int f\left( \vec{d}\tstep{t} \mid \mat{H}\vec{x}, \mat{Q}\tstep[\vec{d}]{t} \right) f\left( \vec{x} \mid \hat{\vec{x}}\tstep{t-1}, \mat{P}\tstep{t-1} \right) d\vec{x}, \\
        &= f\left( \vec{d}\tstep{t} \mid \mat{H}\hat{\vec{x}}\tstep{t-1}, \mat{Q}\tstep[\vec{d}]{t} + \mat{H}\mat{P}\tstep{t-1}\mat{H}\tran \right), \label{eq:art_likelihood}
      \end{split}
    \end{equation}
    which is the value of the probability density function of a Gaussian distribution with mean $\mat{H}\hat{\vec{x}}\tstep{t-1}$ and covariance $\mat{Q}\tstep[\vec{d}]{t} + \mat{H}\mat{P}\tstep{t-1}\mat{H}\tran$ at $\vec{d}\tstep{t}$.
    This expression is used to determine the detection-to-hypothesis association at each step according to equation~\eqref{eq:art_assoc}, as described in the previous section.
    

    \subsection{Arbiter for artifact reporting}
    \label{sec:arbiter_for_artifact_reporting}

\begin{figure}
      \centering
      \adjincludegraphics[width=1.0\textwidth,trim={{0.00\width} {0.0\height} {0.00\width} {0.00\height}},clip]{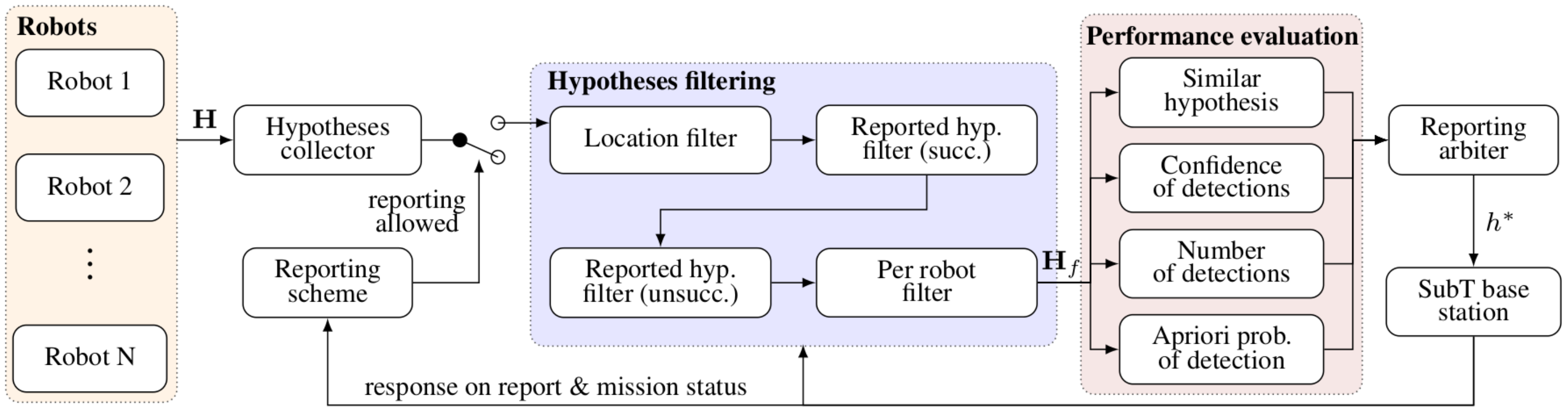}
      \caption{\label{fig:automatic_reporting_scheme}
        Illustration of the automatic reporting process.
      }
    \end{figure}

 In contrast to the system part of the competition, the Virtual Track requires substituting the human operator with an autonomous arbiter for artifact reporting.     
    The main functionality of the autonomous base station resides in collecting the hypotheses from the robots and reporting the location of artifacts. 
    The number of reports in each run is limited and usually lower than the number of hypotheses collected from all robots. Therefore, a subset of hypotheses needs to be chosen so that the expected score is maximized.
    The implemented reporting strategy is based on filtering the collected hypotheses by considering their location and artifact type, followed by evaluating the performance index of particular hypotheses. 
    The entire workflow is illustrated in~\autoref{fig:automatic_reporting_scheme}.

\begin{figure}
      \centering
      \adjincludegraphics[width=0.6\textwidth,trim={{0.00\width} {0.0\height} {0.00\width} {0.00\height}},clip]{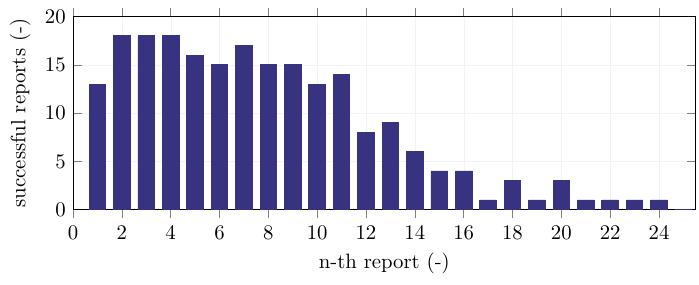}
      \caption{\label{fig:successful_reports_distribution}
        The distribution of successful reports over particular reporting attempts during all runs of the SubT Virtual Track Prize Round. The lower success rate of the first attempt in comparison to later attempts is caused by the early time of the first report, which was allowed \SI{100}{\second} after the start of the mission. By this time, only a single \ac{uav} had already entered the course, and thus the number of available hypotheses to choose from was low.    
      }
    \end{figure}

    The autonomous base station collects the hypotheses from individual robots throughout the entire run.
    The predefined reporting scheme specifies the maximum allowed number of reports at particular time instants of the mission. 
    Most of the reports are saved to the last minutes of the mission when the base station holds most of the information collected from the robots. 
    However, some reports are allowed sooner during the mission to tackle the problem of unreliable communication and prevent a failure to report all hypotheses before the time limit exceeds.  
    When the reporting scheme allows for submitting a report, the collected hypotheses are processed to obtain the best available hypothesis $h^*$ in a set of all collected hypotheses $\mathbf{H}$.
    First, the hypotheses are filtered using information about previous reports, their validity, location, and per robot limits on the number of reports and minimum success rate.
    The final set of filtered hypotheses is obtained as   
    \begin{equation}
      \mathbf{H}_f = \mathbf{H} \setminus \{\mathbf{H}_{\mathrm{area}} \cup \mathbf{H}_{\mathrm{succ}} \cup \mathbf{H}_{\mathrm{unsucc}} \cup \mathbf{H}_{r}\},
    \end{equation}
    where $\mathbf{H}_{\mathrm{area}}$ stands for the hypotheses located outside of the competition course, $\mathbf{H}_{\mathrm{succ}}$ stands for hypotheses in the vicinity of the successful reports of the same artifact class, $\mathbf{H}_{\mathrm{unsucc}}$ contain hypotheses in the vicinity of the unsuccessful reports of the same artifact class, and $\mathbf{H}_{r}$ represents the hypotheses of robots that have exceeded their own limit on reports and concurrently have a low success rate of their submitted hypotheses. 
    The performance index for a hypothesis $h_i$ is computed as
    \begin{equation}
      P(h_i) = \alpha p_r + \beta p_c + \gamma p_n + \delta p_a, 
    \end{equation}

    where the values $p_r$, $p_c$, $p_n$, $p_a$ represent the percentile of particular performance indices of hypothesis $h_i$ among all hypotheses in $\mathbf{H}_f$, and $\alpha$, $\beta$, $\gamma$, $\delta$ are the weight coefficients.
    The particular performance indices are related to the number of robots with a similar hypothesis ($p_r$), the overall confidence of the detections assigned to the hypothesis ($p_c$), the number of detections assigned to the hypothesis ($p_n$), and the apriori probability of detection of a particular object ($p_a$).
    The next hypothesis to be reported $h^*$ is chosen based on the following equation: 
    \begin{equation}
      h^* = \arg \max_{h_i \in \mathbf{H}_f} P(h_i).  
    \end{equation}
    The distribution of successful reports over particular reporting attempts during all runs of the SubT Virtual Track Prize Round is shown in~\autoref{fig:successful_reports_distribution}.

    \section{Mission control}
    \label{sec:mission_control}

The proposed system is designed for fully autonomous operation, so that the rescue team can benefit from the autonomous reconnaissance of the \ac{uav} without the need for any additional personnel operating the \ac{uav}.
The DARPA SubT competition reflects this requirement on autonomy by allowing only robots without human operators to enter the course.
In theory, the robots could be teleoperated~\cite{moniruzzaman2022teleoperation}.
However, this is not scalable with the number of robots.
Moreover, for teleoperation, a reliable communication link between the robot and the operator is required, but is often not available, especially deeper in the subterranean environment where impenetrable walls diminish signal propagation.
Thus, the correct execution of an autonomous mission relies on a state machine that governs the high-level actions of the \ac{uav}.

\subsection{State machine}
The state machine applied in the SubT System Finals consists of 12 fundamental states.  
In the first state, the status of components that are vital to the mission is checked to ensure that the mission will be accomplished.
Both the software components (\textit{localization, mapping, planning, artifact detection, artifact localization, database}) and hardware components (\textit{LiDAR, RGB cameras, depth cameras, mobilicom unit}) are checked prior to the mission.
This component health check is crucial as, while still in the staging area, any potential component failures can be addressed, but it is not possible when the UAV is already flying.

When all components are running correctly, the UAV enables the output of the reference controller, transits to \textit{WAITING FOR TAKEOFF} state, and waits for approval from the safety operator to start the mission.
The approval required to guarantee the safety of the personnel moving in the vicinity of the UAV is given by arming the \ac{uav} and transferring the control of the UAV fully to the onboard computer by toggling the RC switch.  
After the approval to start, the \ac{uav} waits for a specified safety timeout in the \textit{READY FOR TAKEOFF} state while signaling the imminent takeoff by flashing LEDs.
In this state, the approval can be taken back by the safety operator.  
After the timeout elapsed, the \textit{PERFORMING TAKEOFF} state is entered, during which the UAV ascends until reaching the desired takeoff height.

In the next state (\textit{FLYING THROUGH GATE}), the UAV is navigated to a position inside the area to be explored.
Once this position is reached, the space behind the UAV is virtually closed to prevent flight back towards the rescue personnel.
If the rescuers have some prior knowledge about the environment, e.g., they see a door to which they want to send the UAV, they can optionally specify this first position to steer the UAV in that direction.
After reaching this first position or if the flight to the first position is not requested, the UAV enters the \textit{EXPLORATION} state.
In this state, the UAV fulfills the primary mission goals until the upper bound of the estimated time to return is equal to the remaining flight time. 
Then the \ac{uav} initiates returning to the takeoff position in the state \textit{FLYING BACK TO START}.

The return position is the takeoff position by default, but the operator can request any other position (e.g., to serve as a communication retranslation node) to which the UAV tries to return.
After the position is reached, the UAV flies to the nearest safe landing spot as described in~\autoref{sec:landing_spot_detection}, and the \textit{LANDING} state is entered.
The landing is also triggered when the flight time is elapsed during the \textit{FLYING BACK TO START} or \textit{FLYING TO LANDING SPOT} states.
When the UAV lands, it enters the \textit{FINISHED} state, in which it turns off the motors, \acp{led}, \ac{lidar}, and other components except the communication modules to conserve battery power for retranslating communications.

The required communication between the \ac{uav} and its operator during the start of the mission is limited to signals provided by the RC controller and visual signals provided by flashing LEDs. 
This enables very fast deployment of the \ac{uav} that automatically starts all necessary software components once the onboard computer is powered on and provides the information about being prepared to start by a single long flash of LEDs.
After that, the operator can approve the mission by the remote controller without the need for any additional communication or commanding of the UAV.
Following this automated procedure, the \acp{uav} are prepared to start one minute after the battery is plugged in. 

The state machine applied in the Virtual Track of the SubT Challenge differs only in a few states given by the specifics of the simulation environment. 
First, it does not contain the operator commands states that are not available in a virtual environment. 
Second, it contains two additional states, \textit{BACKTRACKING} and \textit{AVOIDING COLLISIONS}.
The \textit{BACKTRACKING} state is entered when the \ac{uav} is stuck in a fog and tries to escape from it by backtracking to the most recent collision-free poses, ignoring the occupied cells in the current map (see~\autoref{sec:fog_detection} for details).    
In the \textit{AVOIDING COLLISIONS} state, the \ac{uav} is avoiding collision with the \acp{uav} of higher priority by stopping the lateral motion and decreasing its altitude.     
We have decided against using collision avoidance in the Systems Track due to the low probability of collision, and high probability of deadlocks in narrow corridors.


    \subsection{Operator commands}
    \label{sec:operator_commands}

    While the \ac{uav} is capable of carrying out the mission on its own in the fully autonomous mode, the operator can intervene by issuing an operator command to influence the behavior of the \ac{uav}.
    All operator commands can be activated only in the \textit{EXPLORATION} state and in the operator command states, in which the \ac{uav} performs its primary goal.
    Allowing operator commands in other states would interfere with the takeoff, returning, and landing processes.
    The commands are transmitted from the operator's base station to the \ac{uav} through the available communication modalities described in~\autoref{sec:communication}. 
    The following commands are available for the operator:

    \begin{itemize}
    
    \item \textbf{Explore to position} ---
    The operator can bias the automatic goal selection process by issuing the \textit{Explore to position} command.
        After the command is received by the \ac{uav}, the currently used reward function for evaluating viewpoints is extended by a term that penalizes the Euclidean distance of the viewpoint from the desired position $\mathbf{p}_D$.
        The term added to the reward function for a viewpoint $\xi$ is simply
    \begin{equation}
      \label{eq:vp_criterion} 
      \Delta R(\xi_\textrm{UAV}, \xi, \mathbf{p}_D) = -c_{oc} \left| \mathbf{p}_{\xi} - \mathbf{p}_D \right|.
    \end{equation}
    Such modification of the reward function causes the viewpoints closer to the desired positions to be preferred over farther viewpoints.
    The assertiveness of reaching the desired position can be controlled by the coefficient $c_{oc}$. If this is set too high, it might force the viewpoints with a minimal distance from obstacles and low information value to be selected.

    \item \textbf{Plan to position} ---
    The \textit{Plan to position} command bypasses the viewpoint selection process and requests the planner to find a path directly to the specified position.
    When the requested position is not reachable, i.e., it is in an occupied or unknown space, the planner will find the path to the closest point using the Euclidean distance heuristic function.
    Thus, this command should be used primarily for reaching an already visited position, e.g., to land there and retranslate communication from robots that are already further in the environment, or to approach a stuck robot to retrieve its data.

  \item \textbf{Set return position} --- 
    Under normal operation, the \ac{uav} returns to the staging area when its battery is depleted.
    The operator can change the return position by issuing the \textit{Set return position} command. 
        This can save valuable flight time of the \ac{uav} when a communication chain is already established.

  \item \textbf{Stop} ---
    The operator can also halt the movement of the \ac{uav} by issuing the \textit{Stop} command.
        This command is useful when the operator wants to inspect an interesting area in more detail, prevent the \ac{uav} from going into a non-informative or dangerous area, or temporarily retranslate communications.  
    Moreover, this command is a prerequisite for calling the \textit{Land} command.

  \item \textbf{Land} ---
    It is possible to land the \ac{uav} prematurely before the end of the mission by issuing the \textit{Land} command. 
        The expected use case involves landing the \ac{uav} at a position advantageous for extending the communication network.	
		Before calling the \textit{Land} command, the \textit{Stop} command must be called to prevent an accidental landing at an incorrect location, due to the arbitrary delay of the command sent through an unreliable network.
		The system does not guarantee landing at the exact specified position, as a safe landing spot is found in the vicinity of the requested position.

    \item \textbf{Return home} ---
    The \textit{Return home} command switches the \ac{uav} to the \textit{returning} state, as defined in~\autoref{sec:single_uav_goal_eval}.
        In this state, the \ac{uav} uses the navigation module to get as close as possible to the specified return position.

    \item \textbf{Resume autonomy} ---
		The last operator command cancels the behavior that was forced by previous operator commands (except \textit{Land} and \textit{Set return position}).
    This causes the \ac{uav} to resume autonomous exploration, start its return, or land (depending on the flight time left).

    \end{itemize}

\subsection{Communication}
\label{sec:communication}

The developed system assumes an unreliable bidirectional low-bandwidth communication network with intermittent dropouts.
It should be mentioned that 2 meshing-capable wireless technologies are used on the hardware level --- \SI{2.3}{\giga\hertz} Mobilicom and \SI{900}{\mega\hertz} motes, with details of both available in~ \cite{roucek2020urban}. 
This paper focuses on high-level usage of the communication network, which is used as a black box, and as such the low-level layers of the communication protocol are not discussed.

\begin{figure}[!htb]
    \centering
    \includegraphics[width=0.7\columnwidth]{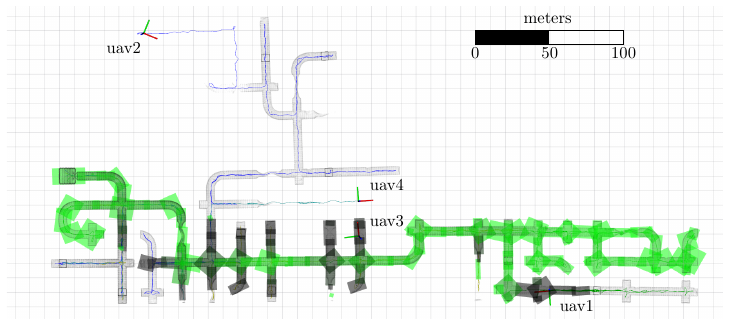}
    \caption{Example of the dispersed exploration of a tunnel system during the first run in world 1 of the virtual track.
    Only LTVMap from \ac{uav}1 is shown for clarity, other \acp{uav} received this map and maps from the other \acp{uav}.
    Instead of exploring again the same places as \ac{uav}1, both \ac{uav}2 and \ac{uav}4 explore previously unvisited corridors.
    Dark parts of LTVMap in this figure are not yet fully explored, so \ac{uav}3 flies to inspect these areas to not miss any potentially hidden artifacts.
    }
    \label{fig:coop_virtual}
\end{figure}

\begin{figure}[!htb]
    \centering
    \includegraphics[width=0.7\columnwidth]{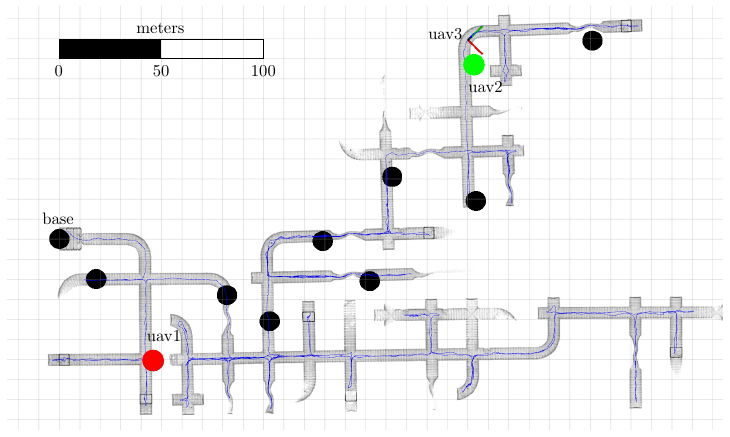}
    \caption{A communication network consisting of a base station and 8 breadcrumbs (black) deployed by the \acp{ugv} and 2 \acp{uav} from the \nth{3} run in world 1 of the virtual track.
    \ac{uav}3 with its trajectory shown in blue could explore further thanks to the deployed communication nodes.
    Without the communication network, the \ac{uav} would have to return to the staging area, thus traveling additional \SI{500}{\meter} from its final landing position.
    }
    \label{fig:comm_chain}
\end{figure}

The developed system benefits from available connections to other agents and the base station in multiple ways.
First, when a robot detects an artifact, the detection with its estimated position is shared over the network instead of returning physically to the base station, thus saving time valuable for the success of the mission.
Second, the individual agents can share the information about the already explored volume in the form of a topological-volumetric map (\acs{ltvmap}) introduced in \autoref{sec:lsegmap}.
The knowledge of other agents' topological-volumetric maps penalized regions already explored by other robots, which encourages splitting of the robot team and covering a larger volume over the same time period as shown in~\autoref{fig:coop_virtual}.
Third, each robot shares its position with the base station, so that the operator has an overview of where all robots are located.
The operator can then influence the future behavior of any robot in the communication range by sending an operator command (\autoref{sec:operator_commands}).
Last, positions of the communication nodes (breadcrumbs or landed \acp{uav}), which form the communication network shown in \autoref{fig:comm_chain}, are sent to be used for returning to the communication range when the remaining flight time is low.

\subsection{Calibrating global reference frame}

\begin{figure}[!htb]
    \centering
    \includegraphics[width=1.0\columnwidth]{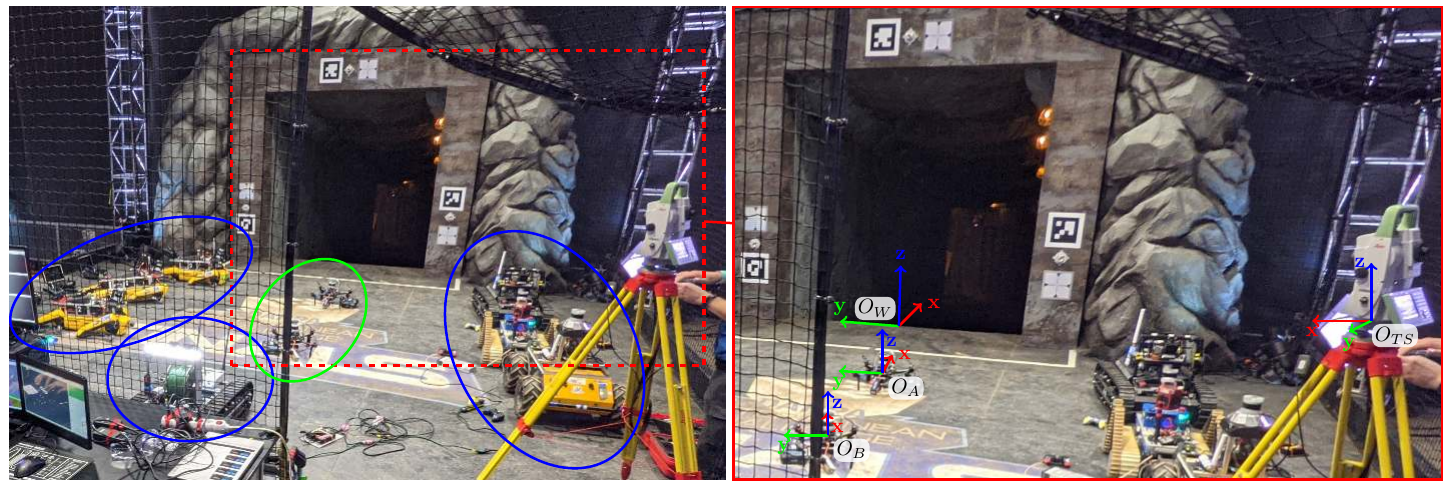}
    \caption{Example robot distribution (7 \ac{ugv} robots in blue, 2 \ac{uav} robots in green) of team CTU-CRAS-NORLAB within the staging area of System Finals environment of \ac{darpa} \ac{subt} Challenge, 2021.
             The Right figure highlights the reference frames of interest --- the world origin $O_W$ together with the origin of the Leica total station $O_{TS}$ used for calibrating local robot origins $O_A$ and $O_B$ within the world.}
    \label{fig:staging_area_calibration}
\end{figure}

The entire navigation system of heterogeneous robots within the CTU-CRAS-NORLAB team is decentralized under the assumption of a shared coordinate frame --- the world coordinate frame $O_W$.
To obtain the transformation of a robot's local origin within the world frame, the staging area of the competition environment provides a set of visual tags and a set of reflective markers, both with precisely known poses within the world (see the markers mounted on the entrance to the environment in~\autoref{fig:staging_area_calibration}).
The reflective markers are used within our 6-\ac{dof} calibration procedure in which a Leica TS16 total station is employed to measure 3D points with sub-millimeter accuracy.
The origin $\mathbf{T}_{TS}^{W}$ of the total station in the world is derived from measuring known in-world marker poses and used in deriving $\mathbf{T}_{B}^{W}$ of a robot $B$.

To calibrate the pose of a single robot $B$ after $\mathbf{T}_{TS}^{W}$ is known, 4 known points on the robot's frame need to be measured, used in estimating $\mathbf{T}_{B}^{W}$, and sent to the information database (see \autoref{sec:communication}) or directly to the robot.
As the number of robots in the CTU-CRAS-NORLAB team deployments reached up to 9 robots per run (see~\autoref{fig:staging_area_calibration}), the overhead for robots-to-world calibration decelerated the rate of robot deployments as well as limited the possibilities for quick in-situ decision making.
To speed up the calibration pipeline for \acp{uav} with limited flight distance (and hence with greater room for calibration errors), just a single \ac{uav} $A$ needs to be calibrated with the total station wherein the initial pose of the remaining \acp{uav} $B$ is estimated from on-board \ac{lidar} data.
The known transformation $\mathbf{T}_A^W$ and pre-takeoff \ac{lidar} data $\mathbf{D}_A$ of a robot $A$ are shared throughout the robots and used to estimate $\mathbf{T}_B^W$.
The transformation $\mathbf{T}_B^A$ is estimated by registering source \ac{lidar} data $\mathbf{D}_B$ onto target data $\mathbf{D}_A$ using \ac{icp} with extremely tight constraints in matching the rotation component of the transformation.
The tight rotation constraints are important as frame-orientation misalignments are the largest source of absolute error during deep deployments.
The pose of robot $B$ in the world is then given by $\mathbf{T}_B^W = \mathbf{T}_B^{A}\mathbf{T}_A^{W}$.

    \section{Hardware platform}
    \label{sec:hardware}

The components of our \ac{sar} \ac{uav} were carefully selected to optimize the flight time and perception capabilities based on years of experience with building aerial robots for research~\cite{ahmad2021autonomous}, competitions~\cite{walter2022fr}, inspection~\cite{silano2021powerline}, documentation~\cite{kratky2021documentation} and aerial filming~\cite{kratky2021aerialfilming}.
All platforms we have designed for diverse tasks and purposes including \ac{darpa} \ac{subt} are presented in~\cite{hert2022hardware}.

 Our platform is built upon the Holybro X500 quadrotor frame.
 The \SI{500}{\milli\meter} frame is made entirely of carbon fiber, therefore it is stiff and light.
 Moreover, the arm length can be changed to accommodate different propellers.
 Our team designed and manufactured a custom \ac{pcb} that replaced the top board of the X500 frame.
 This PCB integrates power supplies for onboard sensors and \ac{led} lights, facilitates communication with our \ac{fcu}, and integrates the XBee-based e-stop receiver.
 We selected MN3510 KV700 motors from T-motor and paired them with \SI{13}{\inch} carbon fiber propellers for large payload capacity and propulsion efficiency.

 The 3D \ac{lidar} was upgraded to the OS0-128 model, which features 128 scanning lines and wide \SI{90}{\degree} vertical field of view, which allows for perceiving the surroundings of the \ac{uav} in the challenging underground environments.
 Despite the wide coverage of the \ac{lidar} sensor, there are still blind spots above and below the \ac{uav} when mounted horizontally.
 To cover these spots, we use two Intel Realsense D435 \ac{rgbd} cameras, facing up and down.
 This enables the \ac{uav} to fly directly upwards, even in cluttered vertical shafts, without risking collision.
 Both of the \ac{rgbd} cameras are also used for mapping and artifact detection.
 Additionally, the bottom facing \ac{rgbd} camera is used for landing site detection.
 The platform is equipped with two (left and right) dedicated artifact detection cameras, the Basler Dart daA1600, and sufficient lighting provided by \ac{led} strips.
 All algorithms run on the onboard Intel NUC i7-10710U \ac{cpu} with 6 physical cores and the detection \ac{cnn} utilizes the integrated Intel UHD \ac{gpu}.

 The high-power Mobilicom MCU-30 wireless communication module provides long-range connection between robots and the base station.
 In some topologically complex areas, even the high-power Mobilicom cannot assure reliable connection between the units, so it is supported by smaller \SI{900}{\mega\hertz} communication motes, which are also dropped as breadcrumbs by the \acp{ugv} to improve the signal range.

 Finally, the large payload capacity of the \ac{uav} allowed us to extend the flight time by using a larger battery.
 We used two 4S \SI{6750}{\milli\ampere\hour} Li-Po batteries in parallel. Instead of a larger battery, two smaller batteries were used due to the \SI{100}{\watt\hour} limit for aircraft transportation.
 This gave the UAV a flight time of \SI{25}{\minute}.

\begin{figure}[!htb]
    \centering
    \adjincludegraphics[height=14em,trim={{0.0\width} {0.01\height} {0.0\width} {0.01\height}},clip]{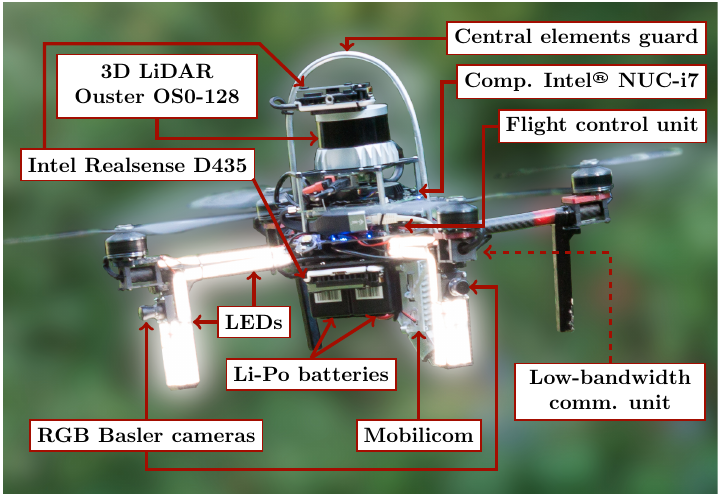}
    \adjincludegraphics[height=14em,trim={{0.0\width} {0.01\height} {0.0\width} {0.01\height}},clip]{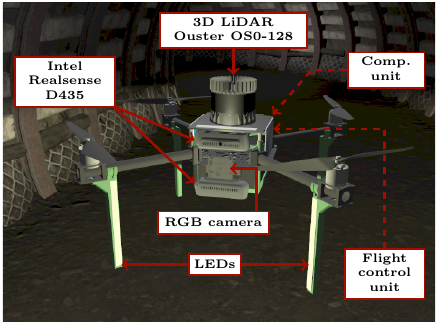}
    \caption{X500 platform used in the Systems Track (left) and Virtual Track model counterpart (right).
    }
    \label{fig:uav_x500_labeled}
\end{figure}

The X500 platform~(\autoref{fig:uav_x500_labeled}) is capable of flying in dense indoor environments, even in tight vertical shafts, while being able to localize itself with the required accuracy.
 It has four different cameras for artifact detection, is able to communicate and form mesh networks with other robots, and possesses a long flight time.
 
 Furthermore, this platform was also replicated in the virtual competition with the same parameters as the physical counterpart.
 All of the teams except for two used the X500 platforms in the Virtual Track due to its long flight time, sufficient sensor suite, and agile dynamics.

    \section{Technical details}
    \label{sec:technical_details}
   
    With a few exceptions, the components of the \ac{uav} software stack deployed in the virtual and systems tracks are equal, yet the available processing powers are not.
    The Virtual Track yields a low real-time simulation factor. Together with the computational capacities of each simulated robot, it provides almost unlimited computational resources for running all algorithms with any desired resolution or maximal settings.
    On the other hand, the simulation-to-world transition requires the algorithms to run on the onboard processing units. This imposes hard requirements on the algorithms' optimization, as well as on minimization of the amount of data transfers and their latency.
    These requirements force us to
    \begin{itemize}
      \item compromise between accuracy and real-time performance in the system design (i.e., cutting out global optimization in on-board running \ac{slam}),
      \item ensure real-time properties for systems handling critical factors of the mission (i.e., \ac{uav} control),
      \item optimize the data flow and the priorities of processing order within the software stack, and
      \item prevent any possible deadlocks from arising from outages of both synchronous, and asynchronous data.
    \end{itemize}
    Ensuring real-time settings for all systems of a robotic deployment is implausible, particularly in complex robotic-research projects where the stack design must allow for the system to function as a whole under limited real-world conditions.
    We summarize the specific aspects of the proposed ROS-based software stack, allowing us to transfer all components to on-board processing capacities.
    Thus, providing full decentralization within a \ac{uav} team.

    Software based on ROS 1 allows for connecting components under a \textit{nodelet manager} in order to group \textit{nodelet} plugins.
    In contrast to \textit{node} configuration, the \textit{nodelets} under a \textit{manager} have shared memory and do not require copying data, a tool useful particularly in the case of passing large maps within the navigation stack. 
    Our deployment stack consists of several \textit{managers}, each of which handles a distinctive part of the system. 
    These include \ac{uav} control, pre-processing of \ac{lidar} data and \ac{slam}, pre-processing of \ac{rgbd} data and dense mapping, navigation and path planning, and perception.
    The data flowing between these \textit{managers} are copied, and thus the rate of sharing is subject to maximal reduction.
    To decrease the temporal and memory demands of algorithms, the resolution of input data and the output maps is decreased as much as possible within the scope and requirements of the desired application.
    The rate of saving data for after-mission analyses is also limited as much as possible, with no post-reconstructable data being recorded at all.

\begin{figure}
      \centering
      \includegraphics[width=1.0\textwidth]{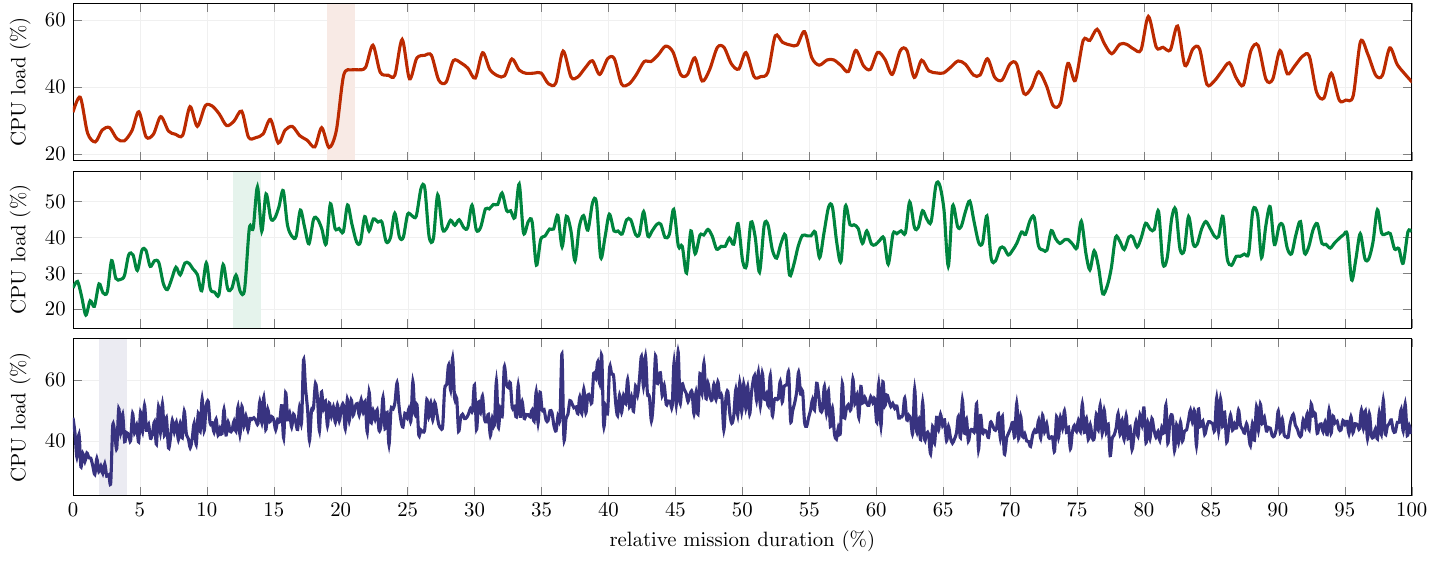}
      \caption{\label{fig:cpu_load} 
       The CPU load of onboard computers of individual \acp{uav} (\uavred{}, \uavgreen{}, \uavblue{}) during the prize round of SubT System Finals.
       The highlighted parts of the graph correspond to the start of processing onboard images by the object detection pipeline.}
    \end{figure}

    In contrast to the system designs for \ac{ugv} platforms, the delays in state estimation and control inputs are a critical subject for reduction.
    This is because excessive delays lead to destabilization of a multi-rotor aerial platform (see analysis on delay feasibility in~\autoref{fig:delay_analysis}) as it is a dynamically unstable system requiring frequent feedback, even for simple hovering.
    The \textit{nodelet managers} handling such critical parts of the system are prioritized at the CPU level, utilizing the negative \texttt{nice} values that prioritize the related processes during CPU scheduling.
    To decrease asynchronous demands on the CPU, non-prioritized components are penalized with positive \texttt{nice}.
    Furthermore, their scheduling is restricted on a predetermined set of threads in a multi-threaded CPU.
    The primary subject of scheduling restriction is the perception pipeline containing a computationally heavy \ac{cnn}, where static allocation reduces its asynchronous influence on the rest of the system at the cost of a limited processing rate. 
    The effect of switching on the perception pipeline is visible in~\autoref{fig:cpu_load}, showing the CPU load of the three deployed \acp{uav} during the \ac{darpa} \ac{subt} System Finals.
    In other validation tests, the CPU load reached up to \SI{90}{\percent} in \SI{1500}{\second} long missions within vast underground environments.
    Such an overloaded CPU results in frequent asynchronous delays, culminating to unpredictable and destructive behavior.

    To limit the power consumption and hence, increase the maximum flight time, unsolicited hardware and software components can be temporarily powered off.
    These include switching off on-board lights in meaningless settings, disabling \ac{cnn} processing when not needed, or powering off the \ac{lidar} in the after-landing phase when the \ac{uav} is serving solely as a retranslation unit for communication.
   
    \section{System deployment}
    \label{sec:experiments}

    Throughout the development of the system presented in this paper, the individual components were extensively tested before integration.
    Deployments of the whole system were less frequent, but allowed testing the interaction of individual modules and verifying the ability to fulfill the primary objective of finding objects of interest in subterranean environments.
    
    \subsection{Continuous field verification}
    \label{sec:field_testing}

\begin{figure} [ht]
    \newcommand{\imheight}{12.50em}
    \newcommand{\xcap}{0.95em}
    \newcommand{\ycap}{0.8em}
    \newcommand{\fillopa}{0.3}
    \centering
    \begin{tikzpicture}
      \node[anchor=south west,inner sep=0] (b) at (0,0) {\adjincludegraphics[width=0.49\textwidth,trim={{0.00\width} {0.0\height} {0.00\width} {0.00\height}},clip]{./fig/photos/x500_pilsen_dust.jpg}};%
    \begin{scope}[x={(b.south east)},y={(b.north west)}]
  \node[fill=black, fill opacity=\fillopa, text=white, text opacity=1.0] at (\xcap, \ycap) {\textbf{(a)}};
      \end{scope}
    \end{tikzpicture}
    \begin{tikzpicture}
      \node[anchor=south west,inner sep=0] (b) at (0,0) {\adjincludegraphics[width=0.49\textwidth,trim={{0.00\width} {0.0\height} {0.00\width} {0.00\height}},clip]{./fig/photos/x500_pilsen_vertical.jpg}};%
    \begin{scope}[x={(b.south east)},y={(b.north west)}]
  \node[fill=black, fill opacity=\fillopa, text=white, text opacity=1.0] at (\xcap, \ycap) {\textbf{(b)}};
      \end{scope}
    \end{tikzpicture}
    \begin{tikzpicture}
      \node[anchor=south west,inner sep=0] (b) at (0,0) {\adjincludegraphics[height=\imheight,trim={{0.00\width} {0.0\height} {0.00\width} {0.00\height}},clip]{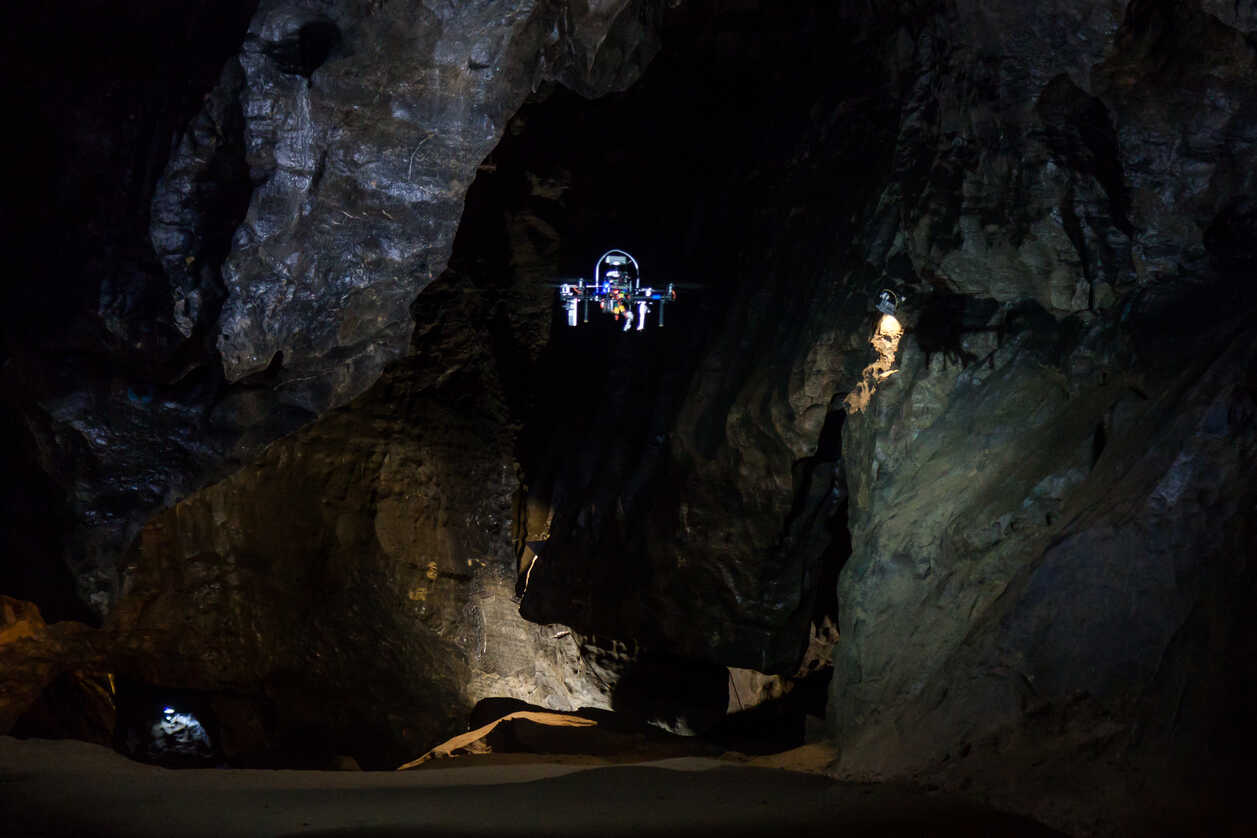}};%
    \begin{scope}[x={(b.south east)},y={(b.north west)}]
  \node[fill=black, fill opacity=\fillopa, text=white, text opacity=1.0] at (\xcap, \ycap) {\textbf{(c)}};
      \end{scope}
    \end{tikzpicture}
    \begin{tikzpicture}
      \node[anchor=south west,inner sep=0] (b) at (0,0) {\adjincludegraphics[height=\imheight,trim={{0.00\width} {0.0\height} {0.00\width} {0.00\height}},clip]{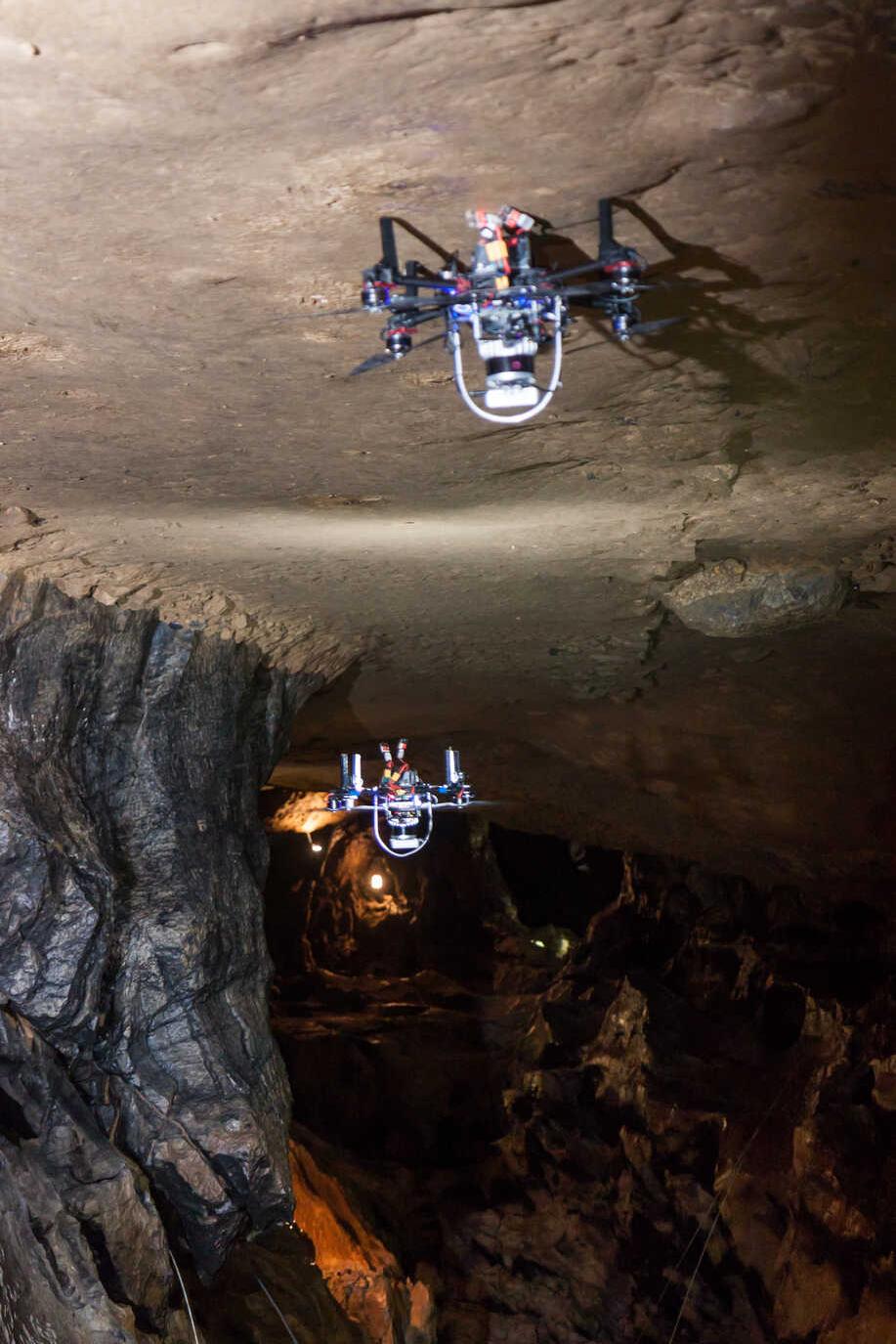}};%
    \begin{scope}[x={(b.south east)},y={(b.north west)}]
  \node[fill=black, fill opacity=\fillopa, text=white, text opacity=1.0] at (\xcap, \ycap) {\textbf{(d)}};
      \end{scope}
    \end{tikzpicture}
    \begin{tikzpicture}
      \node[anchor=south west,inner sep=0] (b) at (0,0) {\adjincludegraphics[height=\imheight,trim={{0.00\width} {0.0\height} {0.00\width} {0.00\height}},clip]{./fig/photos/x500_outside.jpg}};%
    \begin{scope}[x={(b.south east)},y={(b.north west)}]
  \node[fill=black, fill opacity=\fillopa, text=white, text opacity=1.0] at (\xcap, \ycap) {\textbf{(e)}};
      \end{scope}
    \end{tikzpicture}
    \caption{
      The verification of localization and perception in the following scenarios: data degraded by insufficient lighting and whirling dust (a), traversal of vertical narrow passage (b), performance in humid caves (c), multi-robot exploration (d), and scalability with the environment size (e).
  }
    \label{fig:field_testing}
  \end{figure}
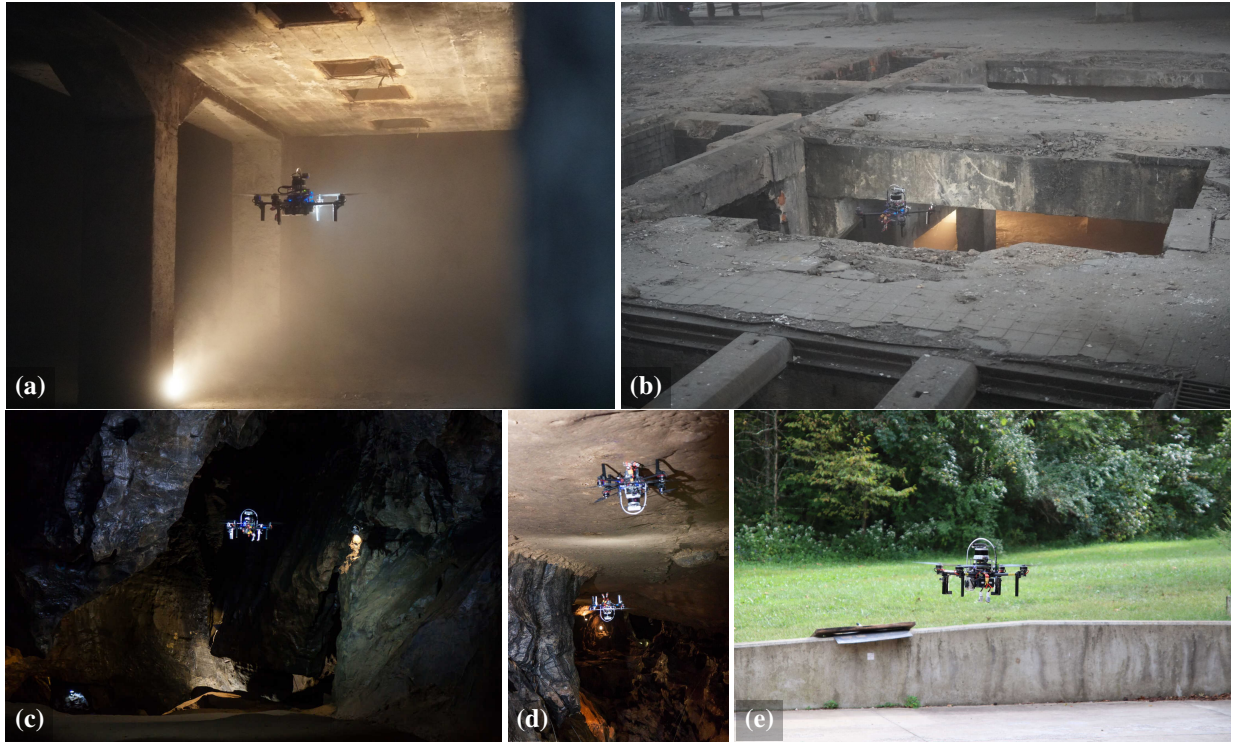

    The \ac{sar} \ac{uav} system was continuously tested to empirically verify the correctness and reliability of the developed algorithms, strategies, and hardware. 
    The \acp{uav} were deployed into diverse types of environments, including historical and industrial buildings of varied levels of disintegration, in humid unstructured caves, a decommissioned underground military fortress, and vast outdoor rural areas.
    Some of these environments are shown in~\autoref{fig:field_testing}.
    Such tests are critical for evaluating the performance under the stochastic influence of real-world conditions, which are typically not modeled in simulations.
    In particular, each perception mode is more or less degraded by ambient lighting or the lack of it, the fog with microscopic condensed droplets of water, smoke or dust particles, reflections on water or smooth surfaces, etc.
    The filtration of \ac{lidar} and depth data from~\autoref{sec:filtering_observation_noise} therefore had to be tuned correctly to prevent the integration of false positives into the map, while keeping the actual obstacles.
    Moreover, the artifact detection system needed to work under a wide range of visibility conditions and chromatic shifts, for which it was necessary to collect artifact datasets from the mentioned environments.

    \subsection{DARPA SubT Final Event Systems Track}
    \label{sec:darpa_systems}

    The final event, which was the culmination of the \ac{darpa} \ac{subt} competition, was organized in the Louisville Mega Cavern in Kentucky on September 23, 2021.
    The course consisted of all three environments from the previous circuits and contained all artifacts from previous events plus \textit{the cube}, which was a new artifact for the final event.
    This section reports on the results achieved by the aerial part of the CTU-CRAS-NORLAB team.
    A total of 40 artifacts were distributed over \SI{880}{\meter} long course, which was divided into 28 smaller sectors to track the team's progress.
    Every robot starts in the staging area, from which a single corridor leads to an intersection that branches into three ways.
    Each of the branches leads to one of the three specific environment types (tunnel, urban, and cave).

    \begin{figure}
      \centering
      \includegraphics[width=1.0\textwidth]{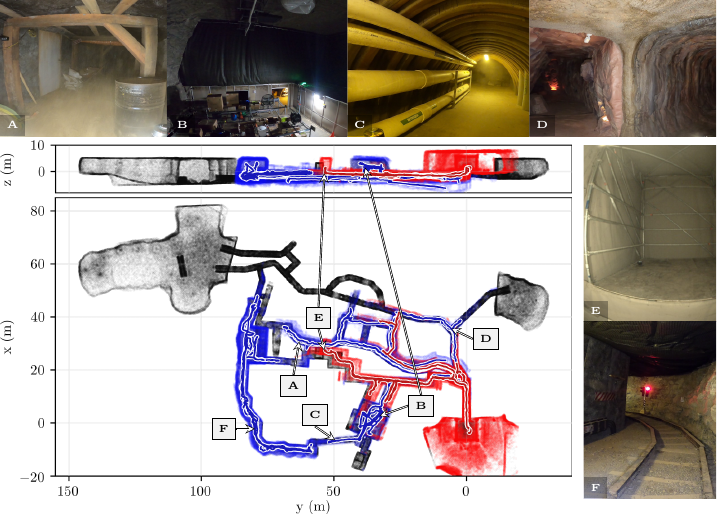}
      \caption{\label{fig:finals_trajs_over_map_overlay} 
        \ac{uav} trajectories and on-board-built maps of the environment from all flights during the prize round (colored in red) and the post-event testing (colored in blue) overlaid over the ground truth map (colored in black).
        The photos from on-board camera highlight the diversity and the narrow confines of the environment.%
      }
    \end{figure}

    \begin{figure}
      \centering
      \includegraphics[width=1.0\textwidth]{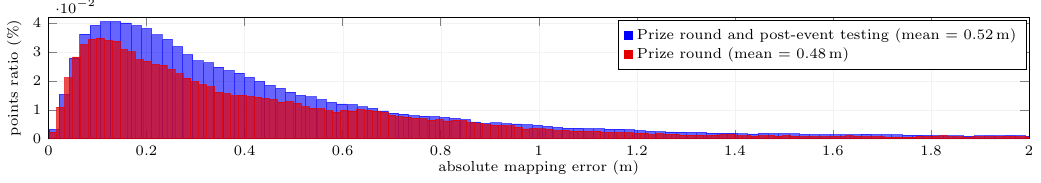}
      \caption{\label{fig:finals_mapping_errors_histogram} 
        Distribution of mapping errors throughout the prize round and the post-event testing flights (colored in red and in blue in~\autoref{fig:finals_trajs_over_map_overlay}) of \ac{darpa} \ac{subt}. 
        The absolute mapping error denotes the distance between the ground truth map and concatenation of DenseMaps built with resolution of \SI{20}{\centi\meter} on-board during particular \ac{uav} flights.
        The error metric is the Euclidean distance between a point from the on-board maps to its closest point in the ground truth map.}
    \end{figure}

    \begin{table}
     \caption{\label{tab:mission_stats}
      The mission statistics from the prize round of the final event.
      The row \textit{mission start} marks the time of takeoff in the \SI{60}{\minute} long SubT mission.
      \textit{Mission end} is the time of landing with the \textit{landing cause} in the last row.%
      }
     \centering
     \tablesize
     \begin{tabular}{l c c c}
       \toprule
       \tablehdg{UAV} & \tablehdg{Red} & \tablehdg{Green} & \tablehdg{Blue} \\
       \midrule
       \tablehdg{Mission start} & 2:00 & 46:30 & 36:00 \\
       \tablehdg{Mission end} & 5:05 & 52:00 & 58:40 \\
       \tablehdg{Flight time} & \SI{180}{\second} & \SI{310}{\second} & \SI{1345}{\second} \\
       \tablehdg{Flight distance} & \SI{69}{\meter} & \SI{119}{\meter} & \SI{304}{\meter} \\
       \tablehdg{Localization accuracy:} &                 &                  &                  \\
       \tablehdg{~~~avg{\textbar}max error in translation (\si{\meter})} & 0.38\,{\textbar}\,0.63 & 0.97\,{\textbar}\,2.66 & - \\
       \tablehdg{~~~avg{\textbar}max error in heading (\si{\degree})}    & 0.64\,{\textbar}\,4.06 & 1.48\,{\textbar}\,5.37 & - \\
       \tablehdg{Sectors explored} & 1 & 3 & 4 \\
       \tablehdg{Sectors entered} & 4 & 6 & 4 \\
       \tablehdg{Safety clearance} & \SI{0.4}{\meter} & \SI{0.11}{\meter} & \SI{0.4}{\meter} \\
       \multirow[t]{2}{*}[0em]{\tablehdg{Landing cause}} & 
       Collision with \acs{ugv} &
       \multirow[c]{2}{*}{\shortstack[c]{Collision with a metal rod\\protruding from the wall}} &
       \multirow[c]{2}{*}{\shortstack[c]{Drained battery after being\\trapped in degraded map}}\\\\
       \bottomrule

     \end{tabular}
    \end{table}


    \begin{figure}
      \centering
      \includegraphics[width=0.90\textwidth]{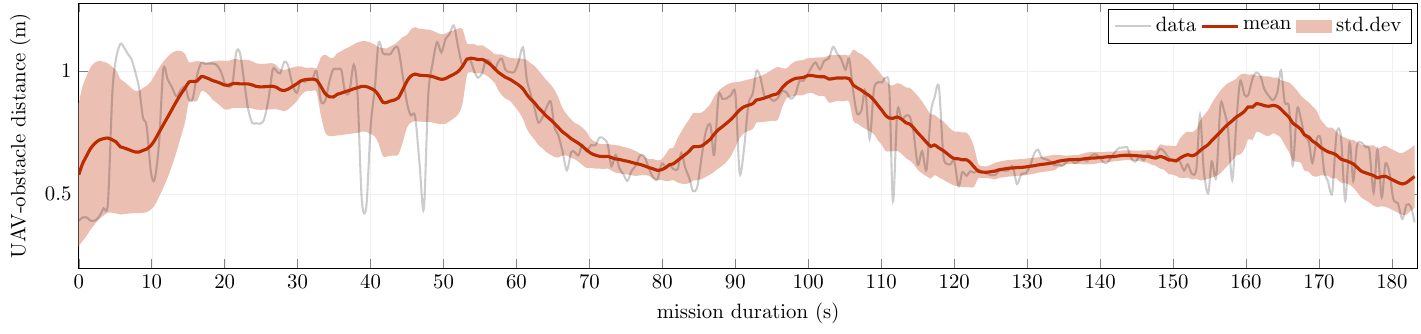}
      \includegraphics[width=0.90\textwidth]{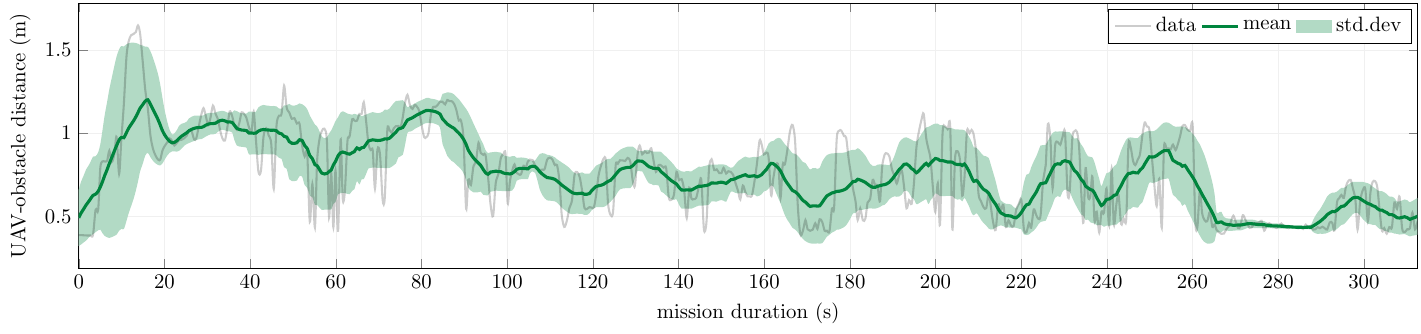}
      \includegraphics[width=0.90\textwidth]{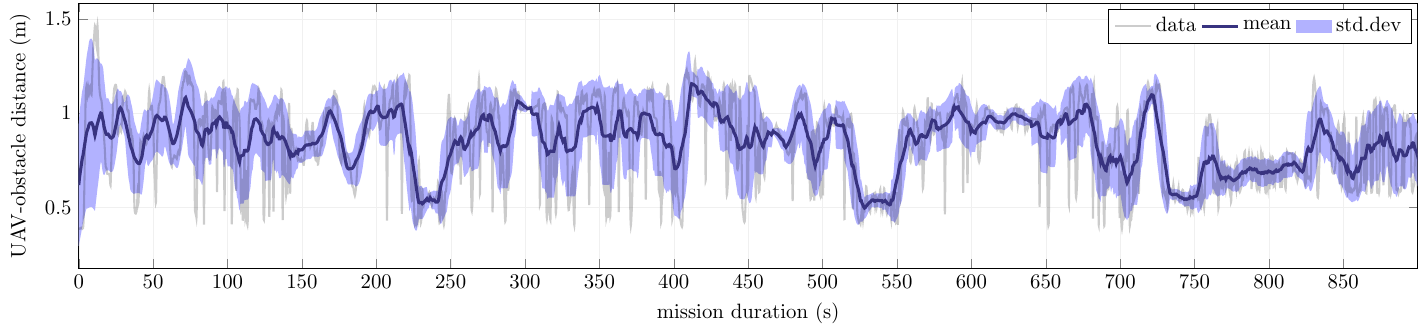}
      \caption{\label{fig:uav_obstacle_dists}
        Distance between the center of the \acp{uav} and the measured nearest obstacle during the prize round of the SubT System Finals. The moving mean and standard deviation are computed over a \SI{10}{\second} long time window.
      }
    \end{figure}

    \subsubsection{UAV deployment summary}
\begin{figure} [ht]
    \newcommand{\xcap}{0.95em}
    \newcommand{\ycap}{0.8em}
    \newcommand{\fillopa}{0.3}
  \centering
    \begin{tikzpicture}
      \node[anchor=south west,inner sep=0] (b) at (0,0) {\includegraphics[height=15em]{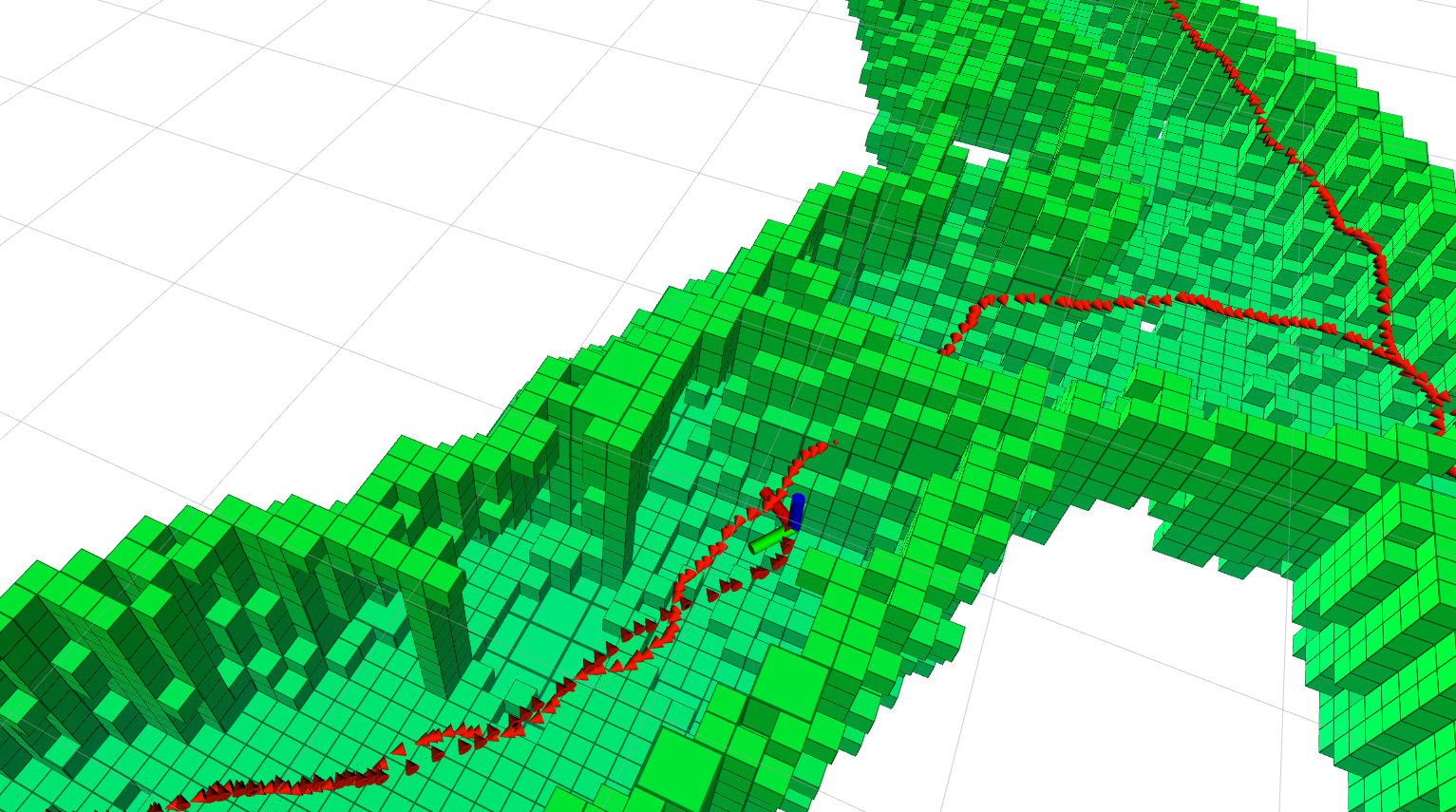}};%
    \begin{scope}[x={(b.south east)},y={(b.north west)}]
      \node[fill=black, fill opacity=\fillopa, text=white, text opacity=1.0] at (\xcap, \ycap) {\textbf{(a)}};
      \end{scope}
    \end{tikzpicture}%
    \begin{tikzpicture}
      \node[anchor=south west,inner sep=0] (b) at (0,0) {\includegraphics[height=15em]{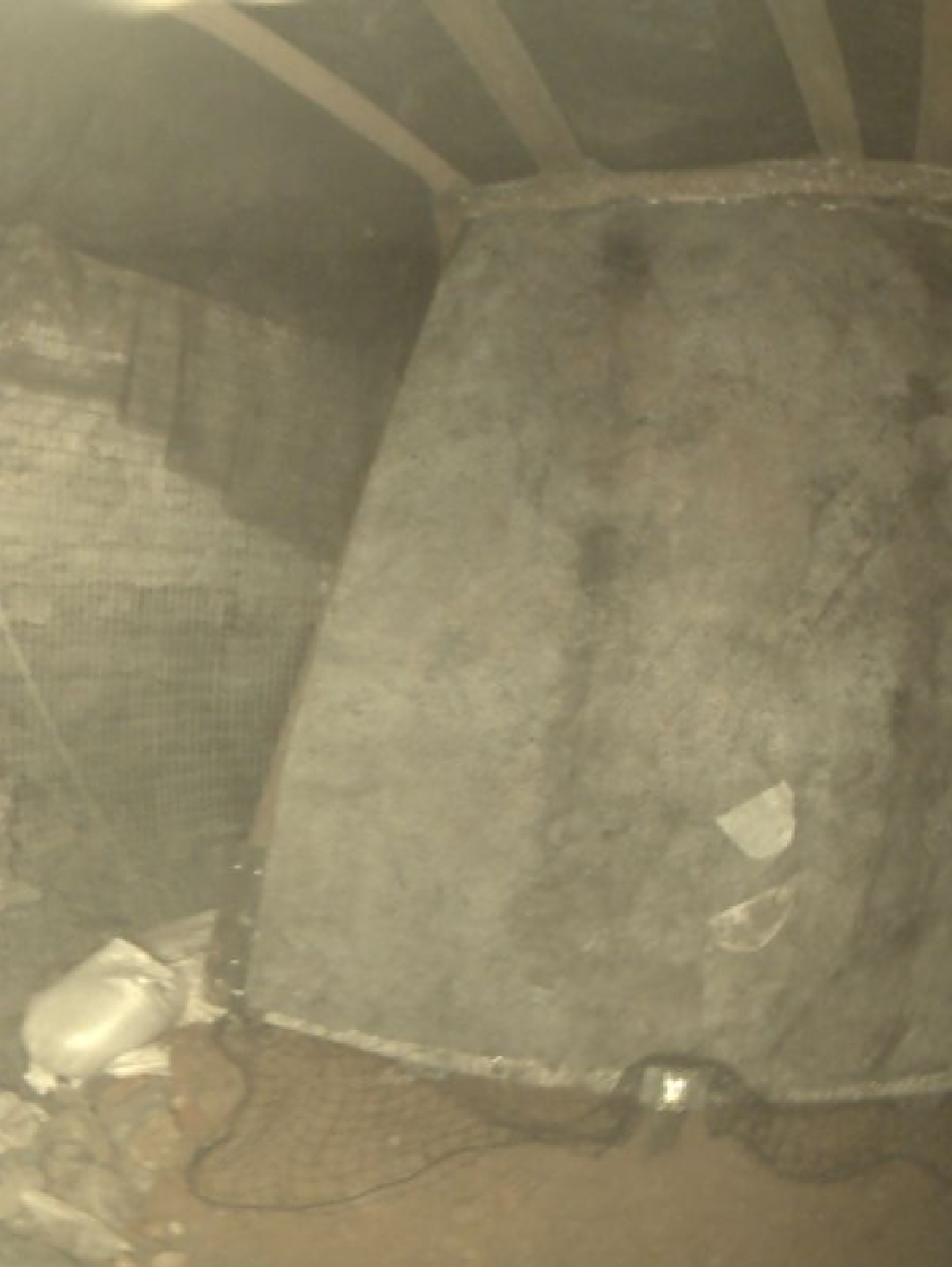}};%
    \begin{scope}[x={(b.south east)},y={(b.north west)}]
      \node[fill=black, fill opacity=\fillopa, text=white, text opacity=1.0] at (\xcap, \ycap) {\textbf{(b)}};
      \end{scope}
    \end{tikzpicture}
  \caption{
    The artificial wall that blocked the way back for \ac{uav} \uavgreen{} in the map (a) and in the camera image (b).
}
  \label{fig:dynamic_wall}
\end{figure}

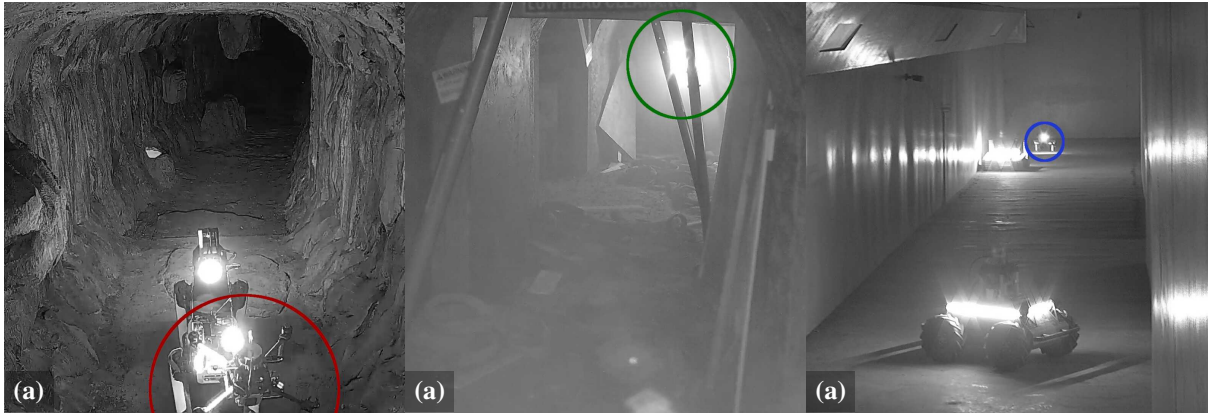
\begin{figure} [ht]
    \newcommand{\xcap}{0.95em}
    \newcommand{\ycap}{0.8em}
    \newcommand{\fillopa}{0.3}
  \centering
    \begin{tikzpicture}
      \node[anchor=south west,inner sep=0] (b) at (0,0) {\adjincludegraphics[height=15.5em,trim={{0.20\width} {0.0\height} {0.25\width} {0.00\height}},clip]{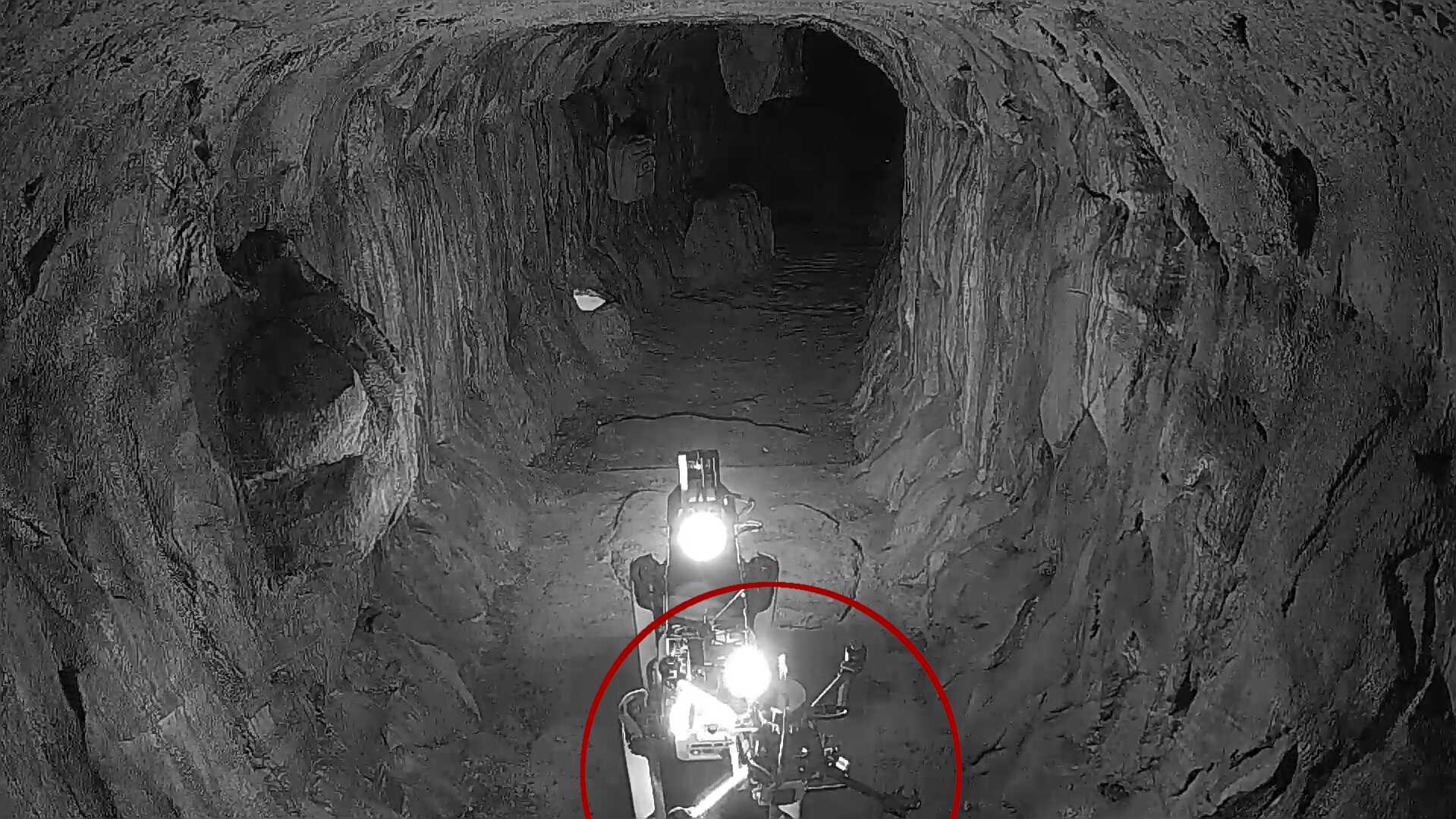}};%
    \begin{scope}[x={(b.south east)},y={(b.north west)}]
      \node[fill=black, fill opacity=\fillopa, text=white, text opacity=1.0] at (\xcap, \ycap) {\textbf{(a)}};
      \end{scope}
    \end{tikzpicture}%
    \begin{tikzpicture}
      \node[anchor=south west,inner sep=0] (b) at (0,0) {\adjincludegraphics[height=15.5em,trim={{0.30\width} {0.0\height} {0.15\width} {0.00\height}},clip]{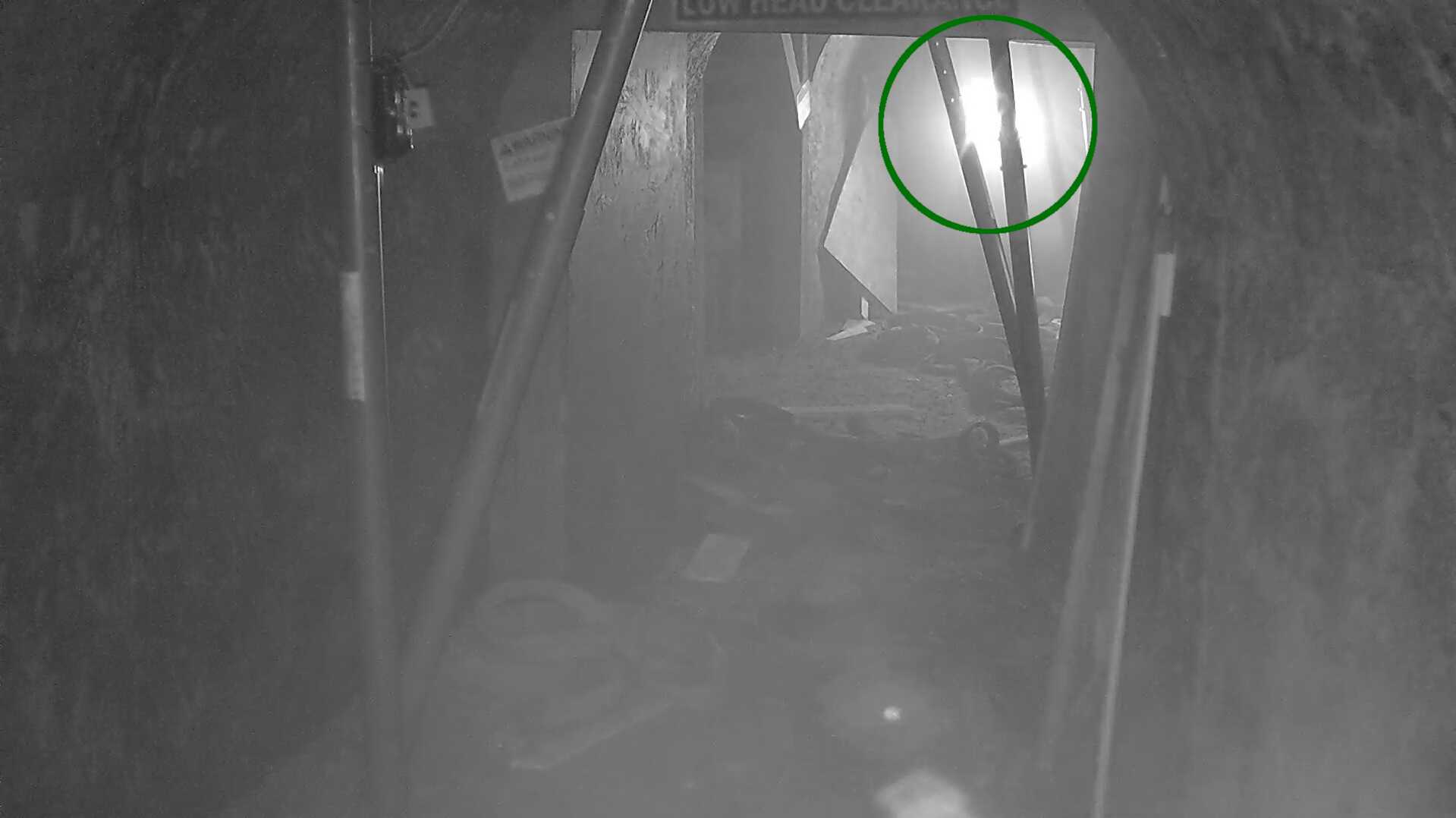}};%
    \begin{scope}[x={(b.south east)},y={(b.north west)}]
      \node[fill=black, fill opacity=\fillopa, text=white, text opacity=1.0] at (\xcap, \ycap) {\textbf{(a)}};
      \end{scope}
    \end{tikzpicture}%
    \begin{tikzpicture}
      \node[anchor=south west,inner sep=0] (b) at (0,0) {\adjincludegraphics[height=15.5em,trim={{0.25\width} {0.0\height} {0.2\width} {0.00\height}},clip]{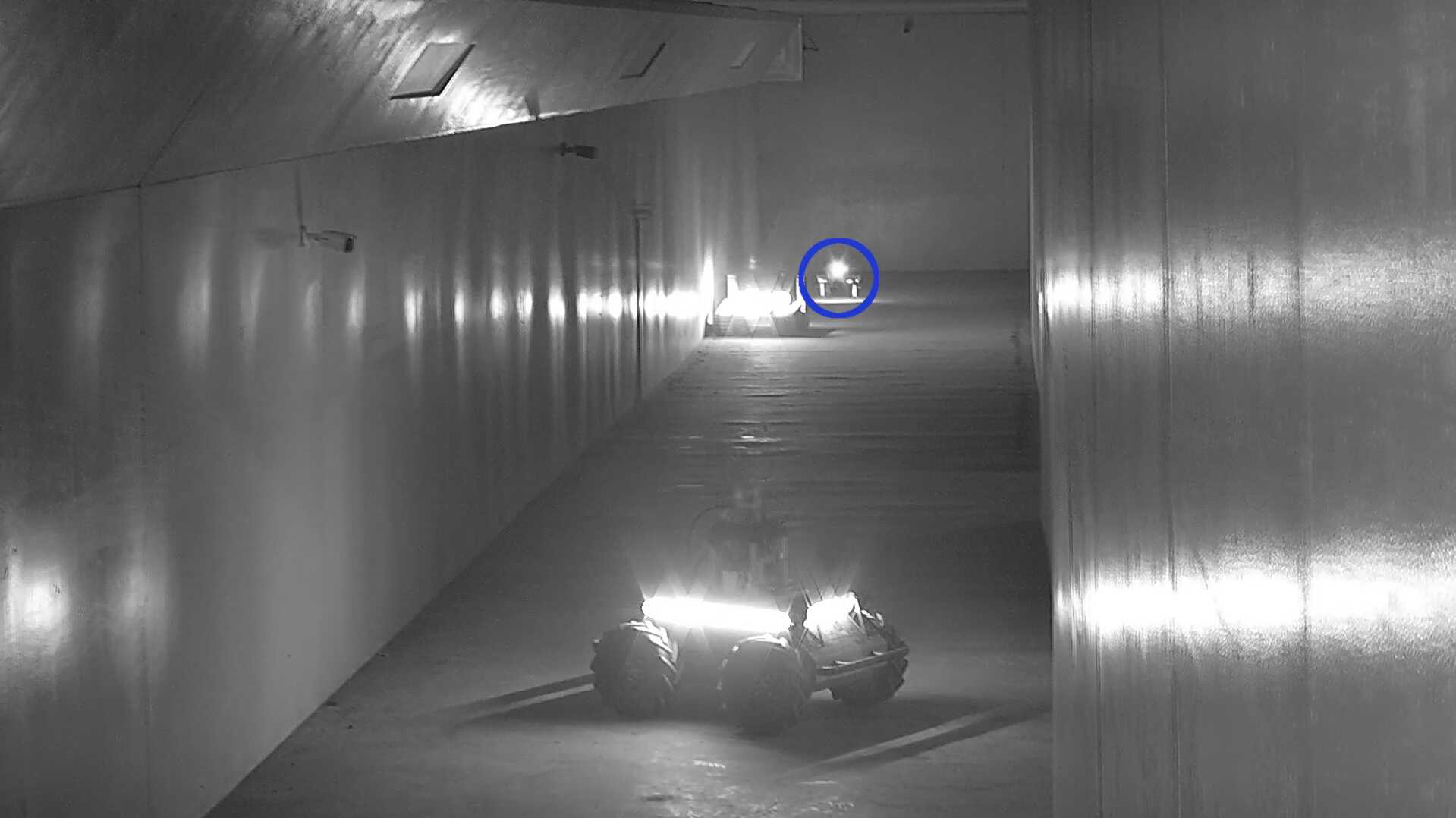}};%
    \begin{scope}[x={(b.south east)},y={(b.north west)}]
      \node[fill=black, fill opacity=\fillopa, text=white, text opacity=1.0] at (\xcap, \ycap) {\textbf{(a)}};
      \end{scope}
    \end{tikzpicture}
  \caption{
    The landing events of all three \acp{uav}.
  The \ac{uav} \uavred{} (a) collided with the Spot \ac{ugv}, \ac{uav} \uavgreen{} (b) hit a metal rod protruding from the wall, and \ac{uav} \uavblue{} (c) landed after its battery was exhausted by hovering while being trapped in a map corrupted by drift in the featureless corridor.
}
  \label{fig:landing_events}
\end{figure}

\begin{figure}
      \centering
      \includegraphics[width=1.0\textwidth]{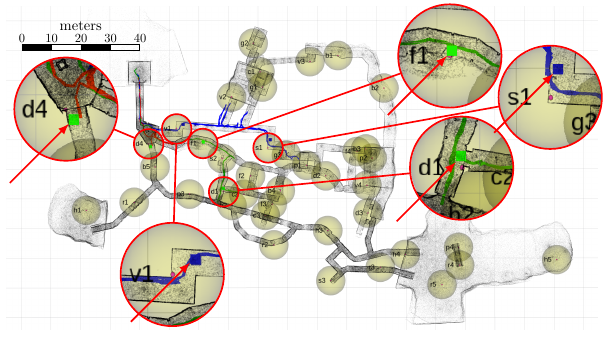}
      \caption{\label{fig:paths} 
        The map of the final event course was obtained by the organizers by scanning the course with a laser scanner station.
        The paths traveled by all three \acp{uav} (\uavred{}, \uavgreen{}, and \uavblue{}) during the final event are depicted by their respective colors.
        The ground truth positions of artifacts are surrounded by a yellow sphere in order to visualize the \SI{5}{\meter} limit for the reported artifact to be counted as a point in the competition.
        The five artifacts that were detected and localized within this \SI{5}{\meter} limit are shown as squares colored by the detecting UAV and highlighted in the magnified sections with red arrows.
      }
    \end{figure}

    Three \acp{uav} in total (\uavred{}, \uavgreen{}, and \uavblue{}) were deployed in the \SI{60}{\minute} long run.
    The \ac{uav} performance is summarized in \autoref{tab:mission_stats} and the flight trajectories are plotted in~\autoref{fig:paths}.
    The first \ac{uav} (\uavred{}) took off just after the first \ac{ugv}, arrived to the first intersection, explored \SI{10}{\meter} of the tunnel section, returned to the intersection, flew to the cave branch where it collided with the Spot \ac{ugv}~(\autoref{fig:landing_events}a).
    The chronologically second deployed \ac{uav} was \uavblue{}, which went into the urban branch where it traveled to a vertical alcove with a phone artifact.
    Then it returned to the start of the urban section, where it hovered until exhausting the battery~(\autoref{fig:landing_events}c), because all viewpoints were blocked in its map corrupted by drift in the featureless urban corridor.
    The last deployed \ac{uav} was \uavgreen{} that explored the tunnel section, where it was blocked by a dynamically added artificial wall~(\autoref{fig:dynamic_wall}).
    After flying through a cluttered tunnel corridor, the \ac{uav} collided with a metal rod protruding from the wall~(\autoref{fig:landing_events}b).

    The maps and the trajectories of all our \ac{uav} flights during the prize round and the post-event testing are shown in~\autoref{fig:finals_trajs_over_map_overlay}, together with summary of the mapping errors from these flights in~\autoref{fig:finals_mapping_errors_histogram}.
    The distance of the \acp{uav} from the nearest obstacle during all flights in the prize round are shown in~\autoref{fig:uav_obstacle_dists}. 

    \subsubsection{Artifact detection discussion}
\begin{figure}
      \centering
      \includegraphics[width=0.33\textwidth]{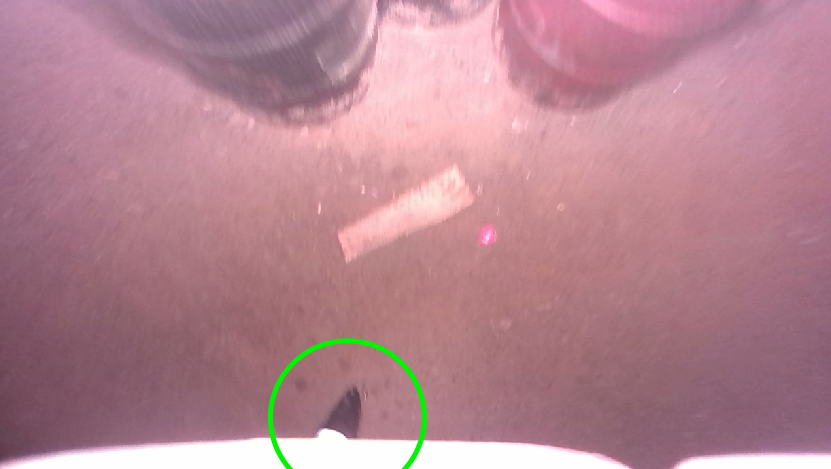}%
      \hfill
      \includegraphics[width=0.33\textwidth]{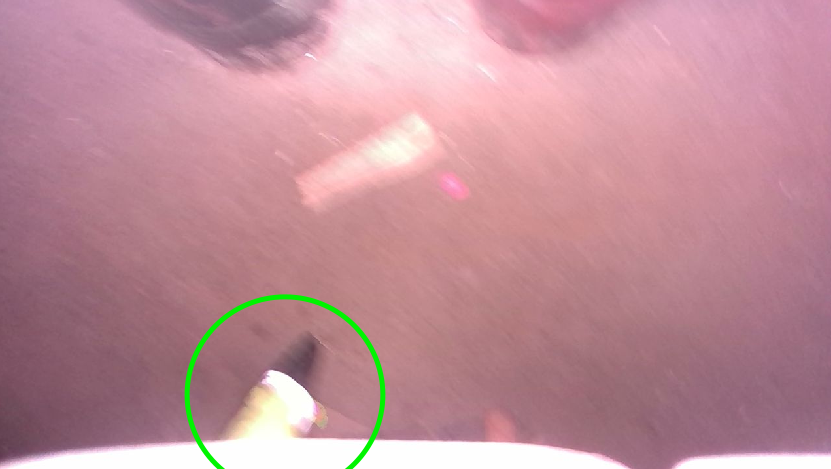}%
      \hfill
      \includegraphics[width=0.33\textwidth]{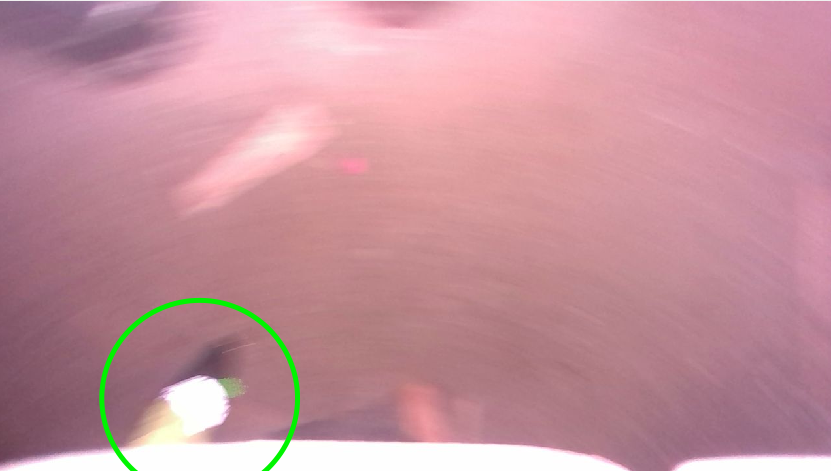}
      \caption{\label{fig:survivor_undetected} 
      The only three image frames of the survivor \textit{s2} captured by the downward-facing camera.
      The artifact was not detected as there is only a small part of the survivor's sleeve visible in the image, which is also degraded by motion blur.
      }
    \end{figure}

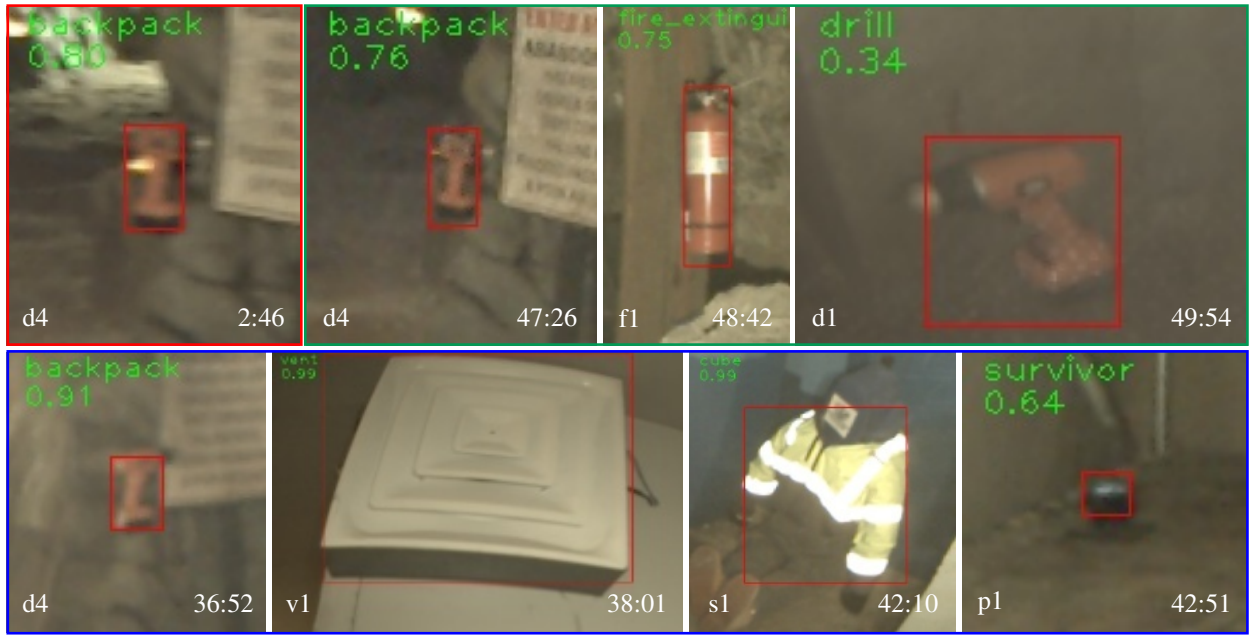
\begin{figure}
      \centering
      \newcommand{\toprowheight}{12.80em}
      \newcommand{\botrowheight}{10.60em}

      \begin{tikzpicture}
        \node[anchor=south west, inner sep=0] at (0,0) (images) {
      \begin{minipage}{\linewidth}
      \includegraphics[height=\toprowheight]{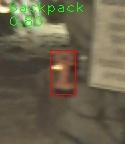}
      \hfill
      \includegraphics[height=\toprowheight]{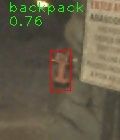}
      \hfill
      \includegraphics[height=\toprowheight]{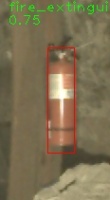}
      \hfill
      \includegraphics[height=\toprowheight]{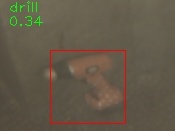}
      \vspace{-10pt}
      \\
      \includegraphics[height=\botrowheight]{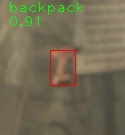}
      \hfill
      \includegraphics[height=\botrowheight]{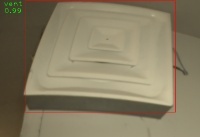}
      \hfill
      \includegraphics[height=\botrowheight]{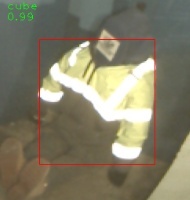}
      \hfill
      \includegraphics[height=\botrowheight]{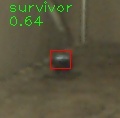}
      \end{minipage}};
        \begin{scope}
          \node[align=center] at (0.4, 4.2) {\color{white}d4};
          \node[align=center] at (4.4, 4.2) {\color{white}d4};
          \node[align=center] at (8.3, 4.2) {\color{white}f1};
          \node[align=center] at (10.9, 4.2) {\color{white}d1};
          \node[align=center] at (0.4, 0.4) {\color{white}d4};
          \node[align=center] at (3.9, 0.4) {\color{white}v1};
          \node[align=center] at (9.5, 0.4) {\color{white}s1};
          \node[align=center] at (13.1, 0.4) {\color{white}p1};

          \node[align=center] at (3.4, 4.2) {\color{white}2:46};
          \node[align=center] at (7.2, 4.2) {\color{white}47:26};
          \node[align=center] at (9.8, 4.2) {\color{white}48:42};
          \node[align=center] at (15.9, 4.2) {\color{white}49:54};
          \node[align=center] at (2.9, 0.4) {\color{white}36:52};
          \node[align=center] at (8.4, 0.4) {\color{white}38:01};
          \node[align=center] at (12.0, 0.4) {\color{white}42:10};
          \node[align=center] at (15.9, 0.4) {\color{white}42:51};

          \draw[draw=red,line width=1pt] (0.03,3.84) rectangle (3.932,8.35);
          \draw[draw=ForestGreen,line width=1pt] (3.985,3.84) rectangle (16.51,8.35);
          \draw[draw=blue,line width=1pt] (0.03,0.0) rectangle (16.51,3.75);

        \end{scope}
      \end{tikzpicture}
      \caption{\label{fig:detections} 
      Images of artifacts detected by the \acp{uav} in the final event.
      The color of the rectangle shows which \ac{uav} detected the artifact and at what mission time as shown in the bottom right corner.
      }
    \end{figure}

    The performance of the artifact detection and localization system is summarized in~\autoref{tab:detection_details}, and the number of artifacts detected by each \ac{uav} in~\autoref{tab:detection_stats}.
   A total of seven artifacts appeared in the camera images, and six artifacts were detected by the \ac{cnn}.
   The detections with estimated bounding boxes from all \acp{uav} are shown in~\autoref{fig:detections},
   The survivor \textit{s2} was seen in three frames of the bottom camera. However, only a small part of the survivor sleeve was visible and the images were further degraded by motion blur, as can be seen in~\autoref{fig:survivor_undetected}. Thus, the \ac{cnn} did not manage to detect the artifact.
   From the six detections, the cellphone artifact \textit{p1} was detected only on one image frame when the \ac{uav} \uavblue{} peeked into the vertical shaft in the urban part.
   However, as explained in~\autoref{sec:object_detection}, a total of four detections are necessary to create a hypothesis and to confirm the position, and thus this single detection was discarded.
   Another missed point was the survivor \textit{s1}, which was detected and localized within the \SI{5}{\meter} limit, but the artifact was labeled as a cube instead of a survivor.
   The hypothesis was merged with a high number of false positives and, consequently, the correct image was not sent to the operator, who could not determine the correct class to report.
   Both vent \textit{v1} and drill \textit{d1} were detected, localized, and correctly labeled.
   The drill \textit{d4} was incorrectly classified as a backpack, nevertheless, the operator reported the correct class based on the detection image.
   All three \acp{uav} detected the \textit{d4} drill, but \ac{uav} \uavgreen{} provided the highest accuracy, which is reported in~\autoref{tab:detection_details}.
   In total, four artifact hypotheses arrived to the base station with sufficient information for obtaining a point for the report.

    \begin{table}
     \caption{\label{tab:detection_stats}
      Statistics of artifact detection for each deployed \ac{uav} from the prize round of the final event.
      The \textit{seen} column yields the number of artifacts that appeared in the image of one of the on-board cameras.
      If the artifact was detected by the \ac{cnn}, it is listed in the \textit{detected} column and the detection is shown in~\autoref{fig:detections}.
      Artifacts that were \textit{confirmed} had enough consistent detections to establish a hypothesis.
      \textit{Confirmed unique} artifacts were not detected by another robot, including \acp{ugv}.
      }
     \centering
     \tablesize
     \begin{tabular}{lcccc}
       \toprule
       \multirow{2}{*}[0em]{\tablehdg{UAV}} & \multicolumn{4}{c}{\tablehdg{Artifacts}}\\\cmidrule(r){2-5}
                                 & \tablehdg{seen} & \tablehdg{detected} & \tablehdg{confirmed} & \tablehdg{confirmed unique}\\
       \midrule
       \tablehdg{Red} & 1 & 1 & 1 & 0\\
       \tablehdg{Green} & 4 & 3 & 3 & 1\\
       \tablehdg{Blue} & 4 & 4 & 3 & 1\\
       \bottomrule
     \end{tabular}
    \end{table}


    \begin{table}
     \caption{\label{tab:detection_details}
      Unique artifacts detected by lightweight \ac{cnn} running on-board \acp{uav} in real time.
      The total error $\mathbf{e_{tot}}$ of the artifact position is the sum of the \ac{uav} localization drift error $\mathbf{e_{loc}}$ and the error of estimating the artifact position $\mathbf{e_{est}}$ from the detected bounding box.
      Artifacts detected by more \acp{uav} are listed only once with values from the most accurate hypothesis among the \acp{uav}.
      The hypothesis was \textit{Confirmed} when more than four images were associated with it.
      Some artifacts were correctly detected and localized, but the wrong label was assigned to them. This is documented in the \textit{Correct class} column. 
      Even with a wrong label, the operator could still deduce the correct class by looking at the image sent with the hypothesis.
      Only one image was sent with each hypothesis, and if it was possible to deduce the correct class, then the image was listed as \textit{Correct image}.
      }
     \centering
     \tablesize

     \begin{tabular}{crcccrrr}
       \toprule
       \tablehdg{Artifact} & \tablehdg{Frames detected} & \tablehdg{Confirmed} & \tablehdg{Correct class} & \tablehdg{Correct image} & $e_{loc}$ (\si{\meter}) & $e_{est}$ (\si{\meter}) & $e_{tot}$ (\si{\meter})\\
       \midrule
       v1 & 27 & $\checkmark$ & $\checkmark$ & $\checkmark$ & 1.94 & 4.61 & 3.08\\
       s1 & 60 & $\checkmark$ & $\times$ & $\times$ & 2.93 & 4.57 & 2.89\\
       p1 & 1  & $\times$   &   $\times$ & $\times$ & - & - & -\\
       d4 & 11 & $\checkmark$ & $\times$ & $\checkmark$ & 0.77 & 1.61 & 1.30\\
       f1 & 13 & $\checkmark$ & $\checkmark$ & $\checkmark$ & 0.85 & 1.33 & 1.31\\
       d1 & 9  & $\checkmark$ & $\checkmark$ & $\checkmark$ & 1.46 & 2.30 & 1.55\\
       \bottomrule
     \end{tabular}
    \end{table}



    \subsection{DARPA SubT Final Event Virtual Track}
    \label{sec:darpa_virtual}

    In parallel to the Systems Track, the competition was also running in the simulated form of the Virtual Track.
    The teams had to submit a solution consisting of docker images of a robotic team put together within a limited budget to buy the robots and their sensory packages.

    The Systems Track included a single run (with two preliminary rounds) conducted in a single world and was therefore focused on the reliability of the robots, which had to overcome challenging terrain with narrow passages and adverse conditions for perception.
    On the other hand, the virtual teams were deployed three times in each of the eight worlds, ranging from vast models of artificially created environments to scanned courses from the previous events, including the final event course.
    Moreover, in the Virtual Track, the whole mission must be fully autonomous and no human interventions are possible.
    The purpose of the virtual event was to evaluate the high-level planning, cooperation, decision-making, and efficient coverage of the large worlds.
    As the cooperative searching strategy is one of the core contributions of this work, we have presented the results from the virtual course here as most of the worlds allowed for efficient deployment and cooperation of the multi-robot teams.

    \subsubsection{Differences from the Systems Track}
    
    The simulation model of the \ac{imu} provides much better data compared to the real sensor with the same parameters. Thus is due to the measurements in the simulation not being corrupted by propeller-induced vibrations, wind gusts, or saturation, as well as having the \ac{imu} rigidly attached to the \ac{uav} body with known extrinsic parameters.
    The higher quality of the simulated data allows for the use of \ac{lidar}-inertial odometry. In addition to the \ac{lidar}, it also relies on the \ac{imu} preintegration in its optimization process, thus providing a smooth and drift-free position estimate, even when there are few geometrically rich features present.
    Specifically, the \ac{liosam}~\cite{liosam2020shan} algorithm was chosen for its low drift and high precision over the \ac{aloam} deployed in the Systems Track.
    Both algorithms are detailed in~\autoref{sec:localization}.

    Reporting of the found artifacts is handled by the operator in the Systems Track, which is not possible in the fully autonomous Virtual Track.
    A virtual artifact reporter algorithm was developed to gather artifact hypotheses from all robots and decide which hypotheses are the most likely to score a point (described in detail in~\autoref{sec:arbiter_for_artifact_reporting}). 

    The control interface of the simulated \ac{uav} was also different from the real one. 
    While the \ac{fcu} of the real \ac{uav} accepted attitude rate commands generated by the \ac{se3} controller, the simulated \ac{uav} was controlled on a higher level by velocity commands. This did not allow for precise control of the \ac{uav} motion, as was the case for the low-level attitude rate control.

    \subsubsection{Virtual Track results}

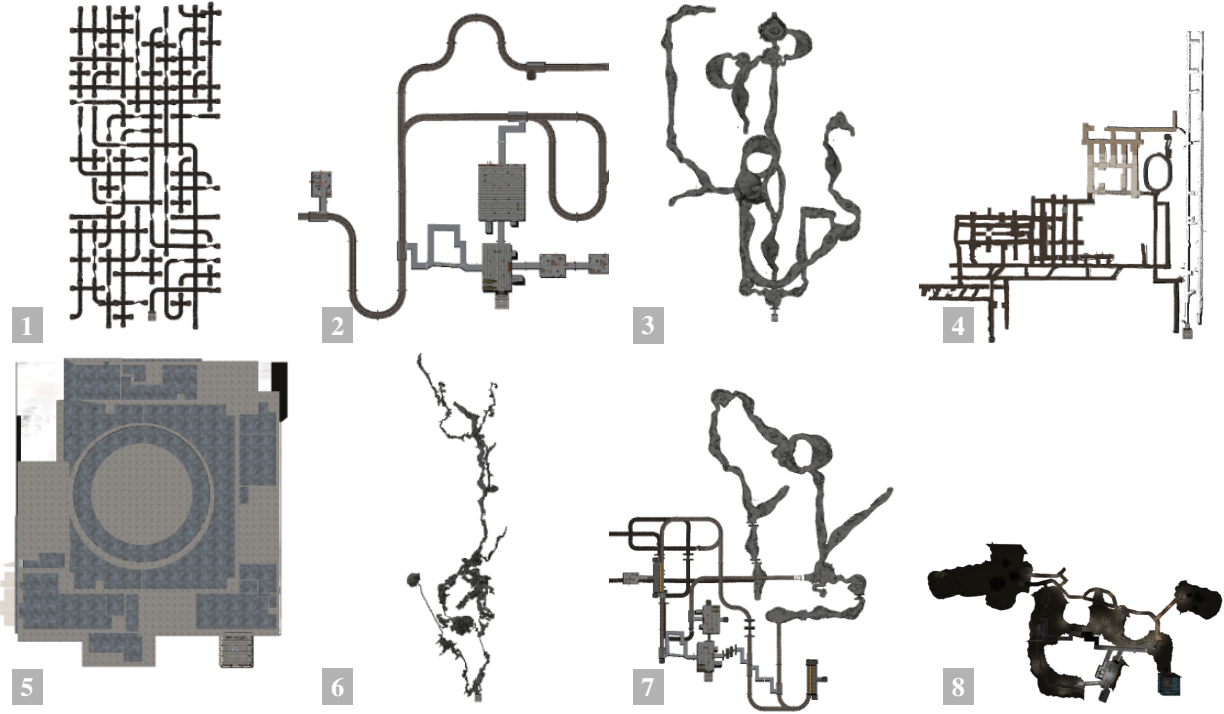
\begin{figure} [ht]
    \newcommand{\imheight}{12.50em}
    \newcommand{\xcap}{1.5em}
    \newcommand{\ycap}{1.0em}
    \newcommand{\fillopa}{0.3}
    \centering

    \begin{tikzpicture}
      \node[anchor=south west,inner sep=0] (b) at (0,0) {\adjincludegraphics[width=0.25\textwidth,trim={{0.30\width} {0.40\height} {0.30\width} {0.15\height}},clip]{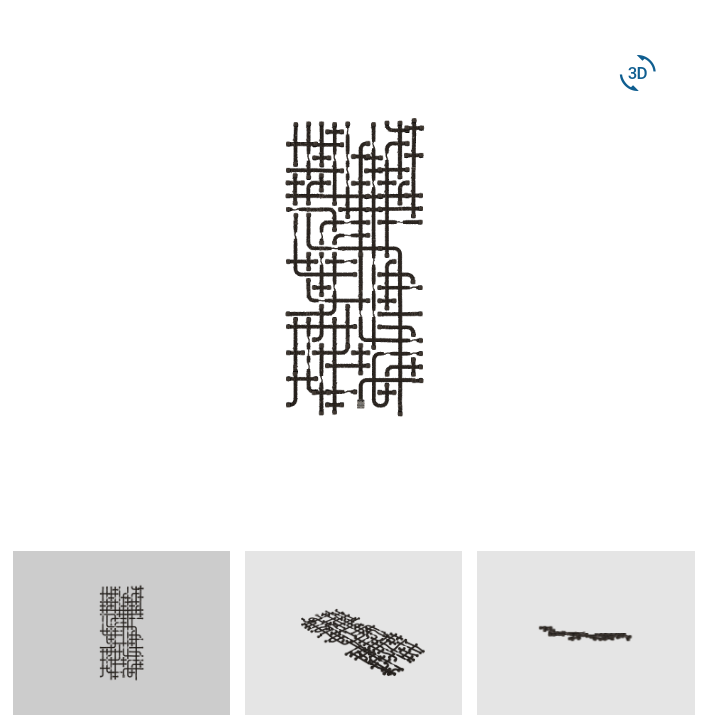}};%
    \begin{scope}[x={(b.south east)},y={(b.north west)}]
  \node[fill=black, fill opacity=\fillopa, text=white, text opacity=1.0] at (\xcap, \ycap) {\textbf{1}};
      \end{scope}
    \end{tikzpicture}%
    \begin{tikzpicture}
      \node[anchor=south west,inner sep=0] (b) at (0,0) {\adjincludegraphics[width=0.25\textwidth,trim={{0.30\width} {0.40\height} {0.30\width} {0.15\height}},clip]{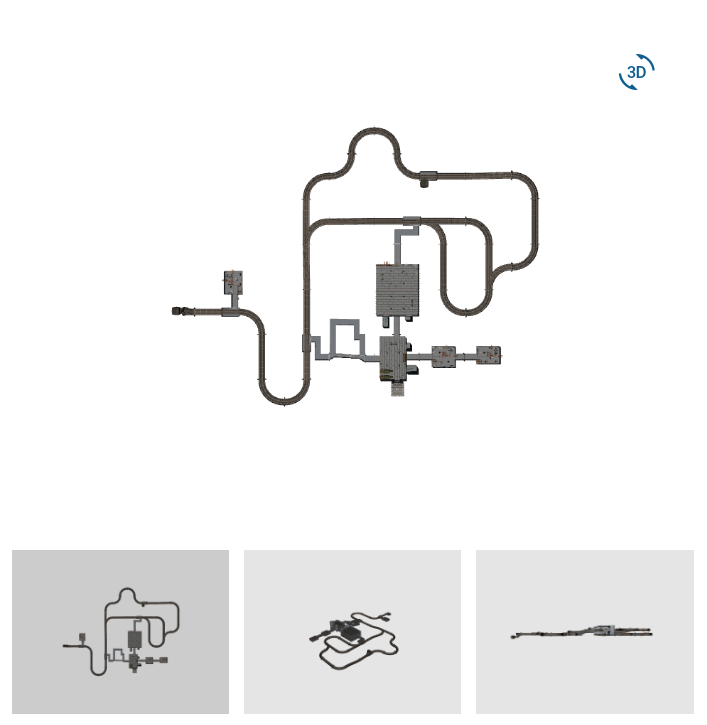}};%
    \begin{scope}[x={(b.south east)},y={(b.north west)}]
  \node[fill=black, fill opacity=\fillopa, text=white, text opacity=1.0] at (\xcap, \ycap) {\textbf{2}};
      \end{scope}
    \end{tikzpicture}%
    \begin{tikzpicture}
      \node[anchor=south west,inner sep=0] (b) at (0,0) {\adjincludegraphics[width=0.25\textwidth,trim={{0.30\width} {0.40\height} {0.30\width} {0.15\height}},clip]{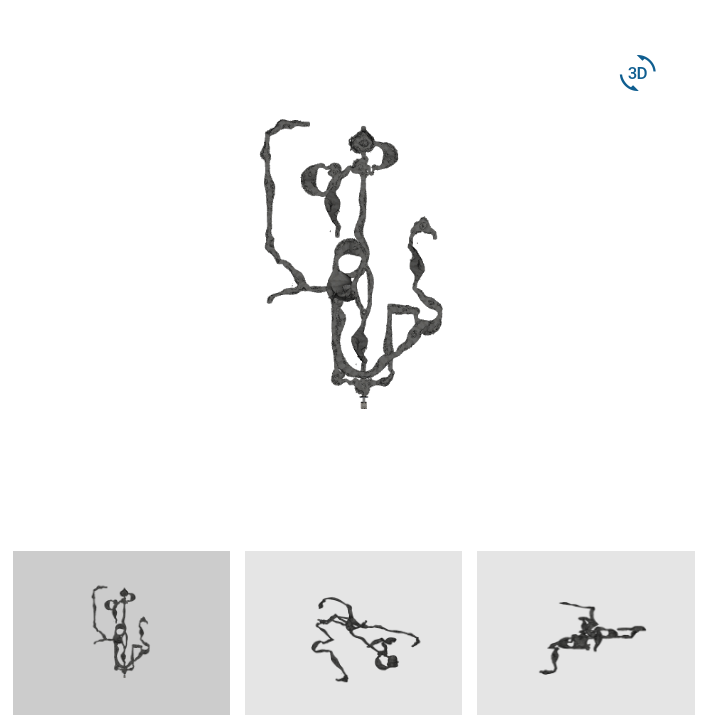}};%
    \begin{scope}[x={(b.south east)},y={(b.north west)}]
  \node[fill=black, fill opacity=\fillopa, text=white, text opacity=1.0] at (\xcap, \ycap) {\textbf{3}};
      \end{scope}
    \end{tikzpicture}%
    \begin{tikzpicture}
      \node[anchor=south west,inner sep=0] (b) at (0,0) {\adjincludegraphics[width=0.25\textwidth,trim={{0.30\width} {0.40\height} {0.25\width} {0.15\height}},clip]{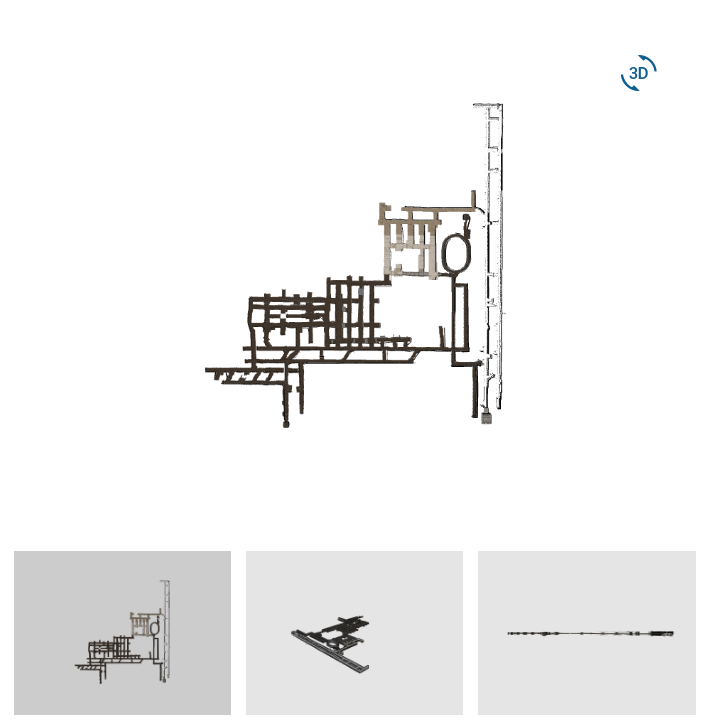}};%
    \begin{scope}[x={(b.south east)},y={(b.north west)}]
  \node[fill=black, fill opacity=\fillopa, text=white, text opacity=1.0] at (\xcap, \ycap) {\textbf{4}};
      \end{scope}
    \end{tikzpicture}

    \begin{tikzpicture}
      \node[anchor=south west,inner sep=0] (b) at (0,0) {\adjincludegraphics[width=0.25\textwidth,trim={{0.30\width} {0.40\height} {0.30\width} {0.15\height}},clip]{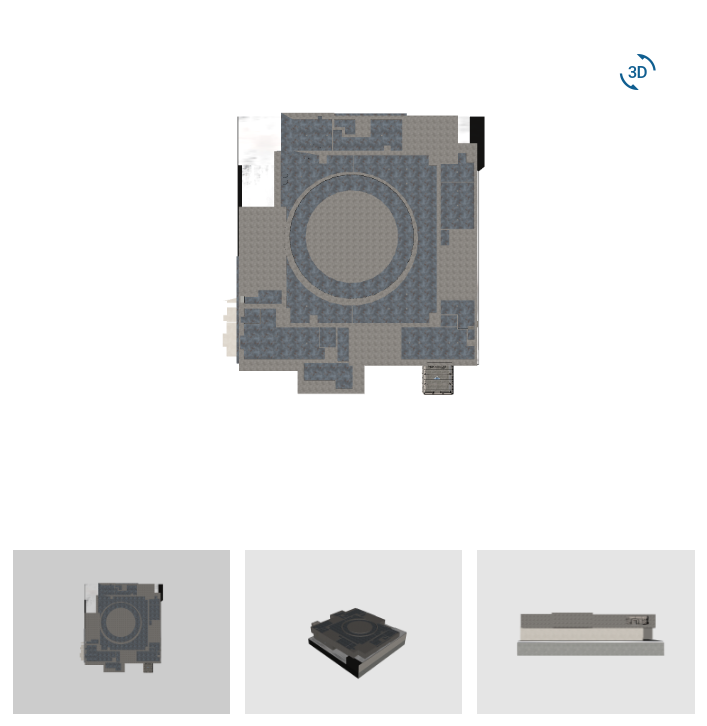}};%
    \begin{scope}[x={(b.south east)},y={(b.north west)}]
  \node[fill=black, fill opacity=\fillopa, text=white, text opacity=1.0] at (\xcap, \ycap) {\textbf{5}};
      \end{scope}
    \end{tikzpicture}%
    \begin{tikzpicture}
      \node[anchor=south west,inner sep=0] (b) at (0,0) {\adjincludegraphics[width=0.25\textwidth,trim={{0.30\width} {0.40\height} {0.30\width} {0.15\height}},clip]{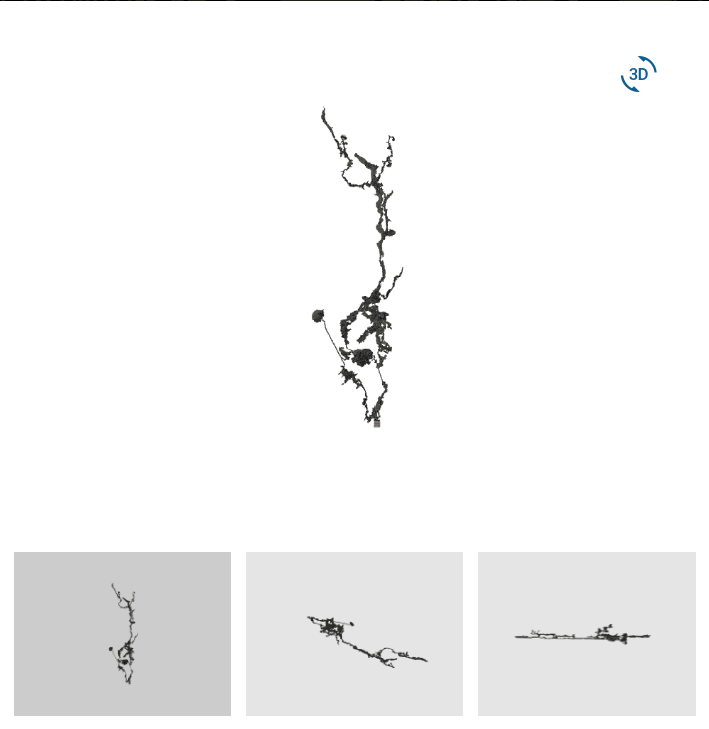}};%
    \begin{scope}[x={(b.south east)},y={(b.north west)}]
  \node[fill=black, fill opacity=\fillopa, text=white, text opacity=1.0] at (\xcap, \ycap) {\textbf{6}};
      \end{scope}
    \end{tikzpicture}%
    \begin{tikzpicture}
      \node[anchor=south west,inner sep=0] (b) at (0,0) {\adjincludegraphics[width=0.25\textwidth,trim={{0.30\width} {0.40\height} {0.25\width} {0.10\height}},clip]{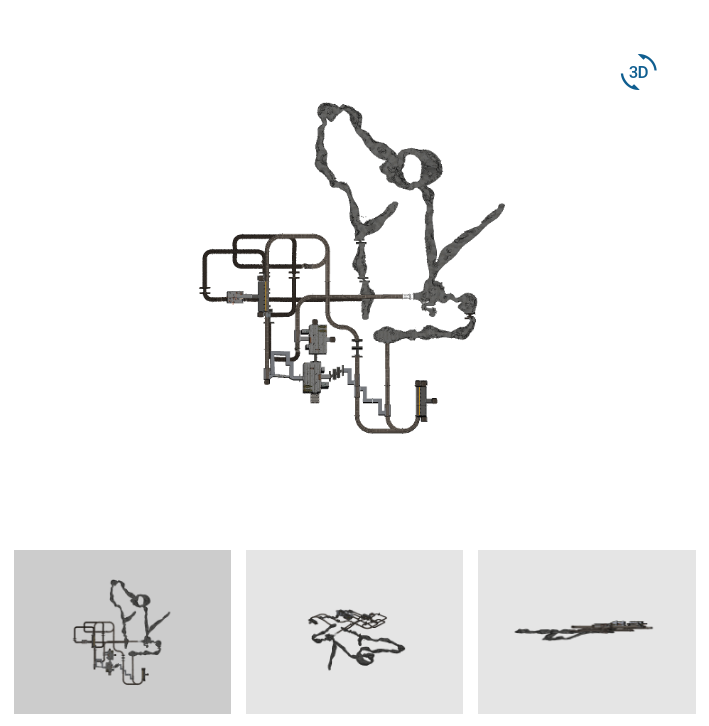}};%
    \begin{scope}[x={(b.south east)},y={(b.north west)}]
  \node[fill=black, fill opacity=\fillopa, text=white, text opacity=1.0] at (\xcap, \ycap) {\textbf{7}};
      \end{scope}
    \end{tikzpicture}%
    \begin{tikzpicture}
      \node[anchor=south west,inner sep=0] (b) at (0,0) {\adjincludegraphics[width=0.25\textwidth,trim={{0.05\width} {0.25\height} {0.05\width} {0.15\height}},clip]{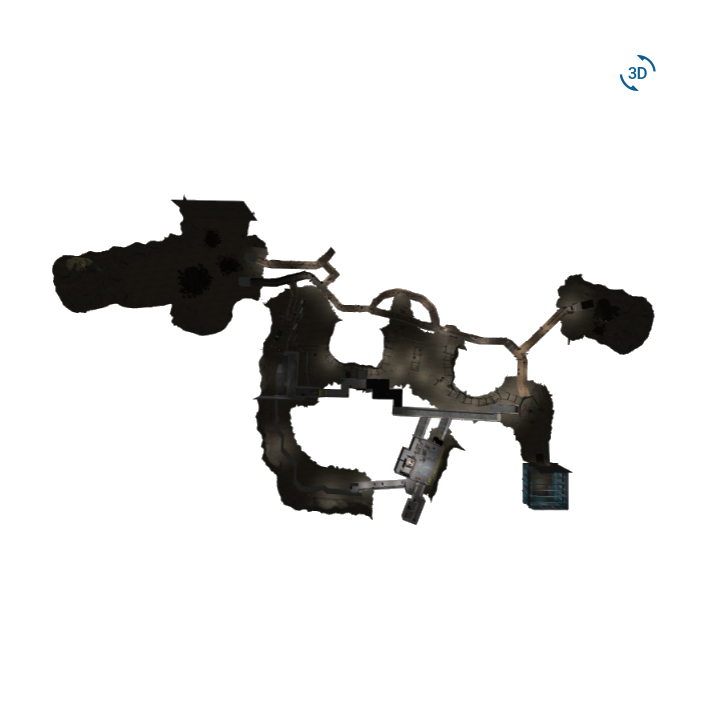}};%
    \begin{scope}[x={(b.south east)},y={(b.north west)}]
  \node[fill=black, fill opacity=\fillopa, text=white, text opacity=1.0] at (\xcap, \ycap) {\textbf{8}};
      \end{scope}
    \end{tikzpicture}%
    \caption{
      All eight worlds used in the Virtual Track of the \ac{darpa} \ac{subt} Finals.
      The worlds 1, 2, 3, and 7 are built from tiles that were used in the preliminary and practice rounds.
      World 4 is the model of the NIOSH research mine, where the tunnel circuit was held.
      Similarly, world 5 corresponds to the model of the location of the urban circuit --- the unfinished Satsop nuclear power plant.
      World 6 is a model of a narrow cave system. World 8 is modeled based on the System Track Finals.
  }
    \label{fig:virtual_worlds}
  \end{figure}

\begin{figure}
      \centering
      \includegraphics[width=0.48\textwidth]{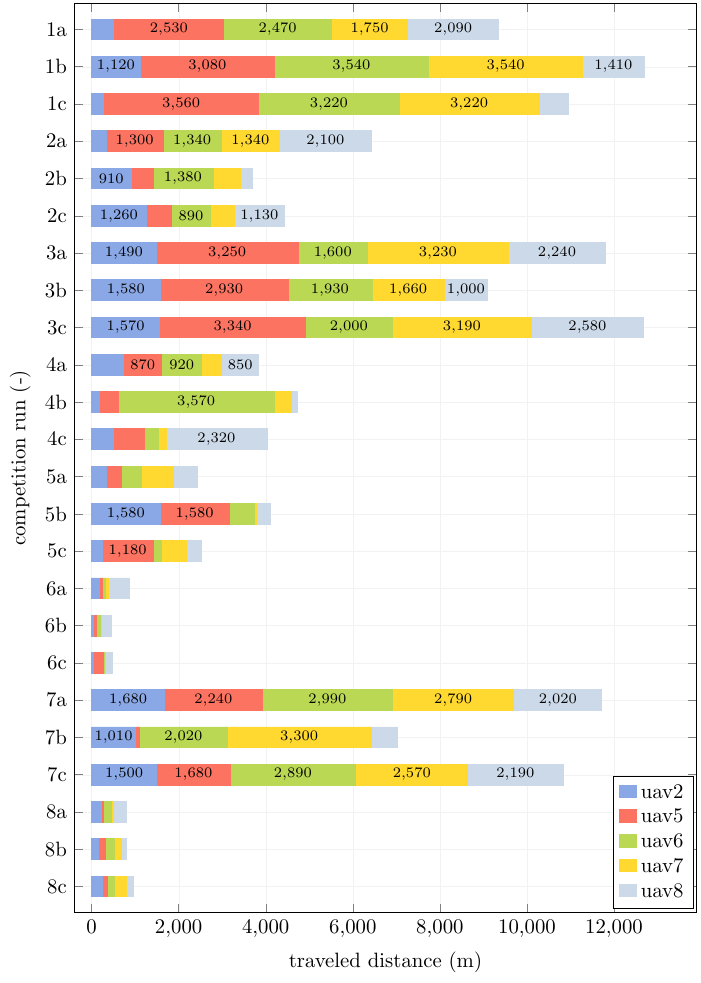}
      \includegraphics[width=0.48\textwidth]{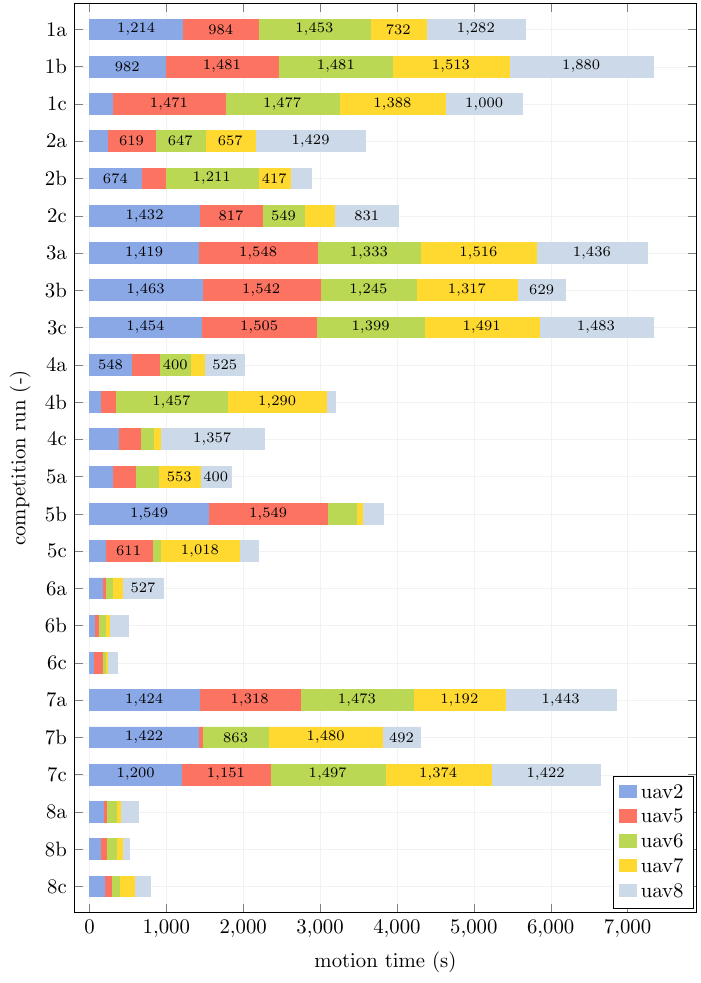}
      \includegraphics[width=0.96\textwidth]{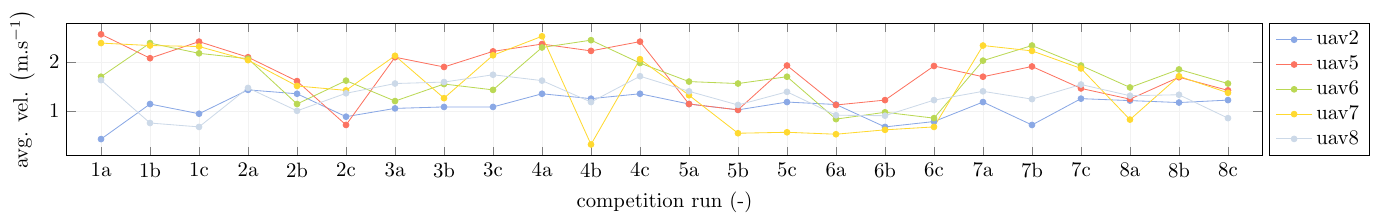}
      \caption{\label{fig:virtual_travel_dist_and_time}
        Overall traveled distance, time of active motion, and average velocity of particular \acp{uav} in all runs of the SubT Virtual Finals. The maximum traveled distance throughout all runs was achieved by UAV5 in run 1c (\SI{3560}{\meter}). The maximum active time was achieved by UAV2 in run 5b (\SI{1539}{\second}). The presented average velocity incorporates the entire flight, including hovering states.
      }
    \end{figure}

\begin{figure}
      \centering
      \includegraphics[width=0.96\textwidth]{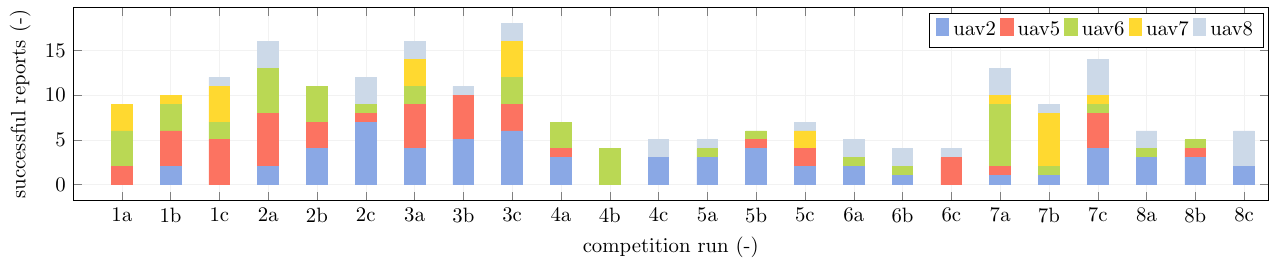}
      \includegraphics[width=0.96\textwidth]{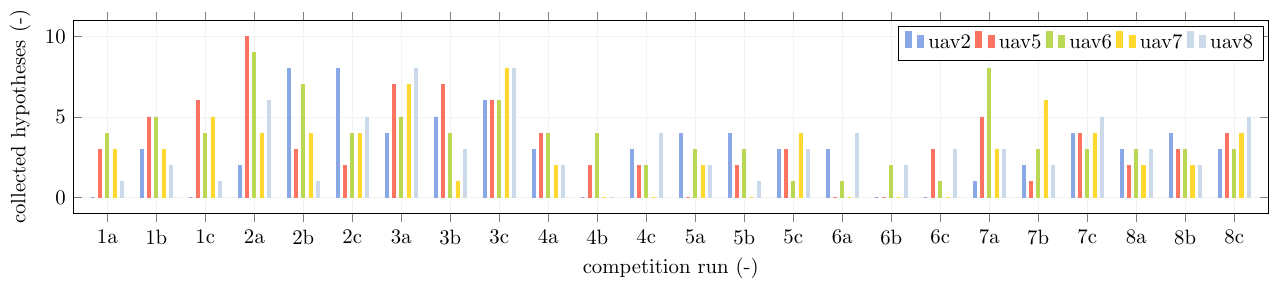}
      \caption{\label{fig:virtual_reports_and_hypotheses}
        Distribution of successful reports among the \acp{uav} in particular runs of SubT Virtual Finals (top) and the number of valid hypotheses collected throughout particular runs by individual robots (bottom). The number of successful reports of individual robots is mostly influenced by the ordering of robots and their delayed starts in the mission.
      }
    \end{figure}

    In the virtual deployment, our team consisted of five \acp{uav} and two \acp{ugv}. The \acp{uav} were the superior platform in the Virtual Track due to their greater movement speed, smaller form-factor, and better mobility to fly over terrain untraversable by the \acp{ugv}.
    We deployed two \acp{ugv} to build a communication network consisting of breadcrumbs dropped at the edges of the wireless signal range. This allowed for the \acp{uav} to maximize the time for searching for artifacts as they could return to the nearest breadcrumb instead of to the base station back at the staging area.
    Our solution achieved \nth{2} place with a total of 215 scored points.
    \autoref{tab:virtual_score} summarizes the points scored by the top three teams on each world of the Virtual Track~(\autoref{fig:virtual_worlds}).
    The lower number of points on worlds 4, 5, 6, and 8 can be explained by the fact that these worlds were not made of the tiles that were used in the qualification and practice worlds.
    The details on traveled distance and collected hypotheses by particular \acp{uav} during all runs of the SubT Virtual Finals are provided in~\autoref{fig:virtual_travel_dist_and_time} and \autoref{fig:virtual_reports_and_hypotheses} respectively.  

    \begin{table}
     \caption{\label{tab:virtual_score}
      The score achieved by the top three teams on each world of the Virtual Track. 
      The reported values are the sums of three runs on each world.
      }
     \centering
     \tablesize
     \begin{tabular}{lccccccccc}
       \toprule
       \tablehdg{World} & \tablehdg{1} & \tablehdg{2} & \tablehdg{3} & \tablehdg{4} & \tablehdg{5} & \tablehdg{6} & \tablehdg{7} & \tablehdg{8} & \tablehdg{total} \\
      \cmidrule{2-10}
       \tablehdg{Dynamo} & 21 & 52 & 48 & 18 & 15 & 11 & 44 & 14 & 223\\
       \tablehdg{CTU-CRAS-NORLAB} & 31 & 39 & 45 & 16 & 18 & 13 & 36 & 17 & 215 \\
       \tablehdg{Coordinated Robotics} & 44 & 41 & 27 & 23 & 17 & 14 & 26 & 20 & 212 \\
       \bottomrule
     \end{tabular}
    \end{table}

    \section{Lessons learned and future work}
    \label{sec:lessons_learned}

    In this section, we present our view on the state of the \ac{sar} \acp{uav}, the lessons learned, which problems are solved, and what areas require more research to achieve reliable performance suitable for deployment as a tool for assisting rescue workers.
    These findings were collected throughout the preparation for as well as during the DARPA SubT Competition, which aimed to push the state of the art of \ac{sar} robotics. Furthermore, this discussion should be of some interest to the community as we highlight aspects that could be explored in future research and development.
In general, most of the individual subproblems, such as localization, mapping, detection, and communication, are solved to the point of being capable of performing an autonomous mission in extremely challenging conditions.
The developed algorithms are now used in actual field deployment instead of just laboratories and simulations, which introduces disturbances, noise, dust, and other detrimental effects that negatively impact the algorithms' performance and reliability.
It is essential to focus on the reliability of the employed methods to make the \acp{uav} a valuable asset to the \ac{sar} team.

The localization method based on 3D \ac{lidar} provides precise position estimates, even under severe degradation by dust. 
However, as proved by the \ac{uav} \uavblue{}, the estimate can begin to drift when the solved optimization is ill-conditioned due to low-variance geometry, typically in long corridors with straight walls.
The unpredictable nature of subterranean environments requires a localization method that is reliable and drift-free under arbitrary conditions.
Solutions based on detecting geometrical degeneracy, and multi-modal fusion of \ac{lidar} and visual methods were described in~\autoref{sec:sota_slam}.
The results seem promising but due to high unpredictability and challenges of subterranean environments more research in localization algorithms is still required for truly robust pose estimation in arbitrary conditions.

In addition to map drift caused by errors in the localization, the volumetric occupancy grid did not contain the smaller obstacles like ropes, cables, and thin poles, which led to the collision of \ac{uav} \uavgreen{} as seen in~\autoref{fig:landing_events}b.
Although some \ac{lidar} rays hit these thin obstacles, the occupied cells generated by these rays were often changed to free when multiple rays that passed through these cells hit the wall behind them.
As a result, the navigation pipeline planned a path through these cells that appeared free, but contained a thin metal pole, causing a collision.
The ability to traverse narrow passages is also impaired since the passages appear narrower than they really are due to grid discretization.
We propose to locally increase the resolution of the grid of DenseMap on demand to approximate the free space more accurately, while keeping the scan integration times bounded.
This approach is however only a partial solution as the need for a more granular resolution might not always be reliably detected.
Consequently, the need arises for a flexible map that is not bound by fixed cell size, similarly to the SphereMap, possibly based on surfel mapping as seen in~\cite{behley2018efficient}, or based on \ac{gmm}~\cite{o2018variable}.

\begin{figure} [ht]
    \newcommand{\imheight}{12.50em}
    \newcommand{\xcap}{0.95em}
    \newcommand{\ycap}{0.8em}
    \newcommand{\fillopa}{0.3}
    \centering

    \begin{tikzpicture}
      \node[anchor=south west,inner sep=0] (b) at (0,0) {\adjincludegraphics[height=\imheight,trim={{0.0\width} {0.2\height} {0.3\width} {0.1\height}},clip]{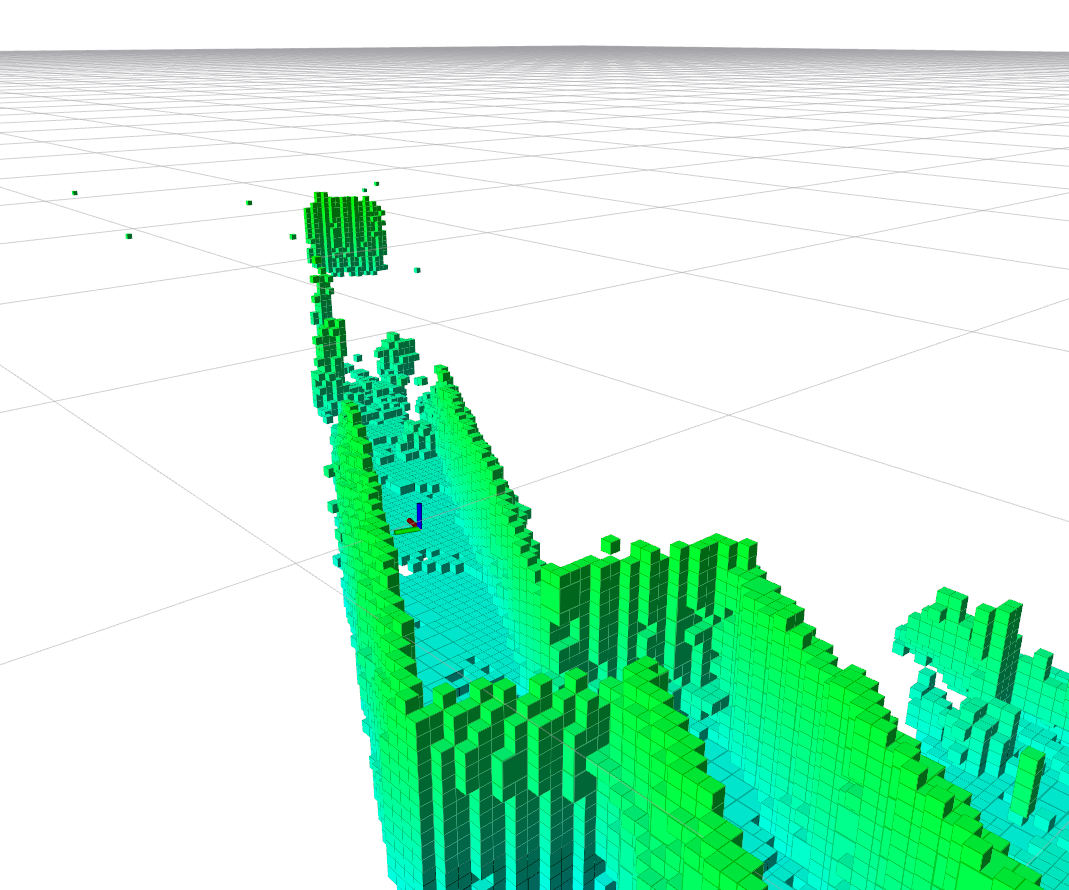}};%
    \begin{scope}[x={(b.south east)},y={(b.north west)}]
      \node[fill=black, fill opacity=\fillopa, text=white, text opacity=1.0] at (\xcap, \ycap) {\textbf{(a)}};
      \end{scope}
    \end{tikzpicture}%
    \begin{tikzpicture}
      \node[anchor=south west,inner sep=0] (b) at (0,0) {\adjincludegraphics[height=\imheight,trim={{0.0\width} {0.0\height} {0.0\width} {0.0\height}},clip]{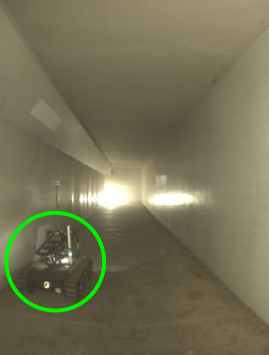}};%
      \node[fill=black, fill opacity=\fillopa, text=white, text opacity=1.0] at (\xcap, \ycap) {\textbf{(b)}};
    \end{tikzpicture}%
    \begin{tikzpicture}
      \node[anchor=south west,inner sep=0] (b) at (0,0) {\adjincludegraphics[height=\imheight,trim={{0.0\width} {0.2\height} {0.4\width} {0.1\height}},clip]{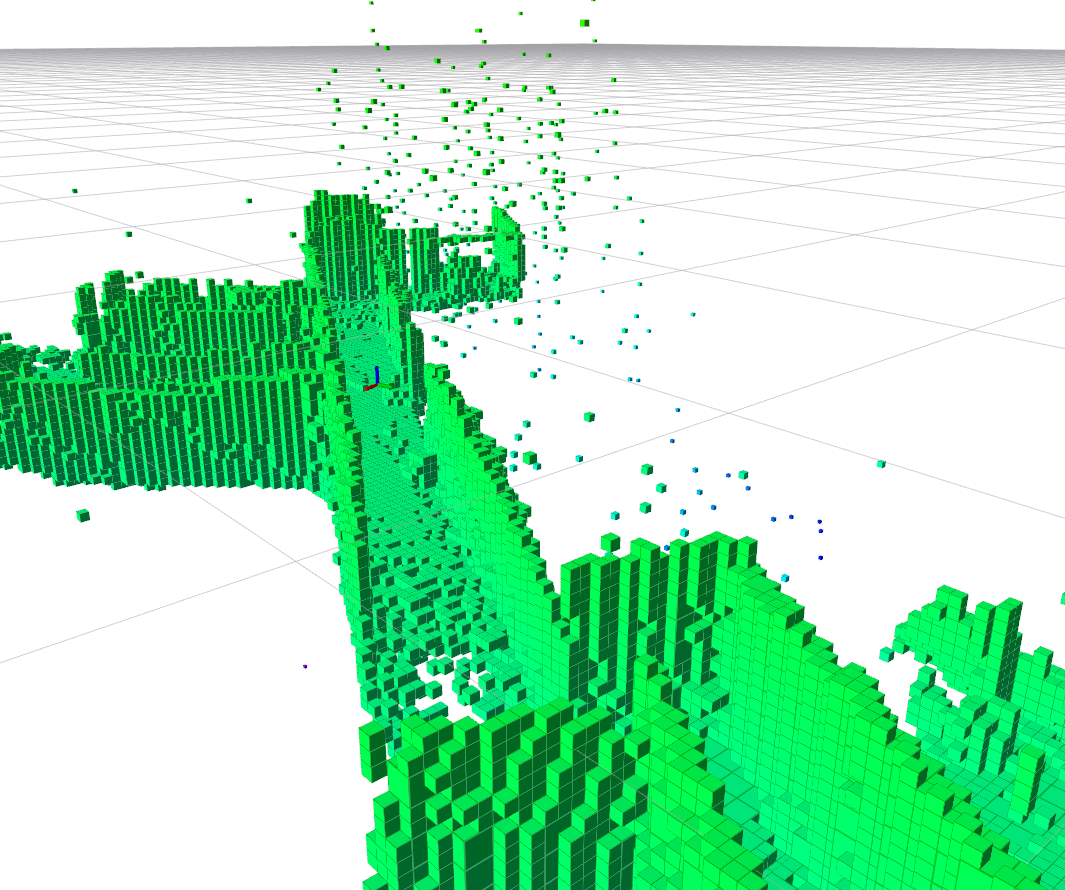}};%
    \begin{scope}[x={(b.south east)},y={(b.north west)}]
      \node[fill=black, fill opacity=\fillopa, text=white, text opacity=1.0] at (\xcap, \ycap) {\textbf{(c)}};
      \end{scope}
    \end{tikzpicture}%
    \caption{
      DenseMap before (a) approaching the \ac{ugv} with LeiShen C16 \ac{lidar} (b) and (c) when it gets corrupted by random points in the \ac{fov} of the \ac{lidar} mounted on the \ac{uav} after flying in close vicinity ($\approx$~\SI{1}{\meter}) to the \ac{ugv}.
      Notice, a few false positives were integrated into the map even when the \ac{uav} was \SI{8}{\meter} away from the \ac{ugv} (a).
  }
    \label{fig:lidar_interference}
  \end{figure}
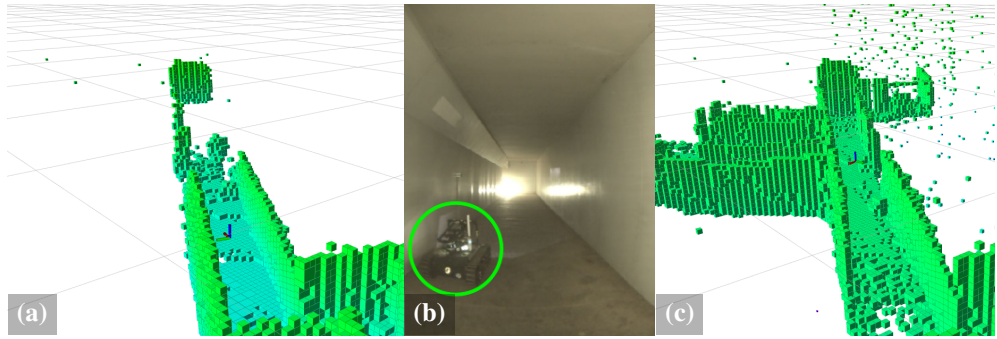

We experienced a surprising issue when our \ac{uav} equipped with the Ouster OS0-128 \ac{lidar} was passing around a \ac{ugv} with LeiShen C16 \ac{lidar}.
The rays emitted by the LeiShen corrupted some of the Ouster measurements, which manifested as points in random distance within the \ac{fov} of the \ac{lidar}.
These false positives were not filtered out by the intensity filter from~\autoref{sec:filtering_observation_noise}, because the intensities fall into the same range of values as true positives.
As a result, the points get integrated into the map, as shown in~\autoref{fig:lidar_interference}. Nevertheless, the performance of the \ac{uav} was not degraded as the navigation pipeline is robust to such sparse noise.
This experience highlights the importance of testing the compatibility of robotic platforms deployed in heterogeneous teams.

The flight time of the \ac{uav} over \SI{20}{\minute} was achieved as the payload was limited only to crucial components.
However, the presence of only a single computational unit without \ac{cnn} acceleration or dedicated \ac{gpu} led to compromises in the artifact detection \ac{cnn}.
Large-size models such as YOLOv3~\cite{redmon2018yolov3} were too slow for achieving satisfactory frame rates on the \ac{cpu}, so lighter models had to be used.
As explained in~\autoref{sec:object_detection}, the lightweight MobileNetV2 \ac{cnn} allowed for lightweight models (\SI{7}{\mega \byte}) that could fit into the cache of the \ac{cpu}.
Furthermore, the OpenVino framework supports accelerating the \ac{cnn} on the \ac{gpu} integrated with the \ac{cpu}, which helped to achieve sufficient frame rates.
Although the lightweight model successfully detected all artifact types, the labeling was not very reliable and many false positives were detected.
This impacted the artifact localization process, as the false positives were fused into the artifact hypotheses, which shifted the estimate further from the true position.
Also, the images of these false positives were sometimes sent as the representative image of the hypothesis.
Thus, the operator could not correctly decide the artifact class when the label produced by the \ac{cnn} was incorrect.
When payload capacity prevents the use of more capable hardware, the issue must be compensated by the sensing strategy.
In contrast to \acp{ugv} the mobility of \acp{uav} allows reaching closer to the artifact to verify the detection. 
Approaches of perception-driven navigation can improve the performance of the lightweight detector by planning a trajectory to inspect the artifact from a closer distance and other angles after the initial detection.

Although our platform is quite compact, it could not pass through all of the narrow passages, even during the post-event testing.
Apart from the discrete map and conservatively set distance from obstacles, the size of the \ac{uav} prevented flying through some of the narrower passages of the circuit. 
As even smaller passages are to be expected during deployment of robots in real \ac{sar} scenarios, the \ac{uav} platforms should be further miniaturized to allow safe traversal of narrow passages.
When such miniaturization is not possible due to, e.g., insufficient payload of smaller platforms, a heterogeneous aerial team consisting of both large and small platforms can be deployed,
In such case, the large platform carrying a \ac{lidar} can command and send position corrections to smaller visually-localized \ac{uav} that can inspect tight narrow passages that are unreachable by the large \ac{uav}.

A mutual collision avoidance module is a necessity for any application where multiple robots share the same workspace.
The developed priority-based module uses the already shared information about the robots' positions when communication is available, as it should since the risk of collision arises when robots are in close proximity.
This module prevented collisions in the Virtual Track, where despite the vastness of most of the worlds, the collisions happened often in the practice runs before implementing the collision avoidance.
We decided against using the collision avoidance module in the Systems Track.
This was done as the robots could easily become deadlocked in tight corridors and also due to the collision probability being quite low because of the delay between each \acp{uav} launch. 
Additionally, the operator could override the full autonomy to prevent collision, if necessary.
Nevertheless, the \ac{uav} \uavred{} collided with a Spot \ac{ugv} shortly after the start of the run, which could have been prevented if collision avoidance was enabled.
A deadlock-free solution based on agent theory approaches can be devised for situations when communication is available, and behavior prediction methods can provide a backup when communication is not possible.


    \section{Conclusion}
    \label{sec:conclusion}

This paper has presented the complex \ac{uav} system deployed in the final round of the \ac{darpa} \ac{subt} Challenge after 3 years of development and testing in numerous real world demanding environments (including gold mine, coal mine, abandoned nuclear power plant, caverns, military fortress, natural caves, old factory hall, subway station, etc.).
Based on these unique opportunities and experience, we have designed both the hardware \ac{uav} platform and the multi-\ac{uav} software with a focus on the exploration of such vast, complicated, and varying environments.

In the Systems Track of \ac{darpa} \ac{subt} Challenge, three \acp{uav} were deployed alongside ground robots into the competition course consisting of a heterogeneous environment of the tunnel, urban, and cave sections, where the aerial team detected and localized four artifacts and traveled \SI{492}{\meter} in total.
The austere conditions of the circuit, such as narrow passages, dust, featureless corridors, and dynamic obstacles, tested the reliability of the system as a whole, including the hardware design of a compact platform with a considerable flight time of \SI{25}{\minute}.
Most of the testing was realized in environments where it performed exceptionally well, including a former brewery where the UAV had to explore an abandoned building with partially collapsed floor and ceiling, or during the exploration of Byci Skala (Bull Rock Cave) in the Moravian Karst cavern system.
Compared to ground robots the \acp{uav} could search a larger volume of space because they could easily fly over any encountered problematic terrain such as mud, water, and rubble and thus had an advantage in the exploration of unknown terrains with unexpected obstacles.
Furthermore, the worlds of the Virtual Track of the competition were also very large; even with our \ac{uav} possessing a \SI{25}{\minute} flight time and fast dynamics, they were not able to reach the furthest parts of some worlds.
Although our system was designed primarily for these large-scale environments, its performance in the challengingly tight corridors of the prize round was also impressive. 
The difficulty of \ac{uav} deployment in such adverse environments motivated numerous achievements beyond the state of the art that are summarized in this paper.
Many lessons were learned in the process that could facilitate and support designing complex robotic systems in similar applications in the future.

A larger team of five aerial robots was deployed in the Virtual Track, alongside two \acp{ugv}.
By employing the proposed cooperative exploration strategies based on topological map sharing, the exploration effort of our team was spread out over a wider area. Together with dynamic flight and reliable artifact detection/localization, this helped to achieve the \nth{2} place with 215 scored points.
Moreover, seven out of the nine participating teams used our X500 \ac{uav}, which was modeled according to the specification of the physical platform thanks to its long flight time, a wide array of sensors, modest size, and reasonable price.

Based on the successful deployment in the \ac{darpa} \ac{subt}, which focused on providing challenging conditions typically encountered during rescue missions in underground environments, we conclude that the presented \ac{uav} system is a valuable addition to teams of first responders, as it can provide situational awareness and even find survivors after a catastrophe without risking the lives of rescuers in dangerous environments.


    \subsubsection*{Acknowledgements}
    \label{sec:acknowledgements}
    We would like to thank the members of the CTU-CRAS-NORLAB team who participated in the design and development of the \ac{uav} and \ac{ugv} hardware platforms, software development, simulations, testing and general support.
    Namely:
Ruslan	Agishev,
Afzal	Ahmad,
Teymur	Azayev,
Jan	Bayer,
Tommy	Bouchard-Lebrun,
Petr	Čížek,
Simon-Pierre	Deschênes,
Jan	Faigl,
Olivier	Gamache,
Alexandre	Guénette,
Bedřich	Himmel,
Jakub	Janoušek,
Tomáš	Krajník,
Vladimír	Kubelka,
Denis	Ouellet,
Tomáš	Petříček,
François	Pomerleau,
Miloš	Prágr,
Tomáš	Rouček,
Vojtěch	Šalanský,
Martin	Škarytka,
Vojtěch	Spurný,
Arsen	Tkachev,
Maxime	Vaidis,
Volodymyr	Zabulskyi,
Karel	Zimmermann,
and Martin	Zoula.

    This work was partially funded 
    by the \acf{darpa}, 
    by the CTU grant no. SGS20/174/OHK3/3T/13, 
    by the Czech Science Foundation (GAČR) under research project no.	20-29531S, 
    by TAČR project no. FW03010020, 
    by the OP VVV funded project CZ.02.1.01/0.0/0.0/16 019/0000765 ``Research Center for Informatics", 
    by the European Union's Horizon 2020 research and innovation programme AERIAL-CORE under grant agreement no. 871479, 
    and by the NAKI II project no. DG18P02OVV069. 

    \bibliographystyle{apalike}
    \bibliography{main}

\begin{thebibliography}{}

\bibitem[Agha et~al., 2021]{agha2021nebula}
Agha, A., Otsu, K., Morrell, B., Fan, D.~D., Thakker, R., Santamaria{-}Navarro,
  A., Kim, S., Bouman, A., Lei, X., Edlund, J.~A., Ginting, M.~F., Ebadi, K.,
  Anderson, M., Pailevanian, T., Terry, E., Wolf, M.~T., Tagliabue, A.,
  Vaquero, T.~S., Palieri, M., Tepsuporn, S., Chang, Y., Kalantari, A., Chavez,
  F., Lopez, B.~T., Funabiki, N., Miles, G., Touma, T., Buscicchio, A.,
  Tordesillas, J., Alatur, N., Nash, J., Walsh, W., Jung, S., Lee, H.,
  Kanellakis, C., Mayo, J., Harper, S., Kaufmann, M., Dixit, A., Correa, G.,
  Lee, C., Gao, J., Merewether, G., Maldonado{-}Contreras, J., Salhotra, G.,
  da~Silva, M.~S., Ramtoula, B., Kubo, Y., Fakoorian, S.~A., Hatteland, A.,
  Kim, T., Bartlett, T., Stephens, A., Kim, L., Bergh, C., Heiden, E., Lew, T.,
  Cauligi, A., Heywood, T., Kramer, A., Leopold, H.~A., Choi, H.~C., Daftry,
  S., Toupet, O., Wee, I., Thakur, A., Feras, M., Beltrame, G., Nikolakopoulos,
  G., Shim, D.~H., Carlone, L., and Burdick, J. (2021).
\newblock Nebula: Quest for robotic autonomy in challenging environments;
  {TEAM} costar at the {DARPA} subterranean challenge.
\newblock {\em CoRR}, abs/2103.11470.

\bibitem[{Ahmad} et~al., 2021]{ahmad2021autonomous}
{Ahmad}, A., {Walter}, V., {Petracek}, P., {Petrlik}, M., {Baca}, T.,
  {Zaitlik}, D., and {Saska}, M. (2021).
\newblock Autonomous aerial swarming in gnss-denied environments with high
  obstacle density.
\newblock In {\em 2021 IEEE International Conference on Robotics and Automation
  (ICRA)}, pages 570--576. IEEE.

\bibitem[Ahmad et~al., 2021]{ahmad2021end}
Ahmad, S., Sunberg, Z.~N., and Humbert, J.~S. (2021).
\newblock End-to-end probabilistic depth perception and 3d obstacle avoidance
  using pomdp.
\newblock {\em Journal of Intelligent \& Robotic Systems}, 103(2):1--18.

\bibitem[Alismail et~al., 2016]{alismail2016direct}
Alismail, H., Kaess, M., Browning, B., and Lucey, S. (2016).
\newblock Direct visual odometry in low light using binary descriptors.
\newblock {\em IEEE Robotics and Automation Letters}, 2(2):444--451.

\bibitem[Alotaibi et~al., 2019]{alotaibi2019lsar}
Alotaibi, E.~T., Alqefari, S.~S., and Koubaa, A. (2019).
\newblock Lsar: Multi-uav collaboration for search and rescue missions.
\newblock {\em IEEE Access}, 7:55817--55832.

\bibitem[Baca et~al., 2018]{baca2018model}
Baca, T., Hert, D., Loianno, G., Saska, M., and Kumar, V. (2018).
\newblock {Model Predictive Trajectory Tracking and Collision Avoidance for
  Reliable Outdoor Deployment of Unmanned Aerial Vehicles}.
\newblock In {\em 2018 IEEE/RSJ International Conference on Intelligent Robots
  and Systems}, pages 1--8. IEEE.

\bibitem[Baca et~al., 2016]{baca2016embedded}
Baca, T., Loianno, G., and Saska, M. (2016).
\newblock {Embedded Model Predictive Control of Unmanned Micro Aerial
  Vehicles}.
\newblock In {\em 2016 IEEE International Conference on Methods and Models in
  Automation and Robotics (MMAR)}, pages 992--997.

\bibitem[Baca et~al., 2021]{baca2021mrs}
Baca, T., Petrlik, M., Vrba, M., Spurny, V., Penicka, R., Hert, D., and Saska,
  M. (2021).
\newblock The mrs uav system: Pushing the frontiers of reproducible research,
  real-world deployment, and education with autonomous unmanned aerial
  vehicles.
\newblock {\em Journal of Intelligent {\&} Robotic Systems}, 102(26):1--28.

\bibitem[Behley and Stachniss, 2018]{behley2018efficient}
Behley, J. and Stachniss, C. (2018).
\newblock Efficient surfel-based slam using 3d laser range data in urban
  environments.
\newblock In {\em Robotics: Science and Systems}, volume 2018, page~59.

\bibitem[Bircher et~al., 2016]{bircher2016receding}
Bircher, A., Kamel, M., Alexis, K., Oleynikova, H., and Siegwart, R. (2016).
\newblock Receding horizon" next-best-view" planner for 3d exploration.
\newblock In {\em 2016 IEEE international conference on robotics and automation
  (ICRA)}, pages 1462--1468. IEEE.

\bibitem[Bl{\"o}chliger et~al., 2018]{blochtlinger2018topomap}
Bl{\"o}chliger, F., Fehr, M., Dymczyk, M., Schneider, T., and Siegwart, R.~Y.
  (2018).
\newblock Topomap: Topological mapping and navigation based on visual slam
  maps.
\newblock {\em 2018 IEEE International Conference on Robotics and Automation
  (ICRA)}, pages 1--9.

\bibitem[Bosse et~al., 2012]{bosse2012zebedee}
Bosse, M., Zlot, R., and Flick, P. (2012).
\newblock Zebedee: Design of a spring-mounted 3-d range sensor with application
  to mobile mapping.
\newblock {\em IEEE Transactions on Robotics}, 28(5):1104--1119.

\bibitem[Burri et~al., 2015]{burri2015realtime}
Burri, M., Oleynikova, H., , Achtelik, M.~W., and Siegwart, R. (2015).
\newblock {Real-Time Visual-Inertial Mapping, Re-localization and Planning
  Onboard MAVs in Unknown Environments}.
\newblock In {\em 2015 IEEE/RSJ International Conference on Intelligent Robots
  and Systems}.

\bibitem[Cadena et~al., 2016]{cadena2016past}
Cadena, C., Carlone, L., Carrillo, H., Latif, Y., Scaramuzza, D., Neira, J.,
  Reid, I., and Leonard, J.~J. (2016).
\newblock Past, present, and future of simultaneous localization and mapping:
  Toward the robust-perception age.
\newblock {\em IEEE Transactions on robotics}, 32(6):1309--1332.

\bibitem[Chen et~al., 2019]{mmdetection}
Chen, K., Wang, J., Pang, J., Cao, Y., Xiong, Y., Li, X., Sun, S., Feng, W.,
  Liu, Z., Xu, J., Zhang, Z., Cheng, D., Zhu, C., Cheng, T., Zhao, Q., Li, B.,
  Lu, X., Zhu, R., Wu, Y., Dai, J., Wang, J., Shi, J., Ouyang, W., Loy, C.~C.,
  and Lin, D. (2019).
\newblock {MMDetection}: Open mmlab detection toolbox and benchmark.
\newblock {\em arXiv preprint arXiv:1906.07155}.

\bibitem[Chen-Lung et~al., 2022]{chenlung2022heterogeneous}
Chen-Lung, L., Jui-Te, H., Huang, C.-I., Zi-Yan, L., Chao-Chun, H., Yu-Yen, H.,
  Siao-Cing, H., Po-Kai, C., Zu~Lin, E., Po-Jui, H., Po-Lin, L., Bo-Hui, W.,
  Lai-Sum, Y., Sheng-Wei, H., MingSian~R., B., and Hsueh-Cheng, W. (2022).
\newblock A heterogeneous unmanned ground vehicle and blimp robot team for
  search and rescue using data-driven autonomy and communication-aware
  navigation.
\newblock {\em Field Robotics}, 2:557--594.

\bibitem[Dang et~al., 2019a]{dang2019explore}
Dang, T., Khattak, S., Mascarich, F., and Alexis, K. (2019a).
\newblock Explore locally, plan globally: A path planning framework for
  autonomous robotic exploration in subterranean environments.
\newblock In {\em 2019 19th International Conference on Advanced Robotics
  (ICAR)}, pages 9--16. IEEE.

\bibitem[Dang et~al., 2020a]{dang2020autonomous}
Dang, T., Mascarich, F., Khattak, S., Nguyen, H., Nguyen, H., Hirsh, S.,
  Reinhart, R., Papachristos, C., and Alexis, K. (2020a).
\newblock Autonomous search for underground mine rescue using aerial robots.
\newblock In {\em 2020 IEEE Aerospace Conference}, pages 1--8. IEEE.

\bibitem[Dang et~al., 2019b]{dang2019gbplanner}
Dang, T., Mascarich, F., Khattak, S., Papachristos, C., and Alexis, K. (2019b).
\newblock Graph-based path planning for autonomous robotic exploration in
  subterranean environments.
\newblock {\em IEEE/RSJ International Conference on Intelligent Robots and
  Systems (IROS)}, pages 3105--3112.

\bibitem[Dang et~al., 2020b]{dang2020graph}
Dang, T., Tranzatto, M., Khattak, S., Mascarich, F., Alexis, K., and Hutter, M.
  (2020b).
\newblock Graph-based subterranean exploration path planning using aerial and
  legged robots.
\newblock {\em Journal of Field Robotics}, 37(8):1363--1388.

\bibitem[Delmerico et~al., 2019]{delmerico2019current}
Delmerico, J., Mintchev, S., Giusti, A., Gromov, B., Melo, K., Horvat, T.,
  Cadena, C., Hutter, M., Ijspeert, A., Floreano, D., et~al. (2019).
\newblock The current state and future outlook of rescue robotics.
\newblock {\em Journal of Field Robotics}, 36(7):1171--1191.

\bibitem[Ebadi et~al., 2020]{ebadi2020lamp}
Ebadi, K., Chang, Y., Palieri, M., Stephens, A., Hatteland, A., Heiden, E.,
  Thakur, A., Funabiki, N., Morrell, B., Wood, S., et~al. (2020).
\newblock Lamp: Large-scale autonomous mapping and positioning for exploration
  of perceptually-degraded subterranean environments.
\newblock In {\em 2020 IEEE International Conference on Robotics and Automation
  (ICRA)}, pages 80--86. IEEE.

\bibitem[Ebadi et~al., 2021]{ebadi2021dare}
Ebadi, K., Palieri, M., Wood, S., Padgett, C., and Agha-mohammadi, A.-a.
  (2021).
\newblock Dare-slam: Degeneracy-aware and resilient loop closing in
  perceptually-degraded environments.
\newblock {\em Journal of Intelligent \& Robotic Systems}, 102(1):1--25.

\bibitem[Fabris et~al., 2021]{fabris2021soft}
Fabris, A., Kirchgeorg, S., and Mintchev, S. (2021).
\newblock A soft drone with multi-modal mobility for the exploration of
  confined spaces.
\newblock In {\em 2021 IEEE International Symposium on Safety, Security, and
  Rescue Robotics (SSRR)}, pages 48--54.

\bibitem[Ginting et~al., 2021]{ginting2021chord}
Ginting, M.~F., Otsu, K., Edlund, J.~A., Gao, J., and Agha-Mohammadi, A.-A.
  (2021).
\newblock Chord: Distributed data-sharing via hybrid ros 1 and 2 for
  multi-robot exploration of large-scale complex environments.
\newblock {\em IEEE Robotics and Automation Letters}, 6(3):5064--5071.

\bibitem[Goodfellow et~al., 2014]{goodfellow2014generative}
Goodfellow, I., Pouget-Abadie, J., Mirza, M., Xu, B., Warde-Farley, D., Ozair,
  S., Courville, A., and Bengio, Y. (2014).
\newblock Generative adversarial nets.
\newblock {\em Advances in neural information processing systems}, 27.

\bibitem[{Hert} et~al., 2022]{hert2022hardware}
{Hert}, D., {Baca}, T., {Petracek}, P., {Kratky}, V., {Spurny}, V., {Petrlik},
  M., {Matous}, V., {Zaitlik}, D., {Stoudek}, P., {Walter}, V., {Stepan}, P.,
  {Horyna}, J., {Pritzl}, V., {Silano}, G., {Bonilla Licea}, D., {Stibinger},
  P., {Penicka}, R., {Nascimento}, T., and {Saska}, M. (2022).
\newblock {MRS Modular UAV Hardware Platforms for Supporting Research in
  Real-World Outdoor and Indoor Environments}.
\newblock In {\em {2022 International Conference on Unmanned Aircraft Systems
  (ICUAS)}}. IEEE.

\bibitem[Hess et~al., 2016]{hess2016real}
Hess, W., Kohler, D., Rapp, H., and Andor, D. (2016).
\newblock Real-time loop closure in 2d lidar slam.
\newblock In {\em 2016 IEEE international conference on robotics and automation
  (ICRA)}, pages 1271--1278. IEEE.

\bibitem[Hornung et~al., 2013]{hornung2013octomap}
Hornung, A., Wurm, K.~M., Bennewitz, M., Stachniss, C., and Burgard, W. (2013).
\newblock {OctoMap: An Efficient Probabilistic 3D Mapping Framework Based on
  Octrees}.
\newblock {\em Autonomous Robots}, 34:189--206.

\bibitem[Huang et~al., 2019]{huang2019duckiefloat}
Huang, Y.-W., Lu, C.-L., Chen, K.-L., Ser, P.-S., Huang, J.-T., Shen, Y.-C.,
  Chen, P.-W., Chang, P.-K., Lee, S.-C., and Wang, H.-C. (2019).
\newblock Duckiefloat: a collision-tolerant resource-constrained blimp for
  long-term autonomy in subterranean environments.
\newblock {\em arXiv preprint arXiv:1910.14275}.

\bibitem[Hudson et~al., 2022]{hudson2021heterogeneous}
Hudson, N., Talbot, F., Cox, M., Williams, J.~L., Hines, T., Pitt, A., Wood,
  B., Frousheger, D., Surdo, K.~L., Molnar, T., Steindl, R., Wildie, M., Sa,
  I., Kottege, N., Stepanas, K., Hern{\'{a}}ndez, E., Catt, G., Docherty, W.,
  Tidd, B., Tam, B., Murrell, S., Bessell, M., Hanson, L., Tychsen{-}Smith, L.,
  Suzuki, H., Overs, L., Kendoul, F., Wagner, G., Palmer, D., Milani, P.,
  O'Brien, M., Jiang, S., Chen, S., and Arkin, R.~C. (2022).
\newblock Heterogeneous ground and air platforms, homogeneous sensing: Team
  {CSIRO} data61's approach to the {DARPA} subterranean challenge.
\newblock {\em Field Robotics}, 2:595--636.

\bibitem[Kasper et~al., 2019]{kasper2019benchmark}
Kasper, M., McGuire, S., and Heckman, C. (2019).
\newblock A benchmark for visual-inertial odometry systems employing onboard
  illumination.
\newblock In {\em 2019 IEEE/RSJ International Conference on Intelligent Robots
  and Systems (IROS)}, pages 5256--5263. IEEE.

\bibitem[Khattak et~al., 2019]{khattak2019robust}
Khattak, S., Mascarich, F., Dang, T., Papachristos, C., and Alexis, K. (2019).
\newblock Robust thermal-inertial localization for aerial robots: A case for
  direct methods.
\newblock In {\em 2019 International Conference on Unmanned Aircraft Systems
  (ICUAS)}, pages 1061--1068. IEEE.

\bibitem[Khattak et~al., 2020]{khattak2020complementary}
Khattak, S., Nguyen, H., Mascarich, F., Dang, T., and Alexis, K. (2020).
\newblock Complementary multi–modal sensor fusion for resilient robot pose
  estimation in subterranean environments.
\newblock In {\em 2020 International Conference on Unmanned Aircraft Systems
  (ICUAS)}, pages 1024--1029.

\bibitem[Kohlbrecher et~al., 2011]{kohlbrecher2011flexible}
Kohlbrecher, S., Meyer, J., von Stryk, O., and Klingauf, U. (2011).
\newblock A flexible and scalable slam system with full 3d motion estimation.
\newblock In {\em Proc. IEEE International Symposium on Safety, Security and
  Rescue Robotics (SSRR)}. IEEE.

\bibitem[Kramer et~al., 2021]{kramer2021vi}
Kramer, A., Kasper, M., and Heckman, C. (2021).
\newblock Vi-slam for subterranean environments.
\newblock In {\em Field and Service Robotics}, pages 159--172. Springer.

\bibitem[{Krátký} et~al., 2021]{kratky2021aerialfilming}
{Krátký}, V., {Alcántara}, A., {Capitán}, J., {Štěpán}, P., {Saska}, M.,
  and {Ollero}, A. (2021).
\newblock Autonomous aerial filming with distributed lighting by a team of
  unmanned aerial vehicles.
\newblock {\em IEEE Robotics and Automation Letters}, 6(4):7580--7587.

\bibitem[Krátký et~al., 2021a]{kratky2021exploration}
Krátký, V., Petráček, P., Báča, T., and Saska, M. (2021a).
\newblock An autonomous unmanned aerial vehicle system for fast exploration of
  large complex indoor environments.
\newblock {\em Journal of Field Robotics}, 38(8):1036--1058.

\bibitem[Krátký et~al., 2021b]{kratky2021documentation}
Krátký, V., Petráček, P., Nascimento, T., Čadilová, M., Škobrtal, M.,
  Stoudek, P., and Saska, M. (2021b).
\newblock Safe documentation of historical monuments by an autonomous unmanned
  aerial vehicle.
\newblock {\em ISPRS International Journal of Geo-Information},
  10(11):738/1--16.

\bibitem[Kulkarni et~al., 2021]{kulkarni2021autonomous}
Kulkarni, M., Dharmadhikari, M., Tranzatto, M., Zimmermann, S., Reijgwart, V.,
  De~Petris, P., Nguyen, H., Khedekar, N., Papachristos, C., Ott, L., et~al.
  (2021).
\newblock Autonomous teamed exploration of subterranean environments using
  legged and aerial robots.
\newblock {\em arXiv preprint arXiv:2111.06482}.

\bibitem[Lajoie et~al., 2020]{lajoie2020door}
Lajoie, P.-Y., Ramtoula, B., Chang, Y., Carlone, L., and Beltrame, G. (2020).
\newblock Door-slam: Distributed, online, and outlier resilient slam for
  robotic teams.
\newblock {\em IEEE Robotics and Automation Letters}, 5(2):1656--1663.

\bibitem[Lee et~al., 2010]{lee2010geometric}
Lee, T. et~al. (2010).
\newblock {Geometric tracking control of a quadrotor UAV on SE(3)}.
\newblock In {\em {2010 IEEE Conference on Decision and Control}}, pages
  5420--5425. IEEE.

\bibitem[Lepetit et~al., 2008]{lepetit2008EPnPAA}
Lepetit, V., Moreno-Noguer, F., and Fua, P. (2008).
\newblock {EPnP: An Accurate O(n) Solution to the PnP Problem}.
\newblock {\em International Journal of Computer Vision}, 81:155--166.

\bibitem[Lindqvist et~al., 2021]{lindqvist2021compra}
Lindqvist, B., Kanellakis, C., Mansouri, S.~S., Agha-mohammadi, A.-a., and
  Nikolakopoulos, G. (2021).
\newblock Compra: A compact reactive autonomy framework for subterranean mav
  based search-and-rescue operations.
\newblock {\em arXiv preprint arXiv:2108.13105}.

\bibitem[Loshchilov and Hutter, 2016]{cosineLR}
Loshchilov, I. and Hutter, F. (2016).
\newblock Sgdr: Stochastic gradient descent with warm restarts.

\bibitem[Lu et~al., 2020]{lu2020see}
Lu, C.~X., Rosa, S., Zhao, P., Wang, B., Chen, C., Stankovic, J.~A., Trigoni,
  N., and Markham, A. (2020).
\newblock See through smoke: robust indoor mapping with low-cost mmwave radar.
\newblock In {\em Proceedings of the 18th International Conference on Mobile
  Systems, Applications, and Services}, pages 14--27.

\bibitem[Martinez-Rozas et~al., 2022]{martinez2022skyeye}
Martinez-Rozas, S., Rey, R., Alejo, D., Acedo, D., Cobano, J.~A.,
  Rodriguez-Ramos, A., Campoy, P., Merino, L., and Caballero, F. (2022).
\newblock An aerial/ground robot team for autonomous firefighting in urban
  gnss-denied scenarios.
\newblock {\em Field Robotics}, 2:241--273.

\bibitem[Moniruzzaman et~al., 2022]{moniruzzaman2022teleoperation}
Moniruzzaman, M., Rassau, A., Chai, D., and Islam, S. M.~S. (2022).
\newblock Teleoperation methods and enhancement techniques for mobile robots: A
  comprehensive survey.
\newblock {\em Robotics and Autonomous Systems}, 150:103973.

\bibitem[Murphy et~al., 2009]{murphy2009mobile}
Murphy, R.~R., Kravitz, J., Stover, S.~L., and Shoureshi, R. (2009).
\newblock Mobile robots in mine rescue and recovery.
\newblock {\em IEEE Robotics \& Automation Magazine}, 16(2):91--103.

\bibitem[Museth, 2013]{museth2013vdb}
Museth, K. (2013).
\newblock Vdb: High-resolution sparse volumes with dynamic topology.
\newblock {\em ACM transactions on graphics (TOG)}, 32(3):1--22.

\bibitem[Musil et~al., 2022]{musil2022spheremap}
Musil, T., Petrlik, M., and Saska, M. (2022).
\newblock {SphereMap: Dynamic Multi-Layer Graph Structure for Rapid
  Safety-Aware UAV Planning}.
\newblock {\em IEEE Robotics and Automation Letters}.
\newblock Under review.

\bibitem[Ohradzansky et~al., 2021]{ohradzansky2021multi}
Ohradzansky, M.~T., Rush, E.~R., Riley, D.~G., Mills, A.~B., Ahmad, S.,
  McGuire, S., Biggie, H., Harlow, K., Miles, M.~J., Frew, E.~W., et~al.
  (2021).
\newblock Multi-agent autonomy: Advancements and challenges in subterranean
  exploration.
\newblock {\em arXiv preprint arXiv:2110.04390}.

\bibitem[Orekhov and Chung, 2022]{orekhov2022darpa}
Orekhov, V. and Chung, T. (2022).
\newblock The darpa subterranean challenge: A synopsis of the circuits stage.
\newblock {\em Field Robotics}, 2:735--747.

\bibitem[O’Meadhra et~al., 2018]{o2018variable}
O’Meadhra, C., Tabib, W., and Michael, N. (2018).
\newblock Variable resolution occupancy mapping using gaussian mixture models.
\newblock {\em IEEE Robotics and Automation Letters}, 4(2):2015--2022.

\bibitem[Palieri et~al., 2020]{palieri2020locus}
Palieri, M., Morrell, B., Thakur, A., Ebadi, K., Nash, J., Chatterjee, A.,
  Kanellakis, C., Carlone, L., Guaragnella, C., and Agha-mohammadi, A.-a.
  (2020).
\newblock Locus: A multi-sensor lidar-centric solution for high-precision
  odometry and 3d mapping in real-time.
\newblock {\em IEEE Robotics and Automation Letters}, 6(2):421--428.

\bibitem[Papachristos et~al., 2017]{papachristos2017uncertainty}
Papachristos, C., Khattak, S., and Alexis, K. (2017).
\newblock Uncertainty-aware receding horizon exploration and mapping using
  aerial robots.
\newblock In {\em 2017 IEEE international conference on robotics and automation
  (ICRA)}, pages 4568--4575. IEEE.

\bibitem[Papachristos et~al., 2019a]{papachristos2019autonomous}
Papachristos, C., Khattak, S., Mascarich, F., and Alexis, K. (2019a).
\newblock Autonomous navigation and mapping in underground mines using aerial
  robots.
\newblock In {\em 2019 IEEE Aerospace Conference}, pages 1--8. IEEE.

\bibitem[Papachristos et~al., 2019b]{papachristos2019localization}
Papachristos, C., Mascarich, F., Khattak, S., Dang, T., and Alexis, K. (2019b).
\newblock Localization uncertainty-aware autonomous exploration and mapping
  with aerial robots using receding horizon path-planning.
\newblock {\em Autonomous Robots}, 43(8):2131--2161.

\bibitem[Petracek et~al., 2021]{petracek2021caves}
Petracek, P., Kratky, V., Petrlik, M., Baca, T., Kratochvil, R., and Saska, M.
  (2021).
\newblock {Large-Scale Exploration of Cave Environments by Unmanned Aerial
  Vehicles}.
\newblock {\em IEEE Robotics and Automation Letters}, 6(4):7596--7603.

\bibitem[Petrl\'{i}k et~al., 2020]{petrlik2020robust}
Petrl\'{i}k, M., B\'{a}\v{c}a, T., He\v{r}t, D., Vrba, M., Krajn\'{i}k, T., and
  Saska, M. (2020).
\newblock {A Robust UAV System for Operations in a Constrained Environment}.
\newblock {\em IEEE Robotics and Automation Letters}, 5(2):2169--2176.

\bibitem[Petrl{\'\i}k et~al., 2021]{petrlik2021lidar}
Petrl{\'\i}k, M., Krajn{\'\i}k, T., and Saska, M. (2021).
\newblock Lidar-based stabilization, navigation and localization for uavs
  operating in dark indoor environments.
\newblock In {\em 2021 International Conference on Unmanned Aircraft Systems
  (ICUAS)}, pages 243--251. IEEE.

\bibitem[Pritzl et~al., 2021]{pritzl2021autonomous}
Pritzl, V., Stepan, P., and Saska, M. (2021).
\newblock Autonomous flying into buildings in a firefighting scenario.
\newblock In {\em 2021 IEEE International Conference on Robotics and Automation
  (ICRA)}, pages 239--245. IEEE.

\bibitem[Pritzl et~al., 2022]{pritzl2022repredictor}
Pritzl, V., Vrba, M., Tortorici, C., Ashour, R., and Saska, M. (2022).
\newblock Adaptive estimation of uav altitude in complex indoor environments
  using degraded and time-delayed measurements with time-varying uncertainties.
\newblock {\em submitted to Robotics and Autonomous Systems}.

\bibitem[Queralta et~al., 2020]{queralta2020collaborative}
Queralta, J.~P., Taipalmaa, J., Pullinen, B.~C., Sarker, V.~K., Gia, T.~N.,
  Tenhunen, H., Gabbouj, M., Raitoharju, J., and Westerlund, T. (2020).
\newblock Collaborative multi-robot search and rescue: Planning, coordination,
  perception, and active vision.
\newblock {\em Ieee Access}, 8:191617--191643.

\bibitem[Redmon and Farhadi, 2018]{redmon2018yolov3}
Redmon, J. and Farhadi, A. (2018).
\newblock Yolov3: An incremental improvement.
\newblock {\em arXiv preprint arXiv:1804.02767}.

\bibitem[Richter et~al., 2016]{richter2016polynomial}
Richter, C., Bry, A., and Roy, N. (2016).
\newblock {Polynomial trajectory planning for aggressive quadrotor flight in
  dense indoor environments}.
\newblock In {\em Robotics Research}, pages 649--666. Springer.

\bibitem[Rogers et~al., 2020a]{rogers2020test}
Rogers, J.~G., Gregory, J.~M., Fink, J., and Stump, E. (2020a).
\newblock Test your slam! the subt-tunnel dataset and metric for mapping.
\newblock In {\em 2020 IEEE International Conference on Robotics and Automation
  (ICRA)}, pages 955--961.

\bibitem[Rogers et~al., 2020b]{rogers2020darpa}
Rogers, J.~G., Schang, A., Nieto-Granda, C., Ware, J., Carter, J., Fink, J.,
  and Stump, E. (2020b).
\newblock The darpa subt urban circuit mapping dataset and evaluation metric.
\newblock In {\em International Symposium on Experimental Robotics}, pages
  391--401. Springer.

\bibitem[Rou{\v{c}}ek et~al., 2019]{roucek2019darpa}
Rou{\v{c}}ek, T., Pecka, M., {\v{C}}{\'\i}{\v{z}}ek, P.,
  Pet{\v{r}}{\'\i}{\v{c}}ek, T., Bayer, J., {\v{S}}alansk{\`y}, V., He{\v{r}}t,
  D., Petrl{\'\i}k, M., B{\'a}{\v{c}}a, T., Spurn{\`y}, V., et~al. (2019).
\newblock Darpa subterranean challenge: Multi-robotic exploration of
  underground environments.
\newblock In {\em International Conference on Modelling and Simulation for
  Autonomous Systems}, pages 274--290. Springer.

\bibitem[Rou{\v{c}}ek et~al., 2020]{roucek2020urban}
Rou{\v{c}}ek, T., Pecka, M., {\v{C}}{\'i}{\v{z}}ek, P.,
  Pet{\v{r}}{\'i}{\v{c}}ek, T., Bayer, J., {\v{S}}alansk{\'y}, V., He{\v{r}}t,
  D., Petrl{\'i}k, M., B{\'a}{\v{c}}a, T., Spurn{\'y}, V., Pomerleau, F.,
  Kubelka, V., Faigl, J., Zimmermann, K., Saska, M., Svoboda, T., and
  Krajn{\'i}k, T. (2020).
\newblock Darpa subterranean challenge: Multi-robotic exploration of
  underground environments.
\newblock In Mazal, J., Fagiolini, A., and Vasik, P., editors, {\em Modelling
  and Simulation for Autonomous Systems}, pages 274--290, Cham. Springer
  International Publishing.

\bibitem[Sandler et~al., 2018]{mobilenet}
Sandler, M., Howard, A., Zhu, M., Zhmoginov, A., and Chen, L.-C. (2018).
\newblock Mobilenetv2: Inverted residuals and linear bottlenecks.
\newblock In {\em Proceedings of the IEEE conference on computer vision and
  pattern recognition}, pages 4510--4520.

\bibitem[Santamaria-Navarro et~al., 2019]{santamaria2019towards}
Santamaria-Navarro, A., Thakker, R., Fan, D.~D., Morrell, B., and
  Agha-mohammadi, A.-a. (2019).
\newblock Towards resilient autonomous navigation of drones.
\newblock In {\em The International Symposium of Robotics Research}, pages
  922--937. Springer.

\bibitem[Scherer et~al., 2022]{scherer2022resilient}
Scherer, S., Agrawal, V., Best, G., Cao, C., Cujic, K., Darnley, R., DeBortoli,
  R., Dexheimer, E., Drozd, B., Garg, R., et~al. (2022).
\newblock Resilient and modular subterranean exploration with a team of roving
  and flying robots.
\newblock {\em Field Robotics}, 2:678–734.

\bibitem[Shakhatreh et~al., 2019]{shakhatreh2019unmanned}
Shakhatreh, H., Sawalmeh, A.~H., Al-Fuqaha, A., Dou, Z., Almaita, E., Khalil,
  I., Othman, N.~S., Khreishah, A., and Guizani, M. (2019).
\newblock Unmanned aerial vehicles (uavs): A survey on civil applications and
  key research challenges.
\newblock {\em Ieee Access}, 7:48572--48634.

\bibitem[Shan et~al., 2020]{liosam2020shan}
Shan, T., Englot, B., Meyers, D., Wang, W., Ratti, C., and Daniela, R. (2020).
\newblock Lio-sam: Tightly-coupled lidar inertial odometry via smoothing and
  mapping.
\newblock In {\em IEEE/RSJ International Conference on Intelligent Robots and
  Systems (IROS)}, pages 5135--5142. IEEE.

\bibitem[Shin et~al., 2020]{shin2020dvl}
Shin, Y.-S., Park, Y.~S., and Kim, A. (2020).
\newblock Dvl-slam: sparse depth enhanced direct visual-lidar slam.
\newblock {\em Autonomous Robots}, 44(2):115--130.

\bibitem[{Silano} et~al., 2021]{silano2021powerline}
{Silano}, G., {Baca}, T., {Penicka}, R., {Liuzza}, D., and {Saska}, M. (2021).
\newblock Power line inspection tasks with multi-aerial robot systems via
  signal temporal logic specifications.
\newblock {\em IEEE Robotics and Automation Letters}, 6(2):4169--4176.

\bibitem[Smith, 2015]{cyclicLR}
Smith, L.~N. (2015).
\newblock Cyclical learning rates for training neural networks.

\bibitem[Spurny et~al., 2021]{spurny2021autonomous}
Spurny, V., Pritzl, V., Walter, V., Petrlik, M., Baca, T., Stepan, P., Zaitlik,
  D., and Saska, M. (2021).
\newblock Autonomous firefighting inside buildings by an unmanned aerial
  vehicle.
\newblock {\em IEEE Access}, 9:15872--15890.

\bibitem[Tardioli et~al., 2019]{tardioli2019ground}
Tardioli, D., Riazuelo, L., Sicignano, D., Rizzo, C., Lera, F., Villarroel,
  J.~L., and Montano, L. (2019).
\newblock Ground robotics in tunnels: Keys and lessons learned after 10 years
  of research and experiments.
\newblock {\em Journal of Field Robotics}, 36(6):1074--1101.

\bibitem[Tomic et~al., 2012]{tomic2012toward}
Tomic, T., Schmid, K., Lutz, P., Domel, A., Kassecker, M., Mair, E., Grixa,
  I.~L., Ruess, F., Suppa, M., and Burschka, D. (2012).
\newblock Toward a fully autonomous uav: Research platform for indoor and
  outdoor urban search and rescue.
\newblock {\em IEEE robotics \& automation magazine}, 19(3):46--56.

\bibitem[Tranzatto et~al., 2022]{tranzatto2022cerberus}
Tranzatto, M., Mascarich, F., Bernreiter, L., Godinho, C., Camurri, M.,
  Khattak, S., Dang, T., Reijgwart, V., L¨oje, J., Wisth, D., Zimmermann, S.,
  Nguyen, H., Fehr, M., Solanka, L., Buchanan, R., Bjelonic, M., Khedekar, N.,
  Valceschini, M., Jenelten, F., and Alexis, K. (2022).
\newblock Cerberus: Autonomous legged and aerial robotic exploration in the
  tunnel and urban circuits of the darpa subterranean challenge.
\newblock {\em Field Robotics}, 2:274--324.

\bibitem[{Vrba} et~al., 2019]{vrba_ral2019}
{Vrba}, M., {Heřt}, D., and {Saska}, M. (2019).
\newblock Onboard marker-less detection and localization of non-cooperating
  drones for their safe interception by an autonomous aerial system.
\newblock {\em IEEE Robotics and Automation Letters}, 4(4):3402--3409.

\bibitem[Walter et~al., 2018]{walter2017mesas}
Walter, V., Nov{\'a}k, T., and Saska, M. (2018).
\newblock Self-localization of unmanned aerial vehicles based on optical flow
  in onboard camera images.
\newblock In {\em Lecture Notes in Computer Science, vol 10756.}, Cham.
  Springer International Publishing.

\bibitem[Walter et~al., 2022]{walter2022fr}
Walter, V., Spurny, V., Petrlik, M., Báca, T., Zaitlík, D., Demkiv, L., , and
  Saska, M. (2022).
\newblock Extinguishing real fires by fully autonomous multirotor uavs in the
  mbzirc 2020 competition.
\newblock {\em Field Robotics}, 2:406--436.

\bibitem[Williams et~al., 2020]{williams2020online}
Williams, J., Jiang, S., O’Brien, M., Wagner, G., Hernandez, E., Cox, M.,
  Pitt, A., Arkin, R., and Hudson, N. (2020).
\newblock Online 3d frontier-based ugv and uav exploration using direct point
  cloud visibility.
\newblock In {\em 2020 IEEE International Conference on Multisensor Fusion and
  Integration for Intelligent Systems (MFI)}, pages 263--270. IEEE.

\bibitem[Xu et~al., 2022]{xu2022fast}
Xu, W., Cai, Y., He, D., Lin, J., and Zhang, F. (2022).
\newblock Fast-lio2: Fast direct lidar-inertial odometry.
\newblock {\em IEEE Transactions on Robotics}.

\bibitem[Zhang et~al., 2016]{zhang2016degeneracy}
Zhang, J., Kaess, M., and Singh, S. (2016).
\newblock On degeneracy of optimization-based state estimation problems.
\newblock In {\em 2016 IEEE International Conference on Robotics and Automation
  (ICRA)}, pages 809--816. IEEE.

\bibitem[Zhang and Singh, 2014]{zhang2014loam}
Zhang, J. and Singh, S. (2014).
\newblock Loam: Lidar odometry and mapping in real-time.
\newblock In {\em Robotics: Science and Systems}, volume~2.

\bibitem[Zhang and Singh, 2015]{zhang2015visual}
Zhang, J. and Singh, S. (2015).
\newblock Visual-lidar odometry and mapping: Low-drift, robust, and fast.
\newblock In {\em 2015 IEEE International Conference on Robotics and Automation
  (ICRA)}, pages 2174--2181. IEEE.

\bibitem[Zhao et~al., 2021]{zhao2021super}
Zhao, S., Zhang, H., Wang, P., Nogueira, L., and Scherer, S. (2021).
\newblock Super odometry: Imu-centric lidar-visual-inertial estimator for
  challenging environments.
\newblock In {\em 2021 IEEE/RSJ International Conference on Intelligent Robots
  and Systems (IROS)}, pages 8729--8736.

\end{thebibliography}

  \end{document}